\documentclass[10pt,twocolumn,letterpaper]{article}

\usepackage[pagenumbers]{cvpr}      %

\usepackage{cuted} %

\usepackage{currfile} %

\usepackage{caption} %
\makeatletter
\def\abstract{%
   \iftoggle{cvprpagenumbers}{}{\thispagestyle{empty}}%
   \centerline{\large\bf Abstract}%
   \vspace*{6pt}\parskip=0pt\noindent%
   \it\ignorespaces%
}

\makeatother

\usepackage{booktabs}
\usepackage{multirow}
\usepackage{placeins}
\makeatletter
\let\@fb@origFloatBarrier\FloatBarrier
\def\FloatBarrier{\@fb@origFloatBarrier
  \ifdim\pagetotal=\z@\ifx\@toplist\@empty\ifx\@dbltoplist\@empty
    \global\@colroom\@colht \global\vsize\@colht
  \fi\fi\fi}
\makeatother

\definecolor{cvprblue}{rgb}{0.21,0.49,0.74}
\usepackage[pagebackref,breaklinks,colorlinks,allcolors=cvprblue]{hyperref}

\title{Knocked Off Course:\ Tracing Uncommanded Motion from Perception to Action in Embodied VLAs}

\author{%
Dan Zlotnikov \qquad Nadav Alali \qquad Ohad Ben-Shahar\\
Ben-Gurion University of the Negev
}

\begin{document}
\crefname{appendix}{App.}{Apps.}
\Crefname{appendix}{App.}{Apps.}
\makeatletter
\let\cvpr@maketitle\@maketitle
\def\@maketitle{\cvpr@maketitle
    \vspace*{-24pt}
    \begin{center}
        \captionsetup{type=figure}
        \includegraphics[width=\textwidth]{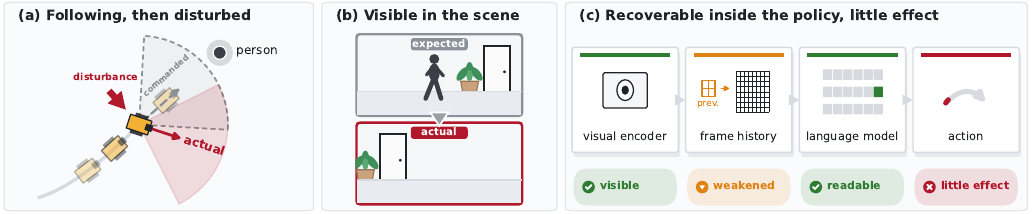}
        \caption{(a) A disturbance turns a person-following robot away from its target. (b) The camera records the realized motion: the target is gone, but the scene shift reveals the disturbance. (c) Inside a tracking VLA, small motion is visible but weakened by history compression, and large rotations are readable at the decision state, yet their decoded direction has little effect on the turn.}
        \label{fig:teaser}
    \end{center}
    }
\makeatother
\maketitle

\begin{abstract}
Vision-Language-Action models (VLAs) enable simultaneous processing of egocentric video and textual target descriptions to produce action commands for embodied tracking platforms. However, these platforms rarely move exactly as commanded, in part due to unexpected environmental disturbances such as wind, unleveled terrain, and road bumps. Here we study how such models handle uncommanded egomotion. First, we explore performance across diverse disturbances by benchmarking three tracking VLAs on three datasets. We find that models differ in which environmental disturbances hurt them most, but all are sensitive to sudden target loss. Tracing egomotion from pixels to emitted actions, we identify two gaps. In representation, motion is reliably recovered through explicit relations between frames, but aggressive compression of past frames weakens small-motion readouts. In action, large rotations do reach the decision state, yet the decoded direction carries only a few percent of a mirrored rotation's effect on the turn. Targeting both gaps, we fine-tune one model, Uni-NaVid, on data amounting to just $1.7\%$ of its original training set in a ladder of analysis-motivated variants. These gain $5$--$22$ points after target-loss without degrading clean success; after a smooth loss, the best one improves success by $72\%$ (relative), recovering $76\%$ of its gap to clean. We also show that targeted fine-tuning improves recovery partly via a text-mediated action dependency, without new architecture, large-scale data collection or training from scratch.
\end{abstract}

\section{Introduction}
\label{sec:intro}

Target tracking has been a core problem of computer vision for decades \cite{yilmaz2006tracking, marvastizadeh2022deeptracking}, evolving from estimating inter-frame motion \cite{lucas1981iterative} to matching a learned target template \cite{bertinetto2016siamfc} and, more recently, to following a target specified in natural language \cite{li2017trackinglanguage}. As large language models matured, integrating them into visual and embodied architectures enabled merging of perception and action, forming Vision-Language-Action models (VLAs) \cite{zitkovich2023rt2,kim2025openvla,ma2024vlasurvey}, which can be applied as embodied visual trackers. As such, these models no longer merely report where a described target appears: they generate the navigation actions that determine the platform's motion and, consequently, its future observations \cite{zhang2025uninavid,wang2025trackvla}.

As tracking systems move toward real-world deployment, they must be evaluated beyond clean, controlled benchmarks. A physical platform rarely executes a command exactly as planned: external forces add uncommanded motion as the platform moves through real world environments. The affected observations are not nuisance noise but a record of the platform's actual movement, which is most useful when the platform loses sight of the target and must react from its memory of earlier frames (\cref{fig:teaser}). It is unclear, however, how well tracking VLAs distinguish commanded motion from these uncommanded residuals when processing visual inputs. Existing studies mostly measure the resulting loss in task performance or train compensatory policies; they do not examine whether, and where, the discrepancy is represented inside the policy. To operate reliably outside controlled conditions, a VLA must detect the departure of its realized motion from commanded actions, and use that information to inform subsequent decisions.

We address this at three levels, starting with an evaluation of three tracking VLAs on three scene datasets in closed-loop simulation, under three environmental disturbances and two target-loss profiles, each along a controlled severity ladder. While success falls with severity in every model, each model fails in its own way, and all three fail worst when the target instantly leaves the view. We then look deeper, into each model's internal representations, tracing uncommanded motion from visual encoding to action, to ask whether failures arise because the signal is unavailable, discarded or ignored, and whether knowing it would let each policy recover. Our analysis finds the motion in every policy's visual features, and two gaps en route to the action. A representation gap: we find that small motions become hard to read when historical frames are compressed. An action gap: large rotations \emph{do} reach the decision state, and we show that knowing their direction would let each policy recover with its own actions, yet the decoded direction barely affects their turn, which follows the target, turning back barely more often than chance after a target is abruptly lost. At the intervention level, we target both gaps with a model adaptation rather than a new architecture. We fine-tune Uni-NaVid on data amounting to $1.7\%$ of its original training set in a ladder of variants that improves success, most after target-loss, by up to $22$ points from the frozen model's state at the loss, without losing clean success; the most complete variant gains most. Beyond performance recovery, fine-tuning changes what drives the turn: out of view the fine-tuned models correct the rotation more often, and those that state their motion act on what they state.

\noindent\textbf{Our contributions are:}
\begin{itemize}
    \item \textbf{A systematic closed-loop robustness benchmark} of three tracking VLAs under environmental disturbances and target-loss, on 2{,}102 episodes in three scene datasets.

    \item \textbf{A pixels-to-actions model internal analysis of uncommanded motion inside VLAs}, following the signal through model layers up to the decision state.

    \item \textbf{A laddered fine-tuning of Uni-NaVid} on minimal data that improves success, most after target-loss, and makes the model use its own motion more to correct its turns.
\end{itemize}

\section{Related Work}
\label{sec:related_work}

\subsection{VLA Robustness}
\label{sec:related_robustness}

Robustness of embodied policies was studied in navigation, where RobustNav perturbs RGB observations and transition dynamics \cite{chattopadhyay2021robustnav}. For manipulation VLAs, LIBERO-Plus varies scene, viewpoint, sensor, language and initial conditions \cite{fei2025liberoplus}, and RobustVLA adds action noise and external forces, finding action execution the most fragile modality \cite{guo2025vlarobustness}. These studies largely assume static goals; none tests whether a mobile VLA keeps tracking a moving target when its body, and with it the camera, is displaced relative to its commands (\cref{tab:related_benchmarks}; details: \cref{app:related}).

\subsection{Representation of Motion}
\label{sec:related_motion_representation}

Standard trackers localize targets in image coordinates, conflating target and camera motion \cite{wu2013benchmark,yan2021stark}, whereas embodied trackers \cite{wang2025trackvla} must separate the two because the policy moves the camera. Frozen visual features encode patch correspondence across nearby views \cite{elbanani2024probe3d}; vision-language models (VLMs) answer poorly about camera and egomotion in video \cite{lin2025camerabench,schaefer2026egodyn}, although ego speed \cite{winter2026speedblindspot} and the direction of object motion \cite{lee2026modirect} are decodable from their hidden states. Navigation VLAs likewise encode collisions they do not avoid \cite{salsa2026} and signal their own path deviation in a few attention heads \cite{navheads2026}. These studies probe or monitor; none tests causally whether a decodable motion signal drives a closed-loop policy's action. NaVid-style VLAs are known to compress historical frames far more than the current one \cite{zhang2024navid,zhang2025uninavid,wang2025trackvla}, and some works add flow-supervised, video-predictive or frame-difference history \cite{yang2026temporalflowvla,miao2026jepavla,na2026evomemvla}, but none models the gap between commanded and realized motion, which we trace from pixels to actions.

\subsection{Mitigating Physical Disturbances}
\label{sec:related_mitigation}

RobustVLA fine-tunes against worst-case action deviations \cite{guo2025vlarobustness}, and Self-Adaptive VLA learns a context from earlier rollouts to compensate persistent hardware errors \cite{zhang2026selfadaptivevla}. Pace-and-Path Correction \cite{zhang2026ppc} and VLA-Corrector \cite{pan2026vlacorrector} adjust action chunks from outside the backbone. In navigation, visual odometry replaces GPS and compass under actuation noise \cite{zhao2021visualodometry}, and TIC-VLA re-anchors decisions with measured egomotion \cite{huang2026ticvla}. None targets a mobile VLA following a moving target or chooses its change by locating where the policy loses the signal; we adapt a released model where our experiments locate the loss, showing that disturbance fine-tuning benefits from attending to the motion signal and its effect on the action.

\begin{table}[htbp]
    \centering
    \footnotesize
    \setlength{\tabcolsep}{4pt}
    \caption{Related benchmarks.}
    \label{tab:related_benchmarks}
    \begin{tabular}{@{}llccccc@{}}
    \toprule
     & Task & \rotatebox{90}{VLA policies} & \rotatebox{90}{Closed loop} & \rotatebox{90}{Moving target} & \rotatebox{90}{\shortstack[l]{Uncommanded\\motion}} & \rotatebox{90}{Target-loss} \\
    \midrule
    RobustNav~\cite{chattopadhyay2021robustnav} & navigation & -- & \checkmark & -- & \checkmark & -- \\
    LIBERO-Plus~\cite{fei2025liberoplus} & manipulation & \checkmark & \checkmark & -- & -- & -- \\
    RobustVLA~\cite{guo2025vlarobustness} & manipulation & \checkmark & \checkmark & -- & \checkmark & -- \\
    CameraBench~\cite{lin2025camerabench} & camera motion & -- & -- & -- & -- & -- \\
    EgoDyn-Bench~\cite{schaefer2026egodyn} & egomotion & \checkmark & -- & -- & -- & -- \\
    EVT-Bench~\cite{wang2025trackvla} & tracking & \checkmark & \checkmark & \checkmark & -- & -- \\
    \textbf{Ours} & \textbf{tracking} & $\pmb{\checkmark}$ & $\pmb{\checkmark}$ & $\pmb{\checkmark}$ & $\pmb{\checkmark}$ & $\pmb{\checkmark}$ \\
    \bottomrule
    \end{tabular}
\end{table}

\section{Benchmark}
\label{sec:benchmark}

The benchmark is designed around four key aspects: disturbances displace the robot, never the pixels, so the camera records realized motion and evidence needed to correct it stays observable; one physical severity ladder, in units of the robot's step, applies to every model; every episode receives the same seeded disturbance in every run and is paired with its own clean run; and two separately tested failure modes: persistent displacement of the body, which the policy can correct while it still sees the target, and disturbance-induced loss of the target from view. Complete protocols and parameters are in \cref{sec:supp_design}.

\subsection{Models and Datasets}
\label{sec:benchmark_models}

\noindent\textbf{Models.} We use released weights, greedy decoding and the authors' EVT-Bench interfaces.
\begin{itemize}
    \item \emph{Uni-NaVid} \cite{zhang2025uninavid}: a 7B NaVid-style policy \cite{zhang2024navid} that compresses each recent historical frame to four tokens and emits discrete commands.
    \item \emph{LightNav-0} \cite{lightnav0}: a 4B generalist whose released vision tower encodes its two latest frames jointly, as one two-frame patch, and which emits short-horizon trajectories.
    \item \emph{OmTrackVLA} \cite{omtrackvla2025}: a 0.6B TrackVLA-style policy \cite{wang2025trackvla} with compressed history and continuous waypoints.
\end{itemize}

Uni-NaVid moves $0.187$\,m or turns $9^\circ$ per step; the waypoint models take smaller steps and never turn more than $9^\circ$.

\textbf{Task and datasets.} EVT-Bench \cite{wang2025trackvla} requires a robot in Habitat \cite{savva2019habitat} to follow a language-specified target. We evaluate all models on the complete EVT-Bench validation splits of Matterport3D (MP3D; 721 episodes) \cite{chang2017matterport3d} and Habitat-Matterport 3D (HM3D; 684) \cite{ramakrishnan2021hm3d}, and on 697 episodes in the Habitat Synthetic Scenes Dataset (HSSD) \cite{khanna2024hssd}, which EVT-Bench does not cover; we build them from the EVT-Bench validation episodes (\cref{sec:supp_datasets}).

\textbf{Episodes.} Episodes last at most 300 steps and end on target collision or prolonged separation.

\subsection{Disturbance Definition}
\label{sec:benchmark_disturbances}

Disturbances displace the robot and camera after its commanded step; \cref{fig:disturbances} in \cref{sec:supp_disturbances} draws them from their definitions.

\paragraph{Environmental disturbances.}
These continuously alter realized motion:
\begin{itemize}
    \item \emph{Current:} constant world-frame translation in one episode-specific direction, as on a slope, a moving floor or in a stream.
    \item \emph{Veer:} fixed-sign body rotation on each translating step, a drive-train bias such as a worn wheel.
    \item \emph{Gust:} zero-mean correlated translation from an Ornstein--Uhlenbeck process \cite{uhlenbeck1930theory}, as from wind or uneven terrain.
\end{itemize}

The common severity ladder S1--S4 ($16$, $24$, $32$ and $48\%$) is relative to the $0.187$\,m translation or $9^\circ$ rotation.

\paragraph{Target-loss disturbances.}
These rotate the body, as a bump or a collision would, so that a centered, actively followed target leaves the field of view by a $10^\circ$, $20^\circ$ or $30^\circ$ margin (severities S1--S3); the side is sampled uniformly.

\begin{itemize}
\item An \emph{instant target-loss} removes the target between consecutive frames.
\item A \emph{smooth target-loss} applies the same rotation over several frames.
\end{itemize}

Their matched total rotation isolates the value of watching the target leave.

\subsection{Metrics}
\label{sec:benchmark_metrics}

We retain EVT-Bench's success rate (SR), tracking rate (TR) and collision rate (CR) \cite{wang2025trackvla}, and summarize each disturbance by the share of clean success it removes, $100\,(1-\mathrm{SR}/\mathrm{SR}_\mathrm{clean})$, which puts models with different clean success on one scale. Exposure differs between models: target-loss events may only fire while the target is tracked and \emph{veer} acts only on translating steps, so the share of episodes with a target-loss and of veered steps is reported with the losses, and target-loss is also scored on those episodes. For target-loss we further record whether it is re-acquired within 20 steps.

\textbf{Statistics.} Every disturbed number is paired per episode with its cell's clean run and carries a 95\% episode-bootstrap interval (5{,}000 resamples, seed 0); repeating a clean run of the target-loss protocol, also in reversed episode order, changes success by under one point (\cref{tab:supp_kick_reference}). Pairing matches episodes and disturbance seeds. Intervals throughout resample episodes, and are pointwise across layers and grids (\cref{sec:supp_metrics}).

\subsection{Benchmark Results}
\label{sec:results_benchmark}

\begin{figure}[htbp]
    \centering
    \includegraphics[width=\columnwidth, trim=0 8.8 0 0.8, clip]{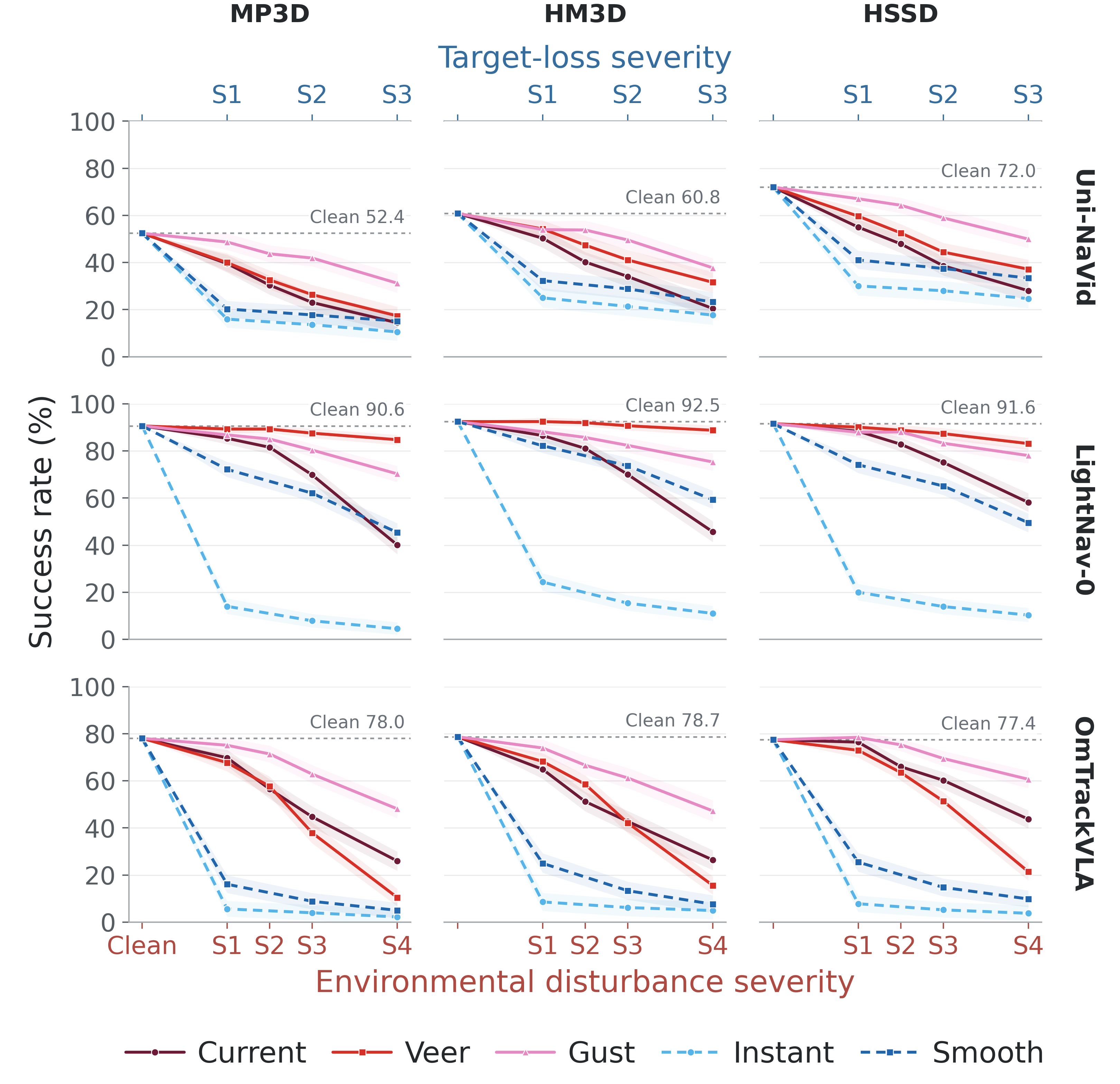}
    \caption{Success falls with disturbance severity. Solid: environmental disturbances against severity (lower axis); dashed: target-loss against its severity (upper axis); dotted: clean success. Bands indicate paired 95\% bootstrap intervals. Detailed analysis: \cref{sec:supp_results}.}
    \label{fig:results_combined}
\end{figure}

The benchmark is our first experimental suite: it characterizes how each model behaves under each disturbance, and its behavioral findings guide the internal analysis of \cref{sec:internal_analysis}. \Cref{fig:results_combined} summarizes the benchmark results; complete results are in \cref{sec:supp_results}.

\subsubsection{Environmental disturbances}
\label{sec:results_environmental}

\begin{figure}[htbp]
    \centering
    \includegraphics[width=\columnwidth, trim=0 6.4 0 9.4, clip]{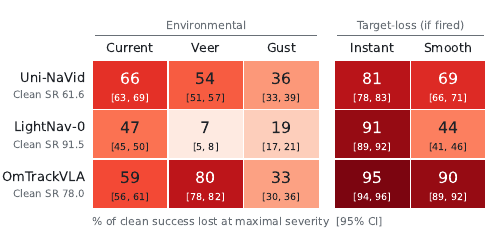}
    \caption{Share of clean success lost at the maximal severity of each disturbance, pooled over datasets (target-loss: if fired); brackets: 95\% paired-bootstrap intervals (full table: \cref{tab:supp_relative_loss}).}
    \label{fig:results_relative_loss}
\end{figure}

Success decreases nearly monotonically with severity, while smaller TR and CR changes show that failures are not explained mainly by early loss or target collision. Which disturbance hurts most depends on the architecture (\cref{fig:results_relative_loss}): LightNav-0 loses the least under all three, while \emph{current} costs Uni-NaVid the most ($66\%$ of its clean success) and \emph{veer} OmTrackVLA ($80\%$), so their ranking flips between disturbances: benchmarking on a single disturbance type would misrank them. \emph{Veer} acts only on translating steps, which make up $57\%$ of Uni-NaVid's steps, $80\%$ of LightNav-0's and $96\%$ of OmTrackVLA's. LightNav-0 is thus exposed to \emph{veer} more often than Uni-NaVid, yet loses only $7\%$ of its clean success against $54\%$: its resilience to \emph{veer} is a property of the policy, not of lower exposure.

\emph{Gust} is consistently less damaging than equal-magnitude persistent \emph{current} at high severity: persistence, not displacement magnitude, makes an environmental disturbance harmful.

\subsubsection{Target-loss disturbances}
\label{sec:results_target_loss}

Instant target-loss causes the largest performance drop for every policy (\cref{fig:results_relative_loss}). On the episodes in which it fires, an instant target-loss at maximal severity costs Uni-NaVid $81\%$ of its clean success, LightNav-0 $91\%$ and OmTrackVLA $95\%$; the weakest clean tracker keeps the most of its baseline. The results suggest that watching the target leave helps: smooth loss is less damaging across models, most for LightNav-0, which keeps the target in view through the rotation more often than the other two (\cref{tab:supp_target_loss_diag}), and least for OmTrackVLA. Together with its small loss under \emph{veer}, this suggests that LightNav-0 counters gradual rotation while it can see the target ($44\%$ lost after smooth loss), but not a rotation that removes the target at once ($91\%$).

\section{Model-Internal Analysis}
\label{sec:internal_analysis}

The benchmark leaves a central question open: success rates cannot tell whether the motion is never encoded, lost on the way, or arrives at the decision and is ignored, and each of these calls for a different remedy. We therefore open the three frozen policies and follow the motion signal from the visual encoder to the action, at every stage. This analysis aims to provide crucial findings for an informed intervention.

\subsection{Design}
\label{sec:internals_design}

All tests share one design: the three frozen policies replay the same recorded trajectories, so they receive identical inputs; every state is read at full dimension from lossless recordings; probes are fitted on clean driving only, in scene-held-out folds; every causal test carries a positive control and a random-direction null; and each test runs on all three datasets. Full details are described in \cref{sec:supp_internals_measure,app:internals}.

\subsubsection{Visual feature probing} 
Convolutional probes predict planar camera motion from the representations of two consecutive frames. Standard probes concatenate the frames; comparison probes receive cosine similarities between their nearby patches \cite{elbanani2024probe3d}, exposing motion held in the relation between frames. Disturbance sensitivity $b$ is the response to imposed motion after controlling for commanded motion ($0$: command only; $1$: imposed motion fully). To isolate history compression, we coarsen only the previous frame, down to the policy's history grid and through its projector, also on real video.

\subsubsection{Motion representation throughout model layers} From \cref{sec:results_rotation_backbone} onward we study large rotations: they can hide the target, leaving egomotion as the only evidence of where it went, and every policy can counter them with its own turns. A linear probe reads the rotation's side at every layer up to the decision state, on \emph{command-free} events: rotations with no turn command at the preceding decision, so the command cannot accidentally reveal the side.

\subsubsection{Sufficiency and use of the rotation's direction} After a target-loss, a scripted controller turns back with the platform's native actions, given the rotation's true direction (oracle) or a coin flip, with identical timing and turn budget. Readability does not establish use. Each command-free rotation is re-rendered with its side reversed and every earlier frame pixel-identical, and interchange \cite{geiger2021causal} copies states of the mirrored run into the original. With $D_e$ the mirror's effect on the turn at event $e$ and $\Delta_e$ a patch's effect, a patch carries the share
\begin{equation}
 M=\frac{\sum_e\operatorname{sign}(D_e)\,\Delta_e}{\sum_e|D_e|}.
\end{equation}
If the action used the direction the probe decodes, patching that component alone would carry the effect (positive control: all positions; null: norm-matched random directions). Steering, erasure and patches by position, image column and attention head complete the test.

\subsection{Internal Analysis Results}
\label{sec:results_internals}

Full analysis of each subsection, including per-dataset results, controls and statistics, is available in \cref{app:internals}.

\subsubsection{Evidence of uncommanded motion in early visual features}
\label{sec:results_motion_evidence}

\begin{figure}[htbp]
    \centering
    \includegraphics[width=0.9\columnwidth, trim=0 1.2 0 3.8, clip]{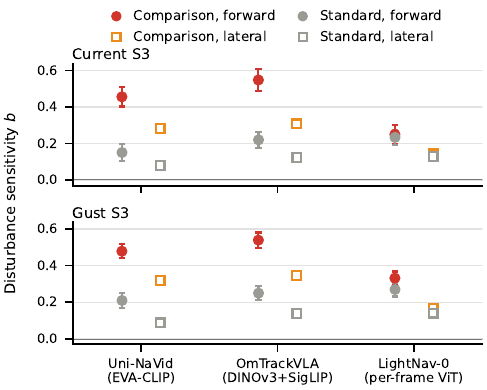}
    \caption{Each policy's visual features carry \emph{current} and \emph{gust} motion, equal-weight mean over datasets. Comparison probe (red/orange) versus standard probe (gray) on the same frame pairs; circles/squares: forward/lateral sensitivity $b$ (0: command only; 1: imposed motion fully). Bars: 95\% intervals. Per dataset results: \cref{app:internals_readers}.}
    \label{fig:internals_motion_evidence}
\end{figure}

On every policy's native feature grid, a comparison probe recovers forward and lateral motion under \emph{current} and \emph{gust} (\cref{fig:internals_motion_evidence}). The signal lives in the correspondence between consecutive frames.

The probe format matters. In Uni-NaVid and OmTrackVLA, a standard probe of the same two frames side by side reads less than the comparison on every dataset: their features hold motion in a relational form that explicit correspondence exposes, as do five diverse frozen image encoders (\cref{tab:app_reader_format}). LightNav-0, whose encoder takes two consecutive frames as one joint input, is the exception: both probes read its per-frame features about equally. We use the comparison probe below, refraining from reporting \emph{veer}: a clean-fitted probe cannot separate uncommanded yaw from command-correlated cues.

\subsubsection{Visual history compression effects}
\label{sec:results_history_compression}

\begin{figure}[htbp]
    \centering
    \includegraphics[width=0.9\columnwidth, trim=0 2.0 0 3.6, clip]{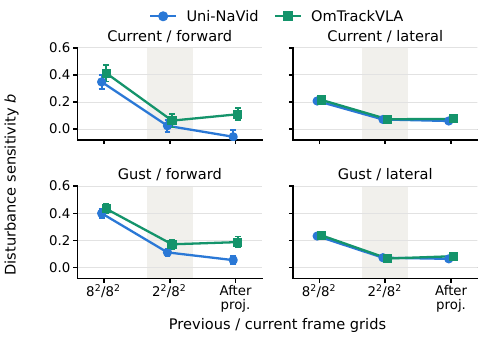}
    \caption{Coarsening only the previous frame (shaded) weakens \emph{current} and \emph{gust} readouts; the projector does not restore them. Bars: 95\% intervals. Per dataset results: \cref{app:internals_compression}.}
    \label{fig:internals_history_compression}
\end{figure}

\begin{figure}[htbp]
    \centering
    \includegraphics[width=\columnwidth, trim=0 2.9 0 3.3, clip]{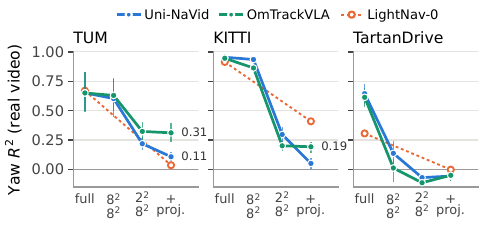}
    \caption{Real-video yaw $R^2$ along the history ladder: each encoder's full grid, both frames at $8\times8$, previous frame at $2\times2$, after the projector. LightNav-0 (dotted): separate-frame versus joint-pair probe, not matched stages. Whiskers: 95\% intervals; full values: \cref{app:internals_real}.}
    \label{fig:internals_real_history}
\end{figure}

Because the evidence lies in frame correspondence, the coarser frame of a pair affects what a comparison can read. Uni-NaVid and OmTrackVLA keep each recent historical frame as a $2\times2$ grid beside an $8\times8$ current frame. Coarsening the previous frame to $2\times2$ removes at least half of the probes' readout in both models on every dataset (\cref{fig:internals_history_compression} shows pooled results), and the native projector does not restore it. We analyze real video, which supports the same effect: on TUM \cite{sturm2012tum} and KITTI \cite{geiger2012kitti}, coarsening the previous frame removes about half or more of the yaw readout; on TartanDrive \cite{triest2022tartandrive} even the $8\times8$ grid loses it (\cref{fig:internals_real_history}).

LightNav-0, the most resilient to \emph{current} and \emph{gust} (\cref{sec:results_environmental}), differs here: its encoder sees the two most recent frames jointly, so the immediate predecessor is never coarsened and there is no coarse-history stage to ablate. 

\subsubsection{Large motion readability at the decision state}
\label{sec:results_rotation_backbone}

\begin{figure}[htbp]
    \centering
    \includegraphics[width=\columnwidth, trim=0 2.4 0 3.8, clip]{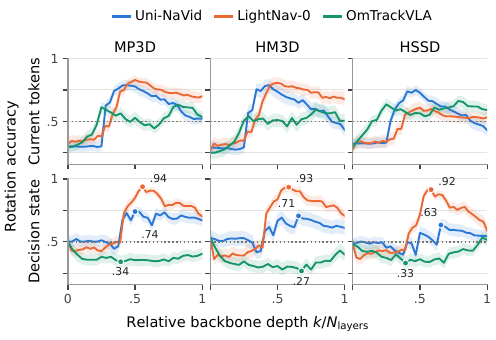}
    \caption{Sign accuracy of clean-fitted linear rotation probes across normalized backbone depth on command-free events: current visual tokens (top) and decision state (bottom; per model: \cref{tab:app_crossmodel_layout}). Below $0.5$: a reversed readout. Shading: 95\% intervals. Dots: each model's decision layer farthest from chance (\cref{tab:app_rotation_xd}).}
    \label{fig:internals_rotation_backbone}
\end{figure}

Unlike small motion, large rotations survive coarse history: against the current frame, a $2\times2$ previous frame gives a rotation's direction on $78$--$85\%$ of events (\cref{app:internals_rot_coarse}). They become readable at intermediate states of every backbone on every dataset (\cref{fig:internals_rotation_backbone}). The direction of a large rotation therefore reaches the decision state of Uni-NaVid and LightNav-0, and OmTrackVLA's through its own planned turn. The profiles hold on all three datasets, differing only in degree, and replays on the waypoint models' own trajectories preserve them (\cref{app:internals_grp_backbone,app:internals_onpolicy}). 

\subsubsection{Recovery is feasible within the existing action space}
\label{sec:results_recovery_feasibility}

\begin{figure}[htbp]
    \centering
    \includegraphics[width=0.9\columnwidth, trim=0 2.3 0 3.8, clip]{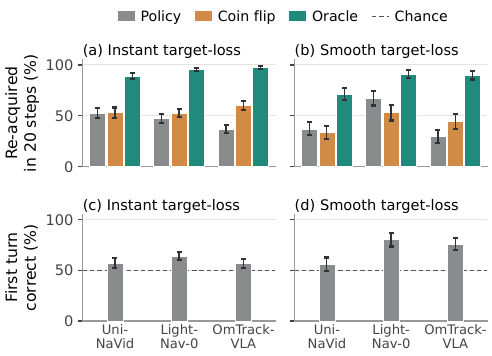}
    \caption{Recovery after target-loss, pooled over datasets. Bars indicate 95\% intervals; per dataset results: \cref{app:internals_functional}.}
    \label{fig:internals_recovery_pooled}
\end{figure}

\begin{figure}[htbp]
    \centering
    \includegraphics[width=0.9\columnwidth, trim=0 2.5 0 3.3, clip]{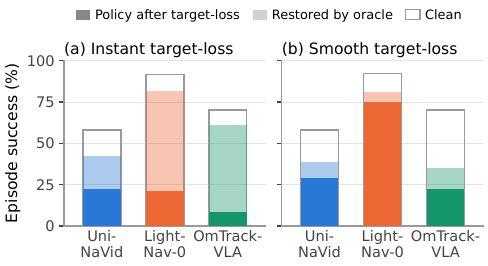}
    \caption{Episode success after target-loss, pooled over datasets. Outline: clean; solid: policy after the loss; light: restored by an oracle controller given the true direction from the simulator.}
    \label{fig:internals_recovery_episode}
\end{figure}

Knowing the rotation's direction significantly improves re-acquisition within 20 steps in all nine model--dataset cells (\cref{fig:internals_recovery_pooled}a,b; after smooth loss the controllers act only once the target is at least $10^\circ$ out of view): each policy's own actions suffice to recover; the policies fail to produce the corrective turn reliably. It also lifts episode success after instant loss, recovering most of LightNav-0's and OmTrackVLA's loss and about half of Uni-NaVid's smaller one; after smooth loss the gains are smaller, but still lift success by $6$--$13$ points (\cref{fig:internals_recovery_episode}).

The direction reaches their decision state (\cref{sec:results_rotation_backbone}), but do the policies act on it? Their first turn after the loss suggests an answer (\cref{fig:internals_recovery_pooled}c,d). After instant loss all three turn back only slightly more often than chance, although the direction is decodable at their decision state: decodability does not establish use. After smooth loss, which shows the target leaving, Uni-NaVid stays near chance, while the other two mostly turn back correctly. A correct turn is not enough: OmTrackVLA turns back on $76\%$ of smooth events but re-acquires the target on only $30\%$, matching its small gain from smooth loss (\cref{sec:results_target_loss}).

\subsubsection{Interchange interventions}
\label{sec:results_rotation_use}

\begin{figure}[htbp]
    \centering
    \includegraphics[width=\columnwidth]{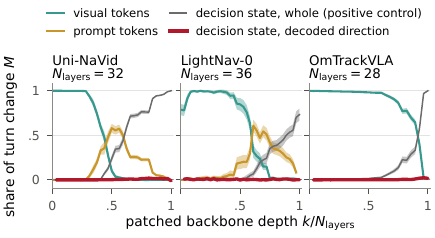}
    \caption{The decoded direction (red) carries under $3\%$ of the mirror's effect on the turn at every layer; the whole decision state, the positive control, reaches $73$--$100\%$. Share $M$ carried by patching each position group at every layer (results shown in MP3D; HM3D and HSSD are provided in \cref{fig:app_patch_route_xd}). Bands: 95\% intervals.}
    \label{fig:internals_patch_search}
\end{figure}

Patching the whole decision state increasingly controls the turn through depth, while its decoded rotation direction alone carries under $3\%$ of the mirror's effect there at any layer (\cref{fig:internals_patch_search}), and at most $5\%$ at the final state with the target out of view (\cref{app:internals_interchange}). Nor does the transfer depend on it: holding it fixed leaves Uni-NaVid's whole-state transfer unchanged (\cref{tab:app_interchange}). Where an edit at the decision state moves the turn at all, steering along the decoded direction never raises the corrective turn, and erasing it changes the share of corrective turns by at most one point (\cref{app:internals_causal}).

The effect follows the target, which mirroring also moves: image columns containing it carry at least $2.5$ times more of the effect per column than the rest, which still carry 28--45\%, and eight late attention heads carry $60\%$ of it in Uni-NaVid and $66\%$ in OmTrackVLA, against under $3\%$ for the decoded direction (\cref{fig:app_patch_heads}); in LightNav-0 it is spread across layers and components (\cref{app:internals_patch_search}). The turn thus follows mainly where the target appears: all three models turn toward a visible target after a rotation; out of view, only OmTrackVLA on MP3D still does, while Uni-NaVid and LightNav-0 turn \emph{with} the rotation more than against it in some cases (\cref{app:internals_mirror}).

\subsubsection{Model Internals Summary}
\label{sec:results_internals_summary}

The analysis locates two gaps. \emph{Representation gap:} uncommanded motion reaches every policy's visual features through frame correspondence, but the coarse history of Uni-NaVid and OmTrackVLA removes most of the small-motion readout. \emph{Action gap:} large motions, such as rotations, reach each decision state, and their direction would suffice for recovery, yet the decoded direction carries almost none of the effect on the decision, which concentrates on the target. These gaps motivate \cref{sec:interventions}, where we design fine-tuning experiments concentrating on our findings.

\section{Interventions}
\label{sec:interventions}

A deployed robot cannot easily be rebuilt around its disturbances, and new sensors change the hardware every robot carries. Training a new VLA from scratch is equally costly, requiring large-scale data collection and compute. Our analysis suggests none of this is needed -- the gaps lie in how egomotion is read and used, not in what the model sees. We therefore fine-tune an existing model from its released weights on a small amount of targeted data, keeping its single camera, native action space and architecture, so the fine-tuned model can replace the released one on the same robot.

\subsection{Fine-Tuning Design}
\label{sec:interventions_recipes}

We adapt Uni-NaVid with four recipes, each adding one change (\cref{tab:ft_recipes}):
\begin{itemize}
    \item \emph{Base}: fine-tunes on clean and rotation episodes.
    \item \emph{Decompress}: adds the previous frame at the current frame's $8\times8$ grid, against the representation gap (\cref{sec:results_history_compression}).
    \item \emph{Say}: targets the action gap (\cref{sec:results_rotation_use}): its generated text first states the robot's own motion and the target's side, then the action.
    \item \emph{Commit}: penalizes repeating a turn when the statements' sides are swapped, committing the action to the statement.
\end{itemize}

\begin{table}[htbp]
    \centering
    \footnotesize
    \setlength{\tabcolsep}{3pt}
    \caption{Fine-tuning recipes. Each adds one change to the row above; all share data, hyperparameters and evaluation; \emph{Commit} adds its swapped copies. CE: cross-entropy on the answer; S, P: the Self and Person statements; swap: the penalty on swapped-statement copies (\cref{app:interventions_training}).}
    \label{tab:ft_recipes}
    \begin{tabular}{@{}llll@{}}
    \toprule
    Recipe & Previous frame & Answer & Loss \\
    \midrule
    Base & $2\times2$ & 4 words & CE \\
    Decompress & $+\,8\times8$ & 4 words & CE \\
    Say & $+\,8\times8$ & S, P, 4 words & CE \\
    Commit & $+\,8\times8$ & S, P, 4 words & CE + swap \\
    \bottomrule
    \end{tabular}
\end{table}

\noindent\textbf{Training and evaluation.} All recipes start from the released weights and share data (amounting to $1.7\%$ of Uni-NaVid's original training set, from the EVT-Bench training split) and hyperparameters. Training contains no environmental disturbance, and its random rotations are drawn independently of the target, unlike the benchmark's target-loss. We evaluate on 200 validation episodes per dataset, whose scenes and targets never occur in training, paired with the frozen model; same-target-loss runs replay the frozen model's actions up to the loss, so all models decide from identical states (\cref{app:interventions_training,app:interventions_eval}).

\subsection{Results}
\label{sec:interventions_results}

\begin{figure}[htbp]
    \centering
    \includegraphics[width=\columnwidth]{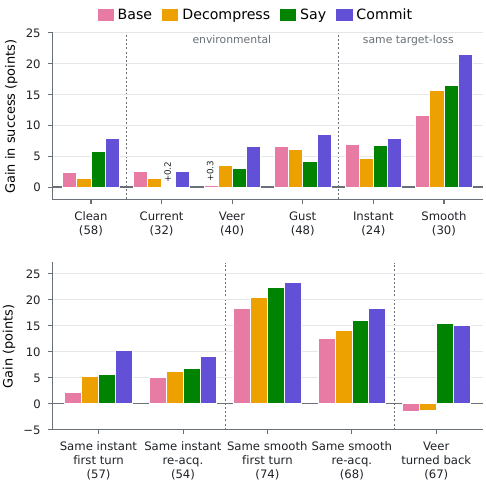}
    \caption{Gain of fine-tuned Uni-NaVid variants over the frozen model in points, on held-out MP3D, HM3D and HSSD validation episodes; in parentheses, the frozen value. Top: success. Bottom: decisions after the same target-loss and under \emph{veer}. Smooth first turn and re-acquisition are counted from the rotation's onset on every event. Full analysis: \cref{tab:supp_ft_pooled,tab:supp_ft_perdataset}.}
    \label{fig:ft_results}
\end{figure}

Every recipe recovers better after target-loss, and the gain is in the decision itself (\cref{fig:ft_results}): from the same state as the frozen model at the target-loss event, success rises by $5$--$8$ points after an instant loss and $12$--$22$ after a smooth one, and the first turn after a smooth loss is correct $18$--$23$ points more often. Part of this gain is recovery training itself: \emph{Base} already gains $+11.7$ points after a smooth loss. Paired against \emph{Base} on the same episodes, \emph{Commit} adds $+10.5$ after a smooth loss ($+5.5$ on clean), and \emph{Say} $+5.2$ and $+3.5$; from the same state after an instant loss neither adds significantly.

No recipe loses clean success. \emph{Commit} does best in this regard ($+7.8$ points), under \emph{veer} and \emph{gust} and after a smooth loss ($+21.5$); under \emph{current} no pooled gain is significant (\cref{app:interventions_results}). Without any environmental disturbance in training, \emph{gust} success rises for every recipe and \emph{veer} success for \emph{Decompress} and \emph{Commit}. \emph{Say} and \emph{Commit} also turn back more of the \emph{veer} than \emph{Base} on MP3D and HM3D.

Importantly, the statement is on the action path: at the identical state after an instant loss, exchanging left and right in a model's own statements reverses its first turn in $68$--$84\%$ of events on all three datasets (\cref{tab:supp_ft_swap}).

\begin{figure}[htbp]
    \centering
    \includegraphics[width=\columnwidth]{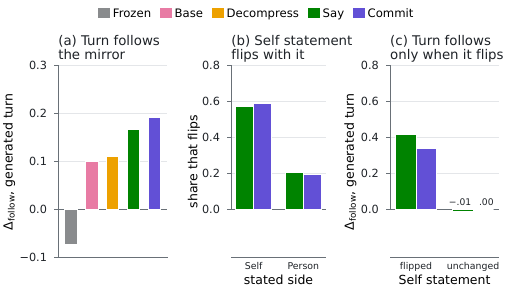}
    \caption{Mirrored rotations with the target out of view ($n=287$, three datasets; 55 on HSSD, outside the fine-tuning scenes). (a) $\Delta_{\text{follow}}$: share of turns following the mirror minus share going with the rotation. (b, c) \emph{Say} and \emph{Commit}: the mirror flips the stated Self side in over half the events, and the turn follows only when it flips. Full analysis: \cref{app:interventions_internals}.}
    \label{fig:ft_internals}
\end{figure}

Fine-tuning also changes how the turn uses knowledge of the robot's egomotion. With the target out of view, the frozen model turns with the rotation, while every recipe's turn, except in one case, turns against it (\cref{fig:ft_internals}a, \cref{tab:supp_ft_mirror}). The decoded direction still carries under $1\%$ of this effect in every recipe; the recipes route it differently. In \emph{Base} and \emph{Decompress} the whole decision state takes up the effect earlier in the network ($2.5$--$7.4\times$ the frozen model's share at $L_{16}$). In \emph{Say} and \emph{Commit} it passes through the statement: the stated Self side flips with the mirror in more than half of the events, the turn follows only when it flips (\cref{fig:ft_internals}b,\,c), and exchanging left and right in the Self line alone moves the turn. Fine-tuning did not make the decoded direction directly drive the action, but gave the motion another path to it.

\section{Impact, Limitations, and Future Work}
\label{sec:limitations}

\textbf{Impact.} Robots that follow targets, in various real-world scenarios, are regularly disturbed, and a tracker that loses its target must either account for its own motion or start a new search, which may be resource-intensive or unsuccessful. Our benchmark and protocol advance the ability to measure such failure before deployment, locate where inside a policy it arises, and the adaptation shows that it can be reduced without a new architecture or large-scale data collection. More robust target following may also make such systems more capable in uses that track people without their consent; deployment should follow the consent and privacy rules of its setting. Conversely, failure modes are exposed more precisely and are easy to trigger on purpose: a bump that turns the target out of view removes $81$--$95\%$ of each model's success, enough to make a following robot lose it.

\textbf{Limitations and future work.} Closed-loop results come from simulation with kinematic disturbances; testing the policies and the adaptation on a physical robot is a natural next step. LightNav-0's resilience to environmental disturbances and smooth target-loss (relative to instant) is not analyzed deeply here; which of its differences underlies it we leave open for investigation. Causal tests intervene on a linear direction at the decision state, so nonlinear or distributed use of the rotation is not excluded. The intervention covers one policy with one seed per recipe and should be extended to waypoint policies. No recipe improves pooled success under \emph{current}, calling for training with environmental disturbances. Finally, the fine-tuned models state an unseen target's side unreliably; inferring where a target went from the robot's motion may be possible using other methods.

\section{Conclusions}
\label{sec:conclusion}

This work investigates effects of uncommanded egomotion on embodied tracking VLAs, demonstrating a mitigation strategy that can be applied to existing models. Our closed-loop benchmark exposes the robot itself to external disturbances rather than manipulating recorded pixels, showing that every policy degrades with severity, differing in behavior under different disturbances; however, all degrade most under instant target-loss, indicating their reliance on the target staying in view. We take a deep dive into model internals, showing that inside these VLAs, uncommanded motion is present in visual features, and stays decodable in deep internal layers, yet remains underused in the models' decision-making patterns. We show that it is possible to fine-tune a model with minimal data, improving recovery after target-loss without losing clean success, making the final decision rely more on the robot's decodable egomotion. This paper establishes robustness to imperfect execution as a located, testable property rather than purely a success score, offering a mitigation strategy that is feasible using no new sensors, architecture or large-scale data, only the policy's own camera and targeted fine-tuning: a practical path toward tracking VLAs that stay on course while their platform is exposed to external disturbances.

\section*{Acknowledgments}
This work uses the Matterport3D~\cite{chang2017matterport3d} and Habitat-Matterport 3D (HM3D)~\cite{ramakrishnan2021hm3d} datasets, provided by Matterport for non-commercial academic use, in the benchmark, the model-internal analysis and the fine-tuning. The data and the information derived from them are subject to the Matterport End User License Agreement for Academic Use of Model Data (\href{http://kaldir.vc.in.tum.de/matterport/MP_TOS.pdf}{Matterport3D agreement}; \href{https://matterport.com/legal/matterport-end-user-license-agreement-academic-use-model-data}{HM3D agreement}).

\section*{LLM Usage Statement}
LLM-based coding agents were used throughout this work, under the authors' direction: Anthropic's Claude, through Claude Code, and OpenAI's Codex.
\begin{itemize}
    \item \emph{Code and experiments.} Coding agents were used in the process of code writing: the benchmark harness and disturbance implementation, the fine-tuning data generation and training scripts, the model-internal recording, probing and intervention analyses, and the evaluation and statistics. They also submitted and monitored high-performance cluster jobs, which were reviewed and approved by the authors.
    \item \emph{Figures and tables.} The agents wrote scripts for generating the figures and tables from produced results, which the authors designed, reviewed and verified for correctness and relevance.
    \item \emph{Text and related work.} The authors used the agents to assist with writing the paper and supplement, mainly for refinement, phrasing, and organization. The agents also reviewed and summarized related literature, which the authors reviewed and verified for correctness and relevance.
    \item \emph{Ideation.} The agents were used sparingly for analyses, controls and wording as options; the research questions, benchmark and disturbance design, experimental choices and claims were decided by the authors.
\end{itemize}
The authors reviewed, edited and take full responsibility for all content.

{
    \small
    \bibliographystyle{ieeenat_fullname}
    \bibliography{main}
}

\clearpage
\setcounter{page}{1}
\renewcommand{\thepage}{S\arabic{page}}
\maketitlesupplementary

\noindent The supplement follows the paper section by section (\cref{tab:supp_map}). \Cref{app:related} extends the related work, \cref{sec:supp_design} defines the benchmark in full, \cref{sec:supp_results} reports every benchmark result behind \cref{fig:results_combined}, \cref{app:internals} gives the complete model-internal analysis, one part per subsection of \cref{sec:results_internals}, each followed by its controls and ablations (\cref{sec:supp_controls,app:internals_controls}), and \cref{app:interventions} details the fine-tuning of \cref{sec:interventions}.

\begin{table}[h]
\centering
\footnotesize
\setlength{\tabcolsep}{3pt}
\caption{Where each part of the paper is detailed.}
\label{tab:supp_map}
\begin{tabular}{@{}llp{0.56\columnwidth}@{}}
\toprule
Paper & Supplement & What the supplement adds \\
\midrule
\S\ref{sec:related_work} & \ref{app:related} & each work's findings and relation to our setting \\
\S\ref{sec:benchmark_models} & \ref{sec:supp_models}, \ref{sec:supp_datasets} & model details, splits, evaluator protocol \\
\S\ref{sec:benchmark_disturbances} & \ref{sec:supp_disturbances} & every family, severity and target-loss rule \\
\S\ref{sec:benchmark_metrics} & \ref{sec:supp_metrics} & metric definitions, exclusion, statistics \\
\S\ref{sec:results_environmental} & \ref{sec:supp_bench}, \ref{sec:supp_env} & all SR, TR and CR values; family ordering \\
\S\ref{sec:results_target_loss} & \ref{sec:supp_bench}, \ref{sec:supp_target} & diagnostics, exposure \\
\S\ref{sec:results_benchmark} & \ref{sec:supp_controls} & controls: reference checks, scene-level resampling, published numbers \\
\S\ref{sec:internals_design} & \ref{sec:supp_internals_measure} & recordings, collections, statistics \\
\S\ref{sec:results_motion_evidence} & \ref{app:internals_readers} & probes, encoders, datasets \\
\S\ref{sec:results_history_compression} & \ref{app:internals_grp_history} & full ladders, real video \\
\S\ref{sec:results_rotation_backbone} & \ref{app:internals_grp_backbone} & full maps, all three datasets \\
\S\ref{sec:results_recovery_feasibility} & \ref{app:internals_functional} & arms and controllers, matching, pooled values, first turn \\
\S\ref{sec:results_rotation_use} & \ref{app:internals_grp_use} & steering, mirror, interchange, components \\
\S\ref{sec:results_internals} & \ref{app:internals_controls} & controls and ablations: frame shuffle, finer grids, projector, label permutation, leak control, own trajectories, HM3D/HSSD patch search \\
\S\ref{sec:interventions} & \ref{app:interventions} & recipes, data, per-dataset results, statements \\
\bottomrule
\end{tabular}
\end{table}

\appendix
\crefalias{section}{appendix}
\crefalias{subsection}{appendix}
\crefalias{subsubsection}{appendix}
\section{Related Work}
\label{app:related}

This section expands \cref{sec:related_work}: what each work shows and how it relates to our setting.

\subsection{Robustness of Embodied Policies and VLAs}
\label{app:related_robustness}

RobustNav \cite{chattopadhyay2021robustnav} is a close precedent for disturbing an embodied
agent's body rather than only its pixels. It corrupts the RGB observations and the transition dynamics (motion bias, drift
and motor failure) of PointNav and ObjectNav agents, and finds that for visual corruptions, augmentation and
self-supervised adaptation recover only part of the loss. Its agents are not VLAs and its goals are static, so no
displacement can remove a moving target from view.

LIBERO-Plus \cite{fei2025liberoplus} varies layout, viewpoint, illumination, background,
sensor noise, language and initialization, and RobustVLA \cite{guo2025vlarobustness} applies 17 perturbations to
OpenVLA, $\pi_0$-FAST and $\pi_0$, among them action noise and an external force applied for a few steps at random intervals; it
finds action execution the most fragile modality. Both establish that VLAs are fragile to physical and observational
perturbations, which we do not claim as new. They differ from our setting in three respects: the disturbance
acts on a manipulator rather than on a mobile robot that carries its camera, RobustVLA's force is transient rather than persistent,
and neither study looks inside the policy.

CameraBench \cite{lin2025camerabench} and EgoDyn-Bench \cite{schaefer2026egodyn}
evaluate whether vision-language models (EgoDyn-Bench also driving VLAs) describe camera motion and egomotion from video, offline and in question
answering. \Cref{tab:related_benchmarks} places them with the closed-loop benchmarks: they test the same perceptual
ability our policies need, but no action depends on the answer.

EVT-Bench \cite{wang2025trackvla} provides the closed-loop task, a robot following a
language-specified moving person, and the episodes we build on; the robot moves as commanded, and no disturbance acts on it.

\subsection{Representation of Motion}
\label{app:related_motion}

Frozen visual foundation models encode patch correspondence across small viewpoint changes
\cite{elbanani2024probe3d}; our comparison probes use this cue, cosine similarities between nearby patches of
consecutive frames, as a probing tool rather than as a new representation. VLMs answer poorly about camera and egomotion
\cite{lin2025camerabench,schaefer2026egodyn}, and EgoDyn-Bench infers that the deficit lies downstream of the
visual encoder.
Interpretability studies find motion that the output misses inside the model: ego speed is decodable inside
VLMs and a driving VLA although their verbal outputs lose it \cite{winter2026speedblindspot}, and video-LLMs answer the
direction of object motion near chance although it is linearly decodable in their encoder, projector and language
model, a gap a frame-difference objective at the projector narrows \cite{lee2026modirect}. We
follow interchange interventions \cite{geiger2021causal} and the distinction between decodable and used information
\cite{elazar2021amnesic}.

A navigation VLA encodes imminent collisions and whether an
obstacle is a person, yet avoids according to neither \cite{salsa2026}, and a few attention heads of a frozen navigation
VLA signal its deviation from the instructed path \cite{navheads2026}. Both probe or monitor; neither intervenes on the signal to test whether it drives
the action.

NaVid gives the current frame 64 visual tokens and each historical frame 4, a budget chosen in
an ablation on success, where 4 tokens per historical frame came close to 16 at a lower inference cost
\cite{zhang2024navid}. Uni-NaVid keeps this split and merges older frames further by similarity
\cite{zhang2025uninavid}, and TrackVLA keeps 64 and 4 tokens \cite{wang2025trackvla}. Manipulation VLAs approach history differently: TemporalFlow-VLA finds a raw
multi-frame baseline largely insensitive to frame order and supervises history queries with robot-surface flow computed from robot state \cite{yang2026temporalflowvla};
JEPA-VLA adds a video-predictive encoder over the two most recent frames \cite{miao2026jepavla}; and EvoMem-VLA learns one
delta token per pair of frames, outperforming naive feature subtraction \cite{na2026evomemvla}. These designs serve
manipulation, where the camera's own uncommanded motion is not at issue.

\subsection{Mitigating Physical Disturbances}
\label{app:related_mitigation}

RobustVLA fine-tunes against worst-case action deviations, with consistency across
perturbed observations \cite{guo2025vlarobustness}. Self-Adaptive VLA trains a context encoder over the policy's own
earlier rollouts, its video, proprioception and actions, under deliberately induced shifts, to compensate persistent
hardware errors; it reports that denser patch tokens bring no significant gain over a compact token per frame
\cite{zhang2026selfadaptivevla}.

Pace-and-Path Correction re-times and offsets action chunks from an
external tracker's estimate of object motion, without training \cite{zhang2026ppc}: there the world moves, not the robot.
VLA-Corrector compares the predicted and observed change of visual features and truncates stale chunks when they
diverge, then steers the replan back toward the intended trajectory \cite{pan2026vlacorrector}, which is closest to an egomotion comparator; its deviations arise in contact-rich manipulation, not from a displaced camera.

Visual odometry replaces GPS and compass for PointGoal agents under actuation noise
\cite{zhao2021visualodometry}, estimating the agent's motion explicitly toward a static goal. TIC-VLA accumulates measured
egomotion while an asynchronous VLM deliberates and re-anchors its decisions \cite{huang2026ticvla}; the motion is the
robot's own motion during latency, measured by odometry, not uncommanded motion inferred from vision. Recovery comes from vision alone; inertial and
odometric sensing are outside our scope.

\FloatBarrier
\section{Benchmark Design}
\label{sec:supp_design}

This section gives the complete definitions for the main paper's benchmark (\cref{sec:benchmark_models,sec:benchmark_disturbances,sec:benchmark_metrics}); the model-internal measurements are defined in \cref{app:internals_data}. Every quantity is defined so that it can be recomputed from the per-episode and per-step records. We will publicly release the HSSD episode set, the disturbance and target-loss code, the model-internals analysis code and the fine-tuning code.

\subsection{Models}
\label{sec:supp_models}

We evaluate three publicly released embodied-tracking VLAs with their released weights, without any fine-tuning, each through the EVT-Bench evaluation loop and greedy decoding.

\paragraph{Uni-NaVid~\cite{zhang2025uninavid}.} A 7B video-language navigation model of the NaVid lineage. An EVA-CLIP~\cite{sun2023evaclip} vision encoder embeds each frame; the current frame is kept as an $8\times8$ grid of 64 visual tokens, while each of the 64 most recent earlier frames is pooled to a $2\times2$ grid of 4 memory tokens and older frames to one token each, merged by similarity, before entering the language model. The model emits navigation words (forward, left, right, backward, stop); the EVT-Bench agent predicts four words per decision and executes the first. Measured on the simulator, an executed forward word moves the robot by $0.187$\,m and a turn word rotates it by $9^\circ$ (medians of the realized steps); these two numbers define the benchmark's reference action (\cref{sec:supp_disturbances}).

\paragraph{LightNav-0~\cite{lightnav0}.} A 4B generalist navigation model built on a Qwen3-VL~\cite{bai2025qwen3vl} visual stack. In the released code, its vision tower has a temporal patch of two frames, so consecutive frames are encoded jointly as one tubelet: the two most recent frames $(t{-}1,t)$ form an $8\times14$ token grid, and the previous pair $(t{-}3,t{-}2)$ is kept at a reduced $4\times7$ grid. The model generates target-position tokens followed by discrete action tokens that decode to a short trajectory of waypoints. We use the authors' own EVT-Bench client unchanged: it converts the first predicted waypoint (forward, lateral, yaw) into the platform's velocity command with the authors' scaling ($0.375$\,m, $0.25$\,m, $\pi/20$\,rad). Unlike Uni-NaVid, LightNav-0 can move laterally.

\paragraph{OmTrackVLA~\cite{omtrackvla2025}.} A 0.6B tracking model of the TrackVLA design: DINOv3~\cite{simeoni2025dinov3} ViT-S/16 and SigLIP~\cite{zhai2023siglip} so400m features feed a Qwen3-0.6B~\cite{yang2025qwen3} language model, which keeps the current frame at $8\times8$ and $2\times2$ and each of up to 31 previous frames at $2\times2$ tokens; a waypoint head reads the final action token. As in the authors' agent, the executed command is the second predicted waypoint divided by the $0.1$\,s control period, each component clipped to $[-1,1]$ by the simulator.

\paragraph{Model choice.} The three models are publicly released checkpoints that run on the same embodied-tracking task (EVT-Bench) in the same simulator through their authors' interfaces, so they face identical episodes and disturbances without retraining, and their open weights let us record and intervene on every internal state. They span history design (a $2\times2$ grid per recent historical frame in Uni-NaVid and OmTrackVLA, from different lineages and visual encoders; joint encoding of the two latest frames in LightNav-0), action interface (navigation words; action tokens decoded to waypoints; continuous waypoints from one action token) and scale (7B, 4B and 0.6B).

\paragraph{Realized steps.} Measured on the reseeded clean runs (\cref{sec:supp_disturbances}), Uni-NaVid moves $0.187$\,m or turns $9.00^\circ$ on every step; LightNav-0 moves a median $0.146$--$0.157$\,m per driving step and turns a median $1.9$--$2.9^\circ$; OmTrackVLA $0.134$--$0.156$\,m and $2.2$--$2.3^\circ$. All three turn at most $9^\circ$ in one step, so veer at maximal severity ($4.3^\circ$) is about half of every model's largest turn, and the translation ladder ($0.03$--$0.09$\,m) is $19$--$67\%$ of the waypoint models' median driving step.

\subsection{Datasets}
\label{sec:supp_datasets}

\paragraph{Task.} All runs use EVT-Bench's single-target tracking task~\cite{wang2025trackvla} in Habitat. A Spot robot with one RGB camera ($90^\circ$ horizontal field of view) must follow one person, specified by a natural-language description. The person walks a scripted path; the robot's actions determine its own future views, so a lost target cannot be recovered without acting.

\paragraph{Episode sets.} We use the complete EVT-Bench validation splits of MP3D (721 episodes in 14 scenes) and HM3D (684 episodes in 87 scenes), and 697 HSSD episodes in 164 scenes; for brevity, all three are called validation splits below. As EVT-Bench releases episodes for MP3D and HM3D only, we generated HSSD episodes by transplanting EVT-Bench single-target validation episodes onto HSSD-hab scenes. Each episode copies a reference episode's layout (robot-to-person offset and facing, the target's walking legs, avatars and instruction; 359 references from MP3D and 338 from HM3D). The layout is rotated by a random angle and anchored at a random point of the scene's largest navmesh island, computed as in the evaluator, which then loads the same navmesh; every point is snapped to that navmesh. A placement is kept only if all agents share one floor, the robot has a line-of-sight proxy to the target (geodesic-to-straight-line distance ratio at most $1.35$), and every walking leg stays reachable and keeps at least $75\%$ of its straight-line length. The instructions describe only the target person, never the scene, so they remain valid after the transplant. To keep the reference's difficulty mix, a random sample of 700 reference episodes is drawn (seed 0) and each is retried across scenes until it fits; the 11 that fit in no scene were replaced by other unused reference episodes, and 3 were dropped because their difficulty stratum had no unused reference left. This yields 697 episodes over 164 of the 167 HSSD scenes that render in habitat-sim (at most 13 per scene, from both of HSSD-hab's scene splits). Their statistics match the reference: mean robot-to-person distance $2.21$ versus $2.23$\,m and first walking leg $7.72$ versus $7.77$\,m. HSSD therefore replicates the benchmark across scene geometry and appearance; the persons, instructions and walking structure follow EVT-Bench. \Cref{tab:supp_episode_stats} gives the episodes' target paths and the disturbance each dataset received. 

\paragraph{Person and instruction per evaluation process.} The EVT-Bench evaluation drivers load the human avatars once per process, from the first episode it runs, and pass that episode's instruction for every later episode, so that the instruction always describes the person actually shown. All our benchmark evaluators reproduce this behavior. Every environmental condition of a model--dataset cell uses the same shards, each starting from the same episode, as that cell's clean run, so the person and the instruction are identical between paired environmental runs.

\begin{table}[htbp]
\centering
\footnotesize
\setlength{\tabcolsep}{3pt}
\begin{tabular}{@{}lccc@{}}
\toprule
 & MP3D & HM3D & HSSD \\
\midrule
Scenes / episodes / IDs & 14 / 721 / 313 & 87 / 684 / 48 & 164 / 697 / 697 \\
Scene type & \shortstack{reconstructed\\real buildings} & \shortstack{scanned\\real interiors} & \shortstack{human-authored\\synthetic interiors} \\
Target path (m) & 15.3 {\tiny [8.3, 23.1]} & 7.6 {\tiny [5.7, 11.1]} & 9.8 {\tiny [6.5, 18.2]} \\
Paths that turn (\%) & 61.3 & 22.7 & 42.5 \\
Their curvature ($^\circ$/m) & 3.7 {\tiny [2.1, 5.5]} & 5.1 {\tiny [3.3, 7.3]} & 4.1 {\tiny [2.2, 6.1]} \\
\midrule
Current, net (steps) & 35.7 {\tiny [25.7, 53.8]} & 27.9 {\tiny [21.8, 35.1]} & 29.9 {\tiny [23.1, 43.4]} \\
Gust, net (steps) & 10.3 {\tiny [6.1, 15.1]} & 8.6 {\tiny [5.6, 13.1]} & 9.0 {\tiny [5.8, 13.8]} \\
Veer, yaw (turn steps) & 29.4 {\tiny [20.2, 47.2]} & 24.1 {\tiny [16.8, 31.8]} & 24.1 {\tiny [17.3, 35.6]} \\
\bottomrule
\end{tabular}
\caption{Episode statistics per dataset, median [interquartile range]. \emph{Target path}: the straight line through the
target's defined waypoints. \emph{Curvature}: total heading change at interior waypoints per meter of path, over the
paths with at least one interior waypoint. Disturbance rows: per episode at maximal severity, over the kept episodes of
the three models, in robot steps ($0.187$\,m) and turn steps ($9^\circ$); \emph{net}: length of the summed realized
displacement; \emph{yaw}: total injected rotation. \emph{IDs}: distinct episode identifiers.}
\label{tab:supp_episode_stats}
\end{table}

\paragraph{Execution per cell.} \Cref{tab:supp_exclusion} lists how each model--dataset cell was run. Within a cell, every environmental run uses one camera configuration and the same processes and episodes. Uni-NaVid and LightNav-0 receive the benchmark's $640\times480$ render. OmTrackVLA receives its native $384\times384$ render at the same $90^\circ$ field of view, as in its original authors' evaluation; the benchmark's target-visibility test is computed on a panoptic render of the same size (\cref{sec:supp_metrics}).

\subsection{Disturbance Definition}
\label{sec:supp_disturbances}

\begin{figure}[htbp]
    \centering
    \includegraphics[width=\columnwidth, trim=0 3.8 0 4.3, clip]{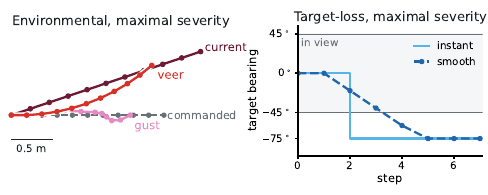}
    \caption{Environmental and target-loss disturbances, illustrated over ten forward steps.}
    \label{fig:disturbances}
\end{figure}

\paragraph{Common mechanism.} Every disturbance is a displacement of the robot base applied after the commanded simulator step and never undone: the robot really ends up where the disturbance takes it, and the camera, mounted on the base, sees the result. Pixels are never modified, and no camera-only perturbation is used, so the image records the realized motion. Each step logs the commanded displacement and the realized one. In the simulator a displacement applied after step $k$ appears in the frame rendered for step $k{+}2$; this one-step render lag is a property of the simulator, identical for every model, family and condition. Disturbances are kinematic: each sets the base pose after the commanded step. \Cref{fig:disturbances} illustrates each family.

\paragraph{Reference action and severity scale.} Severities S1--S4 ($16$, $24$, $32$ and $48\%$) are expressed as a share of a fixed reference action: a $0.187$\,m forward translation and a $9^\circ$ rotation, the medians of Uni-NaVid's realized steps. Each severity is the same displacement in meters or degrees for all three models. For scale, the ladder is comparable to the wheel slip reported on loose or inclined ground: about $20\%$ of each $0.22$\,m step for a rover on sandy slopes of up to $10^\circ$~\cite{angelova2006slip}, a position error of nearly $5$\,m over a $19$\,m cross-slope drive of the Opportunity rover ($26\%$ of the distance;~\cite{maimone2007mer}), and about $80\%$ on flat loose sand~\cite{angelova2006slip} and $125\%$ on a slope steeper than $25^\circ$~\cite{maimone2007mer}.

\paragraph{Seeding.} All random quantities of the environmental disturbances are drawn from a seed that combines a hash of the episode identifier with a fixed disturbance seed (seed 0 for every benchmark run). A given episode therefore receives the same disturbance in every run of every model. EVT-Bench reuses identifiers across scenes, so episodes that share an identifier also share their disturbance draws (MP3D: 721 episodes, 313 identifiers; HM3D: 684 and 48; HSSD's are unique). Resampling identifiers instead of episodes moves the pooled relative-loss bounds by at most $0.5$ points (\cref{tab:supp_scene_boot}).

\paragraph{Environmental families.} Let $s$ be the scale factor of a run (\cref{tab:supp_push_params}).
\begin{itemize}
\item \emph{Current}: a constant world-frame velocity,
\begin{equation}
 \Delta\mathbf p_k = s\,v_0\,(\cos\theta,\sin\theta),
\end{equation}
on every step, including steps on which the robot stops, with $v_0=0.06$\,m and a direction $\theta\sim\mathcal U[0,2\pi)$ drawn once per episode.
\item \emph{Veer}: a body-frame yaw
\begin{equation}
 \Delta\psi_k = \pm s\,\psi_0,
\end{equation}
with $\psi_0=2.5^\circ$, applied only on steps whose command translates the base, with a sign drawn once per episode (equal probabilities).
\item \emph{Gust}: a zero-mean, temporally correlated world-frame velocity. Each horizontal axis follows an Ornstein--Uhlenbeck process~\cite{uhlenbeck1930theory} with correlation time $\tau=4$ steps, discretized exactly and started at its stationary distribution:
\begin{equation}
 \begin{gathered}
 x_k = a\,x_{k-1} + \sigma\sqrt{1-a^2}\,\xi_k,\\
 a=e^{-1/\tau},\qquad \sigma = \frac{s\,r_0}{\sqrt2},
 \end{gathered}
\end{equation}
with $\xi_k\sim\mathcal N(0,1)$, so that the planar RMS is $s\,r_0$, $r_0=0.06$\,m.
\end{itemize}
Current and gust have the same per-step magnitude at the same severity, so their comparison isolates the effect of temporal correlation.

\begin{table}[htbp]
\centering
\footnotesize
\setlength{\tabcolsep}{4pt}
\begin{tabular}{lll}
\toprule
Family & Per step at $s=1$ & Scales $s$ for S1--S4 \\
\midrule
Current & $0.06$\,m, fixed direction & $0.5,\ 0.75,\ 1.0,\ 1.5$ \\
Gust & RMS $0.06$\,m, OU $\tau=4$ & $0.5,\ 0.75,\ 1.0,\ 1.5$ \\
Veer & $\pm2.5^\circ$ while translating & $0.578,\ 0.866,\ 1.155,\ 1.733$ \\
\bottomrule
\end{tabular}
\caption{Environmental disturbance parameters. Severity is the per-step magnitude as a share of the reference action ($0.187$\,m, $9^\circ$): S1--S4 are $16$, $24$, $32$ and $48\%$; for example, S3 ($32\%$) current or gust is $0.06$\,m per step and S3 veer is $2.88^\circ$ per translating step. A step translates when its forward command has a magnitude of at least $0.019$\,m ($10\%$ of the reference step).}
\label{tab:supp_push_params}
\end{table}

\paragraph{Target-loss disturbances.} A target-loss event is an in-place rotation of the base that places the target outside the field of view by a fixed angular margin.
\begin{itemize}
\item \emph{Timing.} A target-loss event becomes due at the 20th action and again 40 actions after the previous event's onset; once due, it fires at the first step after which the target is followed (\cref{sec:supp_metrics}) and lies within $10^\circ$ of the image center. Target-loss events therefore always remove a target that the model was tracking, and the number of events per episode depends on how long the model tracks (\cref{tab:supp_kick_exposure}).
\item \emph{Magnitude.} With target bearing $\beta$ (from the simulator poses; positive to the left), side $\epsilon\in\{-1,+1\}$ and margin $m$, the total rotation is $\beta+\epsilon\,(\mathrm{hfov}/2+m)$, which places the person exactly $m$ beyond the image boundary. The side is drawn from a hash of scene, episode and event index with equal probabilities. The margins $m=10^\circ,20^\circ,30^\circ$ (S1--S3) give total rotations of $45$--$65^\circ$, $55$--$75^\circ$ and $65$--$85^\circ$; all stay below $90^\circ$, so the frames before and after an instant target-loss still overlap. Over all episodes of each split, the mean total rotation of the first event is $54.1$--$55.3^\circ$, $63.8$--$65.3^\circ$ and $73.6$--$75.3^\circ$ at S1--S3.
\item \emph{Instant} target-loss events apply the whole rotation in one step: the target vanishes between two frames, and the only cue is the robot's own image motion.
\item \emph{Smooth} target-loss events apply the same total rotation at $20^\circ$ per step (the last step takes the remainder), in 3--5 steps ($3.1$--$4.4$ on average), so the target slides toward the image edge over several frames before leaving.
\item Smooth target-loss events are \emph{open loop}: their increments are fixed at onset and ignore what the model does. A model that turns against a smooth target-loss can keep the target in view; this outcome is reported (\cref{sec:supp_target}).
\end{itemize}
The definition uses only the camera's field of view and the target's bearing, with no dataset-, scene- or model-specific constants.

\paragraph{Reproducibility of the target-loss runs.} To make a target-loss run identical to its undisturbed counterpart up to the first event, the target-loss protocol reseeds the simulator's random generators and each person's walk controller from (scene, episode) at the start of every episode, and pins the target person and instruction to the dataset's first episode. Without these measures, people walk differently depending on which episodes a process ran before, and target-loss and undisturbed runs diverge before any target-loss event. We call the undisturbed run made under these measures the \emph{reseeded clean} run. It differs from the environmental clean run by construction; their agreement is reported in \cref{tab:supp_kick_reference}.

\paragraph{One reference per cell.} Every disturbed run, environmental or target-loss, is compared with the same clean run of its model--dataset cell, over the same kept episodes. Every curve in a panel of \cref{fig:results_combined} therefore starts at the same point. \Cref{tab:supp_kick_reference} compares the environmental and reseeded clean runs: their success rates agree within $1.4$ points in every cell.

\subsection{Metrics}
\label{sec:supp_metrics}

\paragraph{Following.} At each step the benchmark marks the target as \emph{followed} when the robot is within $3$\,m of the target person and the person is visible, i.e., covers at least 3{,}000 pixels of the panoptic render of the robot camera. The pixel threshold is absolute, so at OmTrackVLA's $384\times384$ render it corresponds to a larger share of the image (2.0\%) than at $640\times480$ (1.0\%).

\paragraph{Departures from native evaluation.} The target-loss runs, and all runs of Uni-NaVid on MP3D, pin the person and instruction to the dataset's first episode (\cref{sec:supp_disturbances}), and Uni-NaVid's MP3D cell takes the reseeded clean run as its clean reference; the fine-tuning evaluation uses each episode's own (\cref{app:interventions_eval}). LightNav-0's repository evaluates with a $120^\circ$ field of view; every model here uses EVT-Bench's $90^\circ$. Within-model paired comparisons are unaffected; comparisons across models mix these differences with architecture.

\paragraph{Episode termination.} An episode ends after 300 steps; when the robot comes within $0.5$\,m of the target person (a collision); after more than 20 consecutive steps at more than $4$\,m from the person, the counter being reset whenever the target is followed (status \emph{lost}); or when the simulator ends the episode.

\paragraph{Success rate (SR).} The benchmark's per-episode success flag. An episode that reaches 300 steps succeeds if the target is followed at its last step. An episode that ends earlier succeeds only if, at its end, the target is followed and the robot is between $1$ and $3$\,m from it. Collisions and lost episodes therefore fail.

\paragraph{Tracking rate (TR).} The per-episode share of executed steps on which the target is followed, averaged over episodes. SR, TR and CR are tracking metrics; they do not measure safety near obstacles or bystanders.

\paragraph{Collision rate (CR).} The share of episodes ended by a collision with the target person.

\paragraph{Target-loss diagnostics.} A target-loss event is the run of consecutive steps with a nonzero target-loss rotation, from onset $k$ to end $e$ ($e=k$ for instant target-loss). The diagnostics use the first event of every episode with a target-loss, i.e., with at least one event.
\begin{itemize}
\item \emph{Not followed after the event}: the target is not followed at step $e{+}1$. For instant target-loss this checks that the rotation removed the target; for smooth target-loss it is also an outcome, because a model may keep the target in view by turning against the rotation.
\item \emph{Re-acquisition}: the target is followed again at some step in $[e{+}1,e{+}20]$; a window cut short by the end of the episode counts as not re-acquired unless the target is followed before the end. The window starts at the step of the not-followed check, so it also counts events in which the target was never lost; after smooth target-loss, re-acquisition can therefore exceed the share not followed.
\end{itemize}
These diagnostics come from the target-loss protocol's own scoring, over all episodes of each split.

\paragraph{Relative loss and exposure.} The \emph{share of clean success lost} by a condition is $100\,(1-\mathrm{SR}/\mathrm{SR}_\mathrm{clean})$ over the cell's kept episodes, pooled over datasets as a ratio of summed successes, with a paired episode bootstrap stratified by dataset. The share of episodes \emph{with a target-loss} (at least one event) and the loss on those episodes alone, against the same clean run, are also reported, as is the share of \emph{veered steps} with the veer losses. Re-acquisition in the exposure table counts the first event of each episode with a target-loss, as in the diagnostics above.

\paragraph{Paired statistics.} All changes are reported in percentage points against the cell's clean run over the same episodes. Confidence intervals are 95\% percentile intervals from a paired episode bootstrap with 5{,}000 resamples (seed 0). Scene- and identifier-level resampling: \cref{sec:supp_ctrl_scene}. Intervals across layers and grids are pointwise.

\paragraph{Episode exclusion.} In some episodes the simulator relocated the robot by several meters after a displacement, a navmesh-snapping bug that places it on the scene's largest navigable island rather than the one it stands on. Episodes with a single-step realized displacement above $0.3$\,m in any environmental run are removed from every run of their cell, clean and target-loss runs included, except a single maximal-severity gust step of $0.332$\,m (\cref{tab:supp_exclusion}). An episode is removed or kept as a whole across conditions, so paired comparisons stay paired.

\FloatBarrier
\section{Benchmark Results}
\label{sec:supp_results}

\subsection{Disturbance Benchmark}
\label{sec:supp_bench}

This section reports every number behind \cref{fig:results_combined}. Each cell has one clean run and 18 disturbed runs: current, veer and gust at severities S1--S4, and instant and smooth target-loss at S1--S3 (\cref{sec:supp_disturbances}). All runs use the complete validation splits, greedy decoding and the kept episodes of their cell. \Cref{tab:supp_env_sr,tab:supp_env_delta,tab:supp_env_tr,tab:supp_env_cr,tab:supp_exclusion} give the environmental results and \cref{tab:supp_target_loss,tab:supp_target_loss_delta,tab:supp_target_loss_diag,tab:supp_kick_exposure} the target-loss results. \Cref{tab:supp_relative_loss} gives the pooled shares of clean success lost behind \cref{fig:results_relative_loss}. The checks of the target-loss protocol, of the intervals and of the clean runs are in \cref{sec:supp_controls}.

\begin{table}[htbp]
\centering
\scriptsize
\setlength{\tabcolsep}{2pt}
\begin{tabular}{@{}lccc@{}}
\toprule
Condition & Uni-NaVid & LightNav-0 & OmTrackVLA \\
Clean SR & 61.6 & 91.5 & 78.0 \\
\midrule
Current & 66.1 {\tiny [63.3, 68.8]} & 47.5 {\tiny [45.0, 49.9]} & 58.7 {\tiny [56.2, 61.2]} \\
Veer & 53.7 {\tiny [50.7, 56.8]} & 6.6 {\tiny [5.2, 8.0]} & 79.7 {\tiny [77.7, 81.6]} \\
Gust & 35.9 {\tiny [32.6, 39.1]} & 18.6 {\tiny [16.5, 20.6]} & 33.2 {\tiny [30.5, 35.9]} \\
Instant target-loss & 71.6 {\tiny [69.0, 74.1]} & 90.6 {\tiny [89.3, 92.0]} & 95.4 {\tiny [94.3, 96.4]} \\
Smooth target-loss & 61.4 {\tiny [58.5, 64.1]} & 44.1 {\tiny [41.7, 46.4]} & 90.4 {\tiny [88.9, 91.9]} \\
\midrule
Instant target-loss, if fired & 80.6 {\tiny [78.0, 83.0]} & 90.6 {\tiny [89.2, 91.9]} & 95.5 {\tiny [94.4, 96.5]} \\
Smooth target-loss, if fired & 68.5 {\tiny [65.7, 71.4]} & 43.6 {\tiny [41.4, 45.9]} & 90.4 {\tiny [88.9, 91.9]} \\
Episodes with a target-loss (\%) & 79.5 / 79.5 & 97.1 / 97.2 & 94.1 / 94.2 \\
Veered steps (\%) & 57.0 & 80.3 & 95.5 \\
\bottomrule
\end{tabular}
\caption{Share of clean success lost, $100\,(1-\mathrm{SR}/\mathrm{SR}_\mathrm{clean})$ (\%) [95\% CI] at maximal severity, pooled over the three datasets as a ratio of summed successes. \emph{Episodes with a target-loss}: instant / smooth. \emph{Veered steps}: share of steps on which veer acted. Paired episode bootstrap stratified by dataset.}
\label{tab:supp_relative_loss}
\end{table}

\subsection{Environmental Disturbances}
\label{sec:supp_env}

\paragraph{Success rate.} \Cref{tab:supp_env_sr} gives SR for every model, dataset, family and severity, and \cref{tab:supp_env_delta} the paired changes from clean. 

\paragraph{Change in success by family.} At S3 and at maximal severity, gust costs less SR than current in all nine cells (\cref{tab:supp_env_delta}).

\paragraph{Tracking and collision rates.} \Cref{tab:supp_env_tr,tab:supp_env_cr} give TR and CR for the same runs. At maximal severity, TR falls less than SR in every cell and family except LightNav-0 under veer.

\begin{table*}[htbp]
\centering
\scriptsize
\setlength{\tabcolsep}{3.2pt}
\begin{tabular}{llrrrrrrrrrrrrrr}
\toprule
\multirow{2}{*}{Model} & \multirow{2}{*}{Dataset} & \multirow{2}{*}{$n$} & \multirow{2}{*}{Clean} & \multicolumn{4}{c}{Current} & \multicolumn{4}{c}{Veer} & \multicolumn{4}{c}{Gust} \\
\cmidrule(lr){5-8}\cmidrule(lr){9-12}\cmidrule(lr){13-16}
 & & & & S1 & S2 & S3 & S4 & S1 & S2 & S3 & S4 & S1 & S2 & S3 & S4 \\
\midrule
\multirow{3}{*}{Uni-NaVid} & MP3D & 721 & 52.4 & 39.5 & 30.4 & 23.0 & 14.4 & 39.9 & 32.6 & 26.4 & 17.3 & 48.7 & 43.7 & 41.9 & 31.2 \\
 & HM3D & 684 & 60.8 & 50.3 & 40.2 & 34.1 & 20.5 & 54.2 & 47.4 & 41.1 & 31.6 & 53.9 & 53.8 & 49.6 & 37.7 \\
 & HSSD & 689 & 72.0 & 55.0 & 47.9 & 38.6 & 28.0 & 59.7 & 52.5 & 44.4 & 37.2 & 67.1 & 64.3 & 58.9 & 49.9 \\
\midrule
\multirow{3}{*}{LightNav-0} & MP3D & 683 & 90.6 & 85.4 & 81.6 & 69.8 & 40.1 & 89.3 & 89.3 & 87.6 & 84.8 & 86.8 & 85.1 & 80.4 & 70.3 \\
 & HM3D & 624 & 92.5 & 86.5 & 81.1 & 70.0 & 45.7 & 92.5 & 92.0 & 90.7 & 88.8 & 88.1 & 85.7 & 82.4 & 75.3 \\
 & HSSD & 688 & 91.6 & 88.4 & 82.8 & 75.1 & 58.1 & 90.1 & 88.8 & 87.4 & 83.1 & 87.9 & 88.1 & 83.3 & 78.1 \\
\midrule
\multirow{3}{*}{OmTrackVLA} & MP3D & 678 & 78.0 & 69.8 & 56.5 & 44.7 & 26.0 & 67.7 & 57.7 & 37.9 & 10.5 & 75.1 & 71.4 & 62.8 & 48.1 \\
 & HM3D & 629 & 78.7 & 64.9 & 51.2 & 42.8 & 26.4 & 68.2 & 58.5 & 42.1 & 15.6 & 73.9 & 66.6 & 61.2 & 47.2 \\
 & HSSD & 691 & 77.4 & 76.4 & 66.0 & 60.2 & 43.7 & 72.9 & 63.5 & 51.4 & 21.4 & 78.4 & 75.3 & 69.5 & 60.6 \\
\bottomrule
\end{tabular}
\caption{Environmental disturbances: success rate (\%) on the full validation splits for every model, dataset, family and severity (S1--S4, \cref{tab:supp_push_params}). $n$: kept episodes (\cref{tab:supp_exclusion}); every run of a cell uses the same $n$ episodes.}
\label{tab:supp_env_sr}
\end{table*}

\begin{table*}[htbp]
\centering
\scriptsize
\setlength{\tabcolsep}{3.2pt}
\begin{tabular}{llcccc}
\toprule
Model & Dataset & S1 & S2 & S3 & S4 \\
\midrule
\multicolumn{6}{l}{\textit{Current}} \\
\multirow{3}{*}{Uni-NaVid} & MP3D & $-$12.9\,[$-$16.6, $-$9.2] & $-$22.1\,[$-$26.1, $-$18.2] & $-$29.4\,[$-$33.4, $-$25.4] & $-$38.0\,[$-$42.0, $-$34.1] \\
 & HM3D & $-$10.5\,[$-$14.5, $-$6.6] & $-$20.6\,[$-$24.7, $-$16.4] & $-$26.8\,[$-$31.0, $-$22.5] & $-$40.4\,[$-$44.4, $-$36.1] \\
 & HSSD & $-$17.0\,[$-$20.5, $-$13.6] & $-$24.1\,[$-$27.9, $-$20.5] & $-$33.4\,[$-$37.4, $-$29.5] & $-$44.0\,[$-$48.2, $-$39.9] \\
\midrule
\multirow{3}{*}{LightNav-0} & MP3D & $-$5.3\,[$-$7.8, $-$2.9] & $-$9.1\,[$-$11.9, $-$6.3] & $-$20.8\,[$-$24.3, $-$17.3] & $-$50.5\,[$-$54.5, $-$46.6] \\
 & HM3D & $-$5.9\,[$-$8.5, $-$3.4] & $-$11.4\,[$-$14.4, $-$8.3] & $-$22.4\,[$-$26.3, $-$18.8] & $-$46.8\,[$-$51.1, $-$42.3] \\
 & HSSD & $-$3.2\,[$-$5.7, $-$0.9] & $-$8.7\,[$-$11.8, $-$5.8] & $-$16.4\,[$-$19.8, $-$13.2] & $-$33.4\,[$-$37.4, $-$29.5] \\
\midrule
\multirow{3}{*}{OmTrackVLA} & MP3D & $-$8.3\,[$-$11.8, $-$4.9] & $-$21.5\,[$-$25.7, $-$17.4] & $-$33.3\,[$-$37.8, $-$28.9] & $-$52.1\,[$-$56.2, $-$48.1] \\
 & HM3D & $-$13.8\,[$-$17.8, $-$9.9] & $-$27.5\,[$-$31.6, $-$23.4] & $-$35.9\,[$-$40.4, $-$31.5] & $-$52.3\,[$-$56.6, $-$48.2] \\
 & HSSD & $-$1.0\,[$-$3.8, $+$1.7] & $-$11.4\,[$-$14.8, $-$8.2] & $-$17.2\,[$-$20.8, $-$13.7] & $-$33.7\,[$-$37.6, $-$30.0] \\
\midrule\midrule
\multicolumn{6}{l}{\textit{Veer}} \\
\multirow{3}{*}{Uni-NaVid} & MP3D & $-$12.5\,[$-$16.2, $-$8.6] & $-$19.8\,[$-$23.9, $-$15.8] & $-$26.1\,[$-$30.2, $-$21.9] & $-$35.1\,[$-$39.0, $-$31.2] \\
 & HM3D & $-$6.6\,[$-$10.4, $-$2.9] & $-$13.5\,[$-$17.5, $-$9.5] & $-$19.7\,[$-$23.8, $-$15.5] & $-$29.2\,[$-$33.6, $-$24.9] \\
 & HSSD & $-$12.3\,[$-$16.0, $-$8.9] & $-$19.4\,[$-$23.1, $-$15.7] & $-$27.6\,[$-$31.3, $-$23.7] & $-$34.8\,[$-$39.0, $-$30.8] \\
\midrule
\multirow{3}{*}{LightNav-0} & MP3D & $-$1.3\,[$-$2.9, $+$0.3] & $-$1.3\,[$-$2.9, $+$0.1] & $-$3.1\,[$-$5.1, $-$1.2] & $-$5.9\,[$-$8.1, $-$3.8] \\
 & HM3D & $+$0.0\,[$-$1.8, $+$1.8] & $-$0.5\,[$-$1.9, $+$1.1] & $-$1.8\,[$-$3.7, $+$0.2] & $-$3.7\,[$-$5.8, $-$1.6] \\
 & HSSD & $-$1.5\,[$-$3.3, $+$0.3] & $-$2.8\,[$-$4.9, $-$0.6] & $-$4.2\,[$-$6.5, $-$1.9] & $-$8.4\,[$-$11.0, $-$6.0] \\
\midrule
\multirow{3}{*}{OmTrackVLA} & MP3D & $-$10.3\,[$-$14.0, $-$6.6] & $-$20.4\,[$-$24.6, $-$16.2] & $-$40.1\,[$-$44.5, $-$35.7] & $-$67.6\,[$-$71.2, $-$63.9] \\
 & HM3D & $-$10.5\,[$-$14.1, $-$7.0] & $-$20.2\,[$-$24.2, $-$16.4] & $-$36.6\,[$-$40.9, $-$32.4] & $-$63.1\,[$-$66.9, $-$59.3] \\
 & HSSD & $-$4.5\,[$-$7.5, $-$1.3] & $-$13.9\,[$-$17.2, $-$10.6] & $-$26.0\,[$-$30.0, $-$22.3] & $-$56.0\,[$-$59.8, $-$52.2] \\
\midrule\midrule
\multicolumn{6}{l}{\textit{Gust}} \\
\multirow{3}{*}{Uni-NaVid} & MP3D & $-$3.7\,[$-$7.1, $-$0.1] & $-$8.7\,[$-$12.3, $-$5.1] & $-$10.5\,[$-$14.0, $-$7.1] & $-$21.2\,[$-$25.2, $-$17.2] \\
 & HM3D & $-$6.9\,[$-$10.4, $-$3.4] & $-$7.0\,[$-$11.0, $-$3.1] & $-$11.3\,[$-$15.2, $-$7.5] & $-$23.1\,[$-$27.3, $-$19.0] \\
 & HSSD & $-$4.9\,[$-$7.8, $-$2.0] & $-$7.7\,[$-$11.0, $-$4.5] & $-$13.1\,[$-$16.7, $-$9.4] & $-$22.1\,[$-$25.8, $-$18.3] \\
\midrule
\multirow{3}{*}{LightNav-0} & MP3D & $-$3.8\,[$-$5.9, $-$1.8] & $-$5.6\,[$-$8.1, $-$3.1] & $-$10.2\,[$-$13.0, $-$7.6] & $-$20.4\,[$-$23.9, $-$17.0] \\
 & HM3D & $-$4.3\,[$-$6.7, $-$1.9] & $-$6.7\,[$-$9.5, $-$4.0] & $-$10.1\,[$-$13.1, $-$7.1] & $-$17.1\,[$-$20.8, $-$13.5] \\
 & HSSD & $-$3.6\,[$-$5.8, $-$1.6] & $-$3.5\,[$-$5.8, $-$1.3] & $-$8.3\,[$-$10.9, $-$5.8] & $-$13.5\,[$-$16.7, $-$10.5] \\
\midrule
\multirow{3}{*}{OmTrackVLA} & MP3D & $-$2.9\,[$-$6.0, $+$0.1] & $-$6.6\,[$-$10.2, $-$3.1] & $-$15.2\,[$-$19.0, $-$11.4] & $-$29.9\,[$-$33.9, $-$26.1] \\
 & HM3D & $-$4.8\,[$-$8.3, $-$1.1] & $-$12.1\,[$-$16.2, $-$8.1] & $-$17.5\,[$-$21.8, $-$13.2] & $-$31.5\,[$-$35.9, $-$27.0] \\
 & HSSD & $+$1.0\,[$-$1.4, $+$3.3] & $-$2.2\,[$-$4.8, $+$0.4] & $-$8.0\,[$-$11.1, $-$4.8] & $-$16.8\,[$-$20.5, $-$13.2] \\
\bottomrule
\end{tabular}
\caption{Environmental disturbances: change in success rate from clean (percentage points), paired over the same episodes, with 95\% bootstrap confidence intervals (5{,}000 episode resamples, seed 0).}
\label{tab:supp_env_delta}
\end{table*}

\begin{table*}[htbp]
\centering
\scriptsize
\setlength{\tabcolsep}{3.2pt}
\begin{tabular}{llrrrrrrrrrrrrr}
\toprule
\multirow{2}{*}{Model} & \multirow{2}{*}{Dataset} & \multirow{2}{*}{Clean} & \multicolumn{4}{c}{Current} & \multicolumn{4}{c}{Veer} & \multicolumn{4}{c}{Gust} \\
\cmidrule(lr){4-7}\cmidrule(lr){8-11}\cmidrule(lr){12-15}
 & & & S1 & S2 & S3 & S4 & S1 & S2 & S3 & S4 & S1 & S2 & S3 & S4 \\
\midrule
\multirow{3}{*}{Uni-NaVid} & MP3D & 65.7 & 62.6 & 60.1 & 57.7 & 54.2 & 61.7 & 59.8 & 56.8 & 53.4 & 65.1 & 64.8 & 63.5 & 60.2 \\
 & HM3D & 75.3 & 71.4 & 68.6 & 64.9 & 58.0 & 71.9 & 69.5 & 67.7 & 62.1 & 73.0 & 72.6 & 70.0 & 64.8 \\
 & HSSD & 79.3 & 74.7 & 71.3 & 67.2 & 60.8 & 74.6 & 71.2 & 67.0 & 62.3 & 77.8 & 77.7 & 74.5 & 70.3 \\
\midrule
\multirow{3}{*}{LightNav-0} & MP3D & 87.6 & 83.8 & 81.5 & 75.8 & 63.5 & 87.4 & 87.2 & 85.8 & 80.9 & 85.2 & 84.8 & 82.9 & 77.8 \\
 & HM3D & 91.4 & 86.5 & 83.6 & 78.5 & 67.3 & 90.9 & 90.0 & 89.0 & 84.9 & 87.3 & 86.1 & 84.4 & 79.8 \\
 & HSSD & 91.6 & 89.2 & 84.6 & 80.5 & 69.5 & 90.7 & 90.5 & 88.9 & 81.6 & 90.7 & 90.2 & 87.9 & 83.3 \\
\midrule
\multirow{3}{*}{OmTrackVLA} & MP3D & 83.0 & 79.0 & 76.3 & 72.0 & 60.9 & 77.2 & 72.9 & 63.2 & 49.6 & 80.9 & 80.3 & 78.2 & 73.0 \\
 & HM3D & 83.5 & 76.7 & 72.5 & 68.4 & 58.5 & 76.6 & 70.6 & 61.6 & 47.1 & 78.9 & 76.8 & 75.6 & 69.3 \\
 & HSSD & 85.2 & 84.4 & 81.8 & 78.4 & 67.1 & 80.6 & 74.7 & 68.1 & 53.2 & 85.8 & 84.8 & 83.2 & 79.7 \\
\bottomrule
\end{tabular}
\caption{Environmental disturbances: tracking rate (\%), same runs and episodes as \cref{tab:supp_env_sr}. TR is the mean per-episode fraction of steps on which the target is followed.}
\label{tab:supp_env_tr}
\end{table*}

\begin{table*}[htbp]
\centering
\scriptsize
\setlength{\tabcolsep}{3.2pt}
\begin{tabular}{llrrrrrrrrrrrrr}
\toprule
\multirow{2}{*}{Model} & \multirow{2}{*}{Dataset} & \multirow{2}{*}{Clean} & \multicolumn{4}{c}{Current} & \multicolumn{4}{c}{Veer} & \multicolumn{4}{c}{Gust} \\
\cmidrule(lr){4-7}\cmidrule(lr){8-11}\cmidrule(lr){12-15}
 & & & S1 & S2 & S3 & S4 & S1 & S2 & S3 & S4 & S1 & S2 & S3 & S4 \\
\midrule
\multirow{3}{*}{Uni-NaVid} & MP3D & 3.3 & 4.2 & 4.3 & 5.3 & 7.2 & 2.9 & 4.2 & 4.0 & 5.5 & 4.9 & 5.1 & 5.7 & 6.7 \\
 & HM3D & 7.9 & 8.6 & 8.8 & 11.1 & 14.3 & 7.9 & 8.6 & 8.8 & 9.2 & 7.6 & 9.8 & 9.6 & 13.7 \\
 & HSSD & 4.6 & 4.2 & 4.6 & 6.0 & 8.3 & 4.2 & 4.6 & 4.8 & 3.6 & 4.4 & 3.5 & 4.4 & 6.1 \\
\midrule
\multirow{3}{*}{LightNav-0} & MP3D & 2.0 & 2.3 & 3.2 & 5.3 & 6.7 & 2.0 & 2.3 & 2.5 & 3.1 & 2.8 & 3.5 & 3.1 & 7.3 \\
 & HM3D & 4.3 & 5.9 & 5.6 & 6.4 & 7.5 & 4.5 & 3.8 & 4.3 & 4.2 & 5.6 & 5.8 & 6.7 & 8.3 \\
 & HSSD & 2.2 & 2.9 & 1.5 & 3.1 & 4.4 & 2.3 & 2.3 & 3.6 & 2.9 & 3.8 & 2.6 & 2.9 & 4.2 \\
\midrule
\multirow{3}{*}{OmTrackVLA} & MP3D & 3.8 & 3.8 & 5.6 & 7.2 & 9.4 & 3.2 & 3.8 & 6.2 & 9.4 & 3.7 & 4.7 & 8.1 & 13.1 \\
 & HM3D & 5.2 & 5.4 & 7.2 & 7.2 & 10.0 & 4.6 & 5.6 & 6.7 & 9.5 & 6.2 & 7.0 & 8.9 & 14.5 \\
 & HSSD & 2.6 & 2.9 & 3.3 & 3.8 & 5.9 & 3.0 & 3.6 & 3.6 & 5.9 & 3.5 & 4.1 & 4.6 & 10.4 \\
\bottomrule
\end{tabular}
\caption{Environmental disturbances: collision rate (\%), same runs and episodes as \cref{tab:supp_env_sr}. CR is the fraction of episodes terminated by a collision with the target person (closer than 0.5\,m).}
\label{tab:supp_env_cr}
\end{table*}

\begin{table*}[htbp]
\centering
\footnotesize
\setlength{\tabcolsep}{3.2pt}
\begin{tabular}{lllrrrrr}
\toprule
Model & Dataset & Processes & Episodes & Excluded & Kept $n$ & Clean SR, all & Clean SR, kept \\
\midrule
\multirow{3}{*}{Uni-NaVid} & MP3D & 1 per run & 721 & 0 & 721 & 52.4 & 52.4 \\
 & HM3D & 1 per run & 684 & 0 & 684 & 60.8 & 60.8 \\
 & HSSD & 5 shards & 697 & 8 & 689 & 71.4 & 72.0 \\
\midrule
\multirow{3}{*}{LightNav-0} & MP3D & 5 shards & 721 & 38 & 683 & 90.2 & 90.6 \\
 & HM3D & 5 shards & 684 & 60 & 624 & 90.6 & 92.5 \\
 & HSSD & 5 shards & 697 & 9 & 688 & 91.2 & 91.6 \\
\midrule
\multirow{3}{*}{OmTrackVLA} & MP3D & 5 shards & 721 & 43 & 678 & 78.5 & 78.0 \\
 & HM3D & 5 shards & 684 & 55 & 629 & 77.5 & 78.7 \\
 & HSSD & 5 shards & 697 & 6 & 691 & 77.2 & 77.4 \\
\bottomrule
\end{tabular}
\caption{Execution and excluded episodes per cell. \emph{Processes}: of the environmental runs, one process per run or five shards; all environmental runs of a cell share camera, processes and episodes. An excluded episode is removed from every run of its cell, clean and target-loss runs included, so paired comparisons stay paired.}
\label{tab:supp_exclusion}
\end{table*}

\subsection{Target-Loss Disturbances}
\label{sec:supp_target}

\paragraph{Success rate.} \Cref{tab:supp_target_loss} gives SR and \cref{tab:supp_target_loss_delta} the paired change from clean for every target-loss condition. 

\paragraph{Target-loss diagnostics.} \Cref{tab:supp_target_loss_diag} gives, for the first event of each episode with a target-loss, the share of episodes in which the target is not followed after the rotation and the share in which it is re-acquired within 20 steps (\cref{sec:supp_metrics}).

\begin{table*}[htbp]
\centering
\scriptsize
\setlength{\tabcolsep}{3.2pt}
\begin{tabular}{llrrrrrrrr}
\toprule
\multirow{2}{*}{Model} & \multirow{2}{*}{Dataset} & \multirow{2}{*}{$n$} & \multirow{2}{*}{Clean} & \multicolumn{3}{c}{Instant} & \multicolumn{3}{c}{Smooth} \\
\cmidrule(lr){5-7}\cmidrule(lr){8-10}
 & & & & S1 & S2 & S3 & S1 & S2 & S3 \\
\midrule
\multirow{3}{*}{Uni-NaVid} & MP3D & 721 & 52.4 & 16.0 & 13.6 & 10.5 & 20.2 & 17.8 & 15.1 \\
 & HM3D & 684 & 60.8 & 25.0 & 21.3 & 17.7 & 32.3 & 28.8 & 23.2 \\
 & HSSD & 689 & 72.0 & 30.0 & 28.0 & 24.7 & 41.1 & 37.4 & 33.4 \\
\midrule
\multirow{3}{*}{LightNav-0} & MP3D & 683 & 90.6 & 14.1 & 7.9 & 4.5 & 72.2 & 62.1 & 45.4 \\
 & HM3D & 624 & 92.5 & 24.4 & 15.4 & 11.1 & 82.2 & 73.7 & 59.3 \\
 & HSSD & 688 & 91.6 & 20.1 & 14.0 & 10.3 & 74.1 & 65.1 & 49.6 \\
\midrule
\multirow{3}{*}{OmTrackVLA} & MP3D & 678 & 78.0 & 5.6 & 4.0 & 2.2 & 16.1 & 8.8 & 5.0 \\
 & HM3D & 629 & 78.7 & 8.6 & 6.2 & 4.9 & 25.0 & 13.4 & 7.6 \\
 & HSSD & 691 & 77.4 & 7.8 & 5.2 & 3.8 & 25.5 & 14.8 & 9.8 \\
\bottomrule
\end{tabular}
\caption{Target-loss disturbances: success rate (\%) on the full validation splits for instant and smooth target-loss at the three severities. Clean is the same run as in \cref{tab:supp_env_sr}; every target-loss run is scored over the cell's $n$ kept episodes.}
\label{tab:supp_target_loss}
\end{table*}

\begin{table*}[htbp]
\centering
\scriptsize
\setlength{\tabcolsep}{2.5pt}
\begin{tabular}{llcccccc}
\toprule
\multirow{2}{*}{Model} & \multirow{2}{*}{Dataset} & \multicolumn{3}{c}{Instant} & \multicolumn{3}{c}{Smooth} \\
\cmidrule(lr){3-5}\cmidrule(lr){6-8}
 & & S1 & S2 & S3 & S1 & S2 & S3 \\
\midrule
\multirow{3}{*}{Uni-NaVid} & MP3D & $-$36.5\,[$-$40.1, $-$32.7] & $-$38.8\,[$-$42.4, $-$35.2] & $-$41.9\,[$-$45.6, $-$38.3] & $-$32.2\,[$-$35.8, $-$28.7] & $-$34.7\,[$-$38.4, $-$31.1] & $-$37.3\,[$-$41.1, $-$33.7] \\
 & HM3D & $-$35.8\,[$-$39.9, $-$31.7] & $-$39.5\,[$-$43.6, $-$35.4] & $-$43.1\,[$-$47.2, $-$39.0] & $-$28.5\,[$-$32.6, $-$24.6] & $-$32.0\,[$-$36.1, $-$27.8] & $-$37.6\,[$-$41.7, $-$33.5] \\
 & HSSD & $-$41.9\,[$-$46.0, $-$37.9] & $-$44.0\,[$-$48.0, $-$39.9] & $-$47.3\,[$-$51.4, $-$43.4] & $-$30.9\,[$-$34.7, $-$27.0] & $-$34.5\,[$-$38.5, $-$30.6] & $-$38.6\,[$-$42.7, $-$34.7] \\
\midrule
\multirow{3}{*}{LightNav-0} & MP3D & $-$76.6\,[$-$79.8, $-$73.2] & $-$82.7\,[$-$85.5, $-$79.8] & $-$86.1\,[$-$88.6, $-$83.3] & $-$18.4\,[$-$21.7, $-$15.2] & $-$28.6\,[$-$32.1, $-$25.0] & $-$45.2\,[$-$49.2, $-$41.3] \\
 & HM3D & $-$68.1\,[$-$71.8, $-$64.4] & $-$77.1\,[$-$80.3, $-$73.7] & $-$81.4\,[$-$84.6, $-$78.2] & $-$10.3\,[$-$13.3, $-$7.2] & $-$18.8\,[$-$22.1, $-$15.2] & $-$33.2\,[$-$37.0, $-$29.3] \\
 & HSSD & $-$71.5\,[$-$75.0, $-$68.0] & $-$77.6\,[$-$80.8, $-$74.3] & $-$81.2\,[$-$84.2, $-$78.2] & $-$17.4\,[$-$20.9, $-$14.2] & $-$26.5\,[$-$30.2, $-$22.8] & $-$42.0\,[$-$46.1, $-$38.1] \\
\midrule
\multirow{3}{*}{OmTrackVLA} & MP3D & $-$72.4\,[$-$75.7, $-$69.0] & $-$74.0\,[$-$77.3, $-$70.6] & $-$75.8\,[$-$78.9, $-$72.6] & $-$61.9\,[$-$65.6, $-$58.1] & $-$69.2\,[$-$72.6, $-$65.6] & $-$73.0\,[$-$76.3, $-$69.8] \\
 & HM3D & $-$70.1\,[$-$73.9, $-$66.3] & $-$72.5\,[$-$76.2, $-$68.8] & $-$73.8\,[$-$77.4, $-$70.1] & $-$53.7\,[$-$57.7, $-$49.4] & $-$65.3\,[$-$69.2, $-$61.5] & $-$71.1\,[$-$74.7, $-$67.2] \\
 & HSSD & $-$69.6\,[$-$73.1, $-$66.1] & $-$72.2\,[$-$75.4, $-$68.9] & $-$73.7\,[$-$76.8, $-$70.3] & $-$52.0\,[$-$56.0, $-$48.0] & $-$62.7\,[$-$66.4, $-$58.9] & $-$67.6\,[$-$71.1, $-$64.0] \\
\bottomrule
\end{tabular}
\caption{Target-loss disturbances: change in success rate from clean (percentage points), paired per episode over the same episodes as \cref{tab:supp_target_loss}, with 95\% bootstrap confidence intervals (5{,}000 episode resamples, seed 0).}
\label{tab:supp_target_loss_delta}
\end{table*}

\begin{table*}[htbp]
\centering
\scriptsize
\setlength{\tabcolsep}{3.2pt}
\begin{tabular}{llcccccc}
\toprule
\multirow{2}{*}{Model} & \multirow{2}{*}{Dataset} & \multicolumn{3}{c}{Instant} & \multicolumn{3}{c}{Smooth} \\
\cmidrule(lr){3-5}\cmidrule(lr){6-8}
 & & S1 & S2 & S3 & S1 & S2 & S3 \\
\midrule
\multirow{3}{*}{Uni-NaVid} & MP3D & 94 / 47 & 99 / 36 & 100 / 27 & 71 / 58 & 83 / 48 & 88 / 39 \\
 & HM3D & 90 / 47 & 99 / 36 & 100 / 27 & 72 / 57 & 84 / 47 & 91 / 38 \\
 & HSSD & 94 / 46 & 99 / 37 & 100 / 28 & 80 / 60 & 89 / 47 & 92 / 41 \\
\midrule
\multirow{3}{*}{LightNav-0} & MP3D & 96 / 47 & 99 / 32 & 100 / 21 & 44 / 94 & 63 / 85 & 83 / 71 \\
 & HM3D & 93 / 52 & 98 / 38 & 100 / 29 & 39 / 93 & 59 / 88 & 77 / 78 \\
 & HSSD & 95 / 43 & 99 / 31 & 100 / 23 & 47 / 91 & 66 / 83 & 83 / 68 \\
\midrule
\multirow{3}{*}{OmTrackVLA} & MP3D & 97 / 36 & 100 / 26 & 100 / 16 & 79 / 59 & 90 / 45 & 92 / 32 \\
 & HM3D & 98 / 31 & 100 / 22 & 100 / 14 & 76 / 56 & 91 / 38 & 95 / 26 \\
 & HSSD & 97 / 26 & 100 / 17 & 100 / 11 & 86 / 55 & 95 / 39 & 96 / 27 \\
\bottomrule
\end{tabular}
\caption{Target-loss diagnostics for the first event of each episode with a target-loss: share of these episodes in which the target is not followed after the rotation / share in which it is re-acquired within 20 steps (\%). Scored over all episodes of the full validation splits (the target-loss protocol's own scoring).}
\label{tab:supp_target_loss_diag}
\end{table*}

\begin{table*}[htbp]
\centering
\scriptsize
\setlength{\tabcolsep}{3.2pt}
\begin{tabular}{lllrrrrrrr}
\toprule
Model & Dataset & Profile & With (\%) & Events / ep. & Clean SR, with & Clean SR, without & Lost, all (\%) & Lost, with (\%) & Re-acquired (\%) \\
\midrule
\multirow{6}{*}{Uni-NaVid} & \multirow{2}{*}{MP3D} & instant & 83.2 & 0.94 & 57.7 & 26.4 & 79.9 & 87.0 & 26.7 \\
 &  & smooth & 83.2 & 0.98 & 57.5 & 27.3 & 71.2 & 77.4 & 39.3 \\
 & \multirow{2}{*}{HM3D} & instant & 75.7 & 0.80 & 68.0 & 38.6 & 70.9 & 78.4 & 26.8 \\
 &  & smooth & 75.6 & 0.82 & 68.1 & 38.3 & 61.8 & 67.0 & 37.9 \\
 & \multirow{2}{*}{HSSD} & instant & 79.4 & 0.90 & 74.6 & 62.0 & 65.7 & 77.0 & 28.3 \\
 &  & smooth & 79.4 & 0.96 & 74.6 & 62.0 & 53.6 & 62.3 & 41.0 \\
\midrule
\multirow{6}{*}{LightNav-0} & \multirow{2}{*}{MP3D} & instant & 97.2 & 1.22 & 92.6 & 21.1 & 95.0 & 95.1 & 20.2 \\
 &  & smooth & 97.4 & 2.32 & 92.5 & 22.2 & 49.9 & 49.8 & 70.7 \\
 & \multirow{2}{*}{HM3D} & instant & 96.6 & 1.17 & 94.9 & 23.8 & 88.0 & 87.9 & 28.5 \\
 &  & smooth & 96.8 & 1.71 & 94.9 & 20.0 & 35.9 & 35.4 & 78.1 \\
 & \multirow{2}{*}{HSSD} & instant & 97.4 & 1.29 & 93.0 & 38.9 & 88.7 & 88.6 & 23.6 \\
 &  & smooth & 97.4 & 1.98 & 92.8 & 44.4 & 45.9 & 45.2 & 68.1 \\
\midrule
\multirow{6}{*}{OmTrackVLA} & \multirow{2}{*}{MP3D} & instant & 94.1 & 1.06 & 81.7 & 20.0 & 97.2 & 97.3 & 15.8 \\
 &  & smooth & 94.0 & 1.20 & 81.6 & 22.0 & 93.6 & 93.7 & 32.2 \\
 & \multirow{2}{*}{HM3D} & instant & 92.7 & 0.99 & 83.4 & 19.6 & 93.7 & 93.6 & 13.7 \\
 &  & smooth & 93.2 & 1.07 & 83.3 & 16.3 & 90.3 & 90.2 & 25.3 \\
 & \multirow{2}{*}{HSSD} & instant & 95.5 & 1.04 & 80.6 & 9.7 & 95.1 & 95.3 & 10.5 \\
 &  & smooth & 95.4 & 1.22 & 80.7 & 9.4 & 87.3 & 87.4 & 27.2 \\
\bottomrule
\end{tabular}
\caption{Target-loss exposure at maximal severity, over each cell's kept episodes. \emph{With}: episodes with at least one target-loss event (a target-loss event fires only while the target is followed and centered); \emph{Events / ep.}: mean number of target-loss events per kept episode; \emph{Clean SR, with / without}: the success of the cell's clean reference on the episodes with and without a target-loss; \emph{Lost}: share of clean success lost on all kept episodes and on episodes with a target-loss only; \emph{Re-acquired}: target followed again within 20 steps of the episode's first event.}
\label{tab:supp_kick_exposure}
\end{table*}

\subsection{Controls}
\label{sec:supp_controls}

\subsubsection{Reference Checks of the Target-Loss Protocol}
\label{sec:supp_ctrl_reference}

\Cref{tab:supp_kick_reference} compares the environmental clean run with the reseeded clean run of the target-loss protocol, and gives that run's repeat and reversed-order variants. The two clean runs differ by at most $1.4$ points of SR.

\begin{table*}[htbp]
\centering
\footnotesize
\setlength{\tabcolsep}{3.5pt}
\begin{tabular}{llrrrcrcr}
\toprule
\multirow{2}{*}{Model} & \multirow{2}{*}{Dataset} & \multirow{2}{*}{$n$} & Clean SR & Clean SR & \multicolumn{2}{c}{Repeat} & \multicolumn{2}{c}{Reversed order} \\
\cmidrule(lr){6-7}\cmidrule(lr){8-9}
 & & & (env.) & (reseeded) & $\Delta$SR [95\% CI] & flips & $\Delta$SR [95\% CI] & flips \\
\midrule
\multirow{3}{*}{Uni-NaVid} & MP3D & 721 & 52.6 & 52.4 & $+$0.1\,[$-$0.3, $+$0.7] & 0.4 & $+$0.0\,[$-$0.8, $+$0.8] & 1.4 \\
 & HM3D & 684 & 60.8 & 60.4 & $-$0.3\,[$-$0.7, $+$0.0] & 0.3 & $-$0.3\,[$-$1.3, $+$0.7] & 1.8 \\
 & HSSD & 697 & 71.4 & 70.7 & $-$0.6\,[$-$1.1, $-$0.1] & 0.6 & $-$0.7\,[$-$1.7, $+$0.1] & 1.6 \\
\midrule
\multirow{3}{*}{LightNav-0} & MP3D & 721 & 90.2 & 89.6 & $-$0.1\,[$-$0.8, $+$0.6] & 1.0 & $+$0.7\,[$-$0.3, $+$1.8] & 2.1 \\
 & HM3D & 684 & 90.6 & 90.6 & $+$0.0\,[$+$0.0, $+$0.0] & 0.0 & $-$0.3\,[$-$1.2, $+$0.4] & 1.2 \\
 & HSSD & 697 & 91.2 & 91.5 & $+$0.0\,[$-$0.9, $+$0.9] & 1.1 & $+$0.1\,[$-$1.0, $+$1.3] & 2.4 \\
\midrule
\multirow{3}{*}{OmTrackVLA} & MP3D & 721 & 78.5 & 77.9 & $+$0.0\,[$+$0.0, $+$0.0] & 0.0 & $-$0.6\,[$-$1.8, $+$0.7] & 3.1 \\
 & HM3D & 684 & 77.5 & 78.1 & $+$0.0\,[$+$0.0, $+$0.0] & 0.0 & $-$0.6\,[$-$2.0, $+$0.7] & 3.5 \\
 & HSSD & 697 & 77.2 & 78.6 & $+$0.0\,[$+$0.0, $+$0.0] & 0.0 & $+$0.3\,[$-$0.9, $+$1.4] & 2.6 \\
\bottomrule
\end{tabular}
\caption{Reference checks of the target-loss protocol, over all episodes ($n$). \emph{Clean SR (env.)}: the cell's environmental clean run, before the episode exclusion; for Uni-NaVid on MP3D, the environmental pipeline's clean run with the person and instruction pinned as in the target-loss protocol ($52.6$), whereas the cell's reference in \cref{tab:supp_env_sr,tab:supp_target_loss} is the reseeded clean run ($52.4$). \emph{Clean SR (reseeded)}: the reseeded clean run (per-episode reseeding, pinned person and instruction). \emph{Repeat} / \emph{Reversed order}: that run repeated, and repeated with the episode order reversed, paired against it ($\Delta$SR in percentage points; 95\% bootstrap CI, 5{,}000 resamples, seed 0); \emph{flips}: episodes whose success changes (\%).}
\label{tab:supp_kick_reference}
\end{table*}

\subsubsection{Scene-Level Resampling}
\label{sec:supp_ctrl_scene}

Episodes of one scene share its geometry, and episodes that share an identifier share
their disturbance draws (\cref{sec:supp_disturbances}). \Cref{tab:supp_scene_boot} repeats the pooled headline intervals
with scenes or identifiers resampled in place of episodes, within each dataset. Pooled, scene resampling scales the
width of the relative-loss intervals by $0.92$--$1.69$ and moves no bound by more than $2.4$ points; identifier
resampling scales it by $0.94$--$1.19$. The oracle-minus-coin-flip intervals scale by $0.89$--$1.12$ pooled. Every interval of \cref{tab:supp_scene_boot} that excludes zero under episode resampling also excludes it
under scene and identifier resampling. For the fine-tuning gains pooled over MP3D, HM3D and HSSD (13, 64 and 86 scenes), scene resampling scales
the intervals by $0.86$--$1.14$; in two of the 56 comparisons the interval no longer excludes zero: \emph{Base} after a
mirrored instant target-loss ($+4.0$ [$-0.2$, $+7.9$]) and \emph{Decompress}'s first turn correct under the same
target-loss, instant ($+5.3$ [$-0.5$, $+10.6$]).

\begin{table*}[htbp]
\centering
\scriptsize
\setlength{\tabcolsep}{3pt}
\begin{tabular}[t]{@{}llcccc@{}}
\toprule
Model & Condition & Lost (\%) & Episodes & Scenes & Identifiers \\
\midrule
Uni-NaVid & Current & 66.1 & [63.3, 68.8] & [62.3, 69.6] & [63.3, 68.8] \\
 & Veer & 53.7 & [50.7, 56.8] & [48.2, 58.6] & [50.8, 56.8] \\
 & Gust & 35.9 & [32.6, 39.1] & [32.5, 38.9] & [32.2, 39.3] \\
 & Instant & 71.6 & [69.0, 74.1] & [67.9, 74.8] & [69.2, 73.9] \\
 & Smooth & 61.4 & [58.5, 64.1] & [57.3, 65.0] & [58.7, 64.1] \\
\addlinespace
LightNav-0 & Current & 47.5 & [45.0, 49.9] & [44.0, 50.7] & [45.1, 49.7] \\
 & Veer & 6.6 & [5.2, 8.0] & [5.0, 8.4] & [5.2, 8.1] \\
 & Gust & 18.6 & [16.5, 20.6] & [16.4, 21.0] & [16.2, 21.1] \\
 & Instant & 90.6 & [89.3, 92.0] & [88.2, 92.5] & [89.3, 92.0] \\
 & Smooth & 44.1 & [41.7, 46.4] & [40.7, 47.0] & [41.5, 46.8] \\
\addlinespace
OmTrackVLA & Current & 58.7 & [56.2, 61.2] & [54.9, 62.2] & [56.2, 61.2] \\
 & Veer & 79.7 & [77.7, 81.6] & [76.4, 82.4] & [77.5, 81.7] \\
 & Gust & 33.2 & [30.5, 35.9] & [30.6, 35.6] & [30.0, 36.3] \\
 & Instant & 95.4 & [94.3, 96.4] & [93.8, 96.6] & [94.3, 96.4] \\
 & Smooth & 90.4 & [88.9, 91.9] & [88.2, 92.1] & [88.8, 91.9] \\
\bottomrule
\end{tabular}\hfill
\begin{tabular}[t]{@{}llcccc@{}}
\toprule
Target-loss & Model & $n$ & Oracle $-$ coin & Episodes & Scenes \\
\midrule
Instant & Uni-NaVid & 368 & 35.9 & [31.1, 40.7] & [31.6, 40.2] \\
 & LightNav-0 & 543 & 42.9 & [38.8, 47.0] & [38.6, 47.2] \\
 & OmTrackVLA & 513 & 37.4 & [33.2, 41.6] & [33.7, 41.2] \\
\addlinespace
Smooth & Uni-NaVid & 219 & 37.4 & [31.2, 43.7] & [30.5, 44.4] \\
 & LightNav-0 & 155 & 38.1 & [30.4, 45.7] & [30.3, 45.9] \\
 & OmTrackVLA & 179 & 45.8 & [38.7, 52.9] & [37.9, 53.7] \\
\bottomrule
\end{tabular}
\caption{Headline intervals under episode, scene and identifier resampling, pooled over the three datasets (95\%,
5{,}000 resamples, seed 0, clusters resampled within each dataset). Left: share of clean success lost at maximal
severity (\cref{tab:supp_relative_loss}). Right: oracle minus coin flip, re-acquisition within 20 steps after the
first target-loss (percentage points; \cref{app:internals_functional}), pooled from the per-dataset intervals.}
\label{tab:supp_scene_boot}
\end{table*}

\FloatBarrier
\subsubsection{Clean Success Against the Published Numbers}
\label{sec:supp_ctrl_published}

The MP3D and HM3D validation splits together are EVT-Bench's 1{,}405 single-target test episodes, on which the models' authors report results. \Cref{tab:supp_published} compares our clean runs with those numbers. Our runs use greedy decoding, one process or five shards and the $90^\circ$ field of view; the table notes give the published settings.

\begin{table}[htbp]
\centering
\footnotesize
\setlength{\tabcolsep}{2.5pt}
\begin{tabular}{@{}llccc@{}}
\toprule
Model & Run & SR & TR & CR \\
\midrule
\multirow{3}{*}{Uni-NaVid} & Clean, env.\ protocol & 56.5 & 70.4 & 5.6 \\
 & Clean, target-loss protocol & 56.3 & 68.9 & 5.2 \\
 & Published$^{a}$ & 53.3 & 67.2 & 12.6 \\
\midrule
\multirow{3}{*}{LightNav-0} & Clean, env.\ protocol & 90.4 & 88.6 & 3.0 \\
 & Clean, target-loss protocol & 90.1 & 88.4 & 2.7 \\
 & Published$^{b}$ & 91.7 & 87.7 & 1.9 \\
\midrule
\multirow{3}{*}{OmTrackVLA} & Clean, env.\ protocol & 78.0 & 82.5 & 4.3 \\
 & Clean, target-loss protocol & 78.0 & 81.2 & 4.7 \\
 & Published$^{c}$ & 81.4 & 82.8 & 5.1 \\
\bottomrule
\end{tabular}
\caption{Clean success rate, tracking rate and collision rate (\%) on EVT-Bench's 1{,}405 single-target test episodes (MP3D 721 + HM3D 684, all episodes, no exclusion), for our two clean protocols and as published by the models' authors. $^{a}$~TrackVLA and Uni-NaVid repositories and \cite{lightnav0}, Tab.~VIII; the official Uni-NaVid script samples actions and runs 30 shards. $^{b}$~\cite{lightnav0}, Tab.~VIII; the authors' repository evaluates with a $120^\circ$ field of view. $^{c}$~The OmTrackVLA repository \cite{omtrackvla2025}, current weights, $384$\,px render as here. For Uni-NaVid, the environmental-protocol row takes MP3D from the cell's clean reference, the reseeded clean run ($52.4$, \cref{tab:supp_env_sr}); the environmental pipeline's own clean run with the same pinned person and instruction ($52.6$) is in \cref{tab:supp_kick_reference}.}
\label{tab:supp_published}
\end{table}

\FloatBarrier
\section{Model-Internal Analysis}
\label{app:internals}

\Cref{app:internals_data} specifies the measurements of \cref{sec:internals_design}; \cref{app:internals_readers,app:internals_grp_history,app:internals_grp_backbone,app:internals_functional,app:internals_grp_use}
give the full results behind the subsections of \cref{sec:results_internals}, in their order; the controls and
ablations are collected in \cref{app:internals_controls}.

\subsection{Model-Internal Measurements}
\label{sec:supp_internals_measure}
\label{app:internals_data}

\subsubsection{Recordings and Readouts}

The internal measurements use recorded trajectories: Uni-NaVid's own closed-loop trajectories, replayed through all three models so that they receive identical inputs, and, for the on-policy check (\cref{app:internals_onpolicy}), LightNav-0's and OmTrackVLA's own closed-loop trajectories. HSSD has no EVT-Bench training episodes, so its collections use scenes of the benchmark's HSSD set; no benchmark number is fitted on them. Each recording stores, per step, the rendered frame, the issued action, the robot and target poses and the realized disturbance, so that every camera motion decomposes into its commanded and uncommanded parts. Camera-motion targets follow the rendered camera, including the one-step render lag, rather than an unshifted difference of simulator poses. Lossless recordings store PNG frames. The exception is Uni-NaVid's MP3D sweep collection, recorded with JPEG frames: it enters the probe-format, frame-shuffle, compression and projector tables (\cref{tab:app_reader_format,tab:app_shuffle,tab:app_compression,tab:app_projector}), each marked ``MP3D JPEG'' with the HM3D and HSSD rows lossless. \Cref{tab:supp_flow} lists the collections.

\paragraph{What is captured.} Every analysis reads the frozen models' own representations at declared token groups: the visual encoder's grids, the memory and navigation tokens, and the language-model states. For the rotation map, each recorded state retains full-dimensional
tokens (\cref{tab:app_crossmodel_layout}): 73 for Uni-NaVid (the decision-state token, four previous-frame memory tokens, four
current-frame memory tokens, and 64 current-navigation tokens), 141 for LightNav-0 and 73 for OmTrackVLA. All
map states are retained: the input and the outputs of all 32, 36 and 28 decoder layers (33, 37 and 29 states). $L_{0}$ denotes the language-model input and $L_{k}$
the output of decoder layer $k-1$. This is full-token capture at the
declared sites, not every text token or every older historical frame in
the prompt. Rotation recordings retain event and pre-event states while
replay advances through the complete observation sequence. LightNav-0 and OmTrackVLA receive the rotation recordings one step per frame, so that their memory evolves as in a live run; their actions are recorded but not executed. Hooks observe and never modify the computation.

\paragraph{No lossy reduction.} Readouts use the captured representations at full dimension, with no PCA, token averaging or spatial pooling added by the analysis; the only spatial reduction is the compression ablation itself, which deliberately changes the retained grid, so any loss of motion information measured is the model's own.

\paragraph{Coverage across policies and domains.}
Table~\ref{tab:app_internals_coverage} distinguishes coverage of a visual-stage observation from coverage
of a policy's complete internal pathway. The real videos do not contain person-following tasks or externally
scheduled target-loss rotations. The probe format, frame shuffle, history compression, decision-state rotation readout, recovery controls, own-trajectory maps,
steering and erasure, rotation mirror, interchange (also split by target visibility), patch search, the grid ladder and the measurement controls (encoder sweep, projector) run on
all three datasets.

\begin{table*}[tbp]
\centering
\footnotesize
\setlength{\tabcolsep}{4pt}
\caption{Evidence coverage. Internal-diagnostic
collections, validation-set recovery controls and the behavioral benchmark are separate collections.}
\label{tab:app_internals_coverage}
\begin{tabular}{@{}p{.13\textwidth}p{.29\textwidth}p{.19\textwidth}p{.32\textwidth}@{}}
\toprule
Domain & Visual evidence / grid tests & Full-depth rotation maps & Action-use / recovery evidence \\
\midrule
MP3D training scenes & Native grids of all three policies: probe format, shuffle, compression, grid ladder; encoder sweep and projector controls & All three policies; own trajectories of LightNav-0 and OmTrackVLA & All three: steering and erasure (a reading for Uni-NaVid and OmTrackVLA), rotation mirror, interchange, patch search \\
HM3D training scenes & Native grids of all three policies: probe format, shuffle, compression, grid ladder; encoder sweep and projector controls & All three policies; own trajectories of LightNav-0 and OmTrackVLA & All three: steering and erasure (a reading for all three), rotation mirror, interchange, patch search \\
HSSD benchmark scenes & Native grids of all three policies: probe format, shuffle, compression, grid ladder; encoder sweep and projector controls & All three policies; own trajectories of LightNav-0 and OmTrackVLA & All three: steering and erasure (a reading for Uni-NaVid only), rotation mirror, interchange, patch search \\
MP3D / HM3D / HSSD validation & -- & -- & All three: $n=200$ recovery per dataset, instant and smooth target-loss (LightNav-0 decoding: \cref{app:internals_functional}) \\
TUM Pioneer & Three native visual interfaces & -- & -- \\
KITTI & Three native visual interfaces & -- & -- \\
TartanDrive & Three native visual interfaces & -- & -- \\
\bottomrule
\end{tabular}
\end{table*}

\begin{table*}[tbp]
\centering
\footnotesize
\setlength{\tabcolsep}{3pt}
\caption{Recording collections and counts behind every analysis, from scenes to the units analyzed (MP3D / HM3D / HSSD where an analysis covers the
three datasets). Exclusions and checks are described in the cited sections.}
\label{tab:supp_flow}
\begin{tabular}{@{}p{.19\textwidth}p{.27\textwidth}p{.15\textwidth}p{.31\textwidth}@{}}
\toprule
Used for & Collection & Scenes / episodes & Units analyzed \\
\midrule
Benchmark (\cref{sec:supp_results}) & validation splits & 14 / 87 / 164 scenes; 721 / 684 / 697 episodes & 624--721 kept episodes per model--dataset cell, each paired with its clean run (\cref{sec:supp_metrics}) \\
Probes: encoder readouts, probe format, shuffle, compression ladders; Uni-NaVid's spatial-grid and projector ablations (\cref{app:internals_readers,app:internals_compression,app:internals_controls}) & MP3D compression sweep (101 episodes per condition: clean, S3 current, gust) and its lossless re-recording (PNG); HM3D and HSSD pairs of the same design, lossless (HM3D training scenes, HSSD benchmark scenes) & 35 / 35 / 49 scenes; five scene folds & clean, current and gust frame pairs: 9{,}657, 7{,}132, 7{,}979 / 6{,}662, 5{,}784, 6{,}099 / 7{,}613, 6{,}587, 7{,}592 \\
Rotation maps of all three models (\cref{app:internals_crossmodel}) & map collections (PNG) & 10 / 9 / 10 scenes; MP3D: 150 episodes ($10\times15$) & clean fitting rows 9{,}267 / 7{,}771 / 11{,}323; imposed rotations 1{,}180 / 1{,}074 / 1{,}177 (MP3D: at least $5^\circ$, median $20^\circ$), of which 486 / 429 / 459 without a turn command (scored) \\
On-policy rotation maps (\cref{app:internals_onpolicy}) & LightNav-0's and OmTrackVLA's own closed-loop recordings, replayed exactly & the map collections' episodes, per model: 150 / 135 / 148 & steps 18{,}537 / 16{,}269 / 19{,}058; imposed rotations LightNav-0 1{,}240 / 1{,}066 / 1{,}337, OmTrackVLA 1{,}071 / 984 / 1{,}281, of which 376 / 362 / 438 and 255 / 234 / 367 with a near-zero turn command \\
Recovery controls: target-recovery feasibility, first-turn readout (\cref{app:internals_functional}) & validation splits; clean, policy, coin-flip and oracle arms after instant and smooth target-loss & 200 episodes per model and dataset & matched first target-loss events per cell: 102--189 instant, 45--92 smooth \\
Rotation mirror (\cref{app:internals_mirror}) & map collections; forks, each an episode re-rendered with one rotation's side reversed, pixel-identical before the flip: 456 / 375 / 418 & the episodes with a command-free rotation: 148 / 132 / 146 & 486 / 421 / 459 command-free events (MP3D, LightNav-0: 485 with a finite turn) \\
Interchange (\cref{app:internals_interchange}) & map collections and their forks & 148 / 132 / 146 episodes & 454 / 372 / 417 eligible events (LightNav-0: 453 / 371 / 417) \\
Patch search (\cref{app:internals_patch_search,app:internals_patch_xd}) & map collections and their forks & 148 / 132 / 146 episodes & 454 / 372 / 417 eligible events (halves A / B: 249 / 205, 176 / 196, 204 / 213) \\
Steering, erasure (\cref{app:internals_causal}) & map collections & 148 / 133 / 146 episodes & 486 / 429 / 459 command-free events \\
Fine-tuning (\cref{app:interventions}) & MP3D, HM3D and HSSD validation episodes & 200 episodes per dataset & every episode, paired with the frozen model \\
\bottomrule
\end{tabular}
\end{table*}

\paragraph{HM3D and HSSD collections, and combining datasets.}
The motion-evidence collections of the three datasets share one design with lossless recordings
(pair counts in \cref{app:internals_readers}). One HM3D scene (16tymPtM7uS) is excluded from every HM3D analysis because replaying it did not
reproduce the recorded decisions: neither Uni-NaVid's replay nor LightNav-0's on-policy replay reproduced them. Each dataset is analyzed separately, with its own folds and
bootstrap. The main-text figures that combine datasets average per-dataset estimates:
\cref{fig:internals_motion_evidence,fig:internals_history_compression} show the equal-weight mean of the three datasets,
with standard error
\begin{equation}
 \mathrm{SE}=\frac{1}{3}\sqrt{\textstyle\sum_{i}\mathrm{SE}_i^2}
\end{equation}
($\mathrm{SE}_i$ from each dataset's 95\% interval);
\cref{fig:internals_recovery_pooled} weights the datasets by event count; the episode-level recovery values pool the
200 episodes of each dataset.

\paragraph{Fitting and temporal alignment.}
Probes are fitted on clean motion and evaluated using five scene- or
group-held-out folds. No disturbed-scene motion labels enter the clean-fit
probe optimization. The visual sweeps exclude recordings flagged by the
displacement and missing-pose checks. Input/output standardization is fitted
on the clean training fold. Some label-free feature-centering references
are computed once on the clean collection.

\paragraph{Statistical units.}
Episode-bootstrap resample counts differ by analysis: 1{,}000 for the disturbance sensitivities $b$
(\cref{sec:internals_design}) and the compression ladders
(\cref{app:internals_readers,app:internals_grp_history}); 2{,}000 for the rotation maps
(\cref{app:internals_crossmodel,app:internals_onpolicy}) and Uni-NaVid's steering and erasure test
(\cref{app:internals_causal}); 5{,}000 for the other analyses, as stated with their results. Frozen Uni-NaVid values that also appear in \cref{app:interventions_internals} carry its intervals (5{,}000 resamples). Every bootstrap uses seed 0.

Intervals are pointwise, not simultaneous bands over all layers, grids,
or encoders. Bootstraps quantify
uncertainty conditional on the out-of-fold predictions; they do not
retrain probes or fully characterize optimization variability.

\subsection{Probing Uncommanded Motion in Visual Features}
\label{app:internals_readers}

\paragraph{Motion targets and residual sensitivity.}
Let $\widehat{\boldsymbol m}_t$ be the probe's predicted camera motion,
$\boldsymbol c_t$ the commanded-motion vector, and $\boldsymbol u_t$
the lag-aligned uncommanded residual. For each translation axis $j$, the
analysis regresses held-out predictions against residual motion and all
command axes:
\begin{equation}
 \widehat m_{t,j}
 = \boldsymbol\alpha_j^{\mathsf T}\boldsymbol c_t
   + b_j u_{t,j} + d_j + \epsilon_{t,j}.
\end{equation}
Here the command terms are written in their equivalent nonredundant
form; the implementation additionally includes the axis-specific command
column. The residual coefficient $b_j$ measures sensitivity under this
probe and distribution. It is neither an information-retention fraction
nor a probability that the policy knows the disturbance. Clean-motion
$R^2$ is reported separately; negative values indicate performance worse
than a constant-mean prediction.

\paragraph{Standard and comparison probes.}
The standard probe receives the two frames' spatial feature grids
concatenated along the channel dimension. The comparison probe instead
receives local cross-frame cosine similarities computed using every
feature channel. For centered features $\widetilde F$, its cost volume is
\begin{equation}
 C_t(p,\Delta)=
 \frac{\langle\widetilde F_t(p),
 \widetilde F_{t-1}(p+\Delta)\rangle}
 {\|\widetilde F_t(p)\|\,
  \|\widetilde F_{t-1}(p+\Delta)\|},
 \quad \Delta\in\{-2,\ldots,2\}^2.
\end{equation}
Out-of-grid entries are zero. For asymmetric grids, previous-frame tokens
are replicated onto the current grid before comparison; this adds no
spatial information. Cost volumes from coarser grids are replicated to
the probe's fixed input size. The correspondence is
computed by the analysis, not by the frozen policy.

\paragraph{Probe capacity and fitting.}
The nonlinear spatial probe uses three convolutional layers and a
fully connected output head. It is trained with AdamW, learning rate
$10^{-3}$, weight decay $10^{-3}$, batch size 512, and at most 60 epochs.
Early stopping uses patience six on an inner validation split comprising
20\% of the training-fold scenes. Full-dimensional linear controls use
ridge regression, solved in primal or exact dual form, with regularization selected by inner scene-grouped
cross-validation. The full-channel standard control retains all
720,896 EVA-CLIP input values or 301,056 pixel input values per frame
pair.

\begin{table*}[tbp]
\centering
\footnotesize
\setlength{\tabcolsep}{3pt}
\caption{Probe format on the same frame pairs (Uni-NaVid's recordings: MP3D JPEG, HM3D and HSSD lossless; $16\times16$ patch grids, final encoder
layer): disturbance readout $b$ [95\% CI] of the full-channel standard probe (the two frames'
features side by side) and of the comparison probe (their cost volume), same convolutional probe.}
\label{tab:app_reader_format}
\begin{tabular}{@{}lllcccccccc@{}}
\toprule
 &  & & \multicolumn{4}{c}{current S3} & \multicolumn{4}{c}{gust S3} \\
\cmidrule(lr){4-7}\cmidrule(lr){8-11}
 &  & & \multicolumn{2}{c}{standard} & \multicolumn{2}{c}{comparison} & \multicolumn{2}{c}{standard} & \multicolumn{2}{c}{comparison} \\
Dataset & Input & Layer & fwd & lat & fwd & lat & fwd & lat & fwd & lat \\
\midrule
MP3D & EVA-CLIP & $L_{39}$ & 0.13 {\tiny[0.04, 0.21]} & 0.09 {\tiny[0.07, 0.12]} & 0.53 {\tiny[0.44, 0.61]} & 0.34 {\tiny[0.31, 0.38]} & 0.20 {\tiny[0.15, 0.24]} & 0.10 {\tiny[0.09, 0.12]} & 0.54 {\tiny[0.48, 0.60]} & 0.40 {\tiny[0.38, 0.43]} \\
 & CLIP & $L_{24}$ & 0.14 {\tiny[0.05, 0.21]} & 0.10 {\tiny[0.07, 0.12]} & 0.57 {\tiny[0.46, 0.68]} & 0.38 {\tiny[0.35, 0.43]} & 0.17 {\tiny[0.13, 0.21]} & 0.12 {\tiny[0.10, 0.13]} & 0.53 {\tiny[0.47, 0.59]} & 0.42 {\tiny[0.40, 0.45]} \\
 & SigLIP & $L_{27}$ & 0.24 {\tiny[0.17, 0.30]} & 0.10 {\tiny[0.08, 0.12]} & 0.54 {\tiny[0.45, 0.63]} & 0.34 {\tiny[0.30, 0.39]} & 0.23 {\tiny[0.18, 0.28]} & 0.13 {\tiny[0.11, 0.14]} & 0.54 {\tiny[0.48, 0.60]} & 0.38 {\tiny[0.36, 0.40]} \\
 & DINOv2 & $L_{24}$ & 0.16 {\tiny[0.05, 0.25]} & 0.14 {\tiny[0.11, 0.16]} & 0.61 {\tiny[0.52, 0.70]} & 0.42 {\tiny[0.38, 0.47]} & 0.18 {\tiny[0.13, 0.22]} & 0.15 {\tiny[0.12, 0.17]} & 0.59 {\tiny[0.53, 0.64]} & 0.48 {\tiny[0.45, 0.50]} \\
 & MAE & $L_{24}$ & 0.25 {\tiny[0.18, 0.31]} & 0.16 {\tiny[0.14, 0.19]} & 0.64 {\tiny[0.52, 0.76]} & 0.49 {\tiny[0.44, 0.54]} & 0.25 {\tiny[0.21, 0.29]} & 0.19 {\tiny[0.17, 0.21]} & 0.60 {\tiny[0.54, 0.67]} & 0.54 {\tiny[0.51, 0.57]} \\
 & pixels & -- & 0.37 {\tiny[0.28, 0.47]} & 0.20 {\tiny[0.18, 0.22]} & 0.51 {\tiny[0.43, 0.60]} & 0.29 {\tiny[0.26, 0.33]} & 0.38 {\tiny[0.33, 0.44]} & 0.24 {\tiny[0.22, 0.26]} & 0.52 {\tiny[0.45, 0.58]} & 0.34 {\tiny[0.32, 0.36]} \\
\midrule
HM3D & EVA-CLIP & $L_{39}$ & 0.23 {\tiny[0.16, 0.30]} & 0.12 {\tiny[0.09, 0.14]} & 0.55 {\tiny[0.49, 0.62]} & 0.36 {\tiny[0.32, 0.39]} & 0.25 {\tiny[0.18, 0.31]} & 0.10 {\tiny[0.08, 0.12]} & 0.58 {\tiny[0.52, 0.64]} & 0.38 {\tiny[0.34, 0.41]} \\
 & CLIP & $L_{24}$ & 0.20 {\tiny[0.13, 0.27]} & 0.15 {\tiny[0.12, 0.17]} & 0.61 {\tiny[0.53, 0.68]} & 0.39 {\tiny[0.36, 0.43]} & 0.25 {\tiny[0.19, 0.31]} & 0.13 {\tiny[0.11, 0.15]} & 0.63 {\tiny[0.57, 0.69]} & 0.40 {\tiny[0.37, 0.43]} \\
 & SigLIP & $L_{27}$ & 0.23 {\tiny[0.17, 0.31]} & 0.12 {\tiny[0.10, 0.14]} & 0.55 {\tiny[0.47, 0.62]} & 0.32 {\tiny[0.29, 0.36]} & 0.24 {\tiny[0.17, 0.31]} & 0.12 {\tiny[0.10, 0.14]} & 0.59 {\tiny[0.53, 0.64]} & 0.34 {\tiny[0.32, 0.37]} \\
 & DINOv2 & $L_{24}$ & 0.25 {\tiny[0.18, 0.32]} & 0.16 {\tiny[0.13, 0.19]} & 0.61 {\tiny[0.52, 0.69]} & 0.44 {\tiny[0.40, 0.48]} & 0.28 {\tiny[0.21, 0.35]} & 0.14 {\tiny[0.12, 0.16]} & 0.64 {\tiny[0.57, 0.70]} & 0.45 {\tiny[0.42, 0.49]} \\
 & MAE & $L_{24}$ & 0.27 {\tiny[0.21, 0.35]} & 0.21 {\tiny[0.18, 0.24]} & 0.60 {\tiny[0.53, 0.67]} & 0.46 {\tiny[0.42, 0.50]} & 0.30 {\tiny[0.24, 0.38]} & 0.20 {\tiny[0.17, 0.23]} & 0.62 {\tiny[0.56, 0.68]} & 0.46 {\tiny[0.43, 0.50]} \\
 & pixels & -- & 0.39 {\tiny[0.31, 0.46]} & 0.23 {\tiny[0.20, 0.25]} & 0.51 {\tiny[0.44, 0.59]} & 0.28 {\tiny[0.24, 0.32]} & 0.46 {\tiny[0.39, 0.54]} & 0.22 {\tiny[0.20, 0.24]} & 0.58 {\tiny[0.51, 0.65]} & 0.27 {\tiny[0.24, 0.29]} \\
\midrule
HSSD & EVA-CLIP & $L_{39}$ & 0.12 {\tiny[0.05, 0.20]} & 0.04 {\tiny[0.03, 0.05]} & 0.28 {\tiny[0.17, 0.38]} & 0.13 {\tiny[0.11, 0.16]} & 0.19 {\tiny[0.10, 0.28]} & 0.05 {\tiny[0.04, 0.07]} & 0.31 {\tiny[0.24, 0.38]} & 0.18 {\tiny[0.15, 0.22]} \\
 & CLIP & $L_{24}$ & 0.14 {\tiny[0.05, 0.23]} & 0.05 {\tiny[0.04, 0.07]} & 0.29 {\tiny[0.18, 0.40]} & 0.15 {\tiny[0.12, 0.17]} & 0.17 {\tiny[0.10, 0.26]} & 0.07 {\tiny[0.06, 0.08]} & 0.32 {\tiny[0.24, 0.39]} & 0.20 {\tiny[0.17, 0.23]} \\
 & SigLIP & $L_{27}$ & 0.08 {\tiny[0.01, 0.18]} & 0.03 {\tiny[0.02, 0.05]} & 0.31 {\tiny[0.19, 0.42]} & 0.12 {\tiny[0.10, 0.15]} & 0.16 {\tiny[0.09, 0.24]} & 0.05 {\tiny[0.03, 0.06]} & 0.32 {\tiny[0.23, 0.39]} & 0.16 {\tiny[0.14, 0.19]} \\
 & DINOv2 & $L_{24}$ & 0.19 {\tiny[0.12, 0.28]} & 0.07 {\tiny[0.05, 0.08]} & 0.32 {\tiny[0.17, 0.46]} & 0.17 {\tiny[0.14, 0.21]} & 0.22 {\tiny[0.13, 0.33]} & 0.08 {\tiny[0.06, 0.09]} & 0.35 {\tiny[0.27, 0.41]} & 0.23 {\tiny[0.19, 0.26]} \\
 & MAE & $L_{24}$ & 0.17 {\tiny[0.10, 0.26]} & 0.09 {\tiny[0.07, 0.11]} & 0.28 {\tiny[0.11, 0.44]} & 0.19 {\tiny[0.16, 0.23]} & 0.20 {\tiny[0.13, 0.27]} & 0.10 {\tiny[0.08, 0.11]} & 0.31 {\tiny[0.24, 0.38]} & 0.25 {\tiny[0.22, 0.29]} \\
 & pixels & -- & 0.18 {\tiny[0.09, 0.28]} & 0.06 {\tiny[0.04, 0.08]} & 0.34 {\tiny[0.26, 0.42]} & 0.10 {\tiny[0.08, 0.13]} & 0.21 {\tiny[0.14, 0.27]} & 0.08 {\tiny[0.07, 0.10]} & 0.28 {\tiny[0.21, 0.34]} & 0.13 {\tiny[0.11, 0.15]} \\
\bottomrule
\end{tabular}
\end{table*}

\paragraph{Probe-format ablation on five encoders.}
The standard probe is the ablation of the comparison probe: the same convolutional probe, frame pairs, targets and
folds, with the two frames' features side by side instead of their explicit cross-frame correspondence. The ablation
runs on Uni-NaVid's recordings of the three datasets for five frozen image encoders (EVA-CLIP, CLIP, SigLIP, DINOv2,
MAE~\cite{sun2023evaclip,radford2021clip,zhai2023siglip,oquab2024dinov2,he2022mae}) and raw pixels
(Table~\ref{tab:app_reader_format}). As a capacity control, the standard probe is also fitted as a full-dimensional
exact ridge (not tabulated); in every encoder and disturbance--axis cell the comparison probe exceeds the better of the
two standard probes by at least $0.23$ on MP3D, $0.20$ on HM3D and $0.08$ on HSSD, where the forward intervals of the
two probes overlap in nine of the ten encoder--disturbance cells. With raw pixels the margin is at least $0.09$, $0.04$ and $0.04$.

\paragraph{The policies' own visual features.}
The main-text figure (\cref{fig:internals_motion_evidence}, the equal-weight mean over the datasets; per dataset in
\cref{tab:app_reader_format_policies}) applies both probes to each policy's own encoder output at its native grid: Uni-NaVid's EVA-CLIP final layer ($16\times16\times1408$), OmTrackVLA's concatenated DINOv3 and SigLIP
features ($24\times24\times1536$) and LightNav-0's per-frame vision features ($16\times28\times1024$), on the lossless
re-recordings used by the cross-model compression sweep. On each dataset all arms share the same frame pairs, scene
folds and targets: MP3D 9{,}657 clean, 7{,}132 current and 7{,}979 gust pairs (35 scenes; five scene folds), HM3D 6{,}662, 5{,}784
and 6{,}099 (35), HSSD 7{,}613, 6{,}587 and 7{,}592 (49). Every input value is kept (stored as fp16, fp32 and bf16 bits,
respectively). EVA-CLIP appears in both this set and \cref{tab:app_reader_format} and agrees within $0.02$ across the JPEG
and lossless recordings on MP3D; on HM3D and HSSD both tables use the same lossless recordings and give the same
values.

\paragraph{Probe-format ablation on each policy.}
The same ablation and exact-ridge control run on each policy's own features on every dataset
(\cref{tab:app_reader_format_policies}). Against the better of the convolutional and exact-ridge standard probes, the
comparison probe's smallest margin over the four disturbance--axis cells is $0.25$--$0.27$ (MP3D), $0.22$--$0.24$
(HM3D) and $0.08$--$0.10$ (HSSD) for Uni-NaVid and OmTrackVLA, and $0.02$, $0.03$ and $-0.06$ for LightNav-0. The
comparison readouts are lower on HSSD than on MP3D and HM3D ($0.13$--$0.39$ against $0.35$--$0.65$ for Uni-NaVid and
OmTrackVLA), and LightNav-0's forward readout under current on HSSD has an interval that includes zero ($0.06$
[$-0.04$, $0.16$]).

\begin{table*}[tbp]
\centering
\footnotesize
\setlength{\tabcolsep}{2.2pt}
\caption{Probe format on each policy's own visual features (lossless recordings, native grids; equal-weight mean in
\cref{fig:internals_motion_evidence}): disturbance readout $b$ [95\% CI] of the full-channel standard probe and of the
comparison probe, same convolutional probe.}
\label{tab:app_reader_format_policies}
\resizebox{\textwidth}{!}{%
\begin{tabular}{@{}llcccccccc@{}}
\toprule
 & & \multicolumn{4}{c}{current S3} & \multicolumn{4}{c}{gust S3} \\
\cmidrule(lr){3-6}\cmidrule(lr){7-10}
 & & \multicolumn{2}{c}{standard} & \multicolumn{2}{c}{comparison} & \multicolumn{2}{c}{standard} & \multicolumn{2}{c}{comparison} \\
Policy & Dataset & fwd & lat & fwd & lat & fwd & lat & fwd & lat \\
\midrule
\multirow{3}{*}{\shortstack[l]{Uni-NaVid\\{\tiny EVA-CLIP L39}}} & MP3D & 0.11 {\tiny[0.01, 0.20]} & 0.09 {\tiny[0.06, 0.11]} & 0.54 {\tiny[0.44, 0.64]} & 0.35 {\tiny[0.31, 0.41]} & 0.20 {\tiny[0.15, 0.24]} & 0.10 {\tiny[0.09, 0.12]} & 0.55 {\tiny[0.48, 0.61]} & 0.40 {\tiny[0.37, 0.42]} \\
 & HM3D & 0.23 {\tiny[0.16, 0.30]} & 0.12 {\tiny[0.09, 0.14]} & 0.55 {\tiny[0.49, 0.62]} & 0.36 {\tiny[0.32, 0.39]} & 0.25 {\tiny[0.18, 0.31]} & 0.10 {\tiny[0.08, 0.12]} & 0.58 {\tiny[0.52, 0.64]} & 0.38 {\tiny[0.34, 0.41]} \\
 & HSSD & 0.12 {\tiny[0.05, 0.20]} & 0.04 {\tiny[0.03, 0.05]} & 0.28 {\tiny[0.17, 0.38]} & 0.13 {\tiny[0.11, 0.16]} & 0.19 {\tiny[0.10, 0.28]} & 0.05 {\tiny[0.04, 0.07]} & 0.31 {\tiny[0.24, 0.38]} & 0.18 {\tiny[0.15, 0.22]} \\
\midrule
\multirow{3}{*}{\shortstack[l]{OmTrackVLA\\{\tiny DINOv3+SigLIP}}} & MP3D & 0.25 {\tiny[0.18, 0.32]} & 0.13 {\tiny[0.11, 0.16]} & 0.65 {\tiny[0.53, 0.77]} & 0.38 {\tiny[0.35, 0.42]} & 0.25 {\tiny[0.20, 0.30]} & 0.17 {\tiny[0.15, 0.19]} & 0.60 {\tiny[0.53, 0.67]} & 0.44 {\tiny[0.42, 0.47]} \\
 & HM3D & 0.26 {\tiny[0.20, 0.33]} & 0.17 {\tiny[0.14, 0.19]} & 0.62 {\tiny[0.55, 0.69]} & 0.39 {\tiny[0.35, 0.42]} & 0.29 {\tiny[0.22, 0.36]} & 0.16 {\tiny[0.13, 0.19]} & 0.62 {\tiny[0.56, 0.69]} & 0.40 {\tiny[0.37, 0.43]} \\
 & HSSD & 0.15 {\tiny[0.05, 0.24]} & 0.07 {\tiny[0.05, 0.09]} & 0.38 {\tiny[0.27, 0.48]} & 0.15 {\tiny[0.12, 0.18]} & 0.21 {\tiny[0.13, 0.28]} & 0.08 {\tiny[0.06, 0.09]} & 0.39 {\tiny[0.30, 0.47]} & 0.19 {\tiny[0.16, 0.23]} \\
\midrule
\multirow{3}{*}{\shortstack[l]{LightNav-0\\{\tiny per-frame ViT}}} & MP3D & 0.32 {\tiny[0.24, 0.38]} & 0.15 {\tiny[0.12, 0.18]} & 0.35 {\tiny[0.26, 0.43]} & 0.17 {\tiny[0.14, 0.20]} & 0.28 {\tiny[0.22, 0.33]} & 0.18 {\tiny[0.16, 0.21]} & 0.38 {\tiny[0.32, 0.44]} & 0.21 {\tiny[0.19, 0.23]} \\
 & HM3D & 0.27 {\tiny[0.20, 0.33]} & 0.18 {\tiny[0.15, 0.21]} & 0.35 {\tiny[0.27, 0.43]} & 0.21 {\tiny[0.18, 0.23]} & 0.32 {\tiny[0.25, 0.40]} & 0.17 {\tiny[0.14, 0.19]} & 0.39 {\tiny[0.32, 0.45]} & 0.21 {\tiny[0.18, 0.24]} \\
 & HSSD & 0.12 {\tiny[0.04, 0.20]} & 0.06 {\tiny[0.04, 0.07]} & 0.06 {\tiny[$-$0.04, 0.16]} & 0.06 {\tiny[0.04, 0.07]} & 0.20 {\tiny[0.14, 0.27]} & 0.07 {\tiny[0.05, 0.09]} & 0.23 {\tiny[0.15, 0.29]} & 0.08 {\tiny[0.07, 0.10]} \\
\bottomrule
\end{tabular}
}
\end{table*}

\FloatBarrier
\subsection{Coarse Visual History}
\label{app:internals_grp_history}

\subsubsection{Compression and Projector Ablations}
\label{app:internals_compression}

\paragraph{Main-text plot and ladders.}
The main-text plot (\cref{fig:internals_history_compression}) averages, with equal weights, the lossless
collections of MP3D, HM3D and HSSD (per dataset and their mean: \cref{tab:app_history_compression_xd}),
and shows only the two models that pool their history. The complete three-model ladder below, Uni-NaVid's finer EVA-CLIP /
pixel ladder and the projector control (\cref{app:internals_finer,app:internals_projector}) run on each dataset; they are not points of that plot.

\begin{table*}[tbp]
\centering
\footnotesize
\setlength{\tabcolsep}{3.5pt}
\caption{Cross-model compression ladder on lossless recordings (101 MP3D, 103 HM3D and 101 HSSD episodes): held-out clean $R^2$ and comparison
readout $b$ [95\% CI]. Uni-NaVid: EVA-CLIP final layer; OmTrackVLA: its concatenated DINOv3 + SigLIP features;
LightNav-0: its per-frame vision features. $^\dagger$Jointly encoded pair, read by the full-channel standard probe (a
different probe; fails the shuffle control for forward motion, see text).}
\label{tab:app_crossmodel_sweep}
\begin{tabular}{llccccc}
\toprule
Model & Rung & Clean $R^2$ fwd / lat & $b$ current fwd & $b$ current lat & $b$ gust fwd & $b$ gust lat \\
\midrule
\multicolumn{7}{l}{\emph{MP3D}} \\
Uni-NaVid & native 16$^2$ & 0.89 / 0.61 & 0.54 {\tiny[0.44, 0.64]} & 0.35 {\tiny[0.31, 0.41]} & 0.55 {\tiny[0.48, 0.61]} & 0.40 {\tiny[0.37, 0.42]} \\
 & both 8$^2$ & 0.81 / 0.53 & 0.37 {\tiny[0.29, 0.44]} & 0.24 {\tiny[0.21, 0.28]} & 0.42 {\tiny[0.36, 0.47]} & 0.27 {\tiny[0.25, 0.29]} \\
 & kept: 2$^2$ + 8$^2$ & 0.24 / 0.14 & 0.05 {\tiny[$-$0.03, 0.13]} & 0.09 {\tiny[0.06, 0.13]} & 0.10 {\tiny[0.05, 0.14]} & 0.06 {\tiny[0.05, 0.07]} \\
 & after projector & 0.15 / 0.09 & $-$0.01 {\tiny[$-$0.08, 0.05]} & 0.06 {\tiny[0.05, 0.07]} & 0.05 {\tiny[0.01, 0.09]} & 0.05 {\tiny[0.04, 0.06]} \\
\addlinespace
OmTrackVLA & native 24$^2$ & 0.94 / 0.68 & 0.65 {\tiny[0.53, 0.77]} & 0.38 {\tiny[0.35, 0.42]} & 0.60 {\tiny[0.53, 0.67]} & 0.44 {\tiny[0.42, 0.47]} \\
 & both 8$^2$ & 0.86 / 0.55 & 0.49 {\tiny[0.38, 0.63]} & 0.26 {\tiny[0.22, 0.30]} & 0.46 {\tiny[0.40, 0.51]} & 0.29 {\tiny[0.27, 0.31]} \\
 & kept: 2$^2$ + 8$^2$ & 0.33 / 0.11 & 0.09 {\tiny[0.00, 0.17]} & 0.08 {\tiny[0.05, 0.13]} & 0.16 {\tiny[0.11, 0.22]} & 0.07 {\tiny[0.06, 0.08]} \\
 & after projector & 0.41 / 0.17 & 0.11 {\tiny[0.03, 0.20]} & 0.08 {\tiny[0.06, 0.09]} & 0.17 {\tiny[0.11, 0.23]} & 0.09 {\tiny[0.07, 0.10]} \\
\addlinespace
LightNav-0 & native 16$\times$28 & 0.78 / 0.43 & 0.35 {\tiny[0.26, 0.43]} & 0.17 {\tiny[0.14, 0.20]} & 0.38 {\tiny[0.32, 0.44]} & 0.21 {\tiny[0.19, 0.23]} \\
 & both 8$\times$14 & 0.87 / 0.62 & 0.46 {\tiny[0.37, 0.56]} & 0.29 {\tiny[0.26, 0.33]} & 0.43 {\tiny[0.37, 0.48]} & 0.35 {\tiny[0.32, 0.37]} \\
 & joint pair$^\dagger$ & 0.88 / 0.25 & 0.20 {\tiny[0.07, 0.33]} & 0.04 {\tiny[0.02, 0.07]} & 0.25 {\tiny[0.17, 0.32]} & 0.07 {\tiny[0.05, 0.09]} \\
\midrule
\multicolumn{7}{l}{\emph{HM3D}} \\
Uni-NaVid & native 16$^2$ & 0.89 / 0.65 & 0.55 {\tiny[0.49, 0.62]} & 0.36 {\tiny[0.32, 0.39]} & 0.58 {\tiny[0.52, 0.64]} & 0.38 {\tiny[0.34, 0.41]} \\
 & both 8$^2$ & 0.81 / 0.54 & 0.47 {\tiny[0.40, 0.53]} & 0.28 {\tiny[0.25, 0.31]} & 0.51 {\tiny[0.45, 0.56]} & 0.28 {\tiny[0.26, 0.31]} \\
 & kept: 2$^2$ + 8$^2$ & 0.22 / 0.17 & 0.09 {\tiny[0.04, 0.14]} & 0.09 {\tiny[0.08, 0.11]} & 0.17 {\tiny[0.11, 0.22]} & 0.09 {\tiny[0.08, 0.11]} \\
 & after projector & 0.17 / 0.11 & 0.02 {\tiny[$-$0.02, 0.08]} & 0.09 {\tiny[0.08, 0.11]} & 0.08 {\tiny[0.02, 0.12]} & 0.08 {\tiny[0.07, 0.09]} \\
\addlinespace
OmTrackVLA & native 24$^2$ & 0.94 / 0.70 & 0.62 {\tiny[0.55, 0.69]} & 0.39 {\tiny[0.35, 0.42]} & 0.62 {\tiny[0.56, 0.69]} & 0.40 {\tiny[0.37, 0.43]} \\
 & both 8$^2$ & 0.86 / 0.58 & 0.51 {\tiny[0.44, 0.59]} & 0.28 {\tiny[0.25, 0.31]} & 0.54 {\tiny[0.48, 0.60]} & 0.28 {\tiny[0.26, 0.31]} \\
 & kept: 2$^2$ + 8$^2$ & 0.23 / 0.14 & 0.12 {\tiny[0.04, 0.20]} & 0.09 {\tiny[0.08, 0.11]} & 0.21 {\tiny[0.15, 0.27]} & 0.08 {\tiny[0.07, 0.09]} \\
 & after projector & 0.39 / 0.19 & 0.12 {\tiny[0.05, 0.18]} & 0.11 {\tiny[0.09, 0.13]} & 0.22 {\tiny[0.17, 0.28]} & 0.11 {\tiny[0.09, 0.13]} \\
\addlinespace
LightNav-0 & native 16$\times$28 & 0.72 / 0.40 & 0.35 {\tiny[0.27, 0.43]} & 0.21 {\tiny[0.18, 0.23]} & 0.39 {\tiny[0.32, 0.45]} & 0.21 {\tiny[0.18, 0.24]} \\
 & both 8$\times$14 & 0.87 / 0.65 & 0.52 {\tiny[0.44, 0.59]} & 0.35 {\tiny[0.32, 0.38]} & 0.55 {\tiny[0.48, 0.61]} & 0.37 {\tiny[0.34, 0.40]} \\
 & joint pair$^\dagger$ & 0.88 / 0.24 & 0.30 {\tiny[0.22, 0.38]} & 0.08 {\tiny[0.05, 0.10]} & 0.29 {\tiny[0.21, 0.37]} & 0.06 {\tiny[0.04, 0.08]} \\
\midrule
\multicolumn{7}{l}{\emph{HSSD}} \\
Uni-NaVid & native 16$^2$ & 0.81 / 0.59 & 0.28 {\tiny[0.17, 0.38]} & 0.13 {\tiny[0.11, 0.16]} & 0.31 {\tiny[0.24, 0.38]} & 0.18 {\tiny[0.15, 0.22]} \\
 & both 8$^2$ & 0.71 / 0.53 & 0.21 {\tiny[0.09, 0.32]} & 0.10 {\tiny[0.07, 0.12]} & 0.28 {\tiny[0.21, 0.34]} & 0.14 {\tiny[0.12, 0.17]} \\
 & kept: 2$^2$ + 8$^2$ & 0.21 / 0.16 & $-$0.06 {\tiny[$-$0.15, 0.03]} & 0.04 {\tiny[0.02, 0.06]} & 0.07 {\tiny[0.01, 0.13]} & 0.07 {\tiny[0.05, 0.08]} \\
 & after projector & 0.19 / 0.11 & $-$0.18 {\tiny[$-$0.32, $-$0.05]} & 0.03 {\tiny[0.01, 0.05]} & 0.04 {\tiny[$-$0.03, 0.10]} & 0.06 {\tiny[0.05, 0.07]} \\
\addlinespace
OmTrackVLA & native 24$^2$ & 0.85 / 0.62 & 0.38 {\tiny[0.27, 0.48]} & 0.15 {\tiny[0.12, 0.18]} & 0.39 {\tiny[0.30, 0.47]} & 0.19 {\tiny[0.16, 0.23]} \\
 & both 8$^2$ & 0.73 / 0.53 & 0.23 {\tiny[0.12, 0.35]} & 0.10 {\tiny[0.08, 0.13]} & 0.30 {\tiny[0.24, 0.36]} & 0.14 {\tiny[0.11, 0.16]} \\
 & kept: 2$^2$ + 8$^2$ & 0.29 / 0.20 & $-$0.02 {\tiny[$-$0.12, 0.08]} & 0.04 {\tiny[0.02, 0.06]} & 0.15 {\tiny[0.08, 0.20]} & 0.05 {\tiny[0.04, 0.06]} \\
 & after projector & 0.43 / 0.20 & 0.10 {\tiny[0.02, 0.18]} & 0.04 {\tiny[0.02, 0.06]} & 0.17 {\tiny[0.09, 0.26]} & 0.06 {\tiny[0.04, 0.07]} \\
\addlinespace
LightNav-0 & native 16$\times$28 & 0.64 / 0.38 & 0.06 {\tiny[$-$0.04, 0.16]} & 0.06 {\tiny[0.04, 0.07]} & 0.23 {\tiny[0.15, 0.29]} & 0.08 {\tiny[0.07, 0.10]} \\
 & both 8$\times$14 & 0.78 / 0.57 & 0.21 {\tiny[0.08, 0.34]} & 0.13 {\tiny[0.11, 0.16]} & 0.31 {\tiny[0.23, 0.38]} & 0.17 {\tiny[0.15, 0.20]} \\
 & joint pair$^\dagger$ & 0.77 / 0.19 & 0.22 {\tiny[0.14, 0.31]} & 0.00 {\tiny[$-$0.01, 0.01]} & 0.14 {\tiny[0.07, 0.21]} & 0.00 {\tiny[$-$0.01, 0.01]} \\
\bottomrule
\end{tabular}
\end{table*}

\begin{table*}[tbp]
\centering
\footnotesize
\setlength{\tabcolsep}{4pt}
\caption{History compression on each dataset and the equal-weight mean of the three, behind \cref{fig:internals_history_compression}: comparison readout $b$ [95\% CI] along the ladder both frames at $8\times8$, the previous frame at $2\times2$ (as the model keeps it) and after the projector. Per dataset: episode bootstrap (1{,}000 resamples). Mean intervals: $\pm1.96\sqrt{\sum_i \mathrm{SE}_i^2}/3$ from the per-dataset intervals (\cref{app:internals_data}). Bold: the equal-weight mean.}
\label{tab:app_history_compression_xd}
\begin{tabular}{@{}lllcccc@{}}
\toprule
Model & Disturbance & Axis & Dataset & $8^2/8^2$ & $2^2/8^2$ & after proj. \\
\midrule
Uni-NaVid & current & fwd & MP3D & 0.37 {\tiny[0.29, 0.44]} & 0.05 {\tiny[$-$0.03, 0.13]} & $-$0.01 {\tiny[$-$0.08, 0.05]} \\
 & &  & HM3D & 0.47 {\tiny[0.40, 0.53]} & 0.09 {\tiny[0.04, 0.14]} & 0.02 {\tiny[$-$0.02, 0.08]} \\
 & &  & HSSD & 0.21 {\tiny[0.09, 0.32]} & $-$0.06 {\tiny[$-$0.15, 0.03]} & $-$0.18 {\tiny[$-$0.32, $-$0.05]} \\
 & &  & \textbf{Mean} & \textbf{0.35 {\tiny[0.30, 0.40]}} & \textbf{0.03 {\tiny[$-$0.02, 0.07]}} & \textbf{$-$0.06 {\tiny[$-$0.11, 0.00]}} \\
\addlinespace
Uni-NaVid & current & lat & MP3D & 0.24 {\tiny[0.21, 0.28]} & 0.09 {\tiny[0.06, 0.13]} & 0.06 {\tiny[0.05, 0.07]} \\
 & &  & HM3D & 0.28 {\tiny[0.25, 0.31]} & 0.09 {\tiny[0.08, 0.11]} & 0.09 {\tiny[0.08, 0.11]} \\
 & &  & HSSD & 0.10 {\tiny[0.07, 0.12]} & 0.04 {\tiny[0.02, 0.06]} & 0.03 {\tiny[0.01, 0.05]} \\
 & &  & \textbf{Mean} & \textbf{0.21 {\tiny[0.19, 0.22]}} & \textbf{0.07 {\tiny[0.06, 0.09]}} & \textbf{0.06 {\tiny[0.05, 0.07]}} \\
\addlinespace
Uni-NaVid & gust & fwd & MP3D & 0.42 {\tiny[0.36, 0.47]} & 0.10 {\tiny[0.05, 0.14]} & 0.05 {\tiny[0.01, 0.09]} \\
 & &  & HM3D & 0.51 {\tiny[0.45, 0.56]} & 0.17 {\tiny[0.11, 0.22]} & 0.08 {\tiny[0.02, 0.12]} \\
 & &  & HSSD & 0.28 {\tiny[0.21, 0.34]} & 0.07 {\tiny[0.01, 0.13]} & 0.04 {\tiny[$-$0.03, 0.10]} \\
 & &  & \textbf{Mean} & \textbf{0.40 {\tiny[0.37, 0.43]}} & \textbf{0.11 {\tiny[0.08, 0.14]}} & \textbf{0.06 {\tiny[0.03, 0.09]}} \\
\addlinespace
Uni-NaVid & gust & lat & MP3D & 0.27 {\tiny[0.25, 0.29]} & 0.06 {\tiny[0.05, 0.07]} & 0.05 {\tiny[0.04, 0.06]} \\
 & &  & HM3D & 0.28 {\tiny[0.26, 0.31]} & 0.09 {\tiny[0.08, 0.11]} & 0.08 {\tiny[0.07, 0.09]} \\
 & &  & HSSD & 0.14 {\tiny[0.12, 0.17]} & 0.07 {\tiny[0.05, 0.08]} & 0.06 {\tiny[0.05, 0.07]} \\
 & &  & \textbf{Mean} & \textbf{0.23 {\tiny[0.22, 0.25]}} & \textbf{0.07 {\tiny[0.07, 0.08]}} & \textbf{0.06 {\tiny[0.06, 0.07]}} \\
\midrule
OmTrackVLA & current & fwd & MP3D & 0.49 {\tiny[0.38, 0.63]} & 0.09 {\tiny[0.00, 0.17]} & 0.11 {\tiny[0.03, 0.20]} \\
 & &  & HM3D & 0.51 {\tiny[0.44, 0.59]} & 0.12 {\tiny[0.04, 0.20]} & 0.12 {\tiny[0.05, 0.18]} \\
 & &  & HSSD & 0.23 {\tiny[0.12, 0.35]} & $-$0.02 {\tiny[$-$0.12, 0.08]} & 0.10 {\tiny[0.02, 0.18]} \\
 & &  & \textbf{Mean} & \textbf{0.41 {\tiny[0.35, 0.47]}} & \textbf{0.06 {\tiny[0.01, 0.12]}} & \textbf{0.11 {\tiny[0.07, 0.15]}} \\
\addlinespace
OmTrackVLA & current & lat & MP3D & 0.26 {\tiny[0.22, 0.30]} & 0.08 {\tiny[0.05, 0.13]} & 0.08 {\tiny[0.06, 0.09]} \\
 & &  & HM3D & 0.28 {\tiny[0.25, 0.31]} & 0.09 {\tiny[0.08, 0.11]} & 0.11 {\tiny[0.09, 0.13]} \\
 & &  & HSSD & 0.10 {\tiny[0.08, 0.13]} & 0.04 {\tiny[0.02, 0.06]} & 0.04 {\tiny[0.02, 0.06]} \\
 & &  & \textbf{Mean} & \textbf{0.21 {\tiny[0.20, 0.23]}} & \textbf{0.07 {\tiny[0.06, 0.09]}} & \textbf{0.08 {\tiny[0.07, 0.09]}} \\
\addlinespace
OmTrackVLA & gust & fwd & MP3D & 0.46 {\tiny[0.40, 0.51]} & 0.16 {\tiny[0.11, 0.22]} & 0.17 {\tiny[0.11, 0.23]} \\
 & &  & HM3D & 0.54 {\tiny[0.48, 0.60]} & 0.21 {\tiny[0.15, 0.27]} & 0.22 {\tiny[0.17, 0.28]} \\
 & &  & HSSD & 0.30 {\tiny[0.24, 0.36]} & 0.15 {\tiny[0.08, 0.20]} & 0.17 {\tiny[0.09, 0.26]} \\
 & &  & \textbf{Mean} & \textbf{0.43 {\tiny[0.40, 0.47]}} & \textbf{0.17 {\tiny[0.14, 0.20]}} & \textbf{0.19 {\tiny[0.15, 0.23]}} \\
\addlinespace
OmTrackVLA & gust & lat & MP3D & 0.29 {\tiny[0.27, 0.31]} & 0.07 {\tiny[0.06, 0.08]} & 0.09 {\tiny[0.07, 0.10]} \\
 & &  & HM3D & 0.28 {\tiny[0.26, 0.31]} & 0.08 {\tiny[0.07, 0.09]} & 0.11 {\tiny[0.09, 0.13]} \\
 & &  & HSSD & 0.14 {\tiny[0.11, 0.16]} & 0.05 {\tiny[0.04, 0.06]} & 0.06 {\tiny[0.04, 0.07]} \\
 & &  & \textbf{Mean} & \textbf{0.24 {\tiny[0.22, 0.25]}} & \textbf{0.07 {\tiny[0.06, 0.07]}} & \textbf{0.08 {\tiny[0.08, 0.09]}} \\
\bottomrule
\end{tabular}
\end{table*}

\begin{table*}[tbp]
\centering
\footnotesize
\setlength{\tabcolsep}{2pt}
\caption{Paired differences along the history ladder of \cref{tab:app_history_compression_xd}: comparison readout $b$ of
the previous frame at $2\times2$, and after the projector, minus both frames at $8\times8$, on the same held-out rows;
equal-weight mean of MP3D, HM3D and HSSD [95\% CI]. Per dataset: paired episode bootstrap refitting $b$ (5{,}000
resamples, seed 0); mean intervals as in \cref{tab:app_history_compression_xd}.}
\label{tab:app_history_paired}
\begin{tabular}{@{}lllcc@{}}
\toprule
Model & Dist. & Axis & $2^2/8^2 - 8^2/8^2$ & after proj. $- 8^2/8^2$ \\
\midrule
Uni-NaVid & current & fwd & $-0.32$ {\tiny[$-0.37$, $-0.27$]} & $-0.40$ {\tiny[$-0.45$, $-0.36$]} \\
 & current & lat & $-0.14$ {\tiny[$-0.15$, $-0.12$]} & $-0.15$ {\tiny[$-0.16$, $-0.13$]} \\
 & gust & fwd & $-0.29$ {\tiny[$-0.32$, $-0.25$]} & $-0.34$ {\tiny[$-0.38$, $-0.31$]} \\
 & gust & lat & $-0.16$ {\tiny[$-0.17$, $-0.15$]} & $-0.17$ {\tiny[$-0.18$, $-0.16$]} \\
\addlinespace
OmTrackVLA & current & fwd & $-0.35$ {\tiny[$-0.40$, $-0.30$]} & $-0.30$ {\tiny[$-0.35$, $-0.25$]} \\
 & current & lat & $-0.14$ {\tiny[$-0.16$, $-0.12$]} & $-0.14$ {\tiny[$-0.16$, $-0.12$]} \\
 & gust & fwd & $-0.26$ {\tiny[$-0.29$, $-0.23$]} & $-0.25$ {\tiny[$-0.28$, $-0.21$]} \\
 & gust & lat & $-0.17$ {\tiny[$-0.18$, $-0.16$]} & $-0.15$ {\tiny[$-0.17$, $-0.14$]} \\
\bottomrule
\end{tabular}
\end{table*}

\paragraph{Compression and projector ablations for the three models, on lossless recordings.}
The grid ladder runs for every model on its own encoder and grids, on lossless (PNG) recordings: every rung
on 101 MP3D, 103 HM3D and 101 HSSD episodes (Table~\ref{tab:app_crossmodel_sweep}), and the history rungs of Uni-NaVid and OmTrackVLA per dataset with their mean
(\cref{tab:app_history_compression_xd}). The rungs are: both frames at the encoder's native grid; both at the
model's current-frame grid; the previous frame as the model keeps it next to the current frame as it keeps it (the
compression ablation: against the previous rung, only the previous frame's grid changes); and, where the model has
one, after the projector (the projector ablation). \Cref{tab:app_history_paired} pairs both ablations with both frames
at $8\times8$ on the same held-out rows; per dataset, all 48 paired differences (two models, two disturbances, two
axes, two ablations, three datasets) have intervals below zero (not tabulated). As a control, within-episode shuffling
(\cref{app:internals_shuffle}) collapses OmTrackVLA's comparison readouts at every recorded depth on every dataset (clean forward
$R^2\le0.13$).

\paragraph{LightNav-0's jointly encoded pair.}
LightNav-0's Qwen3-VL encoder merges frames $t{-}1$ and $t$ into one spatio-temporal patch at the full merged grid, so
its immediate predecessor is not compressed relative to the current frame and there is no previous-frame rung to
ablate. Its rungs (Table~\ref{tab:app_crossmodel_sweep}) are its per-frame encoder features at the native
$16\times28$ grid and at $8\times14$, read by the comparison probe, and the jointly encoded pair, which has no
per-frame tokens and is read by the full-channel standard probe. The joint pair's forward readout fails the shuffle
control: replacing the partner frame of the patch leaves clean forward $R^2$ at $0.47$ (intact $0.88$) on MP3D, $0.43$
($0.88$) on HM3D and $0.34$ ($0.77$) on HSSD.

\subsubsection{Real Video: TUM, KITTI, TartanDrive}
\label{app:internals_real}

\paragraph{Data and independent grouping.}
The real-video readouts use PNG images decoded from the raw sources.
TUM Pioneer contains 134 episodes and 8,197 usable adjacent pairs in 12 sequence/time blocks; KITTI sequences 00--10
contain 234 episodes and 22,728 pairs grouped by 11 drives; TartanDrive contains 144 episodes and 10,753 pairs from
40 recordings across eight days. All three models use the same held-out targets and episode ordering within each
dataset. Episode counts are not counts of independent
environments. TUM and KITTI lack issued commands: these readouts measure total camera motion, not isolated external
disturbances.

\begin{table*}[tbp]
\centering
\footnotesize
\setlength{\tabcolsep}{2.5pt}
\caption{Uni-NaVid on full lossless real-video inputs. Fixed visual sites; previous/current grids are shown before and after the projector. All three motion axes and within-episode shuffled-previous-frame yaw controls are retained. Values: held-out $R^2$ [pointwise $95\%$ episode-bootstrap interval]. Full grid means the encoder\textquotesingle s native pre-pooling grid; it is not standardized across models.}
\label{tab:app_real_compression}
\begin{tabular}{@{}llcccc@{}}
\toprule
Dataset & Visual input & Forward $R^2$ & Lateral $R^2$ & Yaw $R^2$ & Shuffled yaw $R^2$ \\
\midrule
TUM & Full grid & 0.03 [0.00, 0.05] & 0.04 [0.00, 0.45] & 0.65 [0.49, 0.83] & $-$0.01 [$-$0.03, 0.00] \\
TUM & $8^2/8^2$ & $-$0.03 [$-$0.08, 0.01] & 0.01 [$-$0.03, 0.18] & 0.60 [0.45, 0.77] & $-$0.02 [$-$0.04, $-$0.01] \\
TUM & $2^2/8^2$ & 0.00 [$-$0.04, 0.04] & 0.01 [$-$0.01, 0.04] & 0.22 [0.17, 0.27] & $-$0.02 [$-$0.03, 0.00] \\
TUM & $2^2/8^2$ + proj. & $-$0.09 [$-$0.15, $-$0.05] & $-$0.03 [$-$0.22, $-$0.01] & 0.11 [0.07, 0.15] & $-$0.01 [$-$0.03, 0.00] \\
KITTI & Full grid & 0.30 [0.24, 0.37] & 0.64 [0.59, 0.69] & 0.95 [0.95, 0.96] & 0.01 [$-$0.04, 0.06] \\
KITTI & $8^2/8^2$ & 0.22 [0.17, 0.29] & 0.64 [0.58, 0.68] & 0.94 [0.92, 0.94] & 0.03 [$-$0.03, 0.09] \\
KITTI & $2^2/8^2$ & $-$0.03 [$-$0.10, 0.03] & 0.19 [0.13, 0.24] & 0.30 [0.22, 0.36] & 0.07 [0.02, 0.11] \\
KITTI & $2^2/8^2$ + proj. & $-$0.02 [$-$0.09, 0.03] & 0.03 [$-$0.01, 0.06] & 0.05 [0.00, 0.09] & 0.02 [$-$0.01, 0.05] \\
TartanDrive & Full grid & 0.38 [0.33, 0.46] & 0.36 [0.31, 0.40] & 0.64 [0.56, 0.73] & $-$0.08 [$-$0.13, $-$0.06] \\
TartanDrive & $8^2/8^2$ & 0.33 [0.29, 0.39] & 0.15 [0.09, 0.19] & 0.14 [0.01, 0.24] & $-$0.08 [$-$0.13, $-$0.05] \\
TartanDrive & $2^2/8^2$ & 0.01 [$-$0.04, 0.05] & $-$0.07 [$-$0.12, $-$0.04] & $-$0.07 [$-$0.13, $-$0.04] & $-$0.07 [$-$0.12, $-$0.05] \\
TartanDrive & $2^2/8^2$ + proj. & $-$0.04 [$-$0.09, $-$0.01] & $-$0.07 [$-$0.11, $-$0.05] & $-$0.06 [$-$0.10, $-$0.04] & $-$0.05 [$-$0.09, $-$0.03] \\
\bottomrule
\end{tabular}
\end{table*}

\begin{table*}[tbp]
\centering
\footnotesize
\setlength{\tabcolsep}{2.5pt}
\caption{OmTrackVLA on full lossless real-video inputs. Fixed visual sites; previous/current grids are shown before and after the projector. All three motion axes and within-episode shuffled-previous-frame yaw controls are retained. Values: held-out $R^2$ [pointwise $95\%$ episode-bootstrap interval]. Full grid means the encoder\textquotesingle s native pre-pooling grid; it is not standardized across models.}
\label{tab:app_real_omtrack}
\begin{tabular}{@{}llcccc@{}}
\toprule
Dataset & Visual input & Forward $R^2$ & Lateral $R^2$ & Yaw $R^2$ & Shuffled yaw $R^2$ \\
\midrule
TUM & Full grid & 0.05 [0.02, 0.08] & 0.03 [0.00, 0.42] & 0.65 [0.49, 0.83] & $-$0.02 [$-$0.04, $-$0.01] \\
TUM & $8^2/8^2$ & $-$0.01 [$-$0.06, 0.02] & 0.03 [0.00, 0.27] & 0.63 [0.50, 0.78] & $-$0.05 [$-$0.09, $-$0.03] \\
TUM & $2^2/8^2$ & $-$0.02 [$-$0.07, 0.01] & $-$0.02 [$-$0.15, 0.00] & 0.32 [0.25, 0.40] & $-$0.02 [$-$0.04, $-$0.01] \\
TUM & $2^2/8^2$ + proj. & 0.01 [$-$0.03, 0.05] & 0.01 [0.00, 0.14] & 0.31 [0.23, 0.40] & $-$0.06 [$-$0.09, $-$0.04] \\
KITTI & Full grid & 0.47 [0.39, 0.57] & 0.61 [0.55, 0.66] & 0.95 [0.93, 0.95] & 0.07 [0.00, 0.13] \\
KITTI & $8^2/8^2$ & 0.32 [0.24, 0.42] & 0.57 [0.52, 0.61] & 0.86 [0.84, 0.88] & 0.15 [0.08, 0.23] \\
KITTI & $2^2/8^2$ & 0.08 [0.03, 0.13] & 0.13 [0.09, 0.16] & 0.20 [0.16, 0.24] & 0.06 [0.01, 0.11] \\
KITTI & $2^2/8^2$ + proj. & 0.11 [0.02, 0.17] & 0.12 [0.08, 0.15] & 0.19 [0.14, 0.23] & 0.07 [0.02, 0.11] \\
TartanDrive & Full grid & 0.44 [0.40, 0.51] & 0.39 [0.34, 0.44] & 0.61 [0.53, 0.70] & $-$0.12 [$-$0.19, $-$0.07] \\
TartanDrive & $8^2/8^2$ & 0.26 [0.22, 0.31] & 0.08 [0.02, 0.12] & 0.01 [$-$0.11, 0.09] & $-$0.10 [$-$0.20, $-$0.05] \\
TartanDrive & $2^2/8^2$ & $-$0.01 [$-$0.06, 0.02] & $-$0.05 [$-$0.11, $-$0.01] & $-$0.11 [$-$0.19, $-$0.07] & $-$0.14 [$-$0.25, $-$0.09] \\
TartanDrive & $2^2/8^2$ + proj. & $-$0.02 [$-$0.09, 0.02] & $-$0.06 [$-$0.11, $-$0.03] & $-$0.05 [$-$0.10, $-$0.02] & $-$0.12 [$-$0.20, $-$0.07] \\
\bottomrule
\end{tabular}
\end{table*}

\begin{table*}[tbp]
\centering
\footnotesize
\setlength{\tabcolsep}{2.5pt}
\caption{LightNav-0 real-video controls, with both full-dimensional joint probes (CNN and exact ridge) on the native joint grid ($8\times14$). The separate-frame comparison control repeats each frame through the visual path; it is not an upstream state of the native joint-pair run. Values: $R^2$ [pointwise $95\%$ interval].}
\label{tab:app_real_lightnav}
\begin{tabular}{@{}lllcccc@{}}
\toprule
Dataset & Visual input & Probe & Forward $R^2$ & Lateral $R^2$ & Yaw $R^2$ & Shuffled yaw $R^2$ \\
\midrule
TUM & Separate frames & Comparison & 0.01 [$-$0.03, 0.05] & 0.05 [0.01, 0.48] & 0.67 [0.52, 0.83] & $-$0.02 [$-$0.04, $-$0.01] \\
TUM & Joint $8\times14$ & CNN & 0.02 [$-$0.02, 0.06] & $-$0.01 [$-$0.09, 0.00] & 0.04 [$-$0.04, 0.09] & $-$0.17 [$-$0.30, $-$0.08] \\
TUM & Joint $8\times14$ & Ridge & 0.05 [0.01, 0.09] & 0.00 [$-$0.01, 0.00] & 0.02 [0.00, 0.04] & 0.01 [$-$0.01, 0.03] \\
KITTI & Separate frames & Comparison & 0.31 [0.23, 0.41] & 0.59 [0.54, 0.64] & 0.91 [0.90, 0.93] & 0.12 [0.04, 0.19] \\
KITTI & Joint $8\times14$ & CNN & 0.35 [0.29, 0.42] & 0.20 [0.13, 0.26] & 0.41 [0.32, 0.49] & 0.29 [0.20, 0.37] \\
KITTI & Joint $8\times14$ & Ridge & 0.38 [0.31, 0.48] & 0.16 [0.11, 0.20] & 0.30 [0.24, 0.36] & 0.20 [0.14, 0.25] \\
TartanDrive & Separate frames & Comparison & 0.36 [0.31, 0.42] & 0.15 [0.10, 0.19] & 0.31 [0.21, 0.38] & $-$0.14 [$-$0.24, $-$0.08] \\
TartanDrive & Joint $8\times14$ & CNN & 0.17 [0.13, 0.22] & $-$0.03 [$-$0.07, 0.01] & 0.00 [$-$0.08, 0.05] & $-$0.40 [$-$0.69, $-$0.22] \\
TartanDrive & Joint $8\times14$ & Ridge & 0.18 [0.11, 0.24] & $-$0.01 [$-$0.03, 0.01] & 0.00 [$-$0.05, 0.04] & $-$0.04 [$-$0.07, $-$0.02] \\
\bottomrule
\end{tabular}
\end{table*}

\begin{table*}[tbp]
\centering
\footnotesize
\setlength{\tabcolsep}{2.5pt}
\caption{Paired differences from the same held-out predictions and episode resamples. Uni-NaVid and OmTrackVLA: encoder $8^2/8^2$ minus $2^2/8^2$, changing only the previous grid.}
\label{tab:app_real_paired}
\begin{tabular}{@{}llccc@{}}
\toprule
Model & Dataset & Forward $\Delta R^2$ & Lateral $\Delta R^2$ & Yaw $\Delta R^2$ \\
\midrule
Uni-NaVid & TUM & $-$0.03 [$-$0.07, 0.01] & 0.00 [$-$0.04, 0.15] & 0.38 [0.26, 0.52] \\
Uni-NaVid & KITTI & 0.25 [0.21, 0.32] & 0.45 [0.39, 0.51] & 0.64 [0.58, 0.71] \\
Uni-NaVid & TartanDrive & 0.32 [0.26, 0.39] & 0.22 [0.17, 0.26] & 0.21 [0.12, 0.30] \\
OmTrackVLA & TUM & 0.01 [$-$0.03, 0.05] & 0.04 [0.01, 0.42] & 0.31 [0.22, 0.42] \\
OmTrackVLA & KITTI & 0.24 [0.17, 0.34] & 0.44 [0.40, 0.48] & 0.66 [0.62, 0.71] \\
OmTrackVLA & TartanDrive & 0.27 [0.23, 0.34] & 0.13 [0.08, 0.18] & 0.13 [0.05, 0.21] \\
\bottomrule
\end{tabular}
\end{table*}

\paragraph{Fixed sites, full-dimensional probes and uncertainty.}
\Cref{tab:app_real_compression,tab:app_real_omtrack,tab:app_real_lightnav} fix Uni-NaVid's final EVA-CLIP block ($L_{39}$), OmTrackVLA's native concatenated
visual feature, and LightNav-0's merger or native joint pair. The previous-frame ablation is the one of the
simulated ladder ($8^2/8^2$ against $2^2/8^2$, then the projector); \cref{tab:app_real_paired} pairs the two encoder
rungs on the same held-out predictions. The joint LightNav-0
control additionally reports full-dimensional exact ridge, not only a CNN probe.
Intervals use a paired bootstrap over episodes (5,000 resamples, seed 0), recomputing $R^2$ from fixed held-out
predictions without refitting. They cover episode sampling given these sequences, fitted probes and splits, not
training-seed or new-drive uncertainty, and are pointwise, not multiplicity-corrected.

\FloatBarrier
\subsection{Rotation Readability Through the Backbone}
\label{app:internals_grp_backbone}
\label{app:internals_crossmodel}

\begin{figure}[tp]
 \centering
 \includegraphics[width=\linewidth,height=0.55\textheight,keepaspectratio]{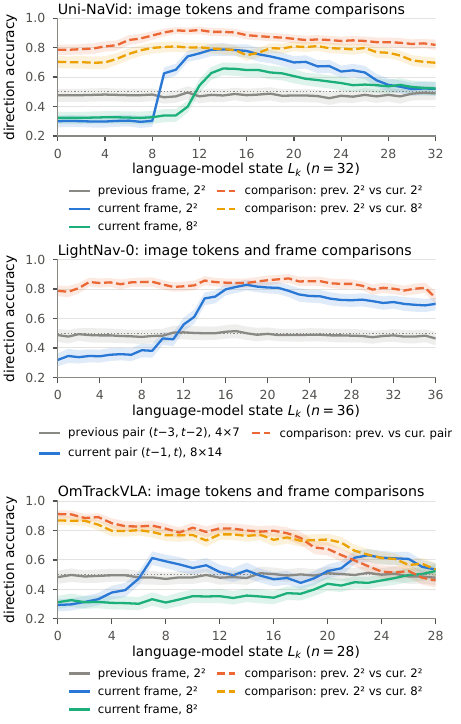}
 \caption{Rotation maps of the captured image-token groups (solid) and
 all-pairs frame comparisons (dashed) of the three models, on the same 486
 command-free MP3D events, across each model's own language-model states.
 Shading: $95\%$ episode-bootstrap intervals (2,000 resamples).}
 \label{fig:app_kickmaps}
\end{figure}

\begin{figure*}[tp]
 \centering
 \begin{minipage}[t]{0.48\textwidth}\centering {\small HM3D}\\[2pt]
 \includegraphics[width=\linewidth,height=0.55\textheight,keepaspectratio]{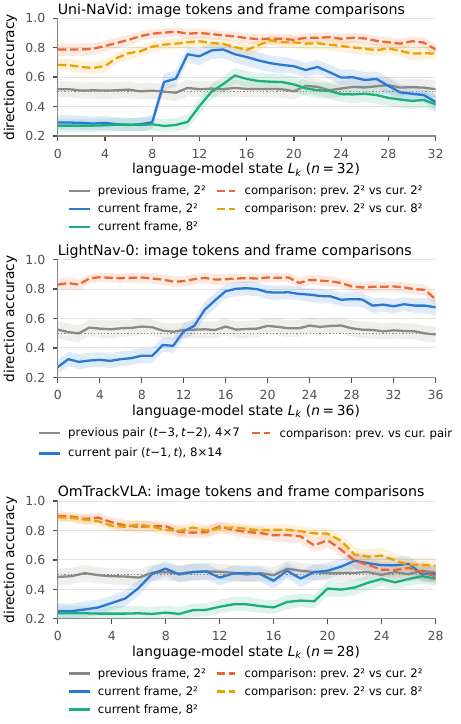}
 \end{minipage}\hfill
 \begin{minipage}[t]{0.48\textwidth}\centering {\small HSSD}\\[2pt]
 \includegraphics[width=\linewidth,height=0.55\textheight,keepaspectratio]{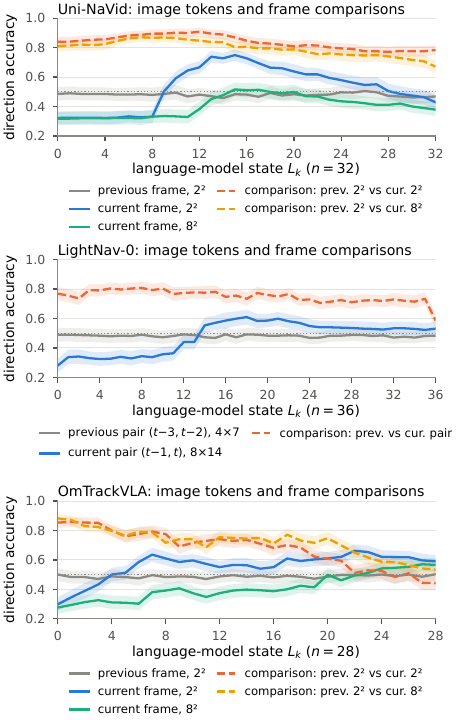}
 \end{minipage}
 \caption{The rotation maps of \cref{fig:app_kickmaps} on HM3D (429 command-free events) and HSSD (459), across each
 model's own language-model states. Shading: $95\%$ episode-bootstrap intervals (2,000 resamples).}
 \label{fig:app_kickmaps_xd}
\end{figure*}

The main-text map (\cref{fig:internals_rotation_backbone}) extracts the current visual group and decision state
from all-state maps recorded and read out identically on MP3D, HM3D and HSSD (\cref{tab:app_rotation_xd}); it
introduces no layer selection or smoothing. The complete per-state maps below (all token groups, comparisons and the
behavior readout) are recorded on all three datasets (\cref{fig:app_kickmaps}: MP3D; \cref{fig:app_kickmaps_xd}: HM3D and HSSD). The groups are
native to each architecture, not equal token budgets. These diagnostic rotations include increments of at least
$5^\circ$; they are not the benchmark's instant or smooth target-loss profiles. Decodability at the decision state is
therefore measured on these rotations, not on the target-loss events of \cref{app:internals_functional}. In OmTrackVLA
the decision-state readout lies below $0.5$: the side is decodable, with the opposite sign. The group curves below and
Table~\ref{tab:app_crossmodel} give the comparisons not in the main-text plot.

\begin{table*}[tbp]
\centering
\footnotesize
\caption{Command-free rotation-direction accuracy of the decision state on each dataset, behind
\cref{fig:internals_rotation_backbone}: at the dataset's own state $k^*$, the state farthest from chance on that
dataset's map, at MP3D's $k^*$ (the same layer on every dataset) and at the final state; 95\%
bootstrap intervals. 0.5 is chance; below 0.5 the
readout is reversed.}
\label{tab:app_rotation_xd}
\begin{tabular}{@{}llrccc@{}}
\toprule
Model & Dataset & Events & At own $k^*$ & At MP3D's $k^*$ & Final layer \\
\midrule
\multirow{3}{*}{Uni-NaVid} & MP3D & 486 & 0.741 [0.701, 0.778] ($L_{16}$) & 0.741 [0.701, 0.778] & 0.671 [0.626, 0.717] ($L_{32}$) \\
 & HM3D & 429 & 0.706 [0.645, 0.761] ($L_{21}$) & 0.667 [0.608, 0.724] & 0.611 [0.556, 0.665] ($L_{32}$) \\
 & HSSD & 459 & 0.634 [0.582, 0.684] ($L_{21}$) & 0.571 [0.517, 0.624] & 0.545 [0.500, 0.592] ($L_{32}$) \\
\midrule
\multirow{3}{*}{LightNav-0} & MP3D & 486 & 0.938 [0.914, 0.959] ($L_{20}$) & 0.938 [0.914, 0.959] & 0.702 [0.655, 0.745] ($L_{36}$) \\
 & HM3D & 429 & 0.935 [0.902, 0.964] ($L_{21}$) & 0.930 [0.887, 0.962] & 0.706 [0.659, 0.752] ($L_{36}$) \\
 & HSSD & 459 & 0.915 [0.885, 0.941] ($L_{21}$) & 0.911 [0.881, 0.939] & 0.590 [0.544, 0.637] ($L_{36}$) \\
\midrule
\multirow{3}{*}{OmTrackVLA} & MP3D & 486 & 0.342 [0.296, 0.391] ($L_{11}$) & 0.342 [0.296, 0.391] & 0.405 [0.358, 0.454] ($L_{28}$) \\
 & HM3D & 429 & 0.268 [0.220, 0.321] ($L_{19}$) & 0.303 [0.254, 0.354] & 0.399 [0.345, 0.452] ($L_{28}$) \\
 & HSSD & 459 & 0.331 [0.282, 0.380] ($L_{11}$) & 0.331 [0.282, 0.380] & 0.516 [0.471, 0.561] ($L_{28}$) \\
\bottomrule
\end{tabular}
\end{table*}

\paragraph{Replay and capture.}
The lossless Uni-NaVid rotation recordings (10 MP3D training scenes, 15
episodes each, PNG frames; HM3D and HSSD: \cref{tab:supp_flow}) are replayed through frozen LightNav-0 and
OmTrackVLA in order, with each episode's own
instruction. Each model reads
exactly the frames Uni-NaVid saw.
\Cref{app:internals_onpolicy} repeats the maps on each model's own
closed-loop trajectories. On the prefill of each selected step (the forward pass over the prompt, before any output token is generated), every token of a few declared
groups is stored at the language-model input and after every decoder layer, in half precision.
Table~\ref{tab:app_crossmodel_layout} lists the groups.

\begin{table}[tbp]
\centering
\footnotesize
\setlength{\tabcolsep}{3pt}
\caption{Captured token groups. ``Decision'' is the position the action is
computed from in each model.}
\label{tab:app_crossmodel_layout}
\begin{tabular}{@{}lp{5.0cm}@{}}
\toprule
Model & Captured groups (tokens) \\
\midrule
Uni-NaVid ($n=32$) & decision: last prompt position, from which the first action word is predicted (1); previous-frame memory, $2\times2$ (4); current-frame memory, $2\times2$ (4); current navigation, $8\times8$ (64) \\
LightNav-0 ($n=36$) & decision: last prompt position, from which the action tokens are generated (1); previous pair $(t{-}3,t{-}2)$ as the model keeps it, $4\times7$ (28); current pair $(t{-}1,t)$, $8\times14$ (112) \\
OmTrackVLA ($n=28$) & decision: the final act token read by the waypoint head (1); previous frame, $2\times2$ (4); current frame, $2\times2$ (4); current frame, $8\times8$ (64) \\
\bottomrule
\end{tabular}
\end{table}

\paragraph{Identical readout procedure.}
Events, folds and probes are the same as in the Uni-NaVid map:
on MP3D, 1,180 imposed rotations, of which 486 have no turn command at the preceding
decision, and clean fitting rows (9,267) from the same scenes (HM3D and HSSD: \cref{tab:supp_flow}); exact dual ridge on
every token group, with the ridge penalty $\alpha$ chosen by inner scene-grouped
cross-validation; accuracy, the mean over events of
\begin{equation}
 \mathbf{1}\bigl[\operatorname{sign}(\hat r-c)=\operatorname{sign}(\text{rotation})\bigr]
\end{equation}
for predicted turn $\hat r$ and commanded turn $c$; episode bootstrap with 2,000
resamples. Frame comparisons use all-pairs cosine similarities between the
previous and current groups. The only adaptation is the token layout, read from each
capture. The leak control reads the decision state one step before the
rotation, before any rotation has occurred; a reading above chance there would come from a pre-event correlate of the
rotation's side, not from the rotation.

\paragraph{The action readout differs by action interface.}
Uni-NaVid emits one of five discrete actions; its corrective preference compares
the log-probabilities of the corrective and opposite turns. LightNav-0 and
OmTrackVLA output continuous waypoints; we use the sign of the turn each model
sends (LightNav-0: the yaw of the first waypoint; OmTrackVLA: the angular
velocity of its agent command). The two quantities are not identical: a
probability preference can favor the corrective turn while the executed action
is ``forward'' (Uni-NaVid executes a corrective turn on only $23.5\%$ of events on MP3D, $26.6\%$ on HM3D and $35.3\%$ on HSSD), whereas a continuous turn has a sign on
nearly every event (LightNav-0 has $|\text{turn}|<10^{-6}$ on $4.5\%$, $4.0\%$ and $4.4\%$ of events,
OmTrackVLA on none). One LightNav-0 event on MP3D and one on HM3D have no finite turn output, giving
$n=485$ and $n=428$ for its behavior readout.

\begin{table*}[tbp]
\centering
\footnotesize
\setlength{\tabcolsep}{4pt}
\caption{Cross-model rotation maps on the command-free events of each dataset (MP3D 486, HM3D 429, HSSD 459). Accuracy
[95\% CI]; $L_k$ in each model's own depth. ``Farthest from chance'' is
$\arg\max_k|\text{acc}_k-0.5|$. Decision-state accuracy: \cref{tab:app_rotation_xd}. \emph{Turn sign against the
rotation}: share of events whose turn opposes the rotation; for Uni-NaVid the sign of its left-versus-right probability
preference, for the waypoint models the sign of the executed turn. Current-frame tokens at $L_0$ read below $0.5$ in
every cell: for Uni-NaVid and OmTrackVLA a single frame contains no frame-to-frame motion, so this readout reflects content of the current frame
that correlates with the rotation's side, with the opposite sign; LightNav-0's current pair also reads below $0.5$ at $L_{0}$.}
\label{tab:app_crossmodel}
\begin{tabular}{lrrr}
\toprule
 & Uni-NaVid & LightNav-0 & OmTrackVLA \\
\midrule
\multicolumn{4}{l}{\emph{MP3D}} \\
Comparison, prev.\ vs cur.\ (coarse), $L_0$ & 0.786 [0.752, 0.820] & 0.788 [0.752, 0.824] & 0.912 [0.881, 0.937] \\
Comparison, prev.\ vs cur.\ (coarse), final state & 0.819 [0.784, 0.855] & 0.741 [0.697, 0.784] & 0.461 [0.413, 0.505] \\
Previous-frame tokens, farthest from chance & 0.457 [0.411, 0.500] & 0.465 [0.419, 0.509] & 0.471 [0.429, 0.514] \\
Current-frame tokens, $L_0$ & 0.300 [0.259, 0.338] & 0.319 [0.275, 0.364] & 0.294 [0.251, 0.334] \\
Current-frame tokens, best state & 0.788 [0.745, 0.831] ($L_{14}$) & 0.829 [0.791, 0.863] ($L_{18}$) & 0.632 [0.587, 0.674] ($L_{23}$) \\
First state with current-frame CI above $0.5$ & $L_{9}$ ($k/n{=}0.28$) & $L_{12}$ (0.33) & $L_{7}$ (0.25) \\
All captured image tokens, best state & 0.786 [0.743, 0.827] ($L_{13}$) & 0.852 [0.814, 0.885] ($L_{18}$) & 0.574 [0.529, 0.619] ($L_{28}$) \\
Decision state, clean yaw $R^2$, best state & 0.49 ($L_{23}$) & 0.77 ($L_{20}$) & 0.24 ($L_{28}$) \\
Leak control, farthest from chance & 0.456 [0.408, 0.505] & 0.443 [0.400, 0.484] & 0.531 [0.481, 0.581] \\
Turn sign against the rotation & 57.8 [53.3, 62.4]\% & 66.4 [62.0, 71.0]\% & 72.4 [68.3, 76.4]\% \\
\midrule
\multicolumn{4}{l}{\emph{HM3D}} \\
Comparison, prev.\ vs cur.\ (coarse), $L_0$ & 0.788 [0.743, 0.829] & 0.830 [0.792, 0.868] & 0.900 [0.866, 0.932] \\
Comparison, prev.\ vs cur.\ (coarse), final state & 0.788 [0.747, 0.827] & 0.730 [0.688, 0.772] & 0.487 [0.436, 0.541] \\
Previous-frame tokens, farthest from chance & 0.545 [0.491, 0.600] & 0.552 [0.492, 0.608] & 0.541 [0.485, 0.595] \\
Current-frame tokens, $L_0$ & 0.291 [0.244, 0.342] & 0.270 [0.227, 0.319] & 0.252 [0.204, 0.307] \\
Current-frame tokens, best state & 0.788 [0.729, 0.837] ($L_{14}$) & 0.807 [0.757, 0.853] ($L_{18}$) & 0.594 [0.537, 0.650] ($L_{22}$) \\
First state with current-frame CI above $0.5$ & $L_{9}$ ($k/n{=}0.28$) & $L_{14}$ (0.39) & $L_{21}$ (0.75) \\
All captured image tokens, best state & 0.783 [0.727, 0.839] ($L_{13}$) & 0.837 [0.796, 0.878] ($L_{19}$) & 0.580 [0.526, 0.633] ($L_{26}$) \\
Decision state, clean yaw $R^2$, best state & 0.49 ($L_{21}$) & 0.81 ($L_{20}$) & 0.29 ($L_{27}$) \\
Leak control, farthest from chance & 0.563 [0.514, 0.612] & 0.555 [0.509, 0.602] & 0.560 [0.508, 0.613] \\
Turn sign against the rotation & 58.0 [52.5, 63.3]\% & 64.3 [59.1, 69.4]\% & 70.4 [65.1, 75.2]\% \\
\midrule
\multicolumn{4}{l}{\emph{HSSD}} \\
Comparison, prev.\ vs cur.\ (coarse), $L_0$ & 0.841 [0.805, 0.873] & 0.769 [0.725, 0.812] & 0.854 [0.818, 0.887] \\
Comparison, prev.\ vs cur.\ (coarse), final state & 0.784 [0.749, 0.817] & 0.586 [0.537, 0.635] & 0.442 [0.395, 0.492] \\
Previous-frame tokens, farthest from chance & 0.458 [0.416, 0.499] & 0.468 [0.425, 0.510] & 0.471 [0.426, 0.513] \\
Current-frame tokens, $L_0$ & 0.322 [0.280, 0.365] & 0.281 [0.236, 0.326] & 0.298 [0.254, 0.347] \\
Current-frame tokens, best state & 0.749 [0.710, 0.787] ($L_{15}$) & 0.610 [0.559, 0.658] ($L_{18}$) & 0.662 [0.616, 0.706] ($L_{22}$) \\
First state with current-frame CI above $0.5$ & $L_{10}$ ($k/n{=}0.31$) & $L_{14}$ (0.39) & $L_{6}$ (0.21) \\
All captured image tokens, best state & 0.760 [0.721, 0.799] ($L_{15}$) & 0.638 [0.591, 0.684] ($L_{18}$) & 0.669 [0.625, 0.710] ($L_{26}$) \\
Decision state, clean yaw $R^2$, best state & 0.42 ($L_{24}$) & 0.71 ($L_{20}$) & 0.31 ($L_{28}$) \\
Leak control, farthest from chance & 0.444 [0.403, 0.484] & 0.415 [0.367, 0.464] & 0.470 [0.424, 0.515] \\
Turn sign against the rotation & 65.4 [60.8, 69.7]\% & 70.8 [65.9, 75.2]\% & 74.5 [70.1, 78.4]\% \\
\bottomrule
\end{tabular}
\end{table*}

\paragraph{Coverage of the maps.}
The maps cover the language model (encoder-side compression: \cref{app:internals_compression}) and, for LightNav-0,
its two most recent frame pairs, not its full history.

\paragraph{Rotation direction from the $2\times2$ previous frame.}
\label{app:internals_rot_coarse}
On the same 486 MP3D events, matching the previous frame against the current one, an image-shift estimate calibrated
on clean driving with no fit to the rotations, recovers the rotation's direction with
$91.6\%$ accuracy from pixels, $85.6\%$ from $8\times8$ features and $79.6\%$ from $2\times2$ features (HM3D, 429
events: $90.7\%$, $90.9\%$ and $77.9\%$; HSSD, 459: $86.7\%$, $86.7\%$ and $85.0\%$); the ridge probe
on the all-pairs comparison of Uni-NaVid's $2\times2$ tokens at $L_0$ reads $0.786$ ($0.788$ on HM3D, $0.841$ on HSSD; \cref{tab:app_crossmodel}).

\subsection{Directional Knowledge and Target Recovery}
\label{app:internals_functional}

\paragraph{Arms and controllers.}
Each frozen policy runs on 200 validation episodes per model on each of MP3D, HM3D and HSSD, in four arms: clean, the
policy after instant target-loss at severity S1 (a $10^\circ$ margin beyond the image edge; \cref{sec:supp_disturbances}),
and the policy with a scripted turn-back controller given the exact (oracle) or a coin-flip
direction of the rotation. The coin-flip arm is the oracle's control: both controllers receive the true timing and magnitude of the rotation and use identical
bounded turn speed and budget within each model; only the direction differs. The coin is a hash of the rotation's
magnitude and a fixed seed, so it does not depend on the rotation's side. Arms, controllers, matching and scoring are the
same on every dataset; Uni-NaVid's target person is pinned to the person of the episode its evaluator runs first on that
dataset.

\begin{table}[tbp]
\centering
\caption{Matched first-event re-acquisition within 20 steps (\%) after instant target-loss on each dataset: policy,
coin-flip and oracle arms, and the paired oracle-minus-coin-flip difference (percentage points, 95\% episode-bootstrap
interval, 5,000 resamples, seed 0).}
\label{tab:app_recovery_xd}
\resizebox{\columnwidth}{!}{%
\begin{tabular}{@{}llrcccc@{}}
\toprule
Model & Dataset & $n$ & Policy & Coin flip & Oracle & Oracle $-$ coin \\
\midrule
\multirow{3}{*}{Uni-NaVid} & MP3D & 144 & 59.7 & 54.9 & 90.3 & $+$35.4\,[27.8, 43.1] \\
 & HM3D & 122 & 46.7 & 53.3 & 88.5 & $+$35.2\,[27.0, 43.4] \\
 & HSSD & 102 & 49.0 & 50.0 & 87.3 & $+$37.3\,[28.4, 47.1] \\
\midrule
\multirow{3}{*}{LightNav-0} & MP3D & 189 & 45.0 & 50.8 & 95.8 & $+$45.0\,[38.1, 51.9] \\
 & HM3D & 175 & 52.6 & 54.3 & 96.6 & $+$42.3\,[34.9, 49.7] \\
 & HSSD & 179 & 44.1 & 53.1 & 94.4 & $+$41.3\,[34.6, 48.6] \\
\midrule
\multirow{3}{*}{OmTrackVLA} & MP3D & 177 & 39.5 & 65.0 & 97.7 & $+$32.8\,[26.0, 40.1] \\
 & HM3D & 167 & 37.7 & 60.5 & 97.0 & $+$36.5\,[29.3, 43.7] \\
 & HSSD & 169 & 33.7 & 54.4 & 97.6 & $+$43.2\,[35.5, 50.9] \\
\bottomrule
\end{tabular}
}
\end{table}

\begin{table}[tbp]
\centering
\caption{As \cref{tab:app_recovery_xd}, after smooth target-loss (first events that leave the target out of view).}
\label{tab:app_recovery_xd_smooth}
\resizebox{\columnwidth}{!}{%
\begin{tabular}{@{}llrcccc@{}}
\toprule
Model & Dataset & $n$ & Policy & Coin flip & Oracle & Oracle $-$ coin \\
\midrule
\multirow{3}{*}{Uni-NaVid} & MP3D & 92 & 35.9 & 29.3 & 72.8 & $+$43.5\,[33.7, 53.3] \\
 & HM3D & 61 & 34.4 & 45.9 & 78.7 & $+$32.8\,[21.3, 44.3] \\
 & HSSD & 66 & 40.9 & 28.8 & 62.1 & $+$33.3\,[22.7, 45.5] \\
\midrule
\multirow{3}{*}{LightNav-0} & MP3D & 50 & 66.0 & 56.0 & 92.0 & $+$36.0\,[22.0, 50.0] \\
 & HM3D & 45 & 75.6 & 53.3 & 91.1 & $+$37.8\,[24.4, 53.3] \\
 & HSSD & 60 & 61.7 & 50.0 & 90.0 & $+$40.0\,[28.3, 51.7] \\
\midrule
\multirow{3}{*}{OmTrackVLA} & MP3D & 69 & 21.7 & 53.6 & 91.3 & $+$37.7\,[27.5, 49.3] \\
 & HM3D & 48 & 33.3 & 45.8 & 87.5 & $+$41.7\,[27.1, 56.2] \\
 & HSSD & 62 & 35.5 & 32.3 & 90.3 & $+$58.1\,[45.2, 69.4] \\
\bottomrule
\end{tabular}
}
\end{table}

\begin{table*}[tbp]
\centering
\footnotesize
\setlength{\tabcolsep}{3pt}
\caption{Recovery pooled over MP3D, HM3D and HSSD, behind \cref{fig:internals_recovery_pooled}: matched first-event re-acquisition within 20 steps per arm, and the paired oracle-minus-coin-flip difference, weighted by event count. Intervals combine the per-dataset intervals (normal approximation, bounded at 100). Per dataset: \cref{tab:app_recovery_xd,tab:app_recovery_xd_smooth}.}
\label{tab:app_recovery_pooled}
\begin{tabular}{@{}llrcccc@{}}
\toprule
 & & & \multicolumn{3}{c}{Re-acquired within 20 steps (\%)} & \\
\cmidrule(lr){4-6}
Profile & Model & $n$ & Policy & Coin flip & Oracle & Oracle $-$ coin (pp) \\
\midrule
Instant & Uni-NaVid & 368 & 52.4 {\tiny[47.4, 57.5]} & 53.0 {\tiny[47.9, 58.1]} & 88.9 {\tiny[85.7, 92.0]} & +35.9 {\tiny[31.1, 40.7]} \\
 & LightNav-0 & 543 & 47.1 {\tiny[43.1, 51.2]} & 52.7 {\tiny[48.5, 56.9]} & 95.6 {\tiny[93.9, 97.3]} & +42.9 {\tiny[38.8, 47.0]} \\
 & OmTrackVLA & 513 & 37.0 {\tiny[32.9, 41.1]} & 60.0 {\tiny[55.8, 64.3]} & 97.5 {\tiny[96.2, 98.8]} & +37.4 {\tiny[33.2, 41.6]} \\
\midrule
Smooth & Uni-NaVid & 219 & 37.0 {\tiny[30.6, 43.3]} & 33.8 {\tiny[27.4, 40.1]} & 71.2 {\tiny[65.2, 77.2]} & +37.4 {\tiny[31.2, 43.7]} \\
 & LightNav-0 & 155 & 67.1 {\tiny[60.0, 74.2]} & 52.9 {\tiny[45.3, 60.5]} & 91.0 {\tiny[86.7, 95.3]} & +38.1 {\tiny[30.4, 45.7]} \\
 & OmTrackVLA & 179 & 29.6 {\tiny[23.1, 36.1]} & 44.1 {\tiny[37.0, 51.2]} & 89.9 {\tiny[85.6, 94.3]} & +45.8 {\tiny[38.7, 52.9]} \\
\bottomrule
\end{tabular}
\end{table*}

\begin{table}[tbp]
\centering
\small
\setlength{\tabcolsep}{3pt}
\caption{Values behind \cref{fig:internals_recovery_episode}: episode SR / TR (\%) on 200 validation episodes per dataset and target-loss profile (policy and oracle arms), and their mean (600 episodes per model). The instant and smooth runs share the clean run except LightNav-0 on HSSD (smooth: $94.0$ / $93.1$) and OmTrackVLA on HM3D ($67.5$ / $76.1$) and HSSD ($76.0$ / $85.2$); the table shows the instant runs' clean. Bold: the mean of the three datasets.}
\label{tab:app_recovery_episode_datasets}
\resizebox{\columnwidth}{!}{%
\begin{tabular}{@{}llccccc@{}}
\toprule
 & & & \multicolumn{2}{c}{Instant target-loss} & \multicolumn{2}{c}{Smooth target-loss} \\
\cmidrule(lr){4-5} \cmidrule(lr){6-7}
Model & Dataset & Clean & Policy & Oracle & Policy & Oracle \\
\midrule
\multirow{4}{*}{Uni-NaVid} & MP3D & 53.5 / 68.7 & 17.0 / 49.4 & 32.0 / 52.3 & 21.0 / 51.4 & 32.0 / 53.4 \\
 & HM3D & 57.0 / 73.9 & 21.5 / 53.4 & 44.0 / 58.9 & 28.0 / 58.5 & 40.0 / 61.1 \\
 & HSSD & 64.0 / 76.9 & 29.5 / 63.0 & 51.5 / 66.6 & 39.0 / 65.3 & 44.5 / 66.4 \\
 & \textbf{Mean} & 58.2 / 73.2 & 22.7 / 55.3 & 42.5 / 59.3 & 29.3 / 58.4 & 38.8 / 60.3 \\
\midrule
\multirow{4}{*}{LightNav-0} & MP3D & 93.0 / 88.9 & 16.5 / 48.1 & 84.5 / 69.9 & 77.0 / 75.1 & 81.0 / 77.2 \\
 & HM3D & 89.5 / 89.9 & 28.5 / 55.8 & 81.5 / 75.5 & 78.0 / 79.7 & 82.5 / 81.4 \\
 & HSSD & 93.0 / 92.9 & 19.0 / 51.2 & 78.5 / 70.7 & 69.5 / 74.5 & 80.0 / 78.1 \\
 & \textbf{Mean} & 91.8 / 90.6 & 21.3 / 51.7 & 81.5 / 72.0 & 74.8 / 76.5 & 81.2 / 78.9 \\
\midrule
\multirow{4}{*}{OmTrackVLA} & MP3D & 67.0 / 78.8 & 9.0 / 45.3 & 57.5 / 65.1 & 14.5 / 53.8 & 29.5 / 59.9 \\
 & HM3D & 67.5 / 76.4 & 8.5 / 46.8 & 58.5 / 65.9 & 22.5 / 55.7 & 34.5 / 60.3 \\
 & HSSD & 76.0 / 84.8 & 8.0 / 50.2 & 68.0 / 68.1 & 29.5 / 61.9 & 41.0 / 65.6 \\
 & \textbf{Mean} & 70.2 / 80.0 & 8.5 / 47.4 & 61.3 / 66.4 & 22.2 / 57.1 & 35.0 / 61.9 \\
\bottomrule
\end{tabular}
}
\end{table}

\begin{table}[tbp]
\centering
\footnotesize
\setlength{\tabcolsep}{2pt}
\caption{First-turn readout on every dataset and pooled over datasets: side-balanced share (\%) of first turns back toward the target [95\% CI] (events); behind panels (c, d) of \cref{fig:internals_recovery_pooled}. Bold: pooled over the three datasets.}
\label{tab:app_first_turn_xd}
\resizebox{\columnwidth}{!}{%
\begin{tabular}{@{}llcc@{}}
\toprule
Model & Dataset & Instant target-loss ($n$) & Smooth target-loss ($n$) \\
\midrule
Uni-NaVid & MP3D & 63.6 {\tiny[55.5, 71.5]} (144) & 54.5 {\tiny[44.2, 64.7]} (92) \\
 & HM3D & 49.6 {\tiny[40.5, 58.1]} (122) & 47.5 {\tiny[34.8, 59.8]} (61) \\
 & HSSD & 57.3 {\tiny[47.1, 66.9]} (102) & 66.4 {\tiny[54.8, 77.6]} (66) \\
 & \textbf{Pooled} & \textbf{57.2 {\tiny[52.2, 62.3]}} (368) & \textbf{56.1 {\tiny[49.6, 62.7]}} (219) \\
\midrule
LightNav-0 & MP3D & 60.4 {\tiny[53.4, 67.3]} (189) & 80.4 {\tiny[70.0, 89.7]} (50) \\
 & HM3D & 65.1 {\tiny[57.9, 72.2]} (175) & 77.5 {\tiny[64.3, 89.6]} (45) \\
 & HSSD & 66.8 {\tiny[59.9, 73.6]} (179) & 81.8 {\tiny[69.8, 93.0]} (60) \\
 & \textbf{Pooled} & \textbf{64.0 {\tiny[60.0, 68.1]}} (543) & \textbf{80.1 {\tiny[73.5, 86.7]}} (155) \\
\midrule
OmTrackVLA & MP3D & 62.8 {\tiny[55.8, 69.8]} (177) & 69.9 {\tiny[58.4, 80.3]} (69) \\
 & HM3D & 55.2 {\tiny[47.3, 62.8]} (167) & 70.8 {\tiny[58.1, 83.1]} (48) \\
 & HSSD & 51.7 {\tiny[43.8, 59.3]} (169) & 86.9 {\tiny[78.2, 95.0]} (62) \\
 & \textbf{Pooled} & \textbf{56.7 {\tiny[52.3, 61.0]}} (513) & \textbf{76.0 {\tiny[69.9, 82.1]}} (179) \\
\bottomrule
\end{tabular}
}
\end{table}

\paragraph{LightNav-0 decoding.}
In all its recovery runs (instant and smooth, on every dataset), LightNav-0's generated action codes are decoded under an
ordered grammar of its code levels: on top of the authors' optional vocabulary mask, each generated position is
restricted to the next field of the trained order (two grounding codes, one code for each of the three
residual-quantized action levels, then the end token), and greedy decoding picks the highest-scoring code within that
field. The mask alone restricts the token set, not the order. The grammar is applied identically to every arm,
including clean.

\paragraph{Event matching and scoring.}
Each episode contributes its first target-loss event, matched when the policy, coin-flip and oracle arms reach it with
the same pre-event trajectory. Uni-NaVid's arms are free rollouts; for LightNav-0 and OmTrackVLA the coin-flip and
oracle arms use a \emph{forced prefix}, replaying the policy's own actions up to the event. An episode is
excluded when it has no target-loss event, its arms reach different events or pre-event trajectories, or its 20-step
window is incomplete. Re-acquisition means the target is followed again (\cref{sec:supp_metrics}) within 20 steps of
the target-loss onset; it is not episode success. Differences are paired per episode with bootstrap intervals (5,000
resamples, seed 0); pooled intervals combine the per-dataset intervals (\cref{tab:app_recovery_pooled}). Matching and episode populations are
model-specific, and no effect is pooled across models.

\paragraph{Smooth target-loss.}
The smooth control uses the benchmark's open-loop smooth target-loss at severity S1 (the same $10^\circ$ margin, reached at $20^\circ$ per step) on the same
episodes. The controllers are off while the rotation is under way. At the first decision whose image shows the completed
rotation they act only if the target is out of view: its bearing is at least $55^\circ$ (the $45^\circ$ half field of
view plus $10^\circ$) and, for LightNav-0 and OmTrackVLA, its panoptic mask is empty. Uni-NaVid's evaluator gives no
reliable target pixel count, so for Uni-NaVid the bearing alone decides; for the other two, selecting by the empty mask
alone gives, on MP3D, the same 50 events for LightNav-0 and, for OmTrackVLA, 72 events that include all 69 (HM3D: 47
that include all 45, and 49 that include all 48; HSSD: 61 that include all 60, and 64 that include all 62), with
identical outcomes on the shared events. Every model's coin-flip and oracle arms
use a forced prefix through the completed rotation. The matched event is each episode's first smooth target-loss event at
whose takeover step the target is not followed and lies at least $55^\circ$ away, scored as after instant target-loss. No
matched event has a prefix mismatch or an inactive controller.

\paragraph{Episode-level SR and TR.}
SR and TR are the evaluator's own per-episode metrics over all 200 episodes per model, including those without a matched
event; TR is averaged over episodes. The forced-prefix arms reproduce the SR and TR of their free rollouts, except
Uni-NaVid's SR after smooth target-loss, which differs by $0.5$ points. \Cref{tab:app_recovery_episode_datasets} (per dataset and their mean) reports the forced-prefix arms, and for
Uni-NaVid after instant target-loss its free rollouts.

\paragraph{The policy's first turn.}
This readout records whether each policy's first turn after the target-loss goes back toward the target, on each
dataset's matched events. The window is the 20 steps from the first frame that shows the rotation (after smooth
target-loss, from the step at which the target is at least $10^\circ$ out of view). A turn is a pure Uni-NaVid turn
action, or for LightNav-0 and OmTrackVLA an executed yaw above $0.1$ standard deviations of the same run's clean yaw; it
is correct when it opposes the rotation. Side-balanced rates average the left- and right-side rates. Values: \cref{tab:app_first_turn_xd} (per dataset and
pooled).

\FloatBarrier
\subsection{Use of the Decoded Direction}
\label{app:internals_grp_use}

\subsubsection{Rotation-Mirror Test}
\label{app:internals_mirror}

\paragraph{Construction.}
Unlike steering and erasure (\cref{app:internals_causal}), which edit a hidden-state direction at one position, the
rotation-mirror test changes the input. For each command-free event (486 on MP3D, 429 on
HM3D, 459 on HSSD), the episode is re-rendered in the simulator with the recorded actions
and that rotation's side reversed, keeping its magnitude, rate and timing. Each
counterfactual is rendered in a fresh process that first replays the scene's
earlier episodes with their recorded actions, and every frame before the flip
is verified pixel-identical to the original recording (all 456 reversed
rotations on MP3D and all 418 on HSSD; on HM3D 375 of 377 pass this check, and the 8 events of the two that do
not are left out, leaving 421). For 454, 372 and 417 events (MP3D, HM3D, HSSD) the decision at the event is the first to see the
rotation, so the mirrored input is the original history plus the mirrored
frame; for the 32, 49 and 42 events on a later frame of a multi-frame rotation, the
intervening mirrored frames are replayed from a memory snapshot. Each model
replays the original recording as for its maps and, at each event, also
predicts from the mirrored input; its memory is then restored, and the original
prediction is recomputed and checked to be unchanged. The original predictions
reproduce each model's recording on 486, 485 and 486 of the 486 MP3D events
(Uni-NaVid, LightNav-0, OmTrackVLA; LightNav-0's one event has no finite turn), on 429, 428 and 429 of the 429 HM3D
events and on all 459 HSSD events.

\paragraph{Readout.}
The turn is Uni-NaVid's $p(\text{left})-p(\text{right})$ for its first action
word and the continuous turn for LightNav-0 and OmTrackVLA. An event
\emph{follows} the visual side if the turn is corrective in both versions
(against the original rotation in the original, against the mirrored rotation
in the mirror), and goes \emph{against} it if the turn goes with the rotation
in both; the readout is
\begin{equation}
 \Delta_{\text{follow}}=P(\text{follow})-P(\text{against}).
\end{equation}
The remaining events keep one turn sign in both versions or do not turn. Episode
bootstrap with 5,000 resamples; differences between models are paired on events.
A secondary split uses whether the target's
segmentation mask is non-empty in the original and in the mirrored decision
frame. With the target in view in both, turning toward where it now appears is
enough to follow the side; with it out of view in both, only the scene's motion
indicates the side. The target is out of view in 43\%, 41\% and 30\% of the original decision
frames (MP3D, HM3D, HSSD); 181, 185 and 234 events have it in view in both versions, 117, 120 and 57 in neither, and
188, 116 and 168 in one only; the subsets hold different events, so a difference between them mixes visibility with the events' other
properties.

\begin{table*}[tbp]
\centering
\footnotesize
\setlength{\tabcolsep}{4pt}
\caption{Rotation-mirror test on the command-free events of each dataset. $\Delta_{\text{follow}}$
[95\% CI]: share turning against the rotation in both versions minus share
turning with it in both.}
\label{tab:app_mirror}
\begin{tabular}{lrrrrr}
\toprule
 & $\Delta_{\text{follow}}$, all events & Follow & Against & Target in view in both & Out of view in both \\
\midrule
\multicolumn{6}{@{}l}{\emph{MP3D: 486 events; target in view in both 181, out of view in both 117}} \\
Uni-NaVid & $+0.084$ [$+0.016$, $+0.152$] & 0.270 & 0.185 & $+0.370$ [$+0.240$, $+0.487$] & $-0.248$ [$-0.356$, $-0.140$] \\
LightNav-0 & $+0.358$ [$+0.296$, $+0.418$] & 0.463 & 0.105 & $+0.635$ [$+0.549$, $+0.712$] & $-0.085$ [$-0.207$, $+0.035$] \\
OmTrackVLA & $+0.436$ [$+0.378$, $+0.491$] & 0.479 & 0.043 & $+0.663$ [$+0.591$, $+0.730$] & $+0.231$ [$+0.126$, $+0.331$] \\
\midrule
\multicolumn{6}{@{}l}{\emph{HM3D: 421 events; target in view in both 185, out of view in both 120}} \\
Uni-NaVid & $+0.166$ [$+0.097$, $+0.237$] & 0.311 & 0.145 & $+0.470$ [$+0.382$, $+0.556$] & $-0.317$ [$-0.411$, $-0.222$] \\
LightNav-0 & $+0.325$ [$+0.257$, $+0.394$] & 0.461 & 0.135 & $+0.638$ [$+0.566$, $+0.709$] & $-0.233$ [$-0.358$, $-0.108$] \\
OmTrackVLA & $+0.451$ [$+0.391$, $+0.510$] & 0.496 & 0.045 & $+0.665$ [$+0.596$, $+0.731$] & $+0.083$ [$-0.023$, $+0.194$] \\
\midrule
\multicolumn{6}{@{}l}{\emph{HSSD: 459 events; target in view in both 234, out of view in both 57}} \\
Uni-NaVid & $+0.316$ [$+0.253$, $+0.380$] & 0.366 & 0.050 & $+0.427$ [$+0.352$, $+0.504$] & $-0.088$ [$-0.235$, $+0.052$] \\
LightNav-0 & $+0.475$ [$+0.411$, $+0.534$] & 0.512 & 0.037 & $+0.590$ [$+0.513$, $+0.664$] & $-0.105$ [$-0.211$, $0.0$] \\
OmTrackVLA & $+0.514$ [$+0.454$, $+0.569$] & 0.538 & 0.024 & $+0.568$ [$+0.502$, $+0.630$] & $+0.070$ [$-0.073$, $+0.213$] \\
\bottomrule
\end{tabular}
\end{table*}

\subsubsection{Interchange Test on the Rotation-Mirror Pairs}
\label{app:internals_interchange}

\begin{figure}[tbp]
 \centering
 \includegraphics[width=\columnwidth]{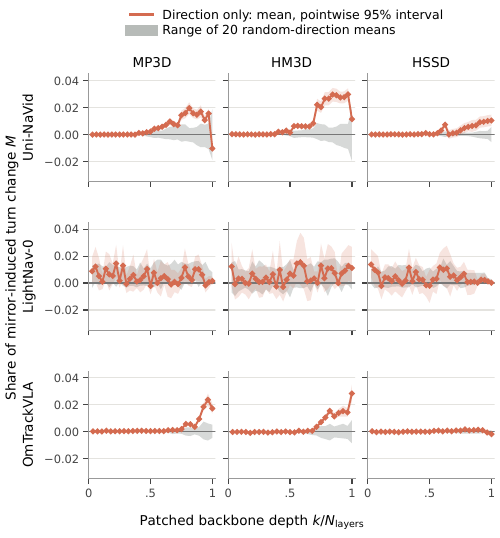}
 \caption{Every-output-layer direction-only effects on a magnified scale, per dataset. Coral: decoded-direction patch, with
 pointwise $95\%$ episode-bootstrap bands; gray: range of the 20 norm-matched random-direction means, not their
 confidence interval. The main-text figure shows the MP3D effects on the whole-state scale.}
 \label{fig:app_interchange}
\end{figure}

The interchange test runs on the rotation-mirror pairs of MP3D, HM3D and HSSD. Whole-decision and direction-only
patches use every decoder output state: 32 Uni-NaVid, 36 LightNav-0 and 28 OmTrackVLA states (\cref{fig:internals_patch_search}: MP3D; \cref{fig:app_interchange}: all three datasets). Input state $L_{0}$ is not an output layer.
\Cref{tab:internals_interchange_final} gives each model's final state. Over the late states where the whole decision
state carries $0.30$--$1.00$ of the mirror's effect (Uni-NaVid, LightNav-0, OmTrackVLA; MP3D $L_{17}$--$L_{32}$,
$L_{27}$--$L_{36}$ and $L_{25}$--$L_{28}$; HM3D $L_{16}$--$L_{32}$, $L_{27}$--$L_{36}$ and $L_{25}$--$L_{28}$; HSSD
$L_{16}$--$L_{32}$, $L_{30}$--$L_{36}$ and $L_{26}$--$L_{28}$), the largest share of the decoded direction alone is
$0.020$, $0.012$ and $0.024$ on MP3D, $0.0298$, $0.013$ and $0.028$ on HM3D, and $0.010$, $0.003$ and $0.001$ on HSSD;
its interval excludes zero at most of these states in Uni-NaVid on all three datasets and in OmTrackVLA on MP3D and
HM3D.

\begin{table}[tbp]
    \centering
    \footnotesize
    \setlength{\tabcolsep}{3pt}
    \caption{Final-state action-effect shares $M$ of the interchange test (\cref{sec:results_rotation_use}), per dataset.}
    \label{tab:internals_interchange_final}
\begin{tabular}{@{}llccc@{}}
\toprule
Model & Dataset & Final state & Whole state & Direction only \\
\midrule
Uni-NaVid & MP3D & $L_{32}$ & 1.000 & $-$0.010 \\
 & HM3D & $L_{32}$ & 1.000 & +0.011 \\
 & HSSD & $L_{32}$ & 1.000 & +0.010 \\
\addlinespace
LightNav-0 & MP3D & $L_{36}$ & 0.732 & +0.002 \\
 & HM3D & $L_{36}$ & 0.630 & +0.011 \\
 & HSSD & $L_{36}$ & 0.887 & 0.000 \\
\addlinespace
OmTrackVLA & MP3D & $L_{28}$ & 1.000 & +0.017 \\
 & HM3D & $L_{28}$ & 1.000 & +0.028 \\
 & HSSD & $L_{28}$ & 1.000 & $-$0.002 \\
\bottomrule
\end{tabular}
\end{table}

\paragraph{Construction.}
The rotation-mirror pairs (\cref{app:internals_mirror}) give, for each event, two inputs that differ only in the side of the
rotation, and therefore two internal states of the same model. The interchange test copies states of the mirrored run
into the original run and asks how much of the mirror's effect on the turn follows. Each episode is replayed once; at
each eligible event (below), the model (1)~predicts from the original input
with every state recorded, (2)~predicts from the mirrored input with every state recorded, (3)~predicts from the
original input with one set of positions at one state replaced by the mirrored states, for every patch condition, and
(4)~predicts from the original input again, which must equal (1) before the replay continues. Patches act on the prefill of the decision (Uni-NaVid: the prefill of its
first answer token; LightNav-0: the prompt before its generated action tokens; OmTrackVLA: its single pass). State $k$
is the input of decoder layer $k$ (state $n$: the last layer's output), so it coincides with map state $L_k$.

\paragraph{Eligible versus valid events.}
The eligible events (454 in 148 MP3D episodes, 372 in 132 HM3D episodes, 417 in 146 HSSD episodes) are those whose decision is the first to see their rotation (\cref{app:internals_mirror}). Uni-NaVid and OmTrackVLA retain all baseline/mirror pairs;
on MP3D and HM3D, LightNav-0 has a nonfinite baseline or mirrored turn on one pair each, leaving 453 and 371 valid. Each patch additionally excludes
nonfinite outputs in that condition: valid counts for LightNav-0 are 452--453 (MP3D), 370--371 (HM3D) and 416--417 (HSSD), and all eligible events for the other models.

\paragraph{Output-layer coverage.}
Whole-state, direction-only and 20 norm-matched random patches are applied at every decoder output
state, each with that state's clean readout direction from the full-dimensional probe; the direction-kept patch likewise. The visual-group patches cover
sampled states, ending at $L_{28}$, $L_{32}$ and $L_{24}$ for Uni-NaVid, LightNav-0 and OmTrackVLA; no final visual-group value is
measured or imputed. At OmTrackVLA's final output the head
reads the action-token state; a patch to other output tokens would require another mixing layer to affect it.
All plotted direction and decision points are
measured, not interpolations over untested layers; lines simply join neighboring output states.

\paragraph{Patch conditions and readout.}
\emph{All positions} from the current frame's first token on (the positive control); the \emph{current-frame image
tokens}; the \emph{whole decision state}; its \emph{decoded rotation direction} alone, a one-dimensional patch
along the clean-fit rotation probe's direction $\boldsymbol g$ that sets the probe's value to the mirror's,
\begin{equation}
 \boldsymbol h\leftarrow\boldsymbol h+\boldsymbol g^{\mathsf T}(\boldsymbol h'-\boldsymbol h)\,
 \frac{\boldsymbol g}{\|\boldsymbol g\|^2};
\end{equation}
the mirrored decision state with that component kept at the original value (the direction ablation of the
whole-state patch); the target-bearing direction alone; and 20 random one-dimensional patches norm-matched per event
to the decoded-direction patch (the random directions of the steering test; control). With $T$ the turn (Uni-NaVid: $\log p(\text{left})-\log p(\text{right})$ of the first answer word~\cite{zhang2024patching,heimersheim2024patching};
the waypoint models: the continuous turn), the mirror's and the patch's effects on event $e$ are
\begin{equation}
 D_e=T^{\text{mirror}}_e-T_e,\qquad \Delta_e=T^{\text{patch}}_e-T_e,
\end{equation}
and the share carried, a signed aggregate in which opposite-sign events cancel, is
\begin{equation}
 M=\frac{\sum_e\operatorname{sign}(D_e)\,\Delta_e}{\sum_e|D_e|},
\end{equation}
with an episode bootstrap (5{,}000 resamples, ratio recomputed per resample).

\paragraph{Event-level effects and target visibility.} \Cref{tab:app_interchange_events} gives the effects per event
on each dataset, including $M$ on the events with the target out of view in both versions (117, 116 and 55), where only
the scene's motion indicates the side. On these events the decoded direction alone carries at most $0.048$ at the final
state; its largest share at a primary state is $0.069$ [$0.009$, $0.150$] (LightNav-0 on HM3D), where the whole state
carries $0.112$.

\begin{table*}[tbp]
\centering
\footnotesize
\setlength{\tabcolsep}{3pt}
\caption{Interchange effects per event on each dataset, at the primary state $k^*$ and the final state (eligible events:
MP3D 454, LightNav-0 453, of which 444 with a nonzero mirror effect $D_e$; HM3D 372, LightNav-0 371, of which 361 with a nonzero mirror effect $D_e$; HSSD 417, LightNav-0 417, of which 390 with a nonzero mirror effect $D_e$). \emph{Toward mirror}: share of events whose patched turn moves
toward the mirrored one, $\operatorname{sign}(D_e)\,\Delta_e>0$, for the whole decision state, the decoded direction and the 20
random directions (mean, range of the 20). \emph{Direction $-$ random}: paired per-event difference of
$\operatorname{sign}(D_e)\,\Delta_e$, normalized by the mean $|D_e|$. \emph{Size}: mean $|\Delta_e|$ of the decoded-direction
patch as a share of the mean $|D_e|$. \emph{Out of view}: $M$ on the events with the target out of view in both
versions (MP3D 117 events in 81 episodes; HM3D 116 events in 71 episodes (LightNav-0 115); HSSD 55 events in 51 episodes), whole state / decoded direction. Intervals: 95\% episode bootstrap, 5{,}000 resamples, seed 0.}
\label{tab:app_interchange_events}
\begin{tabular}{@{}llccccccc@{}}
\toprule
 & & \multicolumn{3}{c}{Toward mirror (\%)} & Direction $-$ random & Size & \multicolumn{2}{c}{Out of view, $M$} \\
\cmidrule(lr){3-5}\cmidrule(l){8-9}
Model & State & Whole & Direction & Random & & (\%) & Whole & Direction \\
\midrule
\multicolumn{9}{@{}l}{\emph{MP3D}} \\
Uni-NaVid & $L_{16}$ & 71.1 & 47.6 & 42.2 (34.6--49.6) & $+0.002$ {\tiny[$+0.001$, $+0.003$]} & 0.6 & $+0.097$ & $+0.006$ \\
 & $L_{32}$ & 100.0 & 29.3 & 36.7 (25.1--63.4) & $-0.008$ {\tiny[$-0.010$, $-0.006$]} & 1.7 & $+1.000$ & $-0.019$ \\
\addlinespace
LightNav-0 & $L_{20}$ & 19.8 & 3.2 & 4.3 (2.7--5.9) & $-0.006$ {\tiny[$-0.013$, $+0.001$]} & 1.2 & $+0.055$ & $+0.006$ \\
 & $L_{36}$ & 76.8 & 0.7 & 0.5 (0.0--1.1) & $-0.002$ {\tiny[$-0.005$, $+0.001$]} & 0.2 & $+0.747$ & $+0.001$ \\
\addlinespace
OmTrackVLA & $L_{11}$ & 55.5 & 54.0 & 51.9 (46.0--59.3) & $0.0$ {\tiny[$0.000$, $+0.001$]} & 0.5 & $+0.022$ & $+0.002$ \\
 & $L_{28}$ & 100.0 & 69.4 & 47.9 (29.1--68.7) & $+0.018$ {\tiny[$+0.013$, $+0.022$]} & 3.3 & $+1.000$ & $+0.014$ \\
\midrule
\multicolumn{9}{@{}l}{\emph{HM3D}} \\
Uni-NaVid & $L_{21}$ & 91.4 & 54.6 & 43.9 (38.7--54.0) & $+0.006$ {\tiny[$+0.004$, $+0.007$]} & 1.2 & $+0.795$ & $+0.016$ \\
 & $L_{32}$ & 100.0 & 66.4 & 36.7 (22.3--68.3) & $+0.014$ {\tiny[$+0.011$, $+0.017$]} & 1.8 & $+1.000$ & $+0.024$ \\
\addlinespace
LightNav-0 & $L_{21}$ & 13.9 & 3.3 & 3.4 (1.9--4.4) & $+0.011$ {\tiny[$-0.001$, $+0.025$]} & 2.7 & $+0.112$ & $+0.069$ \\
 & $L_{36}$ & 70.1 & 1.1 & 0.7 (0.0--1.4) & $+0.007$ {\tiny[$+0.001$, $+0.015$]} & 1.1 & $+0.515$ & $+0.041$ \\
\addlinespace
OmTrackVLA & $L_{19}$ & 72.0 & 55.6 & 47.9 (37.9--58.1) & $+0.001$ {\tiny[$0.0$, $+0.002$]} & 0.5 & $+0.105$ & $+0.001$ \\
 & $L_{28}$ & 100.0 & 79.6 & 46.3 (19.9--76.9) & $+0.029$ {\tiny[$+0.023$, $+0.035$]} & 3.8 & $+1.000$ & $+0.048$ \\
\midrule
\multicolumn{9}{@{}l}{\emph{HSSD}} \\
Uni-NaVid & $L_{21}$ & 95.0 & 44.1 & 44.0 (35.3--49.9) & $0.000$ {\tiny[$-0.003$, $+0.003$]} & 2.1 & $+0.818$ & $+0.018$ \\
 & $L_{32}$ & 100.0 & 55.4 & 40.1 (29.0--54.7) & $+0.011$ {\tiny[$+0.006$, $+0.016$]} & 3.3 & $+1.000$ & $+0.042$ \\
\addlinespace
LightNav-0 & $L_{21}$ & 13.8 & 3.9 & 3.8 (2.6--5.4) & $+0.005$ {\tiny[$-0.001$, $+0.013$]} & 1.6 & $+0.138$ & $+0.040$ \\
 & $L_{36}$ & 79.7 & 0.3 & 0.2 (0.0--0.5) & $0.000$ {\tiny[$-0.001$, $0.0$]} & 0.0 & $+0.358$ & $0.0$ \\
\addlinespace
OmTrackVLA & $L_{11}$ & 58.0 & 50.6 & 52.0 (45.3--56.4) & $0.000$ {\tiny[$0.000$, $0.0$]} & 0.3 & $+0.019$ & $+0.001$ \\
 & $L_{28}$ & 100.0 & 46.0 & 50.4 (41.5--59.2) & $-0.002$ {\tiny[$-0.003$, $0.000$]} & 0.9 & $+1.000$ & $0.000$ \\
\bottomrule
\end{tabular}
\end{table*}

\paragraph{Controls.}
Replay check: share of events on which the replay reproduces the recorded turn (sign identical, $|\Delta|<1\%$ of
$\sigma_T$, the standard deviation of the model's turn $T$ over its undisturbed replay steps); mirror check: share on which the mirrored prediction reproduces the rotation-mirror test's; patch check:
events on which the post-patch prediction equals the original, and share on which every prompt state before the current
frame is bit-identical between the original and mirrored runs (\cref{tab:app_interchange}). The positive control is the
all-positions patch at $k^*$, the decision state whose rotation readout is farthest from chance. Across layers
(\cref{fig:internals_patch_search}), the whole decision state serves as the reference against which the decoded
direction is compared: at the final state it carries $0.73$--$1.00$ of the effect on MP3D and $0.63$--$1.00$ on all
three datasets (\cref{tab:internals_interchange_final}).

\begin{table*}[tbp]
\centering
\footnotesize
\setlength{\tabcolsep}{4pt}
\caption{Interchange test: checks and shares $M$ [95\% CI] of the mirror's effect on the turn, per dataset. Rows from
the primary state on are at $k^*$.}
\label{tab:app_interchange}
\begin{tabular}{@{}lccc@{}}
\toprule
 & Uni-NaVid & LightNav-0 & OmTrackVLA \\
\midrule
\multicolumn{4}{@{}l}{\emph{MP3D}} \\
Eligible events (episodes) & 454 (148) & 454 (148) & 454 (148) \\
Replay reproduces recording & 1.000 & 0.998 & 1.000 \\
Mirror reproduces mirror test & 1.000 & 1.000 & 1.000 \\
Post-patch = base; prefix identical & 454/454; 1.000 & 454/454; 1.000 & 454/454; 1.000 \\
Mirror effect: mean $|D|/\sigma_T$; sign flips & 0.84; 46\% & 1.30; 64\% & 1.14; 51\% \\
Primary state $k^*$ & $L_{16}$ & $L_{20}$ & $L_{11}$ \\
All positions from current frame (positive control) & +1.000 {\tiny[+1.000, +1.000]} & +0.997 {\tiny[+0.967, +1.027]} & +1.000 {\tiny[+1.000, +1.000]} \\
Current-frame image tokens & +0.100 {\tiny[+0.071, +0.130]} & +0.758 {\tiny[+0.678, +0.837]} & +0.987 {\tiny[+0.982, +0.993]} \\
Decision state, whole & +0.239 {\tiny[+0.204, +0.276]} & +0.022 {\tiny[$-$0.007, +0.055]} & +0.010 {\tiny[+0.005, +0.017]} \\
Decoded rotation direction only & +0.002 {\tiny[+0.001, +0.003]} & 0.000 {\tiny[$-$0.007, +0.008]} & +0.001 {\tiny[0.000, +0.001]} \\
Decision state, direction kept & +0.237 {\tiny[+0.201, +0.274]} & +0.029 {\tiny[$-$0.001, +0.063]} & +0.010 {\tiny[+0.004, +0.016]} \\
Target-bearing direction only & +0.008 {\tiny[+0.006, +0.011]} & +0.014 {\tiny[+0.002, +0.029]} & +0.001 {\tiny[+0.001, +0.002]} \\
20 random 1-D directions (means) & $-$0.002 to +0.003 & 0.000 to +0.017 & $-$0.001 to +0.001 \\
\midrule
\multicolumn{4}{@{}l}{\emph{HM3D}} \\
Eligible events (episodes) & 372 (132) & 372 (132) & 372 (132) \\
Replay reproduces recording & 1.000 & 0.997 & 1.000 \\
Mirror reproduces mirror test & 1.000 & 1.000 & 1.000 \\
Post-patch = base; prefix identical & 372/372; 1.000 & 372/372; 1.000 & 372/372; 1.000 \\
Mirror effect: mean $|D|/\sigma_T$; sign flips & 0.86; 44\% & 1.30; 62\% & 1.19; 52\% \\
Primary state $k^*$ & $L_{21}$ & $L_{21}$ & $L_{19}$ \\
All positions from current frame (positive control) & +1.000 {\tiny[+1.000, +1.000]} & +0.958 {\tiny[+0.914, +1.000]} & +1.000 {\tiny[+1.000, +1.000]} \\
Current-frame image tokens & +0.009 {\tiny[0.000, +0.017]} & +0.676 {\tiny[+0.588, +0.768]} & +0.926 {\tiny[+0.908, +0.942]} \\
Decision state, whole & +0.689 {\tiny[+0.666, +0.711]} & +0.046 {\tiny[+0.014, +0.081]} & +0.072 {\tiny[+0.054, +0.091]} \\
Decoded rotation direction only & +0.006 {\tiny[+0.004, +0.007]} & +0.015 {\tiny[$-$0.001, +0.037]} & +0.001 {\tiny[0.000, +0.002]} \\
Decision state, direction kept & +0.682 {\tiny[+0.660, +0.704]} & +0.044 {\tiny[+0.011, +0.083]} & +0.071 {\tiny[+0.053, +0.090]} \\
Target-bearing direction only & +0.079 {\tiny[+0.067, +0.091]} & +0.007 {\tiny[$-$0.001, +0.016]} & +0.001 {\tiny[0.000, +0.002]} \\
20 random 1-D directions (means) & $-$0.002 to +0.004 & $-$0.004 to +0.018 & $-$0.002 to +0.002 \\
\midrule
\multicolumn{4}{@{}l}{\emph{HSSD}} \\
Eligible events (episodes) & 417 (146) & 417 (146) & 417 (146) \\
Replay reproduces recording & 1.000 & 1.000 & 1.000 \\
Mirror reproduces mirror test & 1.000 & 1.000 & 1.000 \\
Post-patch = base; prefix identical & 417/417; 1.000 & 417/417; 1.000 & 417/417; 1.000 \\
Mirror effect: mean $|D|/\sigma_T$; sign flips & 0.77; 42\% & 1.00; 58\% & 1.44; 56\% \\
Primary state $k^*$ & $L_{21}$ & $L_{21}$ & $L_{11}$ \\
All positions from current frame (positive control) & +1.000 {\tiny[+1.000, +1.000]} & +1.018 {\tiny[+0.978, +1.058]} & +1.000 {\tiny[+1.000, +1.000]} \\
Current-frame image tokens & $-$0.006 {\tiny[$-$0.013, +0.001]} & +0.797 {\tiny[+0.713, +0.876]} & +0.998 {\tiny[+0.995, +1.000]} \\
Decision state, whole & +0.713 {\tiny[+0.691, +0.736]} & +0.017 {\tiny[$-$0.007, +0.041]} & +0.004 {\tiny[+0.001, +0.007]} \\
Decoded rotation direction only & 0.000 {\tiny[$-$0.003, +0.003]} & +0.012 {\tiny[+0.002, +0.023]} & 0.000 {\tiny[0.000, 0.000]} \\
Decision state, direction kept & +0.712 {\tiny[+0.690, +0.734]} & +0.017 {\tiny[$-$0.006, +0.040]} & +0.004 {\tiny[+0.001, +0.007]} \\
Target-bearing direction only & +0.059 {\tiny[+0.050, +0.067]} & +0.001 {\tiny[$-$0.009, +0.010]} & 0.000 {\tiny[0.000, 0.000]} \\
20 random 1-D directions (means) & $-$0.001 to +0.001 & 0.000 to +0.011 & 0.000 to +0.001 \\
\bottomrule
\end{tabular}
\end{table*}

\paragraph{Scope.}
The patches act on the prefill only, so content entering through LightNav-0's generated action tokens, which read the
unpatched states, is outside the patched positions; this is why its all-positions share falls below one from $L_{24}$ on
(HSSD: from $L_{28}$).

\subsubsection{Steering and Erasure Ablation at the Decision State}
\label{app:internals_causal}

\paragraph{What is tested.}
These tests ask whether each model's turn uses the decoded rotation direction at its decision state. Recorded rotation
events without a turn command at the preceding decision (486 in 148 MP3D episodes, 429 in 133 HM3D episodes, 459 in 146 HSSD episodes) are replayed with the
state edited at the decision token. \emph{Steering} adds the direction with the sign of the imposed rotation, as if the
rotation were larger; \emph{erasure}, a directional ablation, removes it. If the turn used the direction, steering would raise the corrective
turn and erasure would lower it. Twenty norm-matched random directions show what an arbitrary edit of the same size
does, and steering along the target-bearing direction is the positive control: it shows whether an edit at this token
moves the turn at all. Two checks confirm the replay: without an edit it reproduces the recorded decisions, and the
probe reads the replayed state with the map's direction accuracy (\cref{tab:app_steer_erase}, column ``Replay; map acc.'').

\paragraph{Construction.}
The decision state is the state farthest from chance in each dataset's rotation map (\cref{tab:app_steer_erase}). In
each scene-held-out fold, the clean-fitted linear rotation probe gives a direction $\boldsymbol g$ in the raw state (its
weights mapped through the feature standardization). With $s$ the sign of the imposed rotation and $\sigma_r$ the
standard deviation of the probe's clean training readout, the steering shift is
\begin{equation}
 \delta\boldsymbol h = s\,m\sigma_r\,\frac{\boldsymbol g}{\|\boldsymbol g\|^2},
\end{equation}
which changes the probe's readout by exactly $s m\sigma_r$; the reported dose is $m=3$ (3\,SD). The random directions are norm-matched to this shift per event.
Erasure removes the projection onto $\boldsymbol g$ from three layers before the decision state through the final layer.
The positive control steers along the target-bearing direction with each sign at 3\,SD, and its readout is the
difference in the turn between the two signs. Each edited replay restores the same memory snapshot, so an edit does not
carry into later steps. Uni-NaVid's readouts are the change in corrective log-odds (the corrective against the opposite
turn word) and the corrective preference, the share of events whose corrective word is more probable. The waypoint
models' are the change in the corrective component of the continuous turn,
\begin{equation}
 c=-\operatorname{sign}(\text{rotation})\cdot\text{turn},
\end{equation}
in units of the model's clean-driving turn standard deviation
$\sigma_T$, and the share of events with a corrective turn.

\paragraph{Replay checks and positive control.}
Both replay checks hold in every model--dataset cell (\cref{tab:app_steer_erase}). The positive control moves the
turn, with an interval excluding zero, for Uni-NaVid on all three datasets, for OmTrackVLA on MP3D and HM3D and for
LightNav-0 on HM3D. For LightNav-0 on MP3D and HSSD and OmTrackVLA on HSSD its interval includes zero: an edit at the
decision token does not measurably move their turn there, so these cells give no reading. Steering along the decoded
direction nevertheless moves the corrective turn in two of them, with intervals excluding zero but inside the range of
the random directions: OmTrackVLA on HSSD by $+0.0021$ [$+0.0013$, $+0.0030$] $\sigma_T$ and LightNav-0 on MP3D by
$-0.0109$ [$-0.0267$, $-0.0002$] $\sigma_T$. In the cells with a reading, steering never raises the corrective turn. In Uni-NaVid it lowers the corrective log-odds, beyond the range
of the random directions on HM3D and HSSD ($-0.078$ and $-0.237$), and erasure raises them ($+0.036$ to $+0.077$):
the direction's effect is at most $0.21$ of the positive control's and opposes the correction. In OmTrackVLA,
erasure lowers the corrective turn on MP3D and HM3D ($-0.0148$ [$-0.0208$, $-0.0086$] and $-0.0162$ [$-0.0229$,
$-0.0092$] $\sigma_T$), more than the positive control moves it ($+0.0054$ and $+0.0108$; erasure edits every state from three before $k^*$, the positive control one), while its corrective
share changes from $0.724$ to $0.722$ and from $0.704$ to $0.711$.

\begin{table*}[tbp]
\centering
\footnotesize
\setlength{\tabcolsep}{3pt}
\caption{Steering and erasure (directional ablation) at the decision state. Uni-NaVid: changes in corrective log-odds (positive control:
left-versus-right log-odds); LightNav-0 and OmTrackVLA: changes in the corrective turn in units of $\sigma_T$. Replay:
share of recorded decisions reproduced without an edit; map acc.: the probe's direction accuracy on the replayed state (LightNav-0 on HM3D: $0.937$, against $0.935$ on the map).
Random means: range of the means of 20 norm-matched random directions at 3\,SD, not a null interval. Corrective share: share of events with a
corrective turn, without an edit $\rightarrow$ after erasure. Intervals: 95\% episode bootstrap (Uni-NaVid 2,000
resamples, its positive control and steering 5,000 as in \cref{tab:supp_ft_direction}; the waypoint models 5,000).}
\label{tab:app_steer_erase}
\begin{tabular}{@{}llccccccc@{}}
\toprule
Model & Dataset & State & \begin{tabular}[b]{@{}c@{}}Replay;\\map acc.\end{tabular} & \begin{tabular}[b]{@{}c@{}}Positive\\control\end{tabular} & \begin{tabular}[b]{@{}c@{}}Steer\\3\,SD\end{tabular} & \begin{tabular}[b]{@{}c@{}}20 random\\means\end{tabular} & \begin{tabular}[b]{@{}c@{}}Erasure\\$\Delta$\end{tabular} & \begin{tabular}[b]{@{}c@{}}Corrective\\share\end{tabular} \\
\midrule
Uni-NaVid & MP3D & $L_{16}$ & 1.000; 0.741 & \begin{tabular}[c]{@{}c@{}}$+0.374$\\{\tiny[$+0.339$, $+0.406$]}\end{tabular} & \begin{tabular}[c]{@{}c@{}}$-0.018$\\{\tiny[$-0.021$, $-0.014$]}\end{tabular} & $-0.029$ to $+0.020$ & \begin{tabular}[c]{@{}c@{}}$+0.069$\\{\tiny[$+0.056$, $+0.084$]}\end{tabular} & $0.578\rightarrow0.580$ \\
\addlinespace[3pt]
 & HM3D & $L_{21}$ & 1.000; 0.706 & \begin{tabular}[c]{@{}c@{}}$+2.054$\\{\tiny[$+1.831$, $+2.309$]}\end{tabular} & \begin{tabular}[c]{@{}c@{}}$-0.078$\\{\tiny[$-0.083$, $-0.072$]}\end{tabular} & $-0.052$ to $+0.048$ & \begin{tabular}[c]{@{}c@{}}$+0.077$\\{\tiny[$+0.053$, $+0.101$]}\end{tabular} & $0.580\rightarrow0.585$ \\
\addlinespace[3pt]
 & HSSD & $L_{21}$ & 1.000; 0.634 & \begin{tabular}[c]{@{}c@{}}$+1.150$\\{\tiny[$+1.018$, $+1.285$]}\end{tabular} & \begin{tabular}[c]{@{}c@{}}$-0.237$\\{\tiny[$-0.246$, $-0.227$]}\end{tabular} & $-0.053$ to $+0.062$ & \begin{tabular}[c]{@{}c@{}}$+0.036$\\{\tiny[$+0.014$, $+0.058$]}\end{tabular} & $0.654\rightarrow0.664$ \\
\midrule
LightNav-0 & MP3D & $L_{20}$ & 1.000; 0.938 & \begin{tabular}[c]{@{}c@{}}$+0.0015$\\{\tiny[$-0.0112$, $+0.0160$]}\end{tabular} & \begin{tabular}[c]{@{}c@{}}$-0.0109$\\{\tiny[$-0.0267$, $-0.0002$]}\end{tabular} & $-0.0140$ to $+0.0054$ & \begin{tabular}[c]{@{}c@{}}$+0.0012$\\{\tiny[$-0.0182$, $+0.0214$]}\end{tabular} & $0.664\rightarrow0.665$ \\
\addlinespace[3pt]
 & HM3D & $L_{21}$ & 1.000; 0.937 & \begin{tabular}[c]{@{}c@{}}$+0.0470$\\{\tiny[$+0.0205$, $+0.0789$]}\end{tabular} & \begin{tabular}[c]{@{}c@{}}$0.0000$\\{\tiny[$-0.0123$, $+0.0123$]}\end{tabular} & $-0.0060$ to $+0.0163$ & \begin{tabular}[c]{@{}c@{}}$+0.0146$\\{\tiny[$-0.0070$, $+0.0385$]}\end{tabular} & $0.643\rightarrow0.647$ \\
\addlinespace[3pt]
 & HSSD & $L_{21}$ & 1.000; 0.915 & \begin{tabular}[c]{@{}c@{}}$+0.0062$\\{\tiny[$-0.0020$, $+0.0145$]}\end{tabular} & \begin{tabular}[c]{@{}c@{}}$+0.0032$\\{\tiny[$-0.0062$, $+0.0133$]}\end{tabular} & $-0.0103$ to $+0.0081$ & \begin{tabular}[c]{@{}c@{}}$+0.0055$\\{\tiny[$-0.0068$, $+0.0209$]}\end{tabular} & $0.708\rightarrow0.710$ \\
\midrule
OmTrackVLA & MP3D & $L_{11}$ & 1.000; 0.342 & \begin{tabular}[c]{@{}c@{}}$+0.0054$\\{\tiny[$+0.0040$, $+0.0069$]}\end{tabular} & \begin{tabular}[c]{@{}c@{}}$-0.0018$\\{\tiny[$-0.0026$, $-0.0010$]}\end{tabular} & $-0.0026$ to $+0.0026$ & \begin{tabular}[c]{@{}c@{}}$-0.0148$\\{\tiny[$-0.0208$, $-0.0086$]}\end{tabular} & $0.724\rightarrow0.722$ \\
\addlinespace[3pt]
 & HM3D & $L_{19}$ & 1.000; 0.268 & \begin{tabular}[c]{@{}c@{}}$+0.0108$\\{\tiny[$+0.0076$, $+0.0142$]}\end{tabular} & \begin{tabular}[c]{@{}c@{}}$-0.0035$\\{\tiny[$-0.0055$, $-0.0017$]}\end{tabular} & $-0.0098$ to $+0.0120$ & \begin{tabular}[c]{@{}c@{}}$-0.0162$\\{\tiny[$-0.0229$, $-0.0092$]}\end{tabular} & $0.704\rightarrow0.711$ \\
\addlinespace[3pt]
 & HSSD & $L_{11}$ & 1.000; 0.331 & \begin{tabular}[c]{@{}c@{}}$-0.0007$\\{\tiny[$-0.0017$, $+0.0005$]}\end{tabular} & \begin{tabular}[c]{@{}c@{}}$+0.0021$\\{\tiny[$+0.0013$, $+0.0030$]}\end{tabular} & $-0.0032$ to $+0.0032$ & \begin{tabular}[c]{@{}c@{}}$-0.0008$\\{\tiny[$-0.0034$, $+0.0018$]}\end{tabular} & $0.745\rightarrow0.745$ \\
\bottomrule
\end{tabular}
\end{table*}

\subsubsection{Which Components Carry the Mirror's Effect}
\label{app:internals_patch_search}

\begin{table*}[tbp]
\centering
\footnotesize
\setlength{\tabcolsep}{4pt}
\caption{Components that carry the mirror's effect (stage 1, all events): attention and MLP outputs at the decision
state and at the prompt tokens after the video, every layer; listed are the layers with $|M|$ or $|R|\ge0.10$
($M\,|\,R$).}
\label{tab:app_patch_components}
\begin{tabular}{@{}llp{0.62\textwidth}@{}}
\toprule
Model & Component & Layers with $|M|$ or $|R|\ge0.10$ ($M\,|\,R$) \\
\midrule
Uni-NaVid & decision state, attention & $L_{18}$ $+0.13\,|\,+0.15$; $L_{19}$ $+0.19\,|\,+0.19$; $L_{20}$ $+0.16\,|\,+0.16$; $L_{26}$ $+0.19\,|\,+0.19$ \\
 & decision state, MLP & $L_{25}$ $+0.15\,|\,+0.15$; $L_{28}$ $-0.13\,|\,-0.13$; $L_{31}$ $-0.46\,|\,-0.45$ \\
 & prompt tokens, attention & $L_{11}$ $+0.13\,|\,+0.18$; $L_{13}$ $+0.08\,|\,+0.10$; $L_{14}$ $+0.19\,|\,+0.19$; $L_{15}$ $+0.20\,|\,+0.23$ \\
 & prompt tokens, MLP & $L_{15}$ $+0.15\,|\,+0.16$; $L_{16}$ $+0.12\,|\,+0.13$; $L_{17}$ $+0.12\,|\,+0.12$; $L_{18}$ $+0.11\,|\,+0.10$ \\
\addlinespace
LightNav-0 & prompt tokens, attention & $L_{18}$ $+0.11\,|\,+0.08$; $L_{21}$ $+0.17\,|\,+0.15$; $L_{24}$ $+0.11\,|\,+0.04$ \\
\addlinespace
OmTrackVLA & decision state, attention & $L_{25}$ $+0.27\,|\,+0.27$; $L_{26}$ $+0.50\,|\,+0.50$ \\
\bottomrule
\end{tabular}
\end{table*}

\begin{table*}[tbp]
\centering
\footnotesize
\setlength{\tabcolsep}{3pt}
\caption{Strongest single attention heads (stage 2, all events; LightNav-0: continuous readout $T_c$), ranked by
$(M+R)/2$. Heads were searched only in the layers stage 1 selected on half A (Uni-NaVid: decision-state layers [18, 19, 20, 26], prompt-token layers [11, 14, 15]; LightNav-0: decision-state layers none, prompt-token layers [21]; OmTrackVLA: decision-state layers [25, 26], prompt-token layers none).}
\label{tab:app_patch_heads}
\begin{tabular}{@{}llcc@{}}
\toprule
Model & Head & Brings in, $M$ & Removes, $R$ \\
\midrule
Uni-NaVid & $L_{19}$ h06 (decision) & +0.21 {\tiny[+0.19, +0.23]} & +0.22 {\tiny[+0.20, +0.24]} \\
 & $L_{26}$ h03 (decision) & +0.14 {\tiny[+0.12, +0.15]} & +0.14 {\tiny[+0.12, +0.16]} \\
 & $L_{15}$ h20 (prompt) & +0.07 {\tiny[+0.04, +0.11]} & +0.08 {\tiny[+0.05, +0.11]} \\
 & $L_{18}$ h26 (decision) & +0.06 {\tiny[+0.05, +0.08]} & +0.07 {\tiny[+0.06, +0.09]} \\
\addlinespace
LightNav-0 & $L_{21}$ h18 (prompt) & +0.08 {\tiny[+0.03, +0.12]} & +0.12 {\tiny[+0.07, +0.17]} \\
 & $L_{21}$ h19 (prompt) & +0.05 {\tiny[+0.02, +0.08]} & +0.06 {\tiny[+0.03, +0.10]} \\
 & $L_{21}$ h28 (prompt) & +0.02 {\tiny[0.00, +0.04]} & +0.02 {\tiny[0.00, +0.06]} \\
 & $L_{21}$ h31 (prompt) & +0.04 {\tiny[+0.02, +0.06]} & 0.00 {\tiny[$-$0.02, +0.02]} \\
\addlinespace
OmTrackVLA & $L_{26}$ h00 (decision) & +0.22 {\tiny[+0.20, +0.24]} & +0.22 {\tiny[+0.19, +0.24]} \\
 & $L_{25}$ h05 (decision) & +0.12 {\tiny[+0.10, +0.13]} & +0.11 {\tiny[+0.10, +0.13]} \\
 & $L_{26}$ h05 (decision) & +0.08 {\tiny[+0.07, +0.09]} & +0.07 {\tiny[+0.06, +0.09]} \\
 & $L_{25}$ h04 (decision) & +0.06 {\tiny[+0.05, +0.07]} & +0.07 {\tiny[+0.06, +0.08]} \\
\bottomrule
\end{tabular}
\end{table*}

\begin{figure}[tbp]
\centering
\includegraphics[width=\columnwidth]{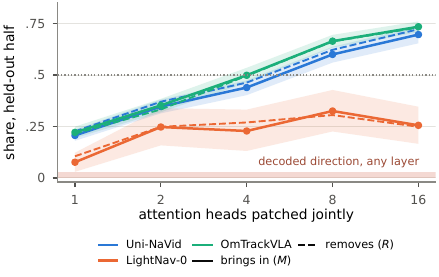}
\caption{Attention heads patched jointly: share of the mirror's effect on the turn that the top 1--16 heads of each model's family (ranked on half A) bring in ($M$, solid) and remove ($R$, dashed) on the held-out half B. Uni-NaVid and OmTrackVLA: decision-state heads; LightNav-0: prompt-token heads, continuous readout $T_c$. Shaded band: the decoded direction's largest share at any layer. Bands: 95\% episode-bootstrap intervals. Values: \cref{tab:app_patch_sets}.}
\label{fig:app_patch_heads}
\end{figure}

\begin{table*}[tbp]
\centering
\footnotesize
\setlength{\tabcolsep}{4pt}
\caption{Head sets patched jointly (stage 3): $M\,|\,R$ on the held-out half B (205 events) and on half A. Sets are
the top-$k$ heads of each family ranked on half A; ``both families'' joins the top 4 or top 8 of each (8 or 16 heads).
On B with 95\% intervals: Uni-NaVid, top 8 decision-state heads: $M$ +0.60 {\tiny[+0.56, +0.64]}, $R$ +0.62 {\tiny[+0.59, +0.66]}; LightNav-0, top 8 prompt-token heads: $M$ +0.33 {\tiny[+0.23, +0.43]}, $R$ +0.31 {\tiny[+0.22, +0.38]}; OmTrackVLA, top 8 decision-state heads: $M$ +0.66 {\tiny[+0.63, +0.70]}, $R$ +0.66 {\tiny[+0.62, +0.71]}.}
\label{tab:app_patch_sets}
\begin{tabular}{@{}lllccccc@{}}
\toprule
Model & Family & Half & 1 head & 2 & 4 & 8 & 16 \\
\midrule
Uni-NaVid & decision-state heads & B & $0.21\,|\,0.23$ & $0.35\,|\,0.37$ & $0.44\,|\,0.46$ & $0.60\,|\,0.62$ & $0.70\,|\,0.72$ \\
 & & A & $0.21\,|\,0.21$ & $0.34\,|\,0.34$ & $0.47\,|\,0.48$ & $0.62\,|\,0.62$ & $0.71\,|\,0.72$ \\
 & prompt-token heads & B & $0.05\,|\,0.08$ & $0.11\,|\,0.15$ & $0.22\,|\,0.26$ & $0.42\,|\,0.48$ & $0.59\,|\,0.66$ \\
 & & A & $0.09\,|\,0.09$ & $0.15\,|\,0.15$ & $0.26\,|\,0.25$ & $0.43\,|\,0.46$ & $0.58\,|\,0.64$ \\
 & both families & B & -- & -- & -- & $0.54\,|\,0.58$ & $0.75\,|\,0.80$ \\
 & & A & -- & -- & -- & $0.57\,|\,0.58$ & $0.74\,|\,0.78$ \\
\addlinespace
LightNav-0 & prompt-token heads, $T_c$ & B & $0.08\,|\,0.11$ & $0.25\,|\,0.25$ & $0.23\,|\,0.27$ & $0.33\,|\,0.31$ & $0.26\,|\,0.25$ \\
 & & A & $0.08\,|\,0.13$ & $0.26\,|\,0.29$ & $0.25\,|\,0.28$ & $0.39\,|\,0.36$ & $0.31\,|\,0.27$ \\
\addlinespace
OmTrackVLA & decision-state heads & B & $0.22\,|\,0.21$ & $0.35\,|\,0.33$ & $0.50\,|\,0.50$ & $0.66\,|\,0.66$ & $0.73\,|\,0.73$ \\
 & & A & $0.21\,|\,0.22$ & $0.34\,|\,0.35$ & $0.49\,|\,0.51$ & $0.65\,|\,0.67$ & $0.73\,|\,0.76$ \\
\bottomrule
\end{tabular}
\end{table*}

\begin{table*}[tbp]
\centering
\footnotesize
\setlength{\tabcolsep}{4pt}
\caption{Person columns versus background columns of the current frame's fine-grid tokens, at the first even-numbered state
(the visual-group patches are run at every second state) where the current frame carries the whole effect ($M\ge0.95$). Person columns overlap the person's bounding box in the
original or the mirrored decision frame (each model's own preprocessing); events without a person column are not
counted. Per column: group share divided by its mean column count (derived).}
\label{tab:app_patch_person}
\begin{tabular}{@{}lccccccc@{}}
\toprule
Model & State & Events & Person, $M\,|\,R$ & Background, $M\,|\,R$ & Columns (person / bg.) & Per column, person / bg. \\
\midrule
Uni-NaVid & $L_{0}$ & 304 & +0.53 {\tiny[+0.45, +0.61]}\,$|$\,+0.57 & +0.28 {\tiny[+0.23, +0.33]}\,$|$\,+0.31 & 3.4 / 4.6 & 2.5$\times$ \\
LightNav-0 & $L_{4}$ & 336 & +0.74 {\tiny[+0.68, +0.81]}\,$|$\,+0.73 & +0.45 {\tiny[+0.38, +0.52]}\,$|$\,+0.43 & 5.0 / 9.0 & 3.0$\times$ \\
OmTrackVLA & $L_{0}$ & 337 & +0.71 {\tiny[+0.66, +0.75]}\,$|$\,+0.70 & +0.32 {\tiny[+0.27, +0.37]}\,$|$\,+0.32 & 3.3 / 4.7 & 3.1$\times$ \\
\bottomrule
\end{tabular}
\end{table*}

This section patches the components and input positions of the interchange test's runs, in three stages.

\paragraph{Events, patches and readout.}
The events, replay, mirrored inputs and turn readout $T$ are those of the interchange test: 454 command-free
rotation-mirror events in 148 MP3D episodes, with patches on the prefill of the first decision that sees the rotation. Every
patch is run in both directions. \emph{Forward} copies a component of the mirrored run into the original run and
measures the share of the mirror's turn change that it brings in ($M$, as in the interchange test); \emph{reverse}
copies the original component into the mirrored run and measures the share of the change that it removes ($R$). The
episodes are split by their index into a half A (249 events) and a held-out half B (205 events); layers and heads
are ranked on A and read on B; the split is by episode, so both halves contain all 10 scenes and the selection is held
out by episode, not by scene. Intervals are 95\% episode-bootstrap intervals
(5{,}000 resamples, seed 0). Because LightNav-0's discrete turn is noisy at the scale of single heads, its selections
use a second, continuous readout, the expected turn under the probabilities of its first generated
action code ($T_c$); the two readouts agree within $0.04$ at every head set. The interchange test's checks
hold at every stage for every model: the replay reproduces the recorded turn on every valid event, the prediction after each
patch is restored exactly, and every prompt state before the current frame is identical between the original and the
mirrored run.

\paragraph{Stages.}
(1)~Route map and components: residual patches of the visual tokens of the current frame, of the prompt tokens after the video (Uni-NaVid: its instruction; LightNav-0: the rest of its prompt; OmTrackVLA has none) and of the decision state at every layer; the attention and MLP
outputs at the decision state and at the prompt tokens at every layer; and the current frame's fine-grid tokens split
into the image columns that hold the person and the remaining columns. (2)~Single attention heads of the layers whose
attention output carries $M\ge0.10$ or $R\ge0.10$ on half A (at most six decision-state and three prompt-token layers).
(3)~Head sets: the top 1, 2, 4, 8 and 16 heads of each family ranked on half A by $(M+R)/2$, and for Uni-NaVid the
union of the top 4 or top 8 of both families, each set patched jointly.

\paragraph{Route.}
\Cref{fig:internals_patch_search} shows the forward shares; as a consistency check, the reverse shares differ from
them by at most $0.09$. The prompt tokens are a route of the computation; the patches do not show that the
instruction's content causes the turn.

\paragraph{Components and heads.}
\Cref{tab:app_patch_components} lists the components with $|M|$ or $|R|\ge0.10$; in OmTrackVLA every MLP carries at
most $0.06$, and in LightNav-0 no decision-state component reaches $0.10$. \Cref{tab:app_patch_heads} gives the
strongest single heads.

\paragraph{Head sets.}
\Cref{fig:app_patch_heads} and \cref{tab:app_patch_sets} give every set on both halves. As a held-out check of the
selection, the shares on halves A and B agree within $0.05$ (Uni-NaVid), $0.03$ (OmTrackVLA) and $0.07$ (LightNav-0).

\paragraph{Person and background columns.}
\Cref{tab:app_patch_person} patches the current frame's fine-grid tokens in two groups, the image columns that overlap
the person and the remaining columns; each group is patched on its own, so the two shares need not add up to the whole
frame's. Per column, the person columns carry $2.5$--$3.1$ times the background columns' share on MP3D, $2.2$--$3.5$
times on HM3D and $10.4$--$22.3$ times on HSSD; the background columns carry $0.28$--$0.45$, $0.23$--$0.34$ and
$0.08$--$0.12$ of the effect (\cref{tab:app_patch_person,tab:app_patch_xd}).

\paragraph{HM3D and HSSD.}
The search is repeated on HM3D and HSSD as an ablation (\cref{app:internals_patch_xd}).

\paragraph{Scope.}
The study covers the first decision after the rotation and rotations of at least $5^\circ$ (median
$20^\circ$). Shares are joint
effects of patched sets and do not add across patches. Heads were searched in the layers stage 1 selected, in sets of
up to 16.

\FloatBarrier
\subsection{Controls and Ablations}
\label{app:internals_controls}

This subsection collects the controls and ablations of the analyses above: of the motion evidence
(\cref{app:internals_shuffle}), of the coarse history
(\cref{app:internals_finer,app:internals_projector,app:internals_real_validation}) and of the rotation maps
(\cref{app:internals_fixed_layer,app:internals_permutation,app:internals_preevent,app:internals_plan,app:internals_onpolicy}),
and the patch search on HM3D and HSSD (\cref{app:internals_patch_xd}).
Checks that are rows of a result table (replay checks, positive controls, random directions, the coin-flip arm)
stay with their tests (\cref{app:internals_functional,app:internals_grp_use}).

\subsubsection{Within-Episode Frame Shuffle}
\label{app:internals_shuffle}

The previous frame is replaced with another frame from the same episode
at least five steps away, while the motion target remains that of the
original consecutive pair. The control keeps the scene, the probe and the fitting unchanged and breaks only the
consecutive-frame correspondence, so a readout that survives it does not come from that correspondence. It runs on
each policy's own features on the lossless recordings of all three datasets (\cref{tab:app_shuffle_policies}) and,
on Uni-NaVid's recordings of the three datasets, on intermediate EVA-CLIP layers and pixels, and on KITTI
(\cref{tab:app_shuffle}; not tabulated: EVA-CLIP $L_{39}$ current-forward $b$ on the MP3D JPEG recordings
$0.53\rightarrow-0.04$).

\begin{table}[tbp]
\centering
\footnotesize
\setlength{\tabcolsep}{4pt}
\caption{Clean-motion $R^2$ after breaking consecutive-frame correspondence (Uni-NaVid's recordings: MP3D JPEG, HM3D and HSSD lossless; KITTI: real video, EVA-CLIP layer $L_{20}$, not the fixed real-video site $L_{39}$ of \cref{tab:app_real_compression}). F/L/Y denote forward, lateral, and yaw.}
\label{tab:app_shuffle}
\begin{tabular}{llrr}
\toprule
Dataset & Input & Original F/L/Y & Shuffled F/L/Y \\
\midrule
MP3D & EVA-CLIP $L_{6}$ & 0.97 / 0.79 / 0.97 & 0.03 / 0.00 / 0.01 \\
 & EVA-CLIP $L_{39}$ & 0.89 / 0.63 / 0.96 & 0.10 / $-$0.01 / 0.03 \\
 & Pixels & 0.87 / 0.58 / 0.94 & 0.00 / 0.00 / 0.00 \\
\midrule
HM3D & EVA-CLIP $L_{6}$ & 0.96 / 0.79 / 0.96 & 0.01 / 0.00 / 0.01 \\
 & EVA-CLIP $L_{39}$ & 0.89 / 0.65 / 0.95 & 0.06 / $-$0.02 / 0.04 \\
 & Pixels & 0.84 / 0.57 / 0.90 & 0.00 / $-$0.03 / 0.02 \\
\midrule
HSSD & EVA-CLIP $L_{6}$ & 0.92 / 0.71 / 0.96 & 0.04 / 0.00 / 0.03 \\
 & EVA-CLIP $L_{39}$ & 0.81 / 0.59 / 0.95 & 0.04 / 0.01 / 0.05 \\
 & Pixels & 0.76 / 0.49 / 0.90 & 0.09 / 0.01 / 0.05 \\
\midrule
KITTI & EVA-CLIP $L_{20}$ & 0.40 / 0.65 / 0.97 & 0.01 / 0.02 / 0.08 \\
\bottomrule
\end{tabular}
\end{table}

\begin{table}[tbp]
    \centering
    \footnotesize
    \setlength{\tabcolsep}{3pt}
    \caption{Frame-shuffle control on each policy's own features. Comparison probe with the consecutive previous frame (intact) or a frame of the same episode $\geq5$ steps away (shuffled); targets unchanged; lossless recordings (Uni-NaVid's intermediate layers and pixels, and its MP3D JPEG run: \cref{tab:app_shuffle}).}
    \label{tab:app_shuffle_policies}
\resizebox{\columnwidth}{!}{%
\begin{tabular}{@{}lllccccc@{}}
\toprule
 & & & Clean $R^2$ & \multicolumn{2}{c}{Current $b$} & \multicolumn{2}{c}{Gust $b$} \\
\cmidrule(lr){4-4}\cmidrule(lr){5-6}\cmidrule(lr){7-8}
Model & Dataset & Prev.\ frame & fwd & fwd & lat & fwd & lat \\
\midrule
Uni-NaVid & MP3D & intact & 0.89 & 0.54 & 0.35 & 0.55 & 0.40 \\
 &  & shuffled & 0.09 & $-$0.02 & 0.00 & 0.02 & 0.00 \\
 & HM3D & intact & 0.89 & 0.55 & 0.36 & 0.58 & 0.38 \\
 &  & shuffled & 0.06 & $-$0.01 & $-$0.01 & 0.01 & 0.00 \\
 & HSSD & intact & 0.81 & 0.28 & 0.13 & 0.31 & 0.18 \\
 &  & shuffled & 0.04 & $-$0.11 & $-$0.01 & 0.01 & 0.00 \\
\midrule
OmTrackVLA & MP3D & intact & 0.94 & 0.65 & 0.38 & 0.60 & 0.44 \\
 &  & shuffled & 0.13 & $-$0.01 & $-$0.01 & 0.03 & $-$0.01 \\
 & HM3D & intact & 0.94 & 0.62 & 0.39 & 0.62 & 0.40 \\
 &  & shuffled & 0.06 & 0.02 & 0.00 & 0.05 & 0.00 \\
 & HSSD & intact & 0.85 & 0.38 & 0.15 & 0.39 & 0.19 \\
 &  & shuffled & 0.04 & $-$0.04 & $-$0.01 & 0.00 & 0.00 \\
\midrule
LightNav-0 & MP3D & intact & 0.78 & 0.35 & 0.17 & 0.38 & 0.21 \\
 &  & shuffled & 0.14 & 0.01 & $-$0.01 & 0.03 & $-$0.01 \\
 & HM3D & intact & 0.72 & 0.35 & 0.21 & 0.39 & 0.21 \\
 &  & shuffled & 0.11 & $-$0.03 & $-$0.01 & 0.03 & 0.00 \\
 & HSSD & intact & 0.64 & 0.06 & 0.06 & 0.23 & 0.08 \\
 &  & shuffled & 0.05 & $-$0.08 & 0.00 & 0.01 & 0.00 \\
\bottomrule
\end{tabular}
}
\end{table}

\subsubsection{Finer-Rung Ablation and Raw Pixels (Uni-NaVid)}
\label{app:internals_finer}

For Uni-NaVid, a sweep on the MP3D JPEG and the HM3D and HSSD lossless recordings (\cref{tab:app_compression}) adds symmetric pairs at $16\times16$, $8\times8$, $4\times4$, and
$2\times2$, followed by the separate native asymmetric condition:
$2\times2$ previous-frame tokens and $8\times8$ current-frame tokens.
The asymmetric condition is not a point of the symmetric ladder; against
the $8\times8$ pair it changes only the previous frame's resolution, with
encoder, observations, targets and fitting fixed.

The lossless re-recording of MP3D reproduces its sweep (a re-recording, not a same-frame control): all 50 comparison
cells agree within $0.017$ in $R^2$ and $0.030$ in $b$. At $2\times2$ and native mixed every $|b|$ is at most $0.09$
on the three datasets, except EVA-CLIP's native mixed gust forward readout on HM3D ($0.17$ [$0.11$, $0.22$]).

\begin{table*}[tbp]
\centering
\footnotesize
\setlength{\tabcolsep}{4pt}
\caption{Complete translational compression ablation (Uni-NaVid's recordings: MP3D JPEG, HM3D and HSSD lossless), with 95\% intervals. Native mixed means previous $2\times2$ plus current $8\times8$.}
\label{tab:app_compression}
\begin{tabular}{lllrrrr}
\toprule
Dataset & Input & Grid & Current F $b$ & Current L $b$ & Gust F $b$ & Gust L $b$ \\
\midrule
MP3D & EVA-CLIP & $16\times16$ & 0.53 [0.44, 0.61] & 0.34 [0.31, 0.38] & 0.54 [0.48, 0.60] & 0.40 [0.38, 0.43] \\
 & EVA-CLIP & $8\times8$ & 0.36 [0.28, 0.43] & 0.24 [0.21, 0.27] & 0.43 [0.37, 0.49] & 0.27 [0.25, 0.29] \\
 & EVA-CLIP & $4\times4$ & 0.17 [0.08, 0.25] & 0.16 [0.13, 0.18] & 0.26 [0.21, 0.32] & 0.16 [0.15, 0.18] \\
 & EVA-CLIP & $2\times2$ & $-$0.04 [$-$0.11, 0.01] & 0.05 [0.03, 0.08] & 0.01 [$-$0.01, 0.04] & 0.05 [0.04, 0.05] \\
 & EVA-CLIP & Native mixed & 0.05 [$-$0.04, 0.13] & 0.08 [0.06, 0.12] & 0.09 [0.05, 0.14] & 0.06 [0.05, 0.07] \\
 & Pixels & $16\times16$ & 0.51 [0.43, 0.60] & 0.29 [0.26, 0.33] & 0.52 [0.45, 0.58] & 0.34 [0.32, 0.36] \\
 & Pixels & $8\times8$ & 0.37 [0.31, 0.43] & 0.18 [0.15, 0.20] & 0.40 [0.35, 0.45] & 0.22 [0.20, 0.23] \\
 & Pixels & $4\times4$ & 0.15 [0.08, 0.22] & 0.10 [0.08, 0.11] & 0.12 [0.08, 0.17] & 0.12 [0.11, 0.14] \\
 & Pixels & $2\times2$ & $-$0.02 [$-$0.04, 0.00] & 0.01 [0.00, 0.02] & $-$0.01 [$-$0.02, 0.00] & 0.01 [0.01, 0.02] \\
 & Pixels & Native mixed & $-$0.03 [$-$0.08, 0.02] & 0.02 [0.00, 0.03] & 0.00 [$-$0.03, 0.03] & 0.03 [0.02, 0.03] \\
\midrule
HM3D & EVA-CLIP & $16\times16$ & 0.55 [0.49, 0.62] & 0.36 [0.32, 0.39] & 0.58 [0.52, 0.64] & 0.38 [0.34, 0.41] \\
 & EVA-CLIP & $8\times8$ & 0.47 [0.40, 0.53] & 0.28 [0.25, 0.31] & 0.51 [0.45, 0.56] & 0.28 [0.26, 0.31] \\
 & EVA-CLIP & $4\times4$ & 0.23 [0.17, 0.29] & 0.17 [0.15, 0.19] & 0.32 [0.26, 0.38] & 0.17 [0.16, 0.19] \\
 & EVA-CLIP & $2\times2$ & 0.00 [$-$0.05, 0.04] & 0.04 [0.03, 0.05] & $-$0.01 [$-$0.03, 0.02] & 0.04 [0.02, 0.05] \\
 & EVA-CLIP & Native mixed & 0.09 [0.04, 0.14] & 0.09 [0.08, 0.11] & 0.17 [0.11, 0.22] & 0.09 [0.08, 0.11] \\
 & Pixels & $16\times16$ & 0.51 [0.44, 0.59] & 0.28 [0.24, 0.32] & 0.58 [0.51, 0.65] & 0.27 [0.24, 0.29] \\
 & Pixels & $8\times8$ & 0.37 [0.30, 0.43] & 0.22 [0.19, 0.24] & 0.42 [0.35, 0.48] & 0.21 [0.19, 0.23] \\
 & Pixels & $4\times4$ & 0.10 [0.04, 0.17] & 0.13 [0.11, 0.15] & 0.15 [0.08, 0.21] & 0.12 [0.11, 0.14] \\
 & Pixels & $2\times2$ & $-$0.01 [$-$0.03, 0.00] & 0.02 [0.01, 0.02] & 0.02 [0.01, 0.03] & 0.01 [0.01, 0.02] \\
 & Pixels & Native mixed & 0.04 [$-$0.01, 0.08] & 0.03 [0.02, 0.05] & 0.06 [0.01, 0.10] & 0.03 [0.02, 0.05] \\
\midrule
HSSD & EVA-CLIP & $16\times16$ & 0.28 [0.17, 0.38] & 0.13 [0.11, 0.16] & 0.31 [0.24, 0.38] & 0.18 [0.15, 0.22] \\
 & EVA-CLIP & $8\times8$ & 0.21 [0.09, 0.32] & 0.10 [0.07, 0.12] & 0.28 [0.21, 0.34] & 0.14 [0.12, 0.17] \\
 & EVA-CLIP & $4\times4$ & 0.08 [$-$0.05, 0.20] & 0.06 [0.05, 0.08] & 0.18 [0.10, 0.25] & 0.11 [0.09, 0.12] \\
 & EVA-CLIP & $2\times2$ & 0.02 [$-$0.06, 0.10] & 0.02 [0.00, 0.04] & 0.02 [$-$0.05, 0.09] & 0.04 [0.03, 0.05] \\
 & EVA-CLIP & Native mixed & $-$0.06 [$-$0.15, 0.03] & 0.04 [0.02, 0.06] & 0.07 [0.01, 0.13] & 0.07 [0.05, 0.08] \\
 & Pixels & $16\times16$ & 0.34 [0.26, 0.42] & 0.10 [0.08, 0.13] & 0.28 [0.21, 0.34] & 0.13 [0.11, 0.15] \\
 & Pixels & $8\times8$ & 0.22 [0.13, 0.29] & 0.10 [0.08, 0.13] & 0.23 [0.17, 0.29] & 0.11 [0.09, 0.13] \\
 & Pixels & $4\times4$ & 0.03 [$-$0.06, 0.13] & 0.04 [0.02, 0.06] & 0.12 [0.06, 0.17] & 0.05 [0.04, 0.07] \\
 & Pixels & $2\times2$ & 0.04 [$-$0.03, 0.12] & $-$0.01 [$-$0.02, 0.01] & 0.06 [0.01, 0.10] & 0.00 [$-$0.01, 0.01] \\
 & Pixels & Native mixed & 0.03 [$-$0.06, 0.10] & 0.00 [$-$0.01, 0.02] & 0.06 [0.02, 0.10] & 0.01 [0.00, 0.02] \\
\bottomrule
\end{tabular}
\end{table*}

\subsubsection{Projector Ablation on the Projected Tokens (Uni-NaVid)}
\label{app:internals_projector}

The comparison is repeated on Uni-NaVid's projected visual tokens, the tokens its language model receives, at the
$8\times8$, $2\times2$ and native mixed grids on the three datasets (Table~\ref{tab:app_projector}). At $2\times2$
and native mixed every $|b|$ is at most $0.09$, except the native mixed current forward readout on HSSD, which is
negative ($-0.18$ [$-0.32$, $-0.05$]).

\begin{table*}[tbp]
\centering
\footnotesize
\setlength{\tabcolsep}{2pt}
\caption{Projector ablation (Uni-NaVid; MP3D JPEG, HM3D and HSSD lossless recordings): comparison readout on the projected tokens. Intervals: 95\% estimates from this fit.}
\label{tab:app_projector}
\begin{tabular}{@{}llrrr@{}}
\toprule
Dataset & Grid & Yaw $R^2$ & Current F $b$ & Current L $b$ \\
\midrule
MP3D & $8\times8$ & 0.94 & 0.33 [0.23, 0.43] & 0.20 [0.17, 0.22] \\
 & $2\times2$ & 0.16 & $-$0.05 [$-$0.11, 0.00] & 0.04 [0.03, 0.07] \\
 & Native mixed & 0.18 & $-$0.03 [$-$0.12, 0.03] & 0.06 [0.05, 0.08] \\
\midrule
HM3D & $8\times8$ & 0.92 & 0.39 [0.32, 0.47] & 0.22 [0.20, 0.25] \\
 & $2\times2$ & 0.15 & $-$0.03 [$-$0.07, 0.01] & 0.04 [0.03, 0.06] \\
 & Native mixed & 0.22 & 0.02 [$-$0.02, 0.08] & 0.09 [0.08, 0.11] \\
\midrule
HSSD & $8\times8$ & 0.93 & 0.14 [0.02, 0.25] & 0.08 [0.06, 0.10] \\
 & $2\times2$ & 0.35 & 0.04 [$-$0.06, 0.13] & 0.02 [0.00, 0.03] \\
 & Native mixed & 0.25 & $-$0.18 [$-$0.32, $-$0.05] & 0.03 [0.01, 0.05] \\
\bottomrule
\end{tabular}
\end{table*}

\subsubsection{Real Video: Temporal and Geometric Validation}
\label{app:internals_real_validation}

Independent visual-odometry checks of the converted recordings give zero-lag yaw correlations of $0.976$, $0.861$
and $0.827$ for TUM, KITTI and TartanDrive, and direction agreement of approximately $100\%$, $99.3\%$ and $97.2\%$
on the selected nontrivial rotations. These validate labels/alignment; they are not probe scores. TartanDrive
intrinsics are approximate.

\subsubsection{Fixed-Layer Replication}
\label{app:internals_fixed_layer}

Each dataset's own $k^*$ is selected on that dataset's map and is therefore
selection-inflated; \cref{tab:app_rotation_xd} also reads HM3D and HSSD at MP3D's $k^*$, a layer not selected on
them, and every dataset at the final state.

\subsubsection{Label-Permutation Control}
\label{app:internals_permutation}

Refitted with the clean fitting targets permuted within scenes, the decision-state rotation probes
score $0.498$, $0.514$ and $0.500$ at the $k^*$ of Uni-NaVid, LightNav-0 and OmTrackVLA on MP3D ($0.531$, $0.529$
and $0.506$ on HM3D and $0.479$, $0.481$ and $0.479$ on HSSD, at each dataset's own $k^*$), and stay within
$0.05$ of $0.5$ at every decision state on every dataset, with no interval excluding chance; the readout does not arise from the probe
fitting arbitrary labels \cite{hewitt2019control}.

\subsubsection{Command-Free Events and Leak Control}
\label{app:internals_preevent}

The rotation maps use the imposed rotations without a turn command at the preceding decision (MP3D 486 of 1{,}180,
HM3D 429 of 1{,}074, HSSD 459 of 1{,}177): commanded
turns predict the rotation's direction on many unrestricted events. Probes are fitted on clean driving, never on
rotation labels. On the complete maps, the leak control, the decision state one step before the rotation, reads $0.41$--$0.56$ at
its state farthest from chance on the three datasets (\cref{tab:app_crossmodel}). Six of the nine intervals exclude
$0.5$: all three models on HM3D from above ($0.555$--$0.563$, lower bounds $0.508$--$0.514$), and from below
LightNav-0 on MP3D ($0.443$ [$0.400$, $0.484$]) and Uni-NaVid and LightNav-0 on HSSD ($0.444$ [$0.403$, $0.484$] and
$0.415$ [$0.367$, $0.464$]). These pre-rotation readouts lie at most $0.09$ from chance, against $0.13$--$0.44$ at the
decision states after the rotation (\cref{tab:app_rotation_xd}). No site's leak-control accuracy has a lower bound above
$0.55$ on any dataset (297, 222 and 261 sites of the three models). Events need not remove the target from view (\cref{app:internals_mirror}).

\subsubsection{Partial Correlations with the Rotation and the Model's Own Turn}
\label{app:internals_plan}

After an imposed rotation, the rotation and the model's corrective turn point opposite ways; partial correlations
separate the two. On MP3D, the readout's partial correlation with the rotation, controlling for the model's own turn, is $+0.66$ [$0.62$, $0.71$] for Uni-NaVid at $L_{16}$ and $+0.84$ [$0.81$, $0.86$] for LightNav-0 at $L_{20}$,
but $-0.20$ [$-0.30$, $-0.09$] for OmTrackVLA at $L_{11}$, whose readout correlates with its own corrective turn
($+0.44$). OmTrackVLA's partial correlation with the rotation is positive only at its final state ($+0.12$ [$0.03$,
$0.21$]). On HM3D and HSSD, at each dataset's $k^*$, it is $+0.66$ [$0.61$, $0.72$] and $+0.59$ [$0.52$, $0.66$] for
Uni-NaVid and $+0.83$ [$0.80$, $0.87$] and $+0.77$ [$0.73$, $0.80$] for LightNav-0; for OmTrackVLA it is $-0.29$
[$-0.40$, $-0.17$] and $-0.02$ [$-0.16$, $0.11$] (correlation with its own turn: $+0.67$ and $+0.50$), positive only
at its final state on HM3D ($+0.12$ [$0.01$, $0.23$]) and from $L_{20}$ on HSSD (final state: $+0.32$ [$0.23$,
$0.40$]).

\subsubsection{On-Policy Rotation Maps}
\label{app:internals_onpolicy}

\paragraph{Closed-loop recordings.}
The cross-model maps replay Uni-NaVid's trajectories. To test whether their
findings depend on that, LightNav-0 and OmTrackVLA each drive the episodes of
the map collections (MP3D 150, HM3D 135, HSSD 148) in their own closed loop under the same rotation protocol: the same
scenes, target persons, instructions and rotation schedule, which is seeded per
episode. Every
frame a model reads is stored losslessly; LightNav-0's client sends lossless
frames, as in the replay maps, instead of its default JPEG. The frames are then
replayed through the capture used for the maps, and the replay reproduces each
live action exactly (float32) on every step: MP3D 18{,}537 (LightNav-0 9{,}986, OmTrackVLA
8{,}551), HM3D 16{,}269 (8{,}512 / 7{,}757), HSSD 19{,}058 (9{,}662 / 9{,}396). On each dataset, LightNav-0 emits an unparseable output on 3 steps; the evaluator repeats
the previous action, as in the benchmark, and the replay reproduces the same
outputs.

\paragraph{Result.} On its own trajectories, LightNav-0's maps match the replay: its decision state reads the side of
near-zero-command rotations at $0.947$, $0.906$ and $0.868$ (replay, command-free: $0.938$, $0.935$, $0.915$), and the
token-group profiles agree (\cref{tab:app_onpolicy}). OmTrackVLA's decision-state readout keeps its reversed sign but
lies nearer chance on its own trajectories ($0.408$, $0.376$, $0.441$ against $0.342$, $0.268$, $0.331$); on HSSD
its interval includes $0.5$ ($[0.382, 0.501]$), and on HM3D so does its all-events readout ($0.485$ [$0.447$, $0.524$]).

\begin{table*}[tbp]
\centering
\footnotesize
\setlength{\tabcolsep}{4pt}
\caption{On-policy versus replay rotation maps, per dataset. Accuracy [95\% CI] for the
compared rows. First row of each dataset: all events; the rows below it: on-policy near-zero-command events and
replay command-free events.}
\label{tab:app_onpolicy}
\begin{tabular}{lrrrr}
\toprule
 & LightNav-0 on-policy & LightNav-0 replay & OmTrackVLA on-policy & OmTrackVLA replay \\
\midrule
\multicolumn{5}{@{}l}{\emph{MP3D}} \\
Decision state, all events ($L_{20}$ / $L_{11}$) & 0.877 [0.852, 0.900] & 0.919 [0.903, 0.935] & 0.568 [0.521, 0.615] & 0.659 [0.626, 0.691] \\
Decision state, near-zero / command-free & 0.947 [0.920, 0.970] & 0.938 [0.914, 0.959] & 0.408 [0.345, 0.472] & 0.342 [0.296, 0.391] \\
Comparison, prev.\ vs cur., $L_0$ / final state & 0.755 / 0.691 & 0.788 / 0.741 & 0.906 / 0.467 & 0.912 / 0.461 \\
Current-frame tokens, $L_0$ / best state & 0.269 / 0.838 ($L_{18}$) & 0.319 / 0.829 ($L_{18}$) & 0.345 / 0.655 ($L_{7}$) & 0.294 / 0.632 ($L_{23}$) \\
All captured image tokens, best state & 0.870 ($L_{18}$) & 0.852 ($L_{18}$) & 0.600 ($L_{28}$) & 0.574 ($L_{28}$) \\
Decision state, final layer & 0.715 & 0.702 & 0.443 & 0.405 \\
Events (all / near-zero or command-free) & 1{,}240 / 376 & 1{,}180 / 486 & 1{,}071 / 255 & 1{,}180 / 486 \\
\midrule
\multicolumn{5}{@{}l}{\emph{HM3D}} \\
Decision state, all events ($L_{21}$ / $L_{19}$) & 0.816 [0.786, 0.843] & 0.899 [0.876, 0.919] & 0.485 [0.447, 0.524] & 0.615 [0.580, 0.647] \\
Decision state, near-zero / command-free & 0.906 [0.867, 0.941] & 0.935 [0.902, 0.964] & 0.376 [0.305, 0.447] & 0.268 [0.220, 0.321] \\
Comparison, prev.\ vs cur., $L_0$ / final state & 0.798 / 0.688 & 0.830 / 0.730 & 0.897 / 0.509 & 0.900 / 0.487 \\
Current-frame tokens, $L_0$ / best state & 0.221 / 0.801 ($L_{18}$) & 0.270 / 0.807 ($L_{18}$) & 0.355 / 0.628 ($L_{26}$) & 0.252 / 0.594 ($L_{22}$) \\
All captured image tokens, best state & 0.840 ($L_{18}$) & 0.837 ($L_{19}$) & 0.607 ($L_{25}$) & 0.580 ($L_{26}$) \\
Decision state, final layer & 0.727 & 0.706 & 0.483 & 0.399 \\
Events (all / near-zero or command-free) & 1{,}066 / 362 & 1{,}074 / 429 & 984 / 234 & 1{,}074 / 429 \\
\midrule
\multicolumn{5}{@{}l}{\emph{HSSD}} \\
Decision state, all events ($L_{21}$ / $L_{11}$) & 0.844 [0.817, 0.870] & 0.920 [0.903, 0.937] & 0.599 [0.562, 0.634] & 0.681 [0.651, 0.708] \\
Decision state, near-zero / command-free & 0.868 [0.833, 0.902] & 0.915 [0.885, 0.941] & 0.441 [0.382, 0.501] & 0.331 [0.282, 0.380] \\
Comparison, prev.\ vs cur., $L_0$ / final state & 0.781 / 0.623 & 0.769 / 0.586 & 0.839 / 0.531 & 0.854 / 0.442 \\
Current-frame tokens, $L_0$ / best state & 0.292 / 0.621 ($L_{21}$) & 0.281 / 0.610 ($L_{18}$) & 0.395 / 0.725 ($L_{22}$) & 0.298 / 0.662 ($L_{22}$) \\
All captured image tokens, best state & 0.662 ($L_{20}$) & 0.638 ($L_{18}$) & 0.700 ($L_{7}$) & 0.669 ($L_{26}$) \\
Decision state, final layer & 0.655 & 0.590 & 0.569 & 0.516 \\
Events (all / near-zero or command-free) & 1{,}337 / 438 & 1{,}177 / 459 & 1{,}281 / 367 & 1{,}177 / 459 \\
\bottomrule
\end{tabular}
\end{table*}

\paragraph{Readout and comparison.}
The clean fitting rows, probes, folds and event definition are those of the
replay maps. The events are each model's own imposed rotations (\cref{tab:app_onpolicy}, last row of each dataset). Continuous turn commands are almost never exactly zero, so the
command-free subset is replaced by a near-zero-command one, $|$commanded
yaw$|<1^\circ$ at the preceding decision. Compared:
the decision state at each model's primary layer, on all events
and on the near-zero subset against the replay's command-free events.

\paragraph{Leak control and comparability.}
The leak control of the replay maps is repeated on the on-policy captures. On MP3D no leak-control lower bound exceeds
$0.55$ at any of LightNav-0's 222 sites; for OmTrackVLA one of 261 sites does (previous-frame versus navigation
comparison at $L_4$, lower bound 0.552), the decision state does not (0.46 at $L_{11}$). On HM3D and HSSD no site of
either model does. The event sets differ between
the models and from the replay (rotation timing within the episode, number of events, commands), so the comparison is
of each model with itself, not on matched events.

\subsubsection{Patch Search on HM3D and HSSD}
\label{app:internals_patch_xd}

\begin{figure}[tbp]
\centering
\includegraphics[width=\columnwidth]{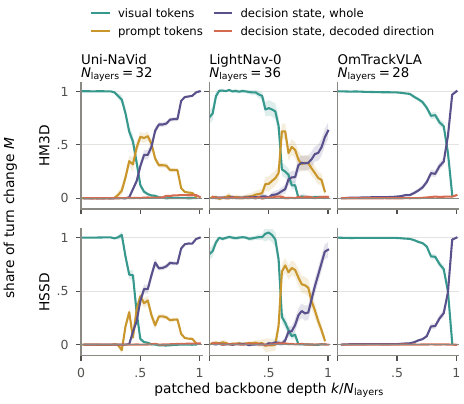}
\caption{Where the mirror's effect on the turn travels on HM3D and HSSD: share $M$ carried by patching each position group at every layer, as \cref{fig:internals_patch_search}. Bands: 95\% episode-bootstrap intervals.}
\label{fig:app_patch_route_xd}
\end{figure}

The patch search of \cref{app:internals_patch_search} is repeated on the rotation-mirror events of HM3D (372 events
in 132 episodes) and HSSD (417 in 146), with the same patches, readout and checks. Stage 1 runs in full. MP3D's head
sets are patched without a new ranking and read on all events (\cref{tab:app_patch_xd}); stages 2 and 3 are also run
with each dataset's own ranking on its half A and read on its half B (\cref{tab:app_patch_own_xd}). The route is that of MP3D
(\cref{fig:app_patch_route_xd}). Components with $|M|$ or $|R|\ge0.10$ on all three datasets: in Uni-NaVid the
decision-state attention at $L_{18}$, $L_{19}$ and $L_{26}$ and the decision-state MLP at $L_{25}$, $L_{28}$ and
$L_{31}$; in OmTrackVLA the decision-state attention at $L_{25}$ and $L_{26}$; in LightNav-0 the prompt-token attention
at $L_{21}$. Uni-NaVid's joint set of four decision-state and four prompt-token heads carries $0.56\,|\,0.56$ on HM3D
and $0.44\,|\,0.45$ on HSSD.

\begin{table*}[tbp]
\centering
\footnotesize
\setlength{\tabcolsep}{4pt}
\caption{The patch search on HM3D (372 events) and HSSD (417), with the MP3D procedure and MP3D's head sets
(\cref{tab:app_patch_sets}); nothing is selected on these datasets, and all shares are on all events. Top: head sets
patched jointly, $M\,|\,R$. $M$ of the top 8 heads of \cref{fig:app_patch_heads}'s family with 95\% intervals, HM3D
and HSSD: Uni-NaVid +0.62 {\tiny[+0.59, +0.64]} and +0.60 {\tiny[+0.57, +0.62]}; LightNav-0 +0.35 {\tiny[+0.29, +0.41]} and +0.25 {\tiny[+0.18, +0.33]}; OmTrackVLA +0.66 {\tiny[+0.64, +0.69]} and +0.70 {\tiny[+0.68, +0.73]}. Bottom: person columns versus background columns, as
\cref{tab:app_patch_person}.}
\label{tab:app_patch_xd}
\begin{tabular}{@{}lllccccc@{}}
\toprule
Model & Family & Dataset & 1 head & 2 & 4 & 8 & 16 \\
\midrule
Uni-NaVid & decision-state heads & HM3D & $0.23\,|\,0.22$ & $0.39\,|\,0.38$ & $0.47\,|\,0.46$ & $0.62\,|\,0.60$ & $0.72\,|\,0.71$ \\
 &  & HSSD & $0.23\,|\,0.22$ & $0.36\,|\,0.36$ & $0.37\,|\,0.38$ & $0.60\,|\,0.60$ & $0.72\,|\,0.72$ \\
 & prompt-token heads & HM3D & $0.07\,|\,0.06$ & $0.13\,|\,0.13$ & $0.22\,|\,0.23$ & $0.43\,|\,0.44$ & $0.60\,|\,0.61$ \\
 &  & HSSD & $-0.08\,|\,-0.05$ & $-0.01\,|\,0.02$ & $0.09\,|\,0.11$ & $0.37\,|\,0.37$ & $0.55\,|\,0.56$ \\
 & both families & HM3D & -- & -- & -- & $0.56\,|\,0.56$ & $0.77\,|\,0.76$ \\
 &  & HSSD & -- & -- & -- & $0.44\,|\,0.45$ & $0.78\,|\,0.78$ \\
\addlinespace
LightNav-0 & prompt-token heads, $T_c$ & HM3D & $0.05\,|\,0.08$ & $0.27\,|\,0.34$ & $0.27\,|\,0.33$ & $0.35\,|\,0.44$ & $0.27\,|\,0.34$ \\
 &  & HSSD & $0.01\,|\,0.02$ & $0.15\,|\,0.17$ & $0.17\,|\,0.16$ & $0.25\,|\,0.22$ & $0.14\,|\,0.16$ \\
\addlinespace
OmTrackVLA & decision-state heads & HM3D & $0.21\,|\,0.21$ & $0.37\,|\,0.36$ & $0.52\,|\,0.51$ & $0.66\,|\,0.66$ & $0.75\,|\,0.74$ \\
 &  & HSSD & $0.25\,|\,0.26$ & $0.35\,|\,0.36$ & $0.52\,|\,0.52$ & $0.70\,|\,0.71$ & $0.77\,|\,0.78$ \\
\bottomrule
\end{tabular}

\vspace{4pt}
\begin{tabular}{@{}llcccccc@{}}
\toprule
Model & Dataset & State & Events & Person, $M\,|\,R$ & Background, $M\,|\,R$ & Columns (person / bg.) & Per column, person / bg. \\
\midrule
Uni-NaVid & HM3D & $L_{0}$ & 234 & +0.76 {\tiny[+0.68, +0.83]}\,$|$\,+0.75 & +0.23 {\tiny[+0.18, +0.29]}\,$|$\,+0.23 & 4.1 / 3.9 & 3.1$\times$ \\
 & HSSD & $L_{0}$ & 325 & +0.90 {\tiny[+0.83, +0.96]}\,$|$\,+0.88 & +0.09 {\tiny[+0.05, +0.13]}\,$|$\,+0.08 & 3.4 / 4.6 & 13.7$\times$ \\
LightNav-0 & HM3D & $L_{2}$ & 256 & +0.84 {\tiny[+0.77, +0.91]}\,$|$\,+0.88 & +0.32 {\tiny[+0.26, +0.38]}\,$|$\,+0.35 & 6.0 / 8.0 & 3.5$\times$ \\
 & HSSD & $L_{2}$ & 361 & +0.95 {\tiny[+0.89, +1.02]}\,$|$\,+0.95 & +0.08 {\tiny[+0.02, +0.14]}\,$|$\,+0.04 & 4.8 / 9.2 & 22.3$\times$ \\
OmTrackVLA & HM3D & $L_{0}$ & 256 & +0.73 {\tiny[+0.69, +0.77]}\,$|$\,+0.76 & +0.34 {\tiny[+0.29, +0.39]}\,$|$\,+0.36 & 3.9 / 4.1 & 2.2$\times$ \\
 & HSSD & $L_{0}$ & 361 & +0.85 {\tiny[+0.82, +0.88]}\,$|$\,+0.83 & +0.12 {\tiny[+0.09, +0.14]}\,$|$\,+0.11 & 3.3 / 4.7 & 10.4$\times$ \\
\bottomrule
\end{tabular}
\end{table*}

\paragraph{Own ranking.}
With each dataset's own ranking, the top 8 decision-state heads carry $0.62\,|\,0.60$ (HM3D) and $0.69\,|\,0.69$
(HSSD) on half B in Uni-NaVid and $0.69\,|\,0.66$ and $0.72\,|\,0.70$ in OmTrackVLA; LightNav-0's top 8 prompt-token
heads carry $0.20\,|\,0.26$ on HM3D and $-0.02\,|\,0.01$ on HSSD, where the heads ranked on half A carry
$M$ $-0.03$ to $-0.01$ on half B. The strongest single head is MP3D's in Uni-NaVid ($L_{19}$ h06) and OmTrackVLA
($L_{26}$ h00) on both datasets. In Uni-NaVid and OmTrackVLA the shares on halves A and B agree within $0.08$ and
$0.06$.

\begin{table*}[tbp]
\centering
\footnotesize
\setlength{\tabcolsep}{4pt}
\caption{The head search on HM3D and HSSD with each dataset's own ranking (stages 2 and 3 of
\cref{app:internals_patch_search}, ranked on the dataset's half A; HM3D halves 176 / 196 events, HSSD 204 / 213). Top: head sets patched jointly, $M\,|\,R$ on the held-out half B and on half A, as
\cref{tab:app_patch_sets}. On B with 95\% intervals: Uni-NaVid, top 8 decision-state heads: HM3D $M$ +0.62 {\tiny[+0.60, +0.65]}, $R$ +0.60 {\tiny[+0.57, +0.64]}; HSSD $M$ +0.69 {\tiny[+0.64, +0.73]}, $R$ +0.69 {\tiny[+0.64, +0.74]}; LightNav-0, top 8 prompt-token heads: HM3D $M$ +0.20 {\tiny[+0.12, +0.29]}, $R$ +0.26 {\tiny[+0.15, +0.37]}; HSSD $M$ $-$0.02 {\tiny[$-$0.05, +0.01]}, $R$ +0.01 {\tiny[$-$0.02, +0.04]}; OmTrackVLA, top 8 decision-state heads: HM3D $M$ +0.69 {\tiny[+0.65, +0.72]}, $R$ +0.66 {\tiny[+0.63, +0.70]}; HSSD $M$ +0.72 {\tiny[+0.68, +0.75]}, $R$ +0.70 {\tiny[+0.66, +0.74]}. Bottom: the two strongest
single heads (all events, ranked by $(M+R)/2$). Decision-state / prompt-token layers searched: HM3D: Uni-NaVid [18, 19, 20, 26] / [11, 14, 15], LightNav-0 [34] / [16, 20, 21], OmTrackVLA [25, 26] / none; HSSD: Uni-NaVid [13, 18, 19, 26] / [11, 13, 14], LightNav-0 none / [24], OmTrackVLA [25, 26] / none.}
\label{tab:app_patch_own_xd}
\begin{tabular}{@{}llllccccc@{}}
\toprule
Model & Family & Dataset & Half & 1 head & 2 & 4 & 8 & 16 \\
\midrule
Uni-NaVid & decision-state heads & HM3D & B & $0.23\,|\,0.22$ & $0.39\,|\,0.38$ & $0.50\,|\,0.48$ & $0.62\,|\,0.60$ & $0.71\,|\,0.70$ \\
 & & & A & $0.22\,|\,0.21$ & $0.39\,|\,0.38$ & $0.48\,|\,0.48$ & $0.62\,|\,0.61$ & $0.72\,|\,0.71$ \\
 &  & HSSD & B & $0.21\,|\,0.21$ & $0.33\,|\,0.34$ & $0.53\,|\,0.54$ & $0.69\,|\,0.69$ & $0.76\,|\,0.78$ \\
 & & & A & $0.24\,|\,0.24$ & $0.37\,|\,0.36$ & $0.57\,|\,0.56$ & $0.71\,|\,0.71$ & $0.79\,|\,0.79$ \\
 & prompt-token heads & HM3D & B & $0.06\,|\,0.04$ & $0.15\,|\,0.12$ & $0.29\,|\,0.24$ & $0.46\,|\,0.44$ & $0.65\,|\,0.61$ \\
 & & & A & $0.07\,|\,0.08$ & $0.12\,|\,0.16$ & $0.24\,|\,0.31$ & $0.42\,|\,0.49$ & $0.61\,|\,0.67$ \\
 &  & HSSD & B & $0.06\,|\,0.08$ & $0.11\,|\,0.14$ & $0.24\,|\,0.27$ & $0.38\,|\,0.40$ & $0.49\,|\,0.52$ \\
 & & & A & $0.09\,|\,0.09$ & $0.14\,|\,0.14$ & $0.28\,|\,0.27$ & $0.41\,|\,0.41$ & $0.53\,|\,0.55$ \\
 & both families & HM3D & B & -- & -- & -- & $0.63\,|\,0.58$ & $0.79\,|\,0.76$ \\
 & & & A & -- & -- & -- & $0.58\,|\,0.61$ & $0.77\,|\,0.79$ \\
 &  & HSSD & B & -- & -- & -- & $0.64\,|\,0.66$ & $0.81\,|\,0.82$ \\
 & & & A & -- & -- & -- & $0.69\,|\,0.69$ & $0.85\,|\,0.85$ \\
\addlinespace
LightNav-0 & prompt-token heads, $T_c$ & HM3D & B & $0.07\,|\,0.06$ & $0.06\,|\,0.08$ & $0.07\,|\,0.06$ & $0.20\,|\,0.26$ & $0.12\,|\,0.15$ \\
 & & & A & $0.02\,|\,0.11$ & $0.02\,|\,0.13$ & $0.04\,|\,0.12$ & $0.15\,|\,0.26$ & $0.09\,|\,0.18$ \\
 &  & HSSD & B & $-0.03\,|\,0.03$ & $-0.01\,|\,0.01$ & $-0.01\,|\,0.01$ & $-0.02\,|\,0.01$ & $-0.02\,|\,0.01$ \\
 & & & A & $0.08\,|\,0.02$ & $0.06\,|\,0.01$ & $0.07\,|\,0.00$ & $0.06\,|\,0.01$ & $0.06\,|\,0.01$ \\
\addlinespace
OmTrackVLA & decision-state heads & HM3D & B & $0.24\,|\,0.23$ & $0.38\,|\,0.36$ & $0.54\,|\,0.52$ & $0.69\,|\,0.66$ & $0.77\,|\,0.75$ \\
 & & & A & $0.19\,|\,0.20$ & $0.35\,|\,0.36$ & $0.50\,|\,0.51$ & $0.64\,|\,0.65$ & $0.72\,|\,0.73$ \\
 &  & HSSD & B & $0.25\,|\,0.26$ & $0.39\,|\,0.40$ & $0.59\,|\,0.59$ & $0.72\,|\,0.70$ & $0.79\,|\,0.77$ \\
 & & & A & $0.25\,|\,0.25$ & $0.38\,|\,0.40$ & $0.55\,|\,0.58$ & $0.69\,|\,0.71$ & $0.76\,|\,0.79$ \\
\bottomrule
\end{tabular}

\vspace{4pt}
\begin{tabular}{@{}lllcc@{}}
\toprule
Model & Dataset & Head & Brings in, $M$ & Removes, $R$ \\
\midrule
Uni-NaVid & HM3D & $L_{19}$ h06 (decision) & +0.23 {\tiny[+0.21, +0.24]} & +0.22 {\tiny[+0.20, +0.23]} \\
 &  & $L_{26}$ h03 (decision) & +0.16 {\tiny[+0.15, +0.18]} & +0.17 {\tiny[+0.15, +0.18]} \\
 & HSSD & $L_{19}$ h06 (decision) & +0.23 {\tiny[+0.21, +0.24]} & +0.22 {\tiny[+0.21, +0.24]} \\
 &  & $L_{18}$ h26 (decision) & +0.14 {\tiny[+0.13, +0.16]} & +0.14 {\tiny[+0.13, +0.16]} \\
\addlinespace
LightNav-0 & HM3D & $L_{21}$ h18 (prompt) & +0.05 {\tiny[+0.01, +0.09]} & +0.08 {\tiny[+0.04, +0.13]} \\
 &  & $L_{34}$ h27 (decision) & +0.05 {\tiny[+0.03, +0.08]} & +0.07 {\tiny[+0.04, +0.10]} \\
 & HSSD & $L_{24}$ h13 (prompt) & +0.02 {\tiny[$-$0.01, +0.06]} & +0.03 {\tiny[0.00, +0.06]} \\
 &  & $L_{24}$ h12 (prompt) & 0.00 {\tiny[$-$0.01, +0.02]} & +0.01 {\tiny[0.00, +0.03]} \\
\addlinespace
OmTrackVLA & HM3D & $L_{26}$ h00 (decision) & +0.21 {\tiny[+0.20, +0.23]} & +0.21 {\tiny[+0.19, +0.23]} \\
 &  & $L_{25}$ h05 (decision) & +0.13 {\tiny[+0.12, +0.15]} & +0.13 {\tiny[+0.12, +0.15]} \\
 & HSSD & $L_{26}$ h00 (decision) & +0.25 {\tiny[+0.23, +0.27]} & +0.26 {\tiny[+0.24, +0.27]} \\
 &  & $L_{26}$ h05 (decision) & +0.11 {\tiny[+0.10, +0.12]} & +0.11 {\tiny[+0.10, +0.12]} \\
\bottomrule
\end{tabular}
\end{table*}

\FloatBarrier
\section{Interventions}
\label{app:interventions}

This section supplements \cref{sec:interventions}: the fine-tuning recipes and data (\cref{app:interventions_training}), the evaluation (\cref{app:interventions_eval}), per-dataset
results (\cref{app:interventions_results}), the statements of the two statement recipes
(\cref{app:interventions_statement}) and the use of the robot's own motion after fine-tuning (\cref{app:interventions_internals}).

\subsection{Fine-Tuning Recipes and Data}
\label{app:interventions_training}

\noindent\textbf{Choice of policy.} \emph{Decompress} acts on the coarse history of \cref{sec:results_history_compression}: Uni-NaVid and OmTrackVLA keep
each recent historical frame as a $2\times2$ grid, whereas LightNav-0 encodes its two latest frames jointly and has no
coarse-history stage to act on. \emph{Say} and \emph{Commit} need actions generated as discrete words after the
prompt: a generated statement can then precede the action words, and \emph{Commit}'s penalty acts on the probability of the
first action word. Uni-NaVid emits such words; OmTrackVLA outputs continuous waypoints, which have no probability over
discrete actions to penalize. Uni-NaVid's discrete actions also make the training labels exact: clean samples carry the frozen
model's own words, with no rounding of a continuous trajectory.

\noindent\textbf{Recipes.} Each recipe keeps everything of the one before it and adds one change,
and is named after that change.
\begin{itemize}
\item \emph{Base} is plain fine-tuning on our clean and fine-tuning disturbance episodes, with Uni-NaVid's
input and answer unchanged; it separates what the data alone does from what the later changes add.

\item \emph{Decompress} undoes, for the latest historical frame, the coarse-history compression of
\cref{sec:results_history_compression}.
\emph{Decompress} adds the previous frame to the navigation stream at the current frame's $8\times8$ grid, next to the
$2\times2$ history.

\item \emph{Say} has the model say its own motion and the target's side before it
acts.
\emph{Say} asks, in one added prompt sentence, for its own motion and the target's
position before the actions, and answers
\begin{quote}\frenchspacing
\texttt{Self: <s>.}\\
\texttt{Person: <p>.}\\
\texttt{Actions: <4 words>},
\end{quote}
for example
\begin{quote}\frenchspacing
\texttt{Self: stayed in place, turned left a lot.}\\
\texttt{Person: out of view, right.}\\
\texttt{Actions: \ldots}.
\end{quote}
\begin{itemize}
\item \texttt{<s>} describes
the motion between the two latest frames: forward (``moved ahead'', ``moved back'', ``stayed in place''; $5$\,cm), turn
(``turned left/right a little'', $3$--$20^\circ$, or ``a lot'') and sideways slide (``slid left/right a little'',
$3$--$10$\,cm, or ``a lot'').
\item \texttt{<p>} is ``ahead'' ($<10^\circ$), ``left'', ``right'', or ``out of view, left/right'' when no target pixel is
visible.
\end{itemize}
The labels come from the simulator's poses at the frames the model sees; the replayed samples keep their
original prompt and answer. At evaluation the executed action is the first word after the last ``Actions:''.

\item \emph{Commit} commits the action to what was said.
\emph{Commit} adds, for each of the 13{,}921 samples whose target is out of view and whose first action is a
turn, a copy with left and right exchanged in both statements and the frames, prompt and actions unchanged; on a copy
the only loss is
\begin{equation}
 \mathcal L=-\log\bigl(1-p(w_1)\bigr)
\end{equation}
on the first action word $w_1$, weight $1$, so the swapped statements themselves are never trained.
\end{itemize}

\noindent\textbf{Data.} \Cref{tab:supp_ft_data} compares our data with Uni-NaVid's original training mixture; the $1.7\%$ of the main text counts samples (102{,}063 of 5.9M), not tokens or compute; \emph{Commit}'s 13{,}921 swapped-statement copies reuse these frames.
Clean following comes from the frozen model's own runs on the EVT-Bench training split (successful episodes whole,
failed ones without their last ten steps). Frames are lossless simulator renders.

\noindent\textbf{Fine-tuning disturbance episodes.} The rotation episodes do not use the benchmark's target-loss rule
(\cref{tab:supp_ft_disturbances}). A fine-tuning disturbance is an uncommanded rotation of the robot drawn independently of the target. After each fine-tuning disturbance a
controller that knows the realized rotation undoes it with Uni-NaVid's own $9^\circ$ turns, the same controller as the
oracle of \cref{sec:results_recovery_feasibility}; its decided actions are the labels, while the executed commands
carry small noise. Half of the episodes are mirrored. The 166 fine-tuning disturbance episodes contain 617 fine-tuning disturbances. Their median total is $49.7^\circ$
(interquartile range $29.1$--$69.3^\circ$); $26.4\%$ are applied in a single frame and $26.1\%$ have a total within the
benchmark's $45$--$65^\circ$. The target is in view at the onset of $69.5\%$ (followed and within $10^\circ$ of the
center for $25.0\%$) and out of view after $49.5\%$ of the 588 fine-tuning disturbances that the episode's end does not cut short. Two
fine-tuning disturbances ($0.3\%$) match an instant target-loss in all four respects at once (target followed and within $10^\circ$ of the center
at onset, a single frame, $45$--$65^\circ$, target out of view after), although single respects are common (a single
frame in $26.4\%$, a $45$--$65^\circ$ total in $26.1\%$); two more match a smooth one (two or more
frames at $15$--$25^\circ$ per frame). The evaluation's target-losses are therefore not held out as a disturbance type, but none of their episodes,
scenes or persons, nor their trigger and size rule, occurs in training.

\begin{table}[tbp]
\centering
\footnotesize
\setlength{\tabcolsep}{3pt}
\caption{Fine-tuning disturbances and the benchmark's target-loss (S1).}
\label{tab:supp_ft_disturbances}
\begin{tabular}{@{}lp{0.36\columnwidth}p{0.36\columnwidth}@{}}
\toprule
 & Fine-tuning disturbance & Target-loss (evaluation) \\
\midrule
Onset & after action 5, then 10--20 actions after the previous fine-tuning disturbance; no condition & due after action 20, then every 40; fires only while the target is followed and within $10^\circ$ of the center \\
Total & uniform $10$--$90^\circ$ & target $10^\circ$ beyond the edge of view ($45$--$65^\circ$) \\
Rate & uniform $5$--$60^\circ$ per frame & one frame (instant) or $20^\circ$ per frame (smooth) \\
Side & fair coin & hash of scene, episode and loss index \\
Target after & in or out of view & always out of view \\
\bottomrule
\end{tabular}
\end{table}

\begin{table}[tbp]
\centering
\footnotesize
\setlength{\tabcolsep}{3pt}
\caption{Training data: Uni-NaVid's original mixture and the fine-tuning data shared by the four recipes.}
\label{tab:supp_ft_data}
\begin{tabular}{@{}lp{0.33\columnwidth}p{0.36\columnwidth}@{}}
\toprule
 & Uni-NaVid (original) & Ours \\
\midrule
Samples & 5.9M & 102{,}063 (1.7\%) \\
Person following & 544k, EVT-Bench training split & 80{,}968 clean (1{,}000 episodes) and 16{,}095 with random rotations (fine-tuning disturbances; 166 episodes, recovery steps counted twice), EVT-Bench training split, 669 scenes \\
Labels & privileged planner & frozen model's own actions; the turns undoing each fine-tuning disturbance \\
Other tasks & VLN 2.4M, ObjectNav 483k, EQA 250k, video QA 2.3M & 5{,}000 replayed released samples (500 \texttimes{} 10) \\
Disturbances & none & fine-tuning disturbances only (\cref{tab:supp_ft_disturbances}) \\
Trained & language model and projector & language-model layers 13--31, final norm and output head \\
\bottomrule
\end{tabular}
\end{table}

\noindent\textbf{Optimization and compute.} Learning rate $10^{-5}$, cosine schedule with $3\%$ warm-up, effective
batch 16 (with gradient accumulation), 10 epochs (63{,}790 steps, 6{,}379 steps of 16 per epoch; \emph{Commit} 72{,}490 steps, 115{,}984 samples per epoch with its
copies; evaluated at the last saved checkpoint, steps 63{,}760 and 72{,}480), loss on every answer token, seed 0. Per
epoch, the language model receives $41.7$--$45.0$, $48.3$--$51.5$, $53.4$--$56.7$ and $61.2$--$64.7$ million tokens for
\emph{Base}, \emph{Decompress}, \emph{Say} and \emph{Commit}; the ranges arise because frames beyond the 64th are merged by visual
similarity in the encoder.

\subsection{Evaluation}
\label{app:interventions_eval}

\noindent\textbf{Splits.} Development: held-out training scenes (MP3D 91 episodes, HM3D 107). Validation: 200 episodes per dataset sampled from the MP3D and HM3D validation splits and the HSSD episodes (\cref{sec:supp_datasets}), reported
for the final checkpoint of each 10-epoch training run.

\noindent\textbf{Conditions.} Clean; current, veer and gust at S3; instant and smooth target-loss at S1; the instant run with every target-loss
mirrored; and same-target-loss runs, which force each model through the frozen model's executed actions up to its
first target-loss, so that every model faces the same state there. Unlike the benchmark's evaluators
(\cref{sec:supp_datasets}), every run loads each episode's own person and instruction. The frozen model's success here therefore differs from
its recovery runs in \cref{app:internals_functional} (\cref{tab:app_recovery_episode_datasets}), which use the
benchmark's evaluators; the model and weights are the same. Abilities outside following were not evaluated.

\noindent\textbf{Readouts.} Success as in \cref{sec:benchmark_metrics}, paired per episode with the frozen model's run
of the same condition (same-target-loss runs: with its target-loss run). \emph{First turn toward the target}: at the
first decision whose frame shows an instant target-loss, with the target out of view, the first turn within 20
steps goes to the target's side (each model's own events). On the same target-losses: \emph{first turn correct}, the first turn within 20 steps from the first decision
whose frame shows the loss goes against the rotation (after an instant loss on events with the target out of view at
that decision; after a smooth loss on every event, counted from the frame that shows the rotation's onset, whether or not
the target is in view there), and \emph{re-acquired} within 20 steps, paired per event. \Cref{app:internals_functional} counts the first turn after a smooth loss from the step at which the target is at least $10^\circ$ out of view, side-balanced, on its own matched events, so the frozen model's share there ($56.1\%$, \cref{tab:app_first_turn_xd}) is lower than here ($73.7\%$). \emph{Share of the veer turned back}: per episode, the commanded rotation opposite to the veer
divided by the total veer, summed over steps. Intervals: 5{,}000 episode-bootstrap resamples, seed 0; they resample evaluation episodes for one trained checkpoint per recipe and do not include training variability.

\subsection{Results per Dataset}
\label{app:interventions_results}

\Cref{tab:supp_ft_pooled} gives \cref{fig:ft_results} in full, with paired differences and intervals, and \cref{tab:supp_ft_perdataset} per dataset.

\noindent\textbf{\emph{Say} and \emph{Commit} against \emph{Base}.}\label{app:interventions_vs_base} Paired with
\emph{Base} on the same validation episodes (pooled over MP3D, HM3D and HSSD, stratified by dataset), \emph{Commit}
adds $+10.5$ points [$+6.7$, $+14.3$] after a smooth target-loss and $+5.5$ [$+2.3$, $+8.7$] on clean episodes, and
\emph{Say} $+5.2$ [$+1.3$, $+8.8$] and $+3.5$ [$+0.3$, $+6.5$]. From the same state at the target-loss, the gain
remains after a smooth loss (\emph{Say} $+4.8$, \emph{Commit} $+9.8$; both intervals exclude 0) but not after an
instant one (both include 0). Both recipes also turn back more of the \emph{veer} than \emph{Base} on MP3D and HM3D
($+0.14$ to $+0.31$ of the veer), but not on HSSD. Intervals are not adjusted for the number of comparisons.

\begin{table*}[tbp]
\centering
\footnotesize
\setlength{\tabcolsep}{2.5pt}
\caption{\Cref{fig:ft_results} in full: fine-tuned Uni-NaVid against the frozen model, pooled over the validation episodes of MP3D, HM3D and HSSD (200 per dataset). Each cell: success rate (\%) or readout, its paired difference from the frozen model (bold: 95\% interval excludes 0) and the interval. \emph{Same target-loss}: each model is forced through the frozen model's actions up to its first target-loss. \emph{First turn}: the first turn within 20 steps after an instant loss that leaves the target out of view (each model's own events, not paired). Frozen readouts: on all of its own events; the same-target-loss differences use the events each model shares with the frozen run up to the loss. Per dataset: \cref{tab:supp_ft_perdataset}.}
\label{tab:supp_ft_pooled}
\begin{tabular}{@{}lccccc@{}}
\toprule
Condition & Frozen & Base & Decompress & Say & Commit \\
\midrule
Clean & 58.3 & 60.7 +2.3 {\tiny[$-$0.8, +5.5]} & 59.7 +1.3 {\tiny[$-$1.7, +4.3]} & 64.2 \textbf{+5.8} {\tiny[+2.3, +9.3]} & 66.2 \textbf{+7.8} {\tiny[+4.7, +11.0]} \\
Current S3 & 32.5 & 35.0 +2.5 {\tiny[$-$0.8, +5.7]} & 33.8 +1.3 {\tiny[$-$1.7, +4.3]} & 32.7 +0.2 {\tiny[$-$3.2, +3.5]} & 35.0 +2.5 {\tiny[$-$0.7, +5.7]} \\
Veer S3 & 40.5 & 40.8 +0.3 {\tiny[$-$2.8, +3.5]} & 44.0 \textbf{+3.5} {\tiny[+0.3, +6.7]} & 43.5 +3.0 {\tiny[$-$0.5, +6.7]} & 47.0 \textbf{+6.5} {\tiny[+2.8, +10.2]} \\
Gust S3 & 48.2 & 54.7 \textbf{+6.5} {\tiny[+2.8, +10.0]} & 54.3 \textbf{+6.2} {\tiny[+2.7, +9.7]} & 52.3 \textbf{+4.2} {\tiny[+0.5, +8.0]} & 56.7 \textbf{+8.5} {\tiny[+4.8, +12.2]} \\
Instant target-loss S1 & 23.7 & 29.8 \textbf{+6.2} {\tiny[+3.0, +9.5]} & 31.3 \textbf{+7.7} {\tiny[+4.2, +11.2]} & 32.5 \textbf{+8.8} {\tiny[+5.0, +12.7]} & 31.5 \textbf{+7.8} {\tiny[+4.2, +11.7]} \\
Instant, mirrored & 24.8 & 28.8 \textbf{+4.0} {\tiny[+0.5, +7.7]} & 32.5 \textbf{+7.7} {\tiny[+4.2, +11.3]} & 35.2 \textbf{+10.3} {\tiny[+6.5, +14.2]} & 32.8 \textbf{+8.0} {\tiny[+4.2, +11.8]} \\
Smooth target-loss S1 & 30.0 & 41.7 \textbf{+11.7} {\tiny[+7.8, +15.5]} & 45.2 \textbf{+15.2} {\tiny[+11.0, +19.2]} & 46.8 \textbf{+16.8} {\tiny[+12.8, +20.8]} & 52.2 \textbf{+22.2} {\tiny[+18.2, +26.3]} \\
Same target-loss, instant & 23.7 & 30.5 \textbf{+6.8} {\tiny[+3.7, +10.2]} & 28.3 \textbf{+4.7} {\tiny[+1.5, +8.0]} & 30.3 \textbf{+6.7} {\tiny[+3.2, +10.3]} & 31.5 \textbf{+7.8} {\tiny[+4.3, +11.3]} \\
Same target-loss, smooth & 30.0 & 41.7 \textbf{+11.7} {\tiny[+8.0, +15.5]} & 45.7 \textbf{+15.7} {\tiny[+12.0, +19.5]} & 46.5 \textbf{+16.5} {\tiny[+12.5, +20.3]} & 51.5 \textbf{+21.5} {\tiny[+17.5, +25.7]} \\
\midrule
\multicolumn{6}{@{}l}{\emph{Decision readouts (\%; veer: fraction)}} \\
First turn toward the target, instant & 56.6 {\tiny[51.6, 61.4]} & 59.6 {\tiny[54.8, 64.4]} & 65.7 {\tiny[60.9, 70.5]} & 66.5 {\tiny[61.7, 71.3]} & 68.6 {\tiny[63.9, 73.3]} \\
Same target-loss, instant: first turn correct & 56.6 {\tiny[51.6, 61.7]} & 58.7 +2.2 {\tiny[$-$2.5, +7.0]} & 61.3 \textbf{+5.3} {\tiny[+0.3, +10.6]} & 62.3 +5.6 {\tiny[0.0, +11.2]} & 67.0 \textbf{+10.1} {\tiny[+4.2, +15.8]} \\
Same target-loss, instant: re-acquired in 20 steps & 53.6 {\tiny[49.0, 58.4]} & 59.6 \textbf{+5.0} {\tiny[+1.5, +8.5]} & 60.8 \textbf{+6.3} {\tiny[+3.0, +9.5]} & 62.8 \textbf{+6.8} {\tiny[+3.0, +10.6]} & 64.5 \textbf{+9.1} {\tiny[+5.3, +13.2]} \\
Same target-loss, smooth: first turn correct & 73.7 {\tiny[69.6, 77.7]} & 91.9 \textbf{+18.2} {\tiny[+14.4, +22.3]} & 94.3 \textbf{+20.3} {\tiny[+16.4, +24.2]} & 95.7 \textbf{+22.2} {\tiny[+18.3, +26.4]} & 97.0 \textbf{+23.2} {\tiny[+18.8, +27.6]} \\
Same target-loss, smooth: re-acquired in 20 steps & 67.9 {\tiny[63.4, 72.4]} & 81.1 \textbf{+12.5} {\tiny[+8.8, +16.5]} & 83.4 \textbf{+14.1} {\tiny[+10.2, +18.4]} & 85.1 \textbf{+16.0} {\tiny[+12.2, +20.0]} & 87.9 \textbf{+18.3} {\tiny[+14.1, +22.6]} \\
Veer: share of the veer turned back & 0.67 {\tiny[0.60, 0.74]} & 0.65 $-$0.01 {\tiny[$-$0.07, +0.04]} & 0.65 $-$0.01 {\tiny[$-$0.07, +0.04]} & 0.83 \textbf{+0.15} {\tiny[+0.09, +0.22]} & 0.82 \textbf{+0.15} {\tiny[+0.09, +0.21]} \\
\bottomrule
\end{tabular}
\end{table*}

\begin{table*}[tbp]
\centering
\footnotesize
\setlength{\tabcolsep}{2.2pt}
\caption{\Cref{tab:supp_ft_pooled} per dataset (200 validation episodes each). Cells as in \cref{tab:supp_ft_pooled}; bold: the paired difference's 95\% interval excludes 0.}
\label{tab:supp_ft_perdataset}
\begin{tabular}{@{}lccccc@{}}
\toprule
Condition & Frozen & Base & Decompress & Say & Commit \\
\midrule
\multicolumn{6}{@{}l}{\textbf{MP3D}} \\
Clean & 49.5 & 53.0 +3.5 {\tiny[$-$2.0, +9.0]} & 48.5 $-$1.0 {\tiny[$-$5.5, +3.5]} & 53.0 +3.5 {\tiny[$-$2.5, +9.0]} & 56.0 \textbf{+6.5} {\tiny[+1.0, +12.0]} \\
Current S3 & 29.5 & 27.0 $-$2.5 {\tiny[$-$7.5, +2.0]} & 28.0 $-$1.5 {\tiny[$-$6.5, +3.5]} & 25.0 $-$4.5 {\tiny[$-$9.5, +0.5]} & 26.5 $-$3.0 {\tiny[$-$8.0, +2.0]} \\
Veer S3 & 31.5 & 27.0 $-$4.5 {\tiny[$-$9.0, 0.0]} & 33.0 +1.5 {\tiny[$-$3.0, +6.5]} & 29.5 $-$2.0 {\tiny[$-$7.0, +2.5]} & 31.5 0.0 {\tiny[$-$5.5, +5.0]} \\
Gust S3 & 39.5 & 45.5 \textbf{+6.0} {\tiny[+0.5, +12.0]} & 47.0 \textbf{+7.5} {\tiny[+1.5, +13.5]} & 43.0 +3.5 {\tiny[$-$2.5, +9.5]} & 50.0 \textbf{+10.5} {\tiny[+5.0, +16.5]} \\
Instant target-loss S1 & 16.5 & 19.5 +3.0 {\tiny[$-$2.0, +8.0]} & 21.0 +4.5 {\tiny[$-$0.5, +10.0]} & 24.5 \textbf{+8.0} {\tiny[+2.5, +13.5]} & 24.5 \textbf{+8.0} {\tiny[+2.5, +14.0]} \\
Instant, mirrored & 16.5 & 20.0 +3.5 {\tiny[$-$2.0, +9.0]} & 20.0 +3.5 {\tiny[$-$2.0, +9.0]} & 22.5 \textbf{+6.0} {\tiny[+0.5, +11.5]} & 20.5 +4.0 {\tiny[$-$2.0, +10.0]} \\
Smooth target-loss S1 & 23.0 & 29.5 \textbf{+6.5} {\tiny[+0.5, +12.5]} & 30.5 \textbf{+7.5} {\tiny[+1.5, +14.0]} & 33.0 \textbf{+10.0} {\tiny[+3.5, +16.5]} & 38.5 \textbf{+15.5} {\tiny[+9.5, +22.0]} \\
Same target-loss, instant & 16.5 & 22.0 +5.5 {\tiny[0.0, +11.0]} & 21.0 +4.5 {\tiny[0.0, +9.5]} & 23.5 \textbf{+7.0} {\tiny[+1.5, +12.5]} & 24.5 \textbf{+8.0} {\tiny[+3.0, +13.0]} \\
Same target-loss, smooth & 23.0 & 32.5 \textbf{+9.5} {\tiny[+3.5, +15.5]} & 35.0 \textbf{+12.0} {\tiny[+6.5, +18.0]} & 33.0 \textbf{+10.0} {\tiny[+4.0, +16.5]} & 39.0 \textbf{+16.0} {\tiny[+9.5, +22.5]} \\
First turn toward the target, instant & 62.1 {\tiny[53.2, 70.2]} & 57.5 {\tiny[48.8, 66.1]} & 73.8 {\tiny[65.9, 81.0]} & 67.5 {\tiny[59.3, 75.6]} & 74.4 {\tiny[66.7, 81.4]} \\
Same target-loss, instant: first turn correct & 62.1 {\tiny[53.2, 70.2]} & 55.6 $-$5.2 {\tiny[$-$13.8, +2.6]} & 66.1 +6.0 {\tiny[$-$4.3, +15.4]} & 58.6 $-$1.7 {\tiny[$-$11.3, +7.8]} & 63.6 +2.6 {\tiny[$-$8.5, +13.7]} \\
Same target-loss, instant: re-acquired in 20 steps & 60.3 {\tiny[52.3, 68.2]} & 57.5 $-$2.8 {\tiny[$-$8.3, +2.8]} & 62.3 +2.8 {\tiny[$-$2.1, +8.3]} & 62.5 +0.7 {\tiny[$-$5.7, +6.4]} & 63.5 +2.8 {\tiny[$-$4.2, +9.7]} \\
Same target-loss, smooth: first turn correct & 71.3 {\tiny[64.3, 78.3]} & 92.1 \textbf{+21.9} {\tiny[+14.6, +29.1]} & 91.9 \textbf{+20.8} {\tiny[+13.4, +28.2]} & 92.7 \textbf{+21.9} {\tiny[+15.2, +28.5]} & 98.0 \textbf{+26.7} {\tiny[+19.3, +34.0]} \\
Same target-loss, smooth: re-acquired in 20 steps & 66.2 {\tiny[58.3, 73.5]} & 79.7 \textbf{+13.1} {\tiny[+6.2, +20.0]} & 83.6 \textbf{+17.5} {\tiny[+11.2, +24.5]} & 82.7 \textbf{+15.2} {\tiny[+8.3, +22.1]} & 87.4 \textbf{+20.0} {\tiny[+13.1, +27.6]} \\
Veer: share of the veer turned back & 0.64 {\tiny[0.54, 0.74]} & 0.61 $-$0.03 {\tiny[$-$0.10, +0.05]} & 0.63 $-$0.01 {\tiny[$-$0.10, +0.07]} & 0.82 \textbf{+0.18} {\tiny[+0.08, +0.28]} & 0.75 \textbf{+0.11} {\tiny[+0.02, +0.20]} \\
\midrule
\multicolumn{6}{@{}l}{\textbf{HM3D}} \\
Clean & 56.0 & 60.5 +4.5 {\tiny[$-$1.0, +10.5]} & 60.5 +4.5 {\tiny[$-$1.5, +10.5]} & 67.0 \textbf{+11.0} {\tiny[+4.0, +17.5]} & 69.0 \textbf{+13.0} {\tiny[+7.0, +19.5]} \\
Current S3 & 26.0 & 35.5 \textbf{+9.5} {\tiny[+3.0, +15.5]} & 32.5 \textbf{+6.5} {\tiny[+0.5, +12.5]} & 34.5 \textbf{+8.5} {\tiny[+2.5, +14.5]} & 32.5 \textbf{+6.5} {\tiny[+1.0, +12.0]} \\
Veer S3 & 41.5 & 47.5 +6.0 {\tiny[0.0, +12.0]} & 49.5 \textbf{+8.0} {\tiny[+1.5, +14.5]} & 52.5 \textbf{+11.0} {\tiny[+3.5, +19.0]} & 56.0 \textbf{+14.5} {\tiny[+7.0, +22.0]} \\
Gust S3 & 43.5 & 54.5 \textbf{+11.0} {\tiny[+4.5, +17.5]} & 53.5 \textbf{+10.0} {\tiny[+3.5, +16.5]} & 51.5 \textbf{+8.0} {\tiny[+0.5, +15.5]} & 53.5 \textbf{+10.0} {\tiny[+3.0, +17.5]} \\
Instant target-loss S1 & 23.0 & 34.5 \textbf{+11.5} {\tiny[+5.5, +17.5]} & 33.0 \textbf{+10.0} {\tiny[+4.0, +16.5]} & 31.5 \textbf{+8.5} {\tiny[+1.5, +15.5]} & 32.5 \textbf{+9.5} {\tiny[+2.5, +17.0]} \\
Instant, mirrored & 26.5 & 31.0 +4.5 {\tiny[$-$2.0, +11.5]} & 33.0 +6.5 {\tiny[$-$0.5, +13.0]} & 35.0 \textbf{+8.5} {\tiny[+1.0, +16.0]} & 37.0 \textbf{+10.5} {\tiny[+3.5, +18.0]} \\
Smooth target-loss S1 & 28.5 & 43.5 \textbf{+15.0} {\tiny[+8.0, +21.5]} & 46.5 \textbf{+18.0} {\tiny[+10.5, +25.5]} & 48.5 \textbf{+20.0} {\tiny[+13.0, +27.0]} & 55.5 \textbf{+27.0} {\tiny[+20.0, +34.5]} \\
Same target-loss, instant & 23.0 & 33.0 \textbf{+10.0} {\tiny[+4.0, +16.0]} & 29.0 \textbf{+6.0} {\tiny[+0.5, +12.0]} & 32.0 \textbf{+9.0} {\tiny[+2.5, +16.0]} & 34.5 \textbf{+11.5} {\tiny[+5.0, +18.5]} \\
Same target-loss, smooth & 28.5 & 44.5 \textbf{+16.0} {\tiny[+10.0, +22.5]} & 46.5 \textbf{+18.0} {\tiny[+11.5, +24.5]} & 51.5 \textbf{+23.0} {\tiny[+16.0, +29.5]} & 53.5 \textbf{+25.0} {\tiny[+18.5, +32.0]} \\
First turn toward the target, instant & 56.7 {\tiny[48.0, 65.4]} & 63.4 {\tiny[55.2, 71.6]} & 63.4 {\tiny[55.2, 71.6]} & 71.9 {\tiny[64.1, 79.7]} & 65.2 {\tiny[57.0, 73.3]} \\
Same target-loss, instant: first turn correct & 56.7 {\tiny[48.0, 65.4]} & 66.4 \textbf{+9.2} {\tiny[+0.8, +18.3]} & 62.6 +4.2 {\tiny[$-$5.0, +13.3]} & 71.5 \textbf{+13.3} {\tiny[+4.2, +22.5]} & 73.3 \textbf{+14.5} {\tiny[+3.4, +24.8]} \\
Same target-loss, instant: re-acquired in 20 steps & 49.6 {\tiny[41.6, 57.7]} & 63.5 \textbf{+11.6} {\tiny[+5.4, +17.8]} & 64.2 \textbf{+13.3} {\tiny[+7.0, +19.5]} & 69.3 \textbf{+16.0} {\tiny[+8.4, +23.7]} & 72.6 \textbf{+19.8} {\tiny[+11.9, +27.8]} \\
Same target-loss, smooth: first turn correct & 73.4 {\tiny[66.2, 79.9]} & 93.8 \textbf{+19.9} {\tiny[+13.7, +27.4]} & 92.6 \textbf{+18.9} {\tiny[+12.8, +25.0]} & 95.2 \textbf{+21.9} {\tiny[+15.1, +29.5]} & 95.7 \textbf{+22.3} {\tiny[+14.4, +30.2]} \\
Same target-loss, smooth: re-acquired in 20 steps & 69.1 {\tiny[61.0, 76.5]} & 83.7 \textbf{+12.5} {\tiny[+5.5, +19.5]} & 83.0 \textbf{+11.6} {\tiny[+3.9, +19.4]} & 88.8 \textbf{+18.3} {\tiny[+11.9, +25.4]} & 88.4 \textbf{+16.4} {\tiny[+9.8, +23.8]} \\
Veer: share of the veer turned back & 0.55 {\tiny[0.42, 0.68]} & 0.50 $-$0.04 {\tiny[$-$0.14, +0.05]} & 0.56 +0.01 {\tiny[$-$0.10, +0.13]} & 0.81 \textbf{+0.26} {\tiny[+0.13, +0.40]} & 0.82 \textbf{+0.27} {\tiny[+0.15, +0.40]} \\
\midrule
\multicolumn{6}{@{}l}{\textbf{HSSD}} \\
Clean & 69.5 & 68.5 $-$1.0 {\tiny[$-$6.5, +4.5]} & 70.0 +0.5 {\tiny[$-$4.5, +5.5]} & 72.5 +3.0 {\tiny[$-$2.5, +8.5]} & 73.5 +4.0 {\tiny[$-$1.0, +9.0]} \\
Current S3 & 42.0 & 42.5 +0.5 {\tiny[$-$5.0, +6.0]} & 41.0 $-$1.0 {\tiny[$-$5.0, +3.0]} & 38.5 $-$3.5 {\tiny[$-$9.5, +2.0]} & 46.0 +4.0 {\tiny[$-$1.5, +9.5]} \\
Veer S3 & 48.5 & 48.0 $-$0.5 {\tiny[$-$6.0, +5.0]} & 49.5 +1.0 {\tiny[$-$3.5, +5.5]} & 48.5 0.0 {\tiny[$-$5.0, +5.0]} & 53.5 +5.0 {\tiny[0.0, +10.0]} \\
Gust S3 & 61.5 & 64.0 +2.5 {\tiny[$-$3.5, +8.5]} & 62.5 +1.0 {\tiny[$-$5.0, +7.0]} & 62.5 +1.0 {\tiny[$-$4.5, +6.5]} & 66.5 +5.0 {\tiny[$-$1.0, +11.0]} \\
Instant target-loss S1 & 31.5 & 35.5 +4.0 {\tiny[$-$1.5, +9.5]} & 40.0 \textbf{+8.5} {\tiny[+2.0, +15.0]} & 41.5 \textbf{+10.0} {\tiny[+3.0, +17.0]} & 37.5 +6.0 {\tiny[$-$0.5, +13.0]} \\
Instant, mirrored & 31.5 & 35.5 +4.0 {\tiny[$-$3.0, +10.5]} & 44.5 \textbf{+13.0} {\tiny[+6.5, +19.5]} & 48.0 \textbf{+16.5} {\tiny[+9.5, +23.0]} & 41.0 \textbf{+9.5} {\tiny[+3.0, +16.0]} \\
Smooth target-loss S1 & 38.5 & 52.0 \textbf{+13.5} {\tiny[+6.5, +20.5]} & 58.5 \textbf{+20.0} {\tiny[+13.5, +26.5]} & 59.0 \textbf{+20.5} {\tiny[+13.5, +28.0]} & 62.5 \textbf{+24.0} {\tiny[+17.0, +31.0]} \\
Same target-loss, instant & 31.5 & 36.5 +5.0 {\tiny[$-$0.5, +10.5]} & 35.0 +3.5 {\tiny[$-$3.0, +10.0]} & 35.5 +4.0 {\tiny[$-$2.5, +10.5]} & 35.5 +4.0 {\tiny[$-$2.5, +10.5]} \\
Same target-loss, smooth & 38.5 & 48.0 \textbf{+9.5} {\tiny[+3.0, +16.0]} & 55.5 \textbf{+17.0} {\tiny[+10.5, +23.5]} & 55.0 \textbf{+16.5} {\tiny[+9.5, +23.5]} & 62.0 \textbf{+23.5} {\tiny[+16.0, +30.5]} \\
First turn toward the target, instant & 51.2 {\tiny[42.4, 60.0]} & 57.5 {\tiny[48.7, 67.3]} & 59.5 {\tiny[50.9, 69.0]} & 58.7 {\tiny[49.0, 68.3]} & 66.1 {\tiny[57.6, 74.6]} \\
Same target-loss, instant: first turn correct & 51.2 {\tiny[42.4, 60.0]} & 54.3 +2.5 {\tiny[$-$3.3, +9.0]} & 55.6 +5.7 {\tiny[$-$2.5, +13.9]} & 56.7 +4.9 {\tiny[$-$4.9, +14.8]} & 64.3 \textbf{+13.2} {\tiny[+4.1, +22.3]} \\
Same target-loss, instant: re-acquired in 20 steps & 50.0 {\tiny[41.5, 58.5]} & 57.8 \textbf{+7.1} {\tiny[+1.6, +12.7]} & 55.6 +3.1 {\tiny[$-$2.4, +8.7]} & 56.4 +3.9 {\tiny[$-$1.6, +10.2]} & 57.3 \textbf{+5.6} {\tiny[+0.8, +11.2]} \\
Same target-loss, smooth: first turn correct & 76.7 {\tiny[69.9, 83.6]} & 89.7 \textbf{+12.7} {\tiny[+5.6, +19.7]} & 98.6 \textbf{+21.3} {\tiny[+14.9, +28.4]} & 99.3 \textbf{+23.0} {\tiny[+15.8, +30.2]} & 97.2 \textbf{+20.4} {\tiny[+12.7, +28.2]} \\
Same target-loss, smooth: re-acquired in 20 steps & 68.7 {\tiny[60.4, 76.1]} & 80.0 \textbf{+11.9} {\tiny[+5.6, +18.3]} & 83.7 \textbf{+13.0} {\tiny[+6.9, +19.1]} & 83.9 \textbf{+14.7} {\tiny[+8.5, +21.7]} & 88.0 \textbf{+18.3} {\tiny[+11.5, +25.2]} \\
Veer: share of the veer turned back & 0.81 {\tiny[0.68, 0.94]} & 0.83 +0.03 {\tiny[$-$0.07, +0.12]} & 0.78 $-$0.03 {\tiny[$-$0.12, +0.06]} & 0.86 +0.03 {\tiny[$-$0.08, +0.14]} & 0.89 +0.08 {\tiny[$-$0.03, +0.19]} \\
\bottomrule
\end{tabular}
\end{table*}

\subsection{The Statements}
\label{app:interventions_statement}

\noindent\textbf{Swap.} At the same-target-loss decision we force the model's own statements with left
and right exchanged before its action words (\cref{tab:supp_ft_swap}; development episodes: all 91 MP3D and 107 HM3D after an
instant loss, 50 per dataset after a smooth one; HSSD validation episodes: the first 100 after an instant loss and the
first 50 after a smooth one).

\begin{table*}[tbp]
\centering
\footnotesize
\setlength{\tabcolsep}{4pt}
\caption{The statement at the same-target-loss decision (final checkpoint; MP3D and HM3D development and HSSD validation episodes, none of whose scenes is in the fine-tuning data). \emph{Swap}: the model's own Self and Person statements with left and right exchanged are forced before the action words; share of matched events whose first turn reverses (\%; $n$: matched events). After a smooth loss the target is still in view at this decision.}
\label{tab:supp_ft_swap}
\begin{tabular}{@{}llcc@{}}
\toprule
Model & Dataset & Swap, instant & Swap, smooth \\
\midrule
Say & MP3D & 81.4 {\tiny[71.4, 90.0]} ($n$ 70) & 15.8 {\tiny[5.3, 28.9]} ($n$ 38) \\
Say & HM3D & 67.9 {\tiny[57.7, 78.2]} ($n$ 78) & 15.8 {\tiny[5.3, 28.9]} ($n$ 38) \\
Say & HSSD & 78.1 {\tiny[68.5, 87.7]} ($n$ 73) & 11.1 {\tiny[2.8, 22.2]} ($n$ 36) \\
Commit & MP3D & 84.1 {\tiny[75.4, 92.8]} ($n$ 69) & 28.9 {\tiny[15.8, 44.7]} ($n$ 38) \\
Commit & HM3D & 75.0 {\tiny[64.5, 84.2]} ($n$ 76) & 15.8 {\tiny[5.3, 28.9]} ($n$ 38) \\
Commit & HSSD & 73.0 {\tiny[62.2, 82.4]} ($n$ 74) & 11.4 {\tiny[2.9, 22.9]} ($n$ 35) \\
\bottomrule
\end{tabular}
\end{table*}

\noindent\textbf{Statement correctness.} \Cref{tab:supp_ft_lines} reads the statements at the first decision that
shows an instant target-loss. In the same-target-loss runs (MP3D, HM3D, HSSD), the Person line of an out-of-view
target names a side in $47$--$61\%$ of events for \emph{Say} and $87$--$96\%$ for \emph{Commit}, the named side is
correct in $60$--$71\%$ and $58$--$68\%$, the Self turn in $49$--$61\%$ and $63$--$72\%$, and the first turn follows the
stated side in $90$--$95\%$ and $84$--$91\%$.

\begin{table*}[tbp]
\centering
\footnotesize
\setlength{\tabcolsep}{4pt}
\caption{The generated statements at the first decision that shows an instant target-loss (validation, final checkpoint). \emph{Names a side}: the Person line of an out-of-view target names left or right (\%). \emph{Side correct}: among those, the named side is the target's (\%; chance 50). \emph{Self turn}: the stated turn since the previous frame has the rotation's side (\%). \emph{Turn follows}: the first turn goes to the stated side of the target (\%).}
\label{tab:supp_ft_lines}
\begin{tabular}{@{}llllccc@{}}
\toprule
Model & Dataset & Runs & Names a side & Side correct & Self turn & Turn follows \\
\midrule
Say & MP3D & mirrored & 59.5 {\tiny[50.8, 68.3]} ($n$ 126) & 60.0 {\tiny[49.3, 70.7]} ($n$ 75) & 53.5 {\tiny[45.2, 61.1]} ($n$ 157) & 82.7 {\tiny[73.3, 90.7]} ($n$ 75) \\
Say & MP3D & same target-loss & 47.0 {\tiny[38.8, 55.2]} ($n$ 134) & 66.7 {\tiny[55.6, 77.8]} ($n$ 63) & 51.5 {\tiny[44.4, 58.6]} ($n$ 169) & 90.5 {\tiny[82.5, 96.8]} ($n$ 63) \\
Say & HM3D & mirrored & 55.6 {\tiny[46.8, 63.5]} ($n$ 126) & 61.4 {\tiny[50.0, 72.9]} ($n$ 70) & 52.4 {\tiny[44.1, 60.7]} ($n$ 145) & 85.7 {\tiny[77.1, 92.9]} ($n$ 70) \\
Say & HM3D & same target-loss & 54.9 {\tiny[46.5, 63.2]} ($n$ 144) & 70.9 {\tiny[60.8, 81.0]} ($n$ 79) & 60.5 {\tiny[53.3, 67.7]} ($n$ 167) & 94.9 {\tiny[89.9, 98.7]} ($n$ 79) \\
Say & HSSD & mirrored & 55.4 {\tiny[45.5, 65.3]} ($n$ 101) & 69.6 {\tiny[57.1, 82.1]} ($n$ 56) & 56.9 {\tiny[48.0, 65.9]} ($n$ 123) & 87.5 {\tiny[78.6, 94.6]} ($n$ 56) \\
Say & HSSD & same target-loss & 61.4 {\tiny[53.6, 69.3]} ($n$ 140) & 60.5 {\tiny[50.0, 69.8]} ($n$ 86) & 48.7 {\tiny[41.1, 56.3]} ($n$ 158) & 93.0 {\tiny[87.2, 97.7]} ($n$ 86) \\
Commit & MP3D & mirrored & 89.3 {\tiny[84.0, 93.9]} ($n$ 131) & 64.1 {\tiny[55.6, 72.6]} ($n$ 117) & 64.2 {\tiny[56.6, 71.1]} ($n$ 159) & 88.0 {\tiny[82.1, 93.2]} ($n$ 117) \\
Commit & MP3D & same target-loss & 87.1 {\tiny[81.3, 92.1]} ($n$ 139) & 58.7 {\tiny[49.6, 66.9]} ($n$ 121) & 62.6 {\tiny[55.0, 69.6]} ($n$ 171) & 84.3 {\tiny[77.7, 90.9]} ($n$ 121) \\
Commit & HM3D & mirrored & 86.9 {\tiny[81.0, 92.0]} ($n$ 137) & 57.1 {\tiny[47.9, 66.4]} ($n$ 119) & 63.5 {\tiny[55.8, 71.2]} ($n$ 156) & 89.1 {\tiny[83.2, 94.1]} ($n$ 119) \\
Commit & HM3D & same target-loss & 86.7 {\tiny[81.1, 92.3]} ($n$ 143) & 67.7 {\tiny[59.7, 75.8]} ($n$ 124) & 71.9 {\tiny[64.7, 78.4]} ($n$ 167) & 88.7 {\tiny[83.1, 93.5]} ($n$ 124) \\
Commit & HSSD & mirrored & 95.7 {\tiny[91.4, 99.1]} ($n$ 116) & 56.8 {\tiny[47.7, 65.8]} ($n$ 111) & 69.6 {\tiny[61.6, 76.8]} ($n$ 138) & 82.0 {\tiny[74.8, 89.2]} ($n$ 111) \\
Commit & HSSD & same target-loss & 95.7 {\tiny[92.1, 98.6]} ($n$ 139) & 57.9 {\tiny[49.6, 66.2]} ($n$ 133) & 63.5 {\tiny[56.0, 71.1]} ($n$ 159) & 91.0 {\tiny[85.7, 95.5]} ($n$ 133) \\
\bottomrule
\end{tabular}
\end{table*}

\subsection{Use of the Robot's Own Motion After Fine-Tuning}
\label{app:interventions_internals}

\noindent\textbf{Setup.} The rotation map, the rotation-mirror test, the interchange test and the steering and erasure
tests of \cref{app:internals_causal,app:internals_mirror,app:internals_interchange} are repeated on the four recipes,
with the frozen model side by side: the same MP3D recordings, the 486 command-free rotations, the 454 rotation-mirror
pairs (the events whose decision is the first to see the rotation; \cref{tab:app_mirror} uses all 486) with their
visibility groups, the same patches, random directions and scorers. The rotation-mirror test is also
run on the HSSD and HM3D rotation-map recordings, with pairs built as on MP3D: 417 on HSSD and 371 on HM3D, where the 8
events of 2 episodes whose re-rendered frames did not match the recording and one event on which \emph{Say}'s answer has no action field
are left out. Its values are given per dataset and pooled over the 1{,}242 pairs, with the episode bootstrap stratified
by dataset; the other tests run on the three datasets and are given per dataset (HM3D: 429 command-free rotations
and 372 pairs; HSSD: 459 and 417). Each recipe runs its final
checkpoint with its validation settings (\emph{Decompress}, \emph{Say} and \emph{Commit} with the previous frame at
$8\times8$), fits its own clean-fitted rotation and target-bearing directions at every state, and takes its primary
state from its own map on each dataset (MP3D $L_{15}$, $L_{16}$, $L_{17}$ and $L_{17}$; HM3D $L_{15}$, $L_{16}$,
$L_{18}$ and $L_{22}$; HSSD $L_{15}$, $L_{16}$, $L_{16}$ and $L_{20}$). For \emph{Say} and \emph{Commit} the turn
$p(\text{left})-p(\text{right})$ is read at the first action word under the model's own statement for the original
frame, held fixed in every condition, and the decision state is the last prompt position; the statement generated
freely on the mirrored frame is recorded as well. The recordings are the frozen model's trajectories, the recorded MP3D
and HM3D scenes belong to the fine-tuning data (the training split) and the HSSD scenes do not.

\noindent\textbf{Rotation-mirror test.} \Cref{tab:supp_ft_mirror}. On the generated turn, every recipe's difference from
the frozen model excludes zero on all events of every dataset. With the target out of view in both versions it does so
on MP3D and HM3D for every recipe, and on HSSD ($n=55$) for \emph{Base}, \emph{Say} and \emph{Commit}
(\emph{Decompress}: $+0.091$ [$0.0$, $+0.185$]).

\begin{figure}[!htbp]
\centering
\includegraphics[width=\columnwidth]{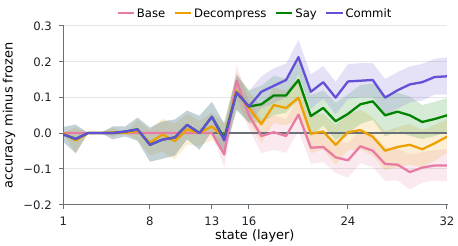}
\caption{The rotation direction decoded at every state after fine-tuning: each recipe's held-out direction accuracy minus the frozen model's, on the 486 command-free MP3D rotations (paired). Bands: 95\% episode-bootstrap intervals, not adjusted for the number of states; exploratory beyond each model's primary state. Values: \cref{tab:supp_ft_decode}.}
\label{fig:supp_ft_decode}
\end{figure}

\begin{figure*}[tbp]
\centering
\includegraphics[width=\textwidth]{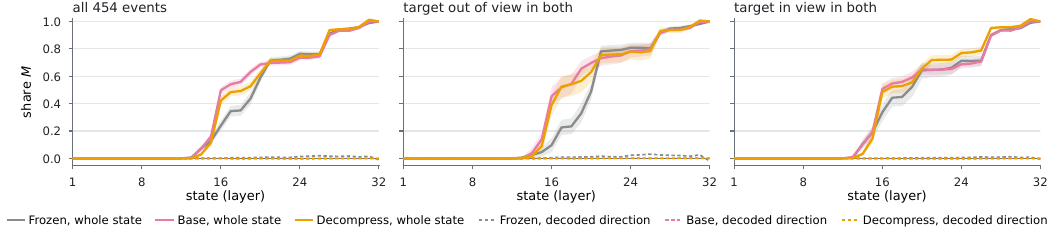}
\caption{Whole-state interchange by state for the frozen model, \emph{Base} and \emph{Decompress}: share $M$ of the mirror's effect on the turn carried by the whole decision state (solid) and by the decoded direction alone (dashed); left to right: all 454 MP3D pairs, those with the target out of view in both versions ($n=117$) and those with it in view in both ($n=167$). Bands: 95\% episode-bootstrap intervals of the whole-state share.}
\label{fig:supp_ft_layers}
\end{figure*}

\noindent\textbf{The decoded direction.} \Cref{tab:supp_ft_direction}. For \emph{Say} and \emph{Commit} the target-bearing positive control moves the
action by at most $0.06$ log-odds on any dataset, against $0.37$--$2.05$ in the frozen model: with the statement held fixed, an edit at the
decision state barely reaches the action word, and the whole state carries at most $0.06$ of the mirror's effect at any
state. \Cref{fig:supp_ft_layers} gives the whole-state share by state for the frozen model, \emph{Base} and
\emph{Decompress} on MP3D. With the target out of view in both versions, the whole state at $L_{16}$ carries $0.45$ and
$0.39$ in \emph{Base} and \emph{Decompress} against $0.10$ in the frozen model on MP3D, $0.51$ and $0.35$ against
$0.07$ on HM3D, and $0.34$ and $0.36$ against $0.13$ on HSSD: $2.5$--$7.4$ times the frozen model's share (ratios of the
unrounded shares). Across states on MP3D
(\cref{fig:supp_ft_decode} and \cref{tab:supp_ft_decode}; exploratory beyond each primary state), \emph{Commit} decodes the
rotation better than the frozen model at every state from $L_{15}$ to $L_{32}$ ($+0.07$ to $+0.21$), \emph{Say} at 12 of the
20 states from $L_{13}$, and \emph{Decompress} at five between $L_{15}$ and $L_{20}$; \emph{Base} does so at $L_{15}$, $L_{16}$
and $L_{20}$ and falls below the frozen model at $L_{14}$ ($-0.06$) and at nine states from $L_{23}$ on ($-0.05$ to $-0.11$). On HM3D and HSSD,
\emph{Commit} and \emph{Say} do so at 20 and 19 of the 20 states, \emph{Decompress} at 8 and 18, and \emph{Base} at two
and one, falling below the frozen model at 12 and 13 states.

\noindent\textbf{The statement route.} \Cref{tab:supp_ft_statement,tab:supp_ft_say}. With the target out of view in
both versions, the stated Self side flips with the mirror on $0.57$ (\emph{Say}) and $0.59$ (\emph{Commit}) of the
pooled events, $0.62$--$0.68$ on MP3D and HM3D, and $0.42$ and $0.30$ on HSSD, the only dataset whose scenes are
not in the fine-tuning data. Exchanging left and right in the Self line of the model's own statement for the original
frame moves the turn toward the mirrored one by $0.9$--$1.3$ (\emph{Say}) and $1.2$--$1.4$ (\emph{Commit}) times the
mirror's own effect under the fixed statement on the three datasets; with the target out of view its interval
excludes zero except for \emph{Commit} on HSSD (53 events). The Person line moves it by $0.4$ and $0.7$ times on MP3D
and $0.4$ and $0.5$ on HSSD (its interval includes zero with the target out of view), and not on HM3D ($-0.03$ and
$-0.03$, intervals including zero; with the target out of view \emph{Commit}'s is $-1.01$ [$-2.01$, $-0.02$]). Both
lines move it by $0.7$--$1.2$ and $1.0$--$2.2$ times.

\begin{table*}[tbp]
\centering
\scriptsize
\setlength{\tabcolsep}{2pt}
\caption{Rotation-mirror test on the fine-tuned models, on the rotation-mirror pairs of \cref{app:internals_mirror} and those built in the same way on HSSD and HM3D (\cref{app:interventions_internals}). $\Delta_{\text{follow}}$: share turning against the rotation in both versions minus share turning with it in both, read from the turn probability $p(\text{left})-p(\text{right})$ of the first action word (for \emph{Say} and \emph{Commit} under the model's own statement for the original frame, held fixed on the mirrored frame) and from the generated turn (the executed action word of a free generation on each frame; forward counts as no turn). \emph{vs frozen}: paired difference; per dataset only this is shown. 95\% episode-bootstrap intervals, 5{,}000 resamples, seed 0, stratified by dataset when pooled.}
\label{tab:supp_ft_mirror}
\begin{tabular}{@{}llcccccc@{}}
\toprule
 & & \multicolumn{3}{c}{Turn probability} & \multicolumn{3}{c}{Generated turn} \\
\cmidrule(lr){3-5}\cmidrule(l){6-8}
Model & & all & out of view & in view & all & out of view & in view \\
\midrule
\multicolumn{8}{@{}l}{\emph{Pooled over MP3D, HSSD and HM3D ($n=1242$; 287 out of view in both, 539 in view in both)}} \\
\midrule
Frozen & $\Delta_{\text{follow}}$ & +0.163 {\tiny[+0.125, +0.200]} & $-$0.244 {\tiny[$-$0.305, $-$0.182]} & +0.404 {\tiny[+0.347, +0.460]} & +0.026 {\tiny[+0.009, +0.042]} & $-$0.073 {\tiny[$-$0.110, $-$0.036]} & +0.065 {\tiny[+0.043, +0.088]} \\
Base & $\Delta_{\text{follow}}$ & +0.536 {\tiny[+0.502, +0.570]} & +0.192 {\tiny[+0.118, +0.268]} & +0.711 {\tiny[+0.670, +0.749]} & +0.213 {\tiny[+0.188, +0.238]} & +0.101 {\tiny[+0.047, +0.157]} & +0.195 {\tiny[+0.162, +0.228]} \\
 & vs frozen & +0.373 {\tiny[+0.331, +0.417]} & +0.436 {\tiny[+0.349, +0.523]} & +0.306 {\tiny[+0.239, +0.377]} & +0.187 {\tiny[+0.160, +0.214]} & +0.174 {\tiny[+0.114, +0.234]} & +0.130 {\tiny[+0.099, +0.161]} \\
Decompress & $\Delta_{\text{follow}}$ & +0.478 {\tiny[+0.446, +0.512]} & +0.122 {\tiny[+0.052, +0.190]} & +0.610 {\tiny[+0.563, +0.656]} & +0.221 {\tiny[+0.195, +0.248]} & +0.111 {\tiny[+0.053, +0.168]} & +0.176 {\tiny[+0.144, +0.208]} \\
 & vs frozen & +0.315 {\tiny[+0.275, +0.357]} & +0.366 {\tiny[+0.284, +0.445]} & +0.206 {\tiny[+0.139, +0.276]} & +0.196 {\tiny[+0.167, +0.224]} & +0.185 {\tiny[+0.122, +0.248]} & +0.111 {\tiny[+0.079, +0.143]} \\
Say & $\Delta_{\text{follow}}$ & +0.210 {\tiny[+0.183, +0.236]} & +0.052 {\tiny[+0.017, +0.090]} & +0.308 {\tiny[+0.267, +0.349]} & +0.263 {\tiny[+0.235, +0.291]} & +0.167 {\tiny[+0.108, +0.228]} & +0.187 {\tiny[+0.153, +0.222]} \\
 & vs frozen & +0.047 {\tiny[+0.007, +0.087]} & +0.296 {\tiny[+0.225, +0.369]} & $-$0.096 {\tiny[$-$0.154, $-$0.037]} & +0.238 {\tiny[+0.207, +0.269]} & +0.240 {\tiny[+0.168, +0.317]} & +0.122 {\tiny[+0.086, +0.158]} \\
Commit & $\Delta_{\text{follow}}$ & +0.130 {\tiny[+0.109, +0.152]} & +0.070 {\tiny[+0.031, +0.108]} & +0.212 {\tiny[+0.173, +0.250]} & +0.362 {\tiny[+0.331, +0.392]} & +0.192 {\tiny[+0.125, +0.257]} & +0.341 {\tiny[+0.298, +0.385]} \\
 & vs frozen & $-$0.033 {\tiny[$-$0.073, +0.006]} & +0.314 {\tiny[+0.239, +0.386]} & $-$0.193 {\tiny[$-$0.255, $-$0.129]} & +0.337 {\tiny[+0.302, +0.370]} & +0.265 {\tiny[+0.187, +0.339]} & +0.276 {\tiny[+0.231, +0.322]} \\
\midrule
\multicolumn{8}{@{}l}{\emph{MP3D ($n=454$; 117 out of view in both, 167 in view in both)}} \\
\midrule
Base & vs frozen & +0.460 {\tiny[+0.383, +0.544]} & +0.444 {\tiny[+0.300, +0.591]} & +0.401 {\tiny[+0.250, +0.561]} & +0.220 {\tiny[+0.177, +0.263]} & +0.145 {\tiny[+0.054, +0.240]} & +0.192 {\tiny[+0.127, +0.252]} \\
Decompress & vs frozen & +0.394 {\tiny[+0.310, +0.486]} & +0.376 {\tiny[+0.233, +0.522]} & +0.263 {\tiny[+0.109, +0.431]} & +0.233 {\tiny[+0.182, +0.285]} & +0.154 {\tiny[+0.055, +0.254]} & +0.156 {\tiny[+0.087, +0.225]} \\
Say & vs frozen & +0.154 {\tiny[+0.086, +0.223]} & +0.282 {\tiny[+0.167, +0.400]} & $-$0.012 {\tiny[$-$0.140, +0.119]} & +0.291 {\tiny[+0.235, +0.346]} & +0.205 {\tiny[+0.096, +0.319]} & +0.174 {\tiny[+0.098, +0.247]} \\
Commit & vs frozen & +0.057 {\tiny[$-$0.011, +0.130]} & +0.308 {\tiny[+0.186, +0.436]} & $-$0.168 {\tiny[$-$0.303, $-$0.024]} & +0.412 {\tiny[+0.351, +0.472]} & +0.291 {\tiny[+0.175, +0.402]} & +0.377 {\tiny[+0.285, +0.466]} \\
\midrule
\multicolumn{8}{@{}l}{\emph{HSSD ($n=417$; 55 out of view in both, 216 in view in both)}} \\
\midrule
Base & vs frozen & +0.264 {\tiny[+0.199, +0.329]} & +0.127 {\tiny[+0.036, +0.226]} & +0.269 {\tiny[+0.177, +0.362]} & +0.153 {\tiny[+0.114, +0.194]} & +0.073 {\tiny[+0.017, +0.148]} & +0.083 {\tiny[+0.040, +0.130]} \\
Decompress & vs frozen & +0.199 {\tiny[+0.138, +0.259]} & +0.073 {\tiny[$-$0.037, +0.185]} & +0.157 {\tiny[+0.070, +0.248]} & +0.153 {\tiny[+0.116, +0.193]} & +0.091 {\tiny[0.000, +0.185]} & +0.079 {\tiny[+0.037, +0.123]} \\
Say & vs frozen & $-$0.125 {\tiny[$-$0.187, $-$0.064]} & +0.055 {\tiny[$-$0.070, +0.185]} & $-$0.167 {\tiny[$-$0.247, $-$0.092]} & +0.206 {\tiny[+0.162, +0.251]} & +0.236 {\tiny[+0.086, +0.400]} & +0.111 {\tiny[+0.065, +0.161]} \\
Commit & vs frozen & $-$0.206 {\tiny[$-$0.272, $-$0.142]} & +0.091 {\tiny[$-$0.035, +0.222]} & $-$0.250 {\tiny[$-$0.333, $-$0.163]} & +0.269 {\tiny[+0.220, +0.315]} & +0.218 {\tiny[+0.093, +0.346]} & +0.218 {\tiny[+0.157, +0.279]} \\
\midrule
\multicolumn{8}{@{}l}{\emph{HM3D ($n=371$; 115 out of view in both, 156 in view in both)}} \\
\midrule
Base & vs frozen & +0.388 {\tiny[+0.318, +0.460]} & +0.574 {\tiny[+0.415, +0.730]} & +0.256 {\tiny[+0.153, +0.366]} & +0.183 {\tiny[+0.127, +0.243]} & +0.252 {\tiny[+0.146, +0.362]} & +0.128 {\tiny[+0.075, +0.186]} \\
Decompress & vs frozen & +0.348 {\tiny[+0.285, +0.410]} & +0.496 {\tiny[+0.364, +0.621]} & +0.212 {\tiny[+0.116, +0.308]} & +0.197 {\tiny[+0.144, +0.252]} & +0.261 {\tiny[+0.153, +0.377]} & +0.109 {\tiny[+0.049, +0.172]} \\
Say & vs frozen & +0.108 {\tiny[+0.031, +0.183]} & +0.426 {\tiny[+0.303, +0.550]} & $-$0.090 {\tiny[$-$0.188, +0.007]} & +0.208 {\tiny[+0.144, +0.270]} & +0.278 {\tiny[+0.154, +0.402]} & +0.083 {\tiny[+0.018, +0.151]} \\
Commit & vs frozen & +0.051 {\tiny[$-$0.019, +0.119]} & +0.426 {\tiny[+0.306, +0.544]} & $-$0.141 {\tiny[$-$0.241, $-$0.037]} & +0.321 {\tiny[+0.254, +0.385]} & +0.261 {\tiny[+0.121, +0.398]} & +0.250 {\tiny[+0.163, +0.343]} \\
\bottomrule
\end{tabular}
\end{table*}

\begin{table*}[tbp]
\centering
\scriptsize
\setlength{\tabcolsep}{2.5pt}
\caption{The decoded rotation direction after fine-tuning, per dataset, at each model's primary decision state $k^*$ (the state farthest from chance in its own rotation map on that dataset). \emph{Decoded}: held-out direction accuracy of the clean-fitted probe at $k^*$ on the command-free rotations (in parentheses, the frozen model at the same state). $M_{\text{dir}}$: share of the mirror's effect on the turn carried by setting only the decoded direction to its mirrored value (rotation-mirror pairs). \emph{Steer}: change in corrective log-odds from a $3$\,SD shift along the direction, with the range of 20 norm-matched random directions; \emph{erase}: change in the corrective share from erasing the direction at states 13--32 (HM3D and HSSD: 18--32; the frozen model's windows, used for every recipe; random erasure in parentheses); \emph{control}: the target-bearing positive control, in log-odds. For \emph{Say} and \emph{Commit} the turn is read under their statement for the original frame, held fixed.}
\label{tab:supp_ft_direction}
\begin{tabular}{@{}lccccccc@{}}
\toprule
Model & $k^*$ & Decoded (frozen) & $M_{\text{dir}}$ & Steer $3$\,SD & Random range & Erase (random) & Control \\
\midrule
\multicolumn{8}{@{}l}{\emph{MP3D (486 command-free rotations, 454 pairs)}} \\
\midrule
Frozen & $L_{16}$ & 0.741 {\tiny[0.701, 0.778]} & +0.002 {\tiny[+0.001, +0.003]} & $-$0.018 {\tiny[$-$0.021, $-$0.014]} & $-$0.029 to +0.020 & +0.002 {\tiny[$-$0.004, +0.010]} (+0.002) & 0.37 {\tiny[0.34, 0.41]} \\
Base & $L_{15}$ & 0.817 {\tiny[0.774, 0.856]} (0.671) & $-$0.001 {\tiny[$-$0.002, $-$0.001]} & $-$0.024 {\tiny[$-$0.029, $-$0.020]} & $-$0.023 to +0.018 & +0.002 {\tiny[0.000, +0.007]} (+0.002) & 0.78 {\tiny[0.70, 0.86]} \\
Decompress & $L_{16}$ & 0.811 {\tiny[0.770, 0.850]} (0.741) & 0.000 {\tiny[$-$0.001, 0.000]} & $-$0.024 {\tiny[$-$0.031, $-$0.017]} & $-$0.068 to +0.050 & +0.012 {\tiny[+0.004, +0.023]} ($-$0.002) & 1.81 {\tiny[1.66, 1.95]} \\
Say & $L_{17}$ & 0.819 {\tiny[0.781, 0.855]} (0.739) & 0.000 {\tiny[0.000, 0.000]} & +0.001 {\tiny[0.000, +0.001]} & 0.000 to +0.001 & 0.000 {\tiny[0.000, 0.000]} (0.000) & 0.06 {\tiny[0.05, 0.07]} \\
Commit & $L_{17}$ & 0.854 {\tiny[0.821, 0.885]} (0.739) & 0.000 {\tiny[0.000, 0.000]} & 0.000 {\tiny[0.000, +0.001]} & 0.000 to 0.000 & 0.000 {\tiny[0.000, 0.000]} (0.000) & 0.03 {\tiny[0.02, 0.05]} \\
\midrule
\multicolumn{8}{@{}l}{\emph{HM3D (429 command-free rotations, 372 pairs)}} \\
\midrule
Frozen & $L_{21}$ & 0.706 {\tiny[0.645, 0.761]} & +0.006 {\tiny[+0.004, +0.007]} & $-$0.078 {\tiny[$-$0.083, $-$0.072]} & $-$0.052 to +0.048 & +0.005 {\tiny[$-$0.005, +0.014]} (0.000) & 2.05 {\tiny[1.83, 2.31]} \\
Base & $L_{15}$ & 0.769 {\tiny[0.714, 0.819]} (0.552) & $-$0.002 {\tiny[$-$0.003, $-$0.001]} & $-$0.045 {\tiny[$-$0.053, $-$0.037]} & $-$0.028 to +0.018 & +0.005 {\tiny[0.000, +0.012]} (0.000) & 2.27 {\tiny[2.00, 2.56]} \\
Decompress & $L_{16}$ & 0.807 {\tiny[0.745, 0.859]} (0.667) & 0.000 {\tiny[$-$0.001, 0.000]} & $-$0.041 {\tiny[$-$0.051, $-$0.031]} & $-$0.077 to +0.051 & $-$0.002 {\tiny[$-$0.010, +0.005]} (+0.005) & 5.54 {\tiny[5.07, 6.00]} \\
Say & $L_{18}$ & 0.844 {\tiny[0.798, 0.887]} (0.646) & 0.000 {\tiny[0.000, 0.000]} & 0.000 {\tiny[0.000, +0.001]} & $-$0.001 to 0.000 & 0.000 {\tiny[0.000, 0.000]} (0.000) & 0.00 {\tiny[0.00, 0.00]} \\
Commit & $L_{22}$ & 0.832 {\tiny[0.778, 0.879]} (0.685) & 0.000 {\tiny[0.000, 0.000]} & 0.000 {\tiny[0.000, +0.001]} & $-$0.001 to 0.000 & 0.000 {\tiny[0.000, 0.000]} (0.000) & 0.00 {\tiny[0.00, 0.00]} \\
\midrule
\multicolumn{8}{@{}l}{\emph{HSSD (459 command-free rotations, 417 pairs)}} \\
\midrule
Frozen & $L_{21}$ & 0.634 {\tiny[0.582, 0.684]} & 0.000 {\tiny[$-$0.003, +0.003]} & $-$0.237 {\tiny[$-$0.246, $-$0.227]} & $-$0.053 to +0.062 & +0.011 {\tiny[0.000, +0.023]} (0.000) & 1.15 {\tiny[1.02, 1.29]} \\
Base & $L_{15}$ & 0.645 {\tiny[0.590, 0.696]} (0.397) & $-$0.001 {\tiny[$-$0.001, 0.000]} & $-$0.022 {\tiny[$-$0.026, $-$0.017]} & $-$0.022 to +0.017 & 0.000 {\tiny[0.000, 0.000]} (0.000) & 0.48 {\tiny[0.42, 0.55]} \\
Decompress & $L_{16}$ & 0.743 {\tiny[0.698, 0.786]} (0.571) & $-$0.001 {\tiny[$-$0.002, $-$0.001]} & $-$0.048 {\tiny[$-$0.054, $-$0.043]} & $-$0.048 to +0.040 & +0.004 {\tiny[0.000, +0.011]} (+0.004) & 1.11 {\tiny[1.01, 1.21]} \\
Say & $L_{16}$ & 0.845 {\tiny[0.807, 0.881]} (0.571) & 0.000 {\tiny[0.000, 0.000]} & +0.001 {\tiny[0.000, +0.002]} & $-$0.002 to +0.003 & 0.000 {\tiny[0.000, 0.000]} (0.000) & 0.00 {\tiny[0.00, 0.00]} \\
Commit & $L_{20}$ & 0.834 {\tiny[0.797, 0.868]} (0.477) & 0.000 {\tiny[0.000, 0.000]} & 0.000 {\tiny[0.000, +0.001]} & 0.000 to +0.001 & 0.000 {\tiny[0.000, 0.000]} (0.000) & 0.00 {\tiny[0.00, 0.00]} \\
\bottomrule
\end{tabular}
\end{table*}

\begin{table*}[tbp]
\centering
\scriptsize
\setlength{\tabcolsep}{2.5pt}
\caption{The rotation direction decoded at every state after fine-tuning (exploratory beyond each model's primary state): held-out direction accuracy of the clean-fitted probe on the 486 command-free MP3D rotations, the frozen model's, and each recipe's with its paired difference from the frozen model. The first 13 layers are not fine-tuned: \emph{Base}, whose input is unchanged, equals the frozen model up to $L_{13}$. 95\% episode-bootstrap intervals (5{,}000 resamples, seed 0), not adjusted for the 20 states compared. Bold: the paired difference's interval excludes 0.}
\label{tab:supp_ft_decode}
\begin{tabular}{@{}lccccc@{}}
\toprule
State & Frozen & Base & Decompress & Say & Commit \\
\midrule
$L_{13}$ & 0.675 {\tiny[0.631, 0.717]} & 0.675 0.000 {\tiny[0.000, 0.000]} & 0.693 +0.019 {\tiny[$-$0.020, +0.057]} & 0.720 \textbf{+0.045} {\tiny[+0.004, +0.087]} & 0.720 \textbf{+0.045} {\tiny[+0.004, +0.087]} \\
$L_{14}$ & 0.728 {\tiny[0.687, 0.769]} & 0.669 \textbf{$-$0.060} {\tiny[$-$0.097, $-$0.023]} & 0.716 $-$0.012 {\tiny[$-$0.052, +0.029]} & 0.714 $-$0.014 {\tiny[$-$0.056, +0.027]} & 0.708 $-$0.021 {\tiny[$-$0.067, +0.026]} \\
$L_{15}$ & 0.671 {\tiny[0.627, 0.712]} & 0.817 \textbf{+0.146} {\tiny[+0.102, +0.191]} & 0.788 \textbf{+0.117} {\tiny[+0.068, +0.169]} & 0.784 \textbf{+0.113} {\tiny[+0.064, +0.162]} & 0.784 \textbf{+0.113} {\tiny[+0.065, +0.162]} \\
$L_{16}$ & 0.741 {\tiny[0.701, 0.778]} & 0.790 \textbf{+0.049} {\tiny[+0.010, +0.089]} & 0.811 \textbf{+0.070} {\tiny[+0.028, +0.114]} & 0.815 \textbf{+0.074} {\tiny[+0.027, +0.120]} & 0.815 \textbf{+0.074} {\tiny[+0.031, +0.117]} \\
$L_{17}$ & 0.739 {\tiny[0.697, 0.781]} & 0.730 $-$0.008 {\tiny[$-$0.057, +0.040]} & 0.763 +0.025 {\tiny[$-$0.028, +0.075]} & 0.819 \textbf{+0.080} {\tiny[+0.031, +0.131]} & 0.854 \textbf{+0.115} {\tiny[+0.070, +0.158]} \\
$L_{18}$ & 0.700 {\tiny[0.653, 0.744]} & 0.702 +0.002 {\tiny[$-$0.047, +0.049]} & 0.778 \textbf{+0.078} {\tiny[+0.028, +0.127]} & 0.805 \textbf{+0.105} {\tiny[+0.050, +0.160]} & 0.831 \textbf{+0.132} {\tiny[+0.083, +0.180]} \\
$L_{19}$ & 0.691 {\tiny[0.650, 0.732]} & 0.683 $-$0.008 {\tiny[$-$0.055, +0.038]} & 0.761 \textbf{+0.070} {\tiny[+0.021, +0.117]} & 0.796 \textbf{+0.105} {\tiny[+0.057, +0.152]} & 0.840 \textbf{+0.148} {\tiny[+0.101, +0.194]} \\
$L_{20}$ & 0.632 {\tiny[0.589, 0.675]} & 0.683 \textbf{+0.051} {\tiny[+0.005, +0.099]} & 0.730 \textbf{+0.099} {\tiny[+0.041, +0.155]} & 0.780 \textbf{+0.148} {\tiny[+0.092, +0.203]} & 0.844 \textbf{+0.212} {\tiny[+0.164, +0.260]} \\
$L_{21}$ & 0.724 {\tiny[0.680, 0.768]} & 0.683 $-$0.041 {\tiny[$-$0.086, +0.004]} & 0.722 $-$0.002 {\tiny[$-$0.057, +0.053]} & 0.772 +0.047 {\tiny[$-$0.009, +0.106]} & 0.840 \textbf{+0.115} {\tiny[+0.067, +0.164]} \\
$L_{22}$ & 0.710 {\tiny[0.669, 0.751]} & 0.671 $-$0.039 {\tiny[$-$0.081, +0.004]} & 0.714 +0.004 {\tiny[$-$0.046, +0.053]} & 0.780 \textbf{+0.070} {\tiny[+0.021, +0.119]} & 0.852 \textbf{+0.142} {\tiny[+0.096, +0.187]} \\
$L_{23}$ & 0.735 {\tiny[0.693, 0.775]} & 0.667 \textbf{$-$0.068} {\tiny[$-$0.111, $-$0.023]} & 0.702 $-$0.033 {\tiny[$-$0.088, +0.022]} & 0.767 +0.033 {\tiny[$-$0.019, +0.084]} & 0.833 \textbf{+0.099} {\tiny[+0.053, +0.143]} \\
$L_{24}$ & 0.695 {\tiny[0.653, 0.739]} & 0.619 \textbf{$-$0.076} {\tiny[$-$0.124, $-$0.029]} & 0.698 +0.002 {\tiny[$-$0.048, +0.052]} & 0.749 +0.053 {\tiny[0.000, +0.108]} & 0.840 \textbf{+0.144} {\tiny[+0.095, +0.193]} \\
$L_{25}$ & 0.681 {\tiny[0.635, 0.727]} & 0.644 $-$0.037 {\tiny[$-$0.084, +0.009]} & 0.689 +0.008 {\tiny[$-$0.046, +0.065]} & 0.761 \textbf{+0.080} {\tiny[+0.025, +0.134]} & 0.827 \textbf{+0.146} {\tiny[+0.100, +0.195]} \\
$L_{26}$ & 0.685 {\tiny[0.642, 0.729]} & 0.636 \textbf{$-$0.049} {\tiny[$-$0.095, $-$0.002]} & 0.675 $-$0.010 {\tiny[$-$0.064, +0.044]} & 0.774 \textbf{+0.088} {\tiny[+0.037, +0.138]} & 0.833 \textbf{+0.148} {\tiny[+0.100, +0.196]} \\
$L_{27}$ & 0.710 {\tiny[0.666, 0.753]} & 0.623 \textbf{$-$0.086} {\tiny[$-$0.134, $-$0.039]} & 0.660 $-$0.049 {\tiny[$-$0.104, +0.004]} & 0.759 +0.049 {\tiny[$-$0.010, +0.107]} & 0.809 \textbf{+0.099} {\tiny[+0.048, +0.151]} \\
$L_{28}$ & 0.698 {\tiny[0.651, 0.742]} & 0.609 \textbf{$-$0.088} {\tiny[$-$0.133, $-$0.044]} & 0.658 $-$0.039 {\tiny[$-$0.096, +0.017]} & 0.757 \textbf{+0.060} {\tiny[+0.006, +0.112]} & 0.817 \textbf{+0.119} {\tiny[+0.068, +0.170]} \\
$L_{29}$ & 0.691 {\tiny[0.648, 0.735]} & 0.582 \textbf{$-$0.109} {\tiny[$-$0.158, $-$0.059]} & 0.658 $-$0.033 {\tiny[$-$0.089, +0.025]} & 0.741 +0.049 {\tiny[$-$0.008, +0.102]} & 0.827 \textbf{+0.136} {\tiny[+0.086, +0.186]} \\
$L_{30}$ & 0.693 {\tiny[0.649, 0.738]} & 0.597 \textbf{$-$0.097} {\tiny[$-$0.139, $-$0.053]} & 0.648 $-$0.045 {\tiny[$-$0.096, +0.009]} & 0.724 +0.031 {\tiny[$-$0.017, +0.078]} & 0.835 \textbf{+0.142} {\tiny[+0.091, +0.193]} \\
$L_{31}$ & 0.683 {\tiny[0.637, 0.730]} & 0.593 \textbf{$-$0.091} {\tiny[$-$0.133, $-$0.048]} & 0.654 $-$0.029 {\tiny[$-$0.078, +0.023]} & 0.722 +0.039 {\tiny[$-$0.009, +0.087]} & 0.840 \textbf{+0.156} {\tiny[+0.107, +0.207]} \\
$L_{32}$ & 0.671 {\tiny[0.626, 0.716]} & 0.580 \textbf{$-$0.091} {\tiny[$-$0.135, $-$0.045]} & 0.660 $-$0.010 {\tiny[$-$0.056, +0.038]} & 0.720 \textbf{+0.049} {\tiny[+0.002, +0.097]} & 0.829 \textbf{+0.158} {\tiny[+0.107, +0.211]} \\
\bottomrule
\end{tabular}
\end{table*}

\begin{table*}[tbp]
\centering
\scriptsize
\setlength{\tabcolsep}{3pt}
\caption{The statement route under the rotation mirror (\emph{Say} and \emph{Commit}; free generation on the original and on the mirrored frame). \emph{Self / Person flips}: share of events whose stated side of the robot's own motion / of the target changes between the two frames, among events whose line names exactly one side on both ($n$). $\Delta_{\text{follow}}$ of the generated turn among events whose Self side flipped and whose Self side did not ($n$). Per dataset, only the events with the target out of view in both are shown. 95\% episode-bootstrap intervals, stratified by dataset when pooled.}
\label{tab:supp_ft_statement}
\begin{tabular}{@{}llcccc@{}}
\toprule
Model & Events & Self flips & Person flips & $\Delta_{\text{follow}}$, Self flipped & $\Delta_{\text{follow}}$, Self unchanged \\
\midrule
\multicolumn{6}{@{}l}{\emph{Pooled over MP3D, HSSD and HM3D ($n=1242$; 287 out of view in both, 539 in view in both)}} \\
\midrule
Say & all & 0.689 {\tiny[0.652, 0.725]} ($n$ 679) & 0.265 {\tiny[0.235, 0.295]} ($n$ 921) & +0.639 {\tiny[+0.588, +0.688]} ($n$ 468) & +0.052 {\tiny[+0.014, +0.091]} ($n$ 211) \\
 & out of view in both & 0.571 {\tiny[0.500, 0.640]} ($n$ 205) & 0.208 {\tiny[0.161, 0.259]} ($n$ 259) & +0.419 {\tiny[+0.295, +0.532]} ($n$ 117) & $-$0.011 {\tiny[$-$0.060, +0.035]} ($n$ 88) \\
 & in view in both & 0.762 {\tiny[0.705, 0.817]} ($n$ 189) & 0.215 {\tiny[0.171, 0.265]} ($n$ 297) & +0.611 {\tiny[+0.530, +0.686]} ($n$ 144) & +0.133 {\tiny[+0.044, +0.227]} ($n$ 45) \\
Commit & all & 0.729 {\tiny[0.698, 0.760]} ($n$ 879) & 0.332 {\tiny[0.303, 0.362]} ($n$ 1036) & +0.627 {\tiny[+0.585, +0.668]} ($n$ 641) & +0.029 {\tiny[0.000, +0.062]} ($n$ 238) \\
 & out of view in both & 0.588 {\tiny[0.526, 0.649]} ($n$ 245) & 0.197 {\tiny[0.152, 0.246]} ($n$ 279) & +0.340 {\tiny[+0.224, +0.452]} ($n$ 144) & 0.000 {\tiny[$-$0.048, +0.050]} ($n$ 101) \\
 & in view in both & 0.850 {\tiny[0.810, 0.887]} ($n$ 293) & 0.405 {\tiny[0.351, 0.457]} ($n$ 368) & +0.671 {\tiny[+0.610, +0.728]} ($n$ 249) & +0.045 {\tiny[0.000, +0.114]} ($n$ 44) \\
\midrule
\multicolumn{6}{@{}l}{\emph{MP3D ($n=454$; 117 out of view in both, 167 in view in both)}} \\
\midrule
Say & out of view in both & 0.618 {\tiny[0.500, 0.728]} ($n$ 76) & 0.231 {\tiny[0.158, 0.309]} ($n$ 104) & +0.298 {\tiny[+0.083, +0.500]} ($n$ 47) & 0.000 {\tiny[$-$0.103, +0.103]} ($n$ 29) \\
Commit & out of view in both & 0.677 {\tiny[0.580, 0.770]} ($n$ 99) & 0.212 {\tiny[0.142, 0.287]} ($n$ 113) & +0.313 {\tiny[+0.154, +0.471]} ($n$ 67) & $-$0.031 {\tiny[$-$0.100, 0.000]} ($n$ 32) \\
\midrule
\multicolumn{6}{@{}l}{\emph{HSSD ($n=417$; 55 out of view in both, 216 in view in both)}} \\
\midrule
Say & out of view in both & 0.420 {\tiny[0.280, 0.566]} ($n$ 50) & 0.085 {\tiny[0.020, 0.174]} ($n$ 47) & +0.524 {\tiny[+0.238, +0.762]} ($n$ 21) & 0.000 {\tiny[0.000, 0.000]} ($n$ 29) \\
Commit & out of view in both & 0.302 {\tiny[0.170, 0.442]} ($n$ 53) & 0.094 {\tiny[0.019, 0.176]} ($n$ 53) & +0.625 {\tiny[+0.353, +0.875]} ($n$ 16) & 0.000 {\tiny[0.000, 0.000]} ($n$ 37) \\
\midrule
\multicolumn{6}{@{}l}{\emph{HM3D ($n=371$; 115 out of view in both, 156 in view in both)}} \\
\midrule
Say & out of view in both & 0.620 {\tiny[0.518, 0.728]} ($n$ 79) & 0.241 {\tiny[0.162, 0.327]} ($n$ 108) & +0.490 {\tiny[+0.327, +0.640]} ($n$ 49) & $-$0.033 {\tiny[$-$0.156, +0.071]} ($n$ 30) \\
Commit & out of view in both & 0.656 {\tiny[0.548, 0.763]} ($n$ 93) & 0.230 {\tiny[0.153, 0.313]} ($n$ 113) & +0.295 {\tiny[+0.102, +0.483]} ($n$ 61) & +0.031 {\tiny[$-$0.100, +0.172]} ($n$ 32) \\
\bottomrule
\end{tabular}
\end{table*}

\begin{table*}[tbp]
\centering
\footnotesize
\setlength{\tabcolsep}{3pt}
\caption{Exchanging left and right in the model's own statement for the original frame, in its Self line, its Person line, or both, on the rotation-mirror pairs of each dataset: share $M$ of the mirror's effect on the turn (the statement held fixed) that the exchange reproduces, over the events whose exchanged line names a side ($n$; all events). 95\% episode-bootstrap intervals.}
\label{tab:supp_ft_say}
\begin{tabular}{@{}llccc@{}}
\toprule
Model & Events & Self & Person & Both \\
\midrule
\multicolumn{5}{@{}l}{\emph{MP3D (454 pairs; 117 out of view in both, 167 in view in both)}} \\
\midrule
Say & all & +0.886 {\tiny[+0.739, +1.034]} ($n$ 316) & +0.376 {\tiny[+0.216, +0.532]} ($n$ 361) & +1.062 {\tiny[+0.874, +1.242]} ($n$ 393) \\
 & out of view in both & +0.795 {\tiny[+0.363, +1.256]} & +0.497 {\tiny[$-$0.041, +0.969]} & +1.100 {\tiny[+0.465, +1.686]} \\
 & in view in both & +0.942 {\tiny[+0.725, +1.135]} & +0.167 {\tiny[+0.007, +0.338]} & +0.885 {\tiny[+0.632, +1.137]} \\
Commit & all & +1.350 {\tiny[+1.094, +1.607]} ($n$ 389) & +0.698 {\tiny[+0.365, +1.056]} ($n$ 389) & +2.174 {\tiny[+1.704, +2.640]} ($n$ 419) \\
 & out of view in both & +1.331 {\tiny[+0.834, +1.862]} & +0.796 {\tiny[$-$0.128, +1.860]} & +2.148 {\tiny[+1.006, +3.487]} \\
 & in view in both & +1.310 {\tiny[+0.995, +1.696]} & +0.508 {\tiny[+0.211, +0.789]} & +1.990 {\tiny[+1.533, +2.479]} \\
\midrule
\multicolumn{5}{@{}l}{\emph{HM3D (372 pairs; 116 out of view in both, 156 in view in both)}} \\
\midrule
Say & all & +0.996 {\tiny[+0.829, +1.171]} ($n$ 236) & $-$0.028 {\tiny[$-$0.268, +0.206]} ($n$ 309) & +0.708 {\tiny[+0.454, +0.960]} ($n$ 329) \\
 & out of view in both & +0.793 {\tiny[+0.443, +1.150]} & $-$0.190 {\tiny[$-$0.978, +0.643]} & +0.525 {\tiny[$-$0.329, +1.420]} \\
 & in view in both & +0.922 {\tiny[+0.678, +1.192]} & +0.029 {\tiny[$-$0.085, +0.138]} & +0.604 {\tiny[+0.396, +0.824]} \\
Commit & all & +1.187 {\tiny[+0.941, +1.434]} ($n$ 304) & $-$0.032 {\tiny[$-$0.348, +0.289]} ($n$ 335) & +1.026 {\tiny[+0.579, +1.445]} ($n$ 349) \\
 & out of view in both & +0.726 {\tiny[+0.304, +1.100]} & $-$1.009 {\tiny[$-$2.006, $-$0.017]} & $-$0.633 {\tiny[$-$1.865, +0.558]} \\
 & in view in both & +1.208 {\tiny[+0.923, +1.519]} & +0.377 {\tiny[+0.190, +0.563]} & +1.381 {\tiny[+1.041, +1.735]} \\
\midrule
\multicolumn{5}{@{}l}{\emph{HSSD (417 pairs; 55 out of view in both, 216 in view in both)}} \\
\midrule
Say & all & +1.347 {\tiny[+1.122, +1.581]} ($n$ 269) & +0.364 {\tiny[+0.230, +0.508]} ($n$ 329) & +1.231 {\tiny[+1.007, +1.457]} ($n$ 369) \\
 & out of view in both & +1.734 {\tiny[+0.657, +2.973]} & +0.870 {\tiny[$-$0.201, +2.156]} & +2.541 {\tiny[+0.900, +4.392]} \\
 & in view in both & +1.106 {\tiny[+0.894, +1.338]} & +0.241 {\tiny[+0.101, +0.377]} & +0.843 {\tiny[+0.629, +1.061]} \\
Commit & all & +1.198 {\tiny[+0.902, +1.480]} ($n$ 341) & +0.515 {\tiny[+0.270, +0.748]} ($n$ 376) & +1.833 {\tiny[+1.388, +2.253]} ($n$ 394) \\
 & out of view in both & +0.040 {\tiny[$-$2.719, +1.882]} & $-$1.081 {\tiny[$-$3.201, +0.503]} & $-$0.863 {\tiny[$-$5.347, +2.411]} \\
 & in view in both & +1.103 {\tiny[+0.833, +1.377]} & +0.620 {\tiny[+0.419, +0.824]} & +1.701 {\tiny[+1.365, +2.039]} \\
\bottomrule
\end{tabular}
\end{table*}

\end{document}